\documentclass[twoside,11pt]{article}

\usepackage{jmlr2e}
\usepackage[utf8]{inputenc}

\usepackage{microtype}
\usepackage{graphicx}
\usepackage{booktabs} % for professional tables

\usepackage{graphicx}
\usepackage{times}   % assumes new font selection scheme installed
\usepackage{amsmath, amssymb} % assumes amsmath package installed
\usepackage{mathtools}
\usepackage{ntheorem}
\usepackage{bm}
\usepackage{algorithm}
\usepackage{algpseudocode}
\usepackage{subcaption}
\usepackage{epstopdf}
\usepackage{stfloats}

\usepackage{enumerate}

\usepackage{tikz}
\usetikzlibrary{calc}
\tikzset{mynode/.style={draw,circle, minimum size = 0.7cm}}
\usetikzlibrary{positioning,shapes,arrows}
\usepackage{xcolor}

\usepackage{pgfplots}
\usetikzlibrary{external}
\ShortHeadings{SLEIPNIR: Accurate Feature Expansion for GPR with Derivatives (PREPRINT)}{Angelis, Wenk, Schölkopf, Bauer and Krause}
\firstpageno{1}

\begin{document}
\newtheorem{theorem}{Theorem}
\newtheorem{example}[theorem]{Example}
\newtheorem{remark}[theorem]{Remark}
\newtheorem{corollary}[theorem]{Corollary}
\newtheorem{definition}[theorem]{Definition}
\newtheorem{lemma}[theorem]{Lemma}
\newtheorem{proposition}[theorem]{Proposition}
\newtheorem{proof}{Proof}
%\input{theoremsetup.tex}
% removed all theorem and proof stuff from sty file.
\title{SLEIPNIR: Deterministic and Provably Accurate Feature Expansion for Gaussian Process Regression with Derivatives}

\author{\name Emmanouil Angelis\thanks{The first two authors contributed equally.}
		\email angelise@ethz.ch \\
		\addr Learning and Adaptive Systems Group\\
		ETH Zürich\\
		\name Philippe Wenk\footnotemark[1]
		\email wenkph@ethz.ch\\ 
	    \addr Learning and Adaptive Systems Group\\
	    ETH Zürich and Max Planck ETH Center for Learning Systems
        \\
        \name Bernhard Schölkopf \email bernhard.schoelkopf@tuebingen.mpg.de\\
        \addr Empirical Inference Group\\
        Max Planck Institute for Intelligent Systems, Tübingen
        \\
        \name Stefan Bauer \email stefan.bauer@tuebingen.mpg.de\\
        \addr Empirical Inference Group\\
        Max Planck Institute for Intelligent Systems, Tübingen
        \\
        \name Andreas Krause \email krausea@ethz.ch\\
        \addr Learning and Adaptive Systems Group\\
        ETH Zürich
        }

\maketitle
% sty file has been edited such that editor is no longer required

\begin{abstract}%   <- trailing '%' for backward compatibility of .sty file
Gaussian processes are an important regression tool with excellent analytic properties which allow for direct integration of derivative observations. However, vanilla GP methods scale cubically in the amount of observations. In this work, we propose a novel approach for scaling GP regression with derivatives based on quadrature Fourier features. We then prove deterministic, non-asymptotic and exponentially fast decaying error bounds which apply for both the approximated kernel as well as the approximated posterior. To furthermore illustrate the practical applicability of our method, we then apply it to ODIN, a recently developed algorithm for ODE parameter inference. In an extensive experiments section, all results are empirically validated, demonstrating the speed, accuracy, and practical applicability of this approach.
\end{abstract}

\section{Introduction}
Gaussian process (GP) regression \citep{RasmussenBook} is an important machine learning model with many desirable properties. Due to their inherently Bayesian nature, GPs naturally provide uncertainty estimates which are of crucial importance in Bayesian optimization \citep{mockus2012bayesian} or active learning \citep{seo2000gaussian}. Furthermore, since a derivative of a GP is again a GP, there is a natural extension to derivative observations \citep{solak2003derivative}, which can be leveraged in the context of probabilistic numerics \citep{hennig2015probabilistic} or Gaussian-process-based gradient matching \citep{calderhead2009accelerating, wenk2019odin}. Despite these appealing properties, a clear drawback of classic GP regression is the fact that they scale cubic in the amount of observation points.

\paragraph{Scaling standard GP regression} For standard, unconstrained GP regression, there exist several ideas on how to tackle this problem. One family of approaches focuses on summarizing the data set with a fixed amount of pseudo-observations, the so-called inducing points \citep{quinonero2005unifying, snelson2006sparse, titsias2009variational}. Complementary to these methods, special structure in either the kernel function \citep{wilson2014fast} or the input data \citep{cunningham2008fast} can be exploited to speed up the necessary matrix-vector multiplications needed to perform GP regression. \citet{wilson2015kernel} combine these two ideas, creating an efficient approximation scheme linear in the amount of observations. Independently, \citet{sarkka2013spatiotemporal} propose a SDE-based reformulation and connect GP regression to Kalman filtering. Finally, there is a family of approaches approximating the kernel function via a finite dimensional scalar product of feature vectors. These features have been obtained using the Nyström method \citep{williams2001using}, MC samples \citep{rahimi2008random}, sparse spectrum approximations \citep{quia2010sparse}, variational optimization \citep{hensman2017variational} or a quadrature scheme \citep{mutny2018efficient}. Furthermore, \citet{solin2014hilbert} develop a deterministic feature expansion based on the Fourier transform of the Laplace operator. While they are able to provide deterministic, non-asymptotic error bounds, their error decays linearly with the the size of the domain of the operator expansion $L$, which cannot grow faster than the amount of features. This essentially means that their approximation decays at best linearly.

\paragraph{Scaling GP regression with derivatives} In the context of Gaussian process regression with derivatives though, scalable methods seem to have received little attention \citep{eriksson2018scaling}, despite their practical relevance. While some approaches like inducing points could be applied by just using the same inducing points, it is not obvious how such changes would affect the quality of the resulting estimates. Similarly, an empirical extension of the approach of \citet{solin2014hilbert} to the derivative case is presented by \citet{solin2018modeling}, without quantifying the errors of the approximation scheme. For random Fourier features (RFF), \citet{szabo2018kernel} present a rigorous theoretical analysis. However, due to the probabilistic nature of RFF, it is not possible to provide deterministic, non-asymptotic guarantees for any given data set of fixed size.

To obtain deterministic (i.e. hold with probability 1) and non-asymptotic (i.e. for data sets with a fixed amount of observations) error bounds, we thus turn to quadratic Fourier features. \citet{mutny2018efficient} recently derived exponentially fast decaying bounds for standard GP regression. Building on their work, we derive approximations and bounds for kernel derivatives. As we will demonstrate, these bounds can be used directly to quantify the absolute error of the predictive posterior mean and covariance of a GP with derivative information, leading to deterministic, non-asymptotic error bounds that can guide the choice of the amount of features needed to obtain a desired accuracy.

However, as we shall demonstrate, our bounds are so powerful that they can easily be extended to more applications. We will demonstrate this using ODE-informed regression ($\operatorname{ODIN}$) as an example. In ODE-Informed regression, Gaussian processes are used to infer the parameters of a system of ODEs governing the dynamics of a time-continuous system that is observed at discrete time points under additive Gaussian noise. The idea to use GPs for this task goes back to the pioneering work of \citet{calderhead2009accelerating}, whose theoretical models were later refined by \citet{dondelinger2013ode} and \citet{wenk2018fast}. As shown e.g. by \citet{gorbach2017scalable} and \citet{abbati2019ares}, such GP based inference schemes scale almost linearly in the dimension of the ODE. Since applications of these algorithms involve deducing novel scientific knowledge \citep{dony2019parametric, macdonald2015gradient}, any approximation scheme would require a rigorous error analysis. Fortunately, our algorithm leads to deterministic, finite sample error bounds for the risk it optimizes, again guiding the choice of the complexity of the approximation for a desired accuracy level.

\paragraph{Contributions:} In summary, we
\begin{itemize}
	\item extend the QFF framework to Gaussian process regression with derivative information, naming the extension $\operatorname{SLEIPNIR}$,
	\item derive, prove and empirically validate deterministic, finite-sample guarantees for the accuracy of said approximation,
	\item demonstrate how these theoretical insights can be used to control the approximation error in GPR with derivatives,
	\item expand these insights to consistently reduce the cubic run time of ODE informed regression \citep{wenk2019odin},
	\item verify the theoretical results empirically on four different systems with both locally linear and more involved, nonlinear dynamics, demonstrating a significant reduction in computational complexity without noticeable loss of accuracy.\\
\end{itemize}
All code needed to recreate our results can be found at \url{https://github.com/sdi1100041/SLEIPNIR} to facilitate future research and reproducibility.
\section{Background}
\label{sec:Background}
In this section, we provide a high-level overview of the background for our work. For an in-depth introduction, see \citet{RasmussenBook} (GPs), \citet{rahimi2008random} (RFF) and \citet{mutny2018efficient} (QFF).

\subsection{Feature Expansions for normal GPR}
Unfortunately, any reasoning in such a model requires the calculation and inversion of the covariance matrices $\bm{C_\phi}$ and $\bm{A}$. Without any tricks or approximations, this will scale as $\mathcal{O}(N^3)$. Random Fourier features (RFF) introduced by \citet{rahimi2008random} and quadrature Fourier features (QFF) introduced by \citet{mutny2018efficient} are two approximations that can be used to reduce this complexity for standard GP regression. Both approaches are based on the following observation: Any kernel represents a scalar product in a potentially infinite dimensional feature space, but if we allow for a small error, this could be approximated by a finite-dimensional feature vector.

For readability, we will introduce all concepts using a kernel with scalar inputs $k(t_i, t_j)$. However, it should be noted that this is by no means a necessary condition, and the concept generalizes nicely to higher dimensional inputs.

For a stationary, scalar kernel, Bochner's theorem \citep[see e.g.][]{rudin1976principles} guarantees the existence of a density $p(\omega)$ such that we can write
\begin{equation}
\label{eq:Bochner}
k(|t_i - t_j|) = \int_{-\infty}^{\infty}p(\omega)\left(\begin{matrix} \cos(\omega t_i) \\ \sin(\omega t_i) \end{matrix} \right)^T \left(\begin{matrix} \cos(\omega t_j) \\ \sin(\omega t_j) \end{matrix} \right) d \omega.
\end{equation}
This equation can be interpreted as a scalar product of infinitely many features given by $\cos(\omega t_i)$ and $\sin(\omega t_i)$. But how do we find the most important features such that the kernel $k$ can be reasonably well approximated by a finite feature vector
\begin{equation}
k(|t_i - t_j|) \approx \bm{\phi}(t_i)^T \bm{\phi}(t_j)?
\end{equation}

To obtain random Fourier features, \citet{rahimi2008random} propose two different Monte Carlo sampling schemes. The first approximation just obtains MC samples from $p(\omega)$ and for each sample adds $[\sin(\omega t_i) \cos(\omega t_i)]$ to the feature vector $\bm{\phi}(t_i)$. In the second scheme, they observe that
\begin{equation}
\left(\begin{matrix} \cos(\omega t_i) \\ \sin(\omega t_i) \end{matrix} \right)^T \left(\begin{matrix} \cos(\omega t_j)\\ \sin(\omega t_j) \end{matrix} \right)
=
\mathbb{E}_{b} \{\sqrt{2} \cos(\omega t_i + b) \},
\end{equation}
where $b \sim \text{Unif}([0, 2\pi])$. Thus, one can obtain a second approximation by sampling concurrently $b$ and $\omega$ and for each sample adding $\cos(\omega t_i + b)$ to the feature vector $\bm{\phi}(t_i)$. Accentuating the presence of the bias term b, we will refer to this approximation in the following as RFF-B, while we refer to the first one as RFF.

One main drawback of RFFs however is the fact that the intermediate sampling step makes it impossible to obtain any deterministic bounds on their approximation quality.
\citet{mutny2018efficient} thus propose to approximate the integral of Equation \eqref{eq:Bochner} using Hermitian quadrature. While RFF and RFF-B use sampling to determine the locations of the $\omega_i$, in the context of Hermitian quadrature, these locations are fully determined by the functional form of the kernel. Thus, quadrature Fourier features (QFF) are deterministic in nature, leading to a deterministic, theoretical analysis. Furthermore, as a numerical integration scheme, they are especially efficient in low dimensions. As we shall see both in our theoretical and empirical analysis, similar properties can be achieved if we want to include derivative observations.

\subsection{Gaussian Process Regression with Derivatives}
\label{sec:GPRD_Intro}
Assume there exists a scalar-valued function $x(t)$. At $N$ distinct time points $\bm{t}=[t_0, \dots, t_{N-1}]$, we obtain noisy observations of the function itself and its derivatives, assuming Gaussian noise with standard deviations $\sigma$ and $\sqrt{\gamma}$. Following \citet{wenk2019odin}, these observations are represented as the vectors $\bm{y}$ and $\bm{F}$. Using GP regression, we aim to find estimates for $\mathbf{x}=[x(t_0), \dots, x(t_{N-1})]$ given $\bm{y}$ and $\bm{F}$. In GP regression, we assume that $x(t)$ is drawn from a Gaussian process with a kernel $k(t_i, t_j)$ that is parameterized by hyper-parameters $\bm{\phi}$. For fixed $\bm{\phi}$, this leads to tractable Gaussian priors over the function $\bm{x}$ and its derivatives $\bm{\dot{x}}$. These priors can be combined with the observation models to obtain the generative model summarized in Figure \ref{fig:GPDerivGenModel}.

\begin{figure*}[h]
	\centering
	\scalebox{0.9}{
		\begin{minipage}{0.5\textwidth}
			\begin{alignat*}{2}
			&p(\bm{x} \mid \bm{\phi}) &&= \mathcal{N}(\bm{x} \mid \mathbf{0}, \bm{C}_{\bm{\phi}})\\
			&p(\bm{\dot{x}} \mid \bm{x}, \bm{\phi}) &&= \mathcal{N}(\bm{\dot{x}} \mid \bm{D} \bm{x}, \bm{A})\\
			&p(\bm{y} \mid \bm{x}, \sigma) &&= \mathcal{N}(\bm{y} | \bm{x}, \sigma^2 \bm{I}) \\
			&p(\bm{F} \mid \bm{\dot{x}}, \bm{\gamma}) &&= \mathcal{N}(\bm{F} \mid \bm{\dot{x}}, \gamma \bm{I}) 
			\end{alignat*}
		\end{minipage}\hfill
		\begin{minipage}{0.5\textwidth}
			\centering
			\begin{tikzpicture}[
			node distance=0.5 cm and 0.5 cm,
			mynode/.style={draw,circle,minimum size=0.9cm}  % needs to be that big for F_1 to fit
			]
			\node[mynode](sigma) {$\sigma$};
			\node[mynode, below=of sigma] (y){$\mathbf{y}$};
			\node[mynode, right=of y] (x){$\mathbf{x}$};
			\node[mynode, right=of x] (xDot){$\mathbf{\dot{x}}$};
			\node[mynode, right=of xDot] (F){$\mathbf{F}$};
			\node[mynode, above=of F] (gamma){$\gamma$};
			\node[mynode, at = ($(sigma)!0.5!(gamma)$)] (phi){$\boldsymbol{\phi}$};
			\path
			(phi) edge[-latex] (x)
			(x) edge[-latex] (y)
			(sigma) edge[-latex] (y)
			(phi) edge[-latex] (xDot)
			(x) edge[-latex] (xDot)
			(xDot) edge[-latex] (F)
			(gamma) edge[-latex] (F)
			;
			\end{tikzpicture}
	\end{minipage}}
	\vspace{-0.1cm}
	\caption{Generative model for Gaussian process regression with derivative observations.}
	\vspace{-0.4cm}
	\label{fig:GPDerivGenModel}
\end{figure*}

Here, the matrices $\bm{C_\phi}$, $\bm{D}$ and $\bm{A}$ are fully determined by the kernel $k$ and the choice of $\bm{t}$. Since all relevant probability densities are Gaussian, the posteriors can be calculated analytically, as they will be Gaussian as well. In the rest of this paper, we will focus on the predictive posterior mean and variance for both state and variance, i.e. 
\begin{equation}
p(x(\tau) | \bm{y}, \bm{F}, \bm{\phi}, \gamma) = \mathcal{N}(x(\tau) | \mu(\tau), \Sigma(\tau))
\label{eq:GPRD_post_state}
\end{equation}
and
\begin{equation}
p(x'(\tau) | \bm{y}, \bm{F}, \bm{\phi}, \gamma) = \mathcal{N}(x'(\tau) | \mu'(\tau), \Sigma'(\tau)).
\label{eq:GPRD_post_deriv}
\end{equation}
$p(\bm{x(\tau)} | \bm{y}, \bm{F}, \bm{\phi}, \gamma)$
These quantities are of particular importance for any task involving smoothing or interpolation. As we shall demonstrate, our scheme allows for exponentially fast decaying error bounds for these quantities as well. For more details regarding notation, please refer to Section \ref{subsec:AppendixNotation}.

\subsection{Parameter Inference for ODEs with GPs}
One important application of this regression technique is parameter inference of differential equations. To simplify notation, we briefly recap the theory for one dimensional systems of ODEs, noting that there is no significant difference to the multidimensional case.

Assume that we are given the noisy observations $\bm{y}$ of a dynamical system with known parametric form $\dot{x} = f(x, \bm{\theta})$. The goal is to infer the unknown parameters $\bm{\theta}$. In $\operatorname{ODIN}$, this equation is now used as a constraint for the probabilistic model in Figure \ref{fig:GPDerivGenModel}. Parameter and states are then found by minimizing the following objective, where $\bm{F}$ has already been substituted by the constraints:

\begin{align}
\mathcal{R}(\bm{x}, &\bm{F}, \bm{y}) = \bm{x}^{T}\bm{C}_{\bm{\phi}}^{-1}\bm{x} \label{eq:riskPrior}\\
&+ (\bm{x} - \bm{y})^T \sigma^{-2}(\bm{x} - \bm{y})\label{eq:riskObs} \\
&+ (\bm{F} - \bm{D}\bm{x})^{T} (\bm{A} + \gamma \bm{I})^{-1}(\bm{F} - \bm{D}\bm{x}). \label{eq:riskDerivs}
\end{align}

To infer the hyper-parameters from data, \citet{wenk2019odin} propose a preprocessing step, in which for each dimension, $\operatorname{ODIN}$ maximizes the marginal likelihood by solving
\begin{align}
&\text{max } &p(\bm{y} | \bm{\phi}, \sigma) = \mathcal{N}(\bm{y} \mid \bm{0}, \bm{C_\phi} + \sigma^2 \bm{I}) \label{eq:marginalObs}\\
&\text{w.r.t. } &\bm{\phi}, \sigma 
\end{align}

The model mismatch parameter $\gamma$ can either be hand-tuned (as done e.g. by \citet{wenk2018fast} or \citet{gorbach2017scalable}) or inferred at run time using $\operatorname{ODIN}$ as well. In this case, the risk term of Equations \eqref{eq:riskPrior} - \eqref{eq:riskDerivs} is extended by the additional summand $\log\det(\bm{A} + \gamma \bm{I})$, representing the contribution of the determinant term of the conditional $p(\bm{F} | \bm{x}, \bm{y}, \bm{\phi}, \gamma)$. In this work, we will refer to this case whenever we speak about learning $\gamma$. Clearly, $\operatorname{ODIN}$ scales similarly to standard GP regression with derivatives cubically in the amount of observations due to the inversions of $\bm{C_\phi}$ and $\bm{A}$. Again, all quantities that are not defined in detail can be looked up in Section \ref{subsec:AppendixNotation}.

\section{QFF for Derivative Information}
\label{sec:QFF}
For standard GP regression, the main computational challenge lies in calculating any terms involving the covariance matrix $\bm{C_\phi}$. In the context of Hermitian quadrature, the work of \citet{mutny2018efficient} effectively reduces this complexity from $\mathcal{O}(N^3)$ to $\mathcal{O}(N M^2 + M^3)$, where $M << N$ is a fixed constant representing the amount of features used in the feature expansion. % In the context of $\operatorname{ODIN}$, this method can be directly applied to reduce the run time complexity of the terms given by Equations \eqref{eq:riskDerivs} and \eqref{eq:marginalObs}.

However, as soon as we introduce derivative information, we obtain an additional term depending on the matrices $\bm{D}$ and $\bm{A}$. Similar to $\bm{C_\phi}$, the size of these matrices grows linearly with the amount of observations $N$, again leading to a computational complexity of $\mathcal{O}(N^3)$. Since the calculation of these matrices involves the first two derivatives of the kernel, we need to extend the standard QFF framework to include such derivatives.

Since the functional form of the kernel dictates the QFF features, any analysis would need to be repeated for every kernel. We will thus restrict ourselves to the RBF kernel, even though it could in principle be extended to any stationary kernel. The RBF kernel is defined as $k(t_i, t_j) = k(t_i - t_j) = k(r_{ij}) = \rho \exp(-\frac{r_{ij}^2}{2l^2})$, where $r_{ij} \coloneqq t_i - t_j$. Here, the variance $\rho$ and lengthscale $l$ are the hyper-parameters of the kernel, which need to be learned in a preprocessing step. The RBF kernel is known for its excellent smoothing properties. Since scaling up the amount of observations is mainly relevant for densely observed trajectories, this makes it an ideal choice for the experiments we will show in Section \ref{sec:ExperimentsNew}.

\subsection{Approximate Derivative of the Kernel}
Since both the integral and the derivative operator are linear, we can use their commutativity to obtain new quadrature schemes for the kernel derivatives. For ease of readability, let us define $I(f(\omega)) :=\int_{-\infty}^{+\infty} e^{-\omega^2} f(\omega)  d\omega $ and fix $\rho=\sqrt{\pi}$. This assumption is simply to avoid cluttered notation and will could be lifted by introducing a constant factor instead. For ease of notation, let $r \coloneqq t_i - t_j$.

Applying Bochner's theorem \eqref{eq:Bochner} to the RBF kernel with $\rho=\sqrt{\pi}$, we get, similarly to \citet{mutny2018efficient},
\begin{equation}
\label{eq:RBF}
k(t_i,t_j) = I(\cos(\omega r \frac{\sqrt 2}{l})).
%k(x-y) = k(\delta)= \sqrt{\pi} e^{-\frac{\delta^2}{2l^2}}= \int_{-\infty}^{+\infty} e^{-\omega^2}\cos(\omega \delta\frac{\sqrt 2}{l})d\omega =
\end{equation}
Using the linearity of the integral operator, we can differentiate the above equality to obtain
\begin{equation}
\frac{ \partial }{t_i} k(t_i,t_j) 
= \frac{d  }{d r} I(\cos(\omega r\frac{\sqrt 2}{l}))= I(- \frac{\sqrt 2}{l} \omega \sin(\omega r \frac{\sqrt 2}{l})). \label{eq:integrals1}
\end{equation}

Similarly, 
\begin{align}
\frac{ \partial }{\partial t_j} k(t_i,t_j) &= I(\omega \frac{\sqrt 2}{l} \sin(\omega r\frac{\sqrt 2}{l}))
\end{align}
and
\begin{align}
\frac{ \partial^2 }{\partial t_i \partial t_j} k(t_i,t_j) &= I(\omega^2 \frac{2}{l^2} \cos(\omega r\frac{\sqrt 2}{l})). \label{eq:integrals2}
\end{align}

These calculations reduce the problem of approximating the kernel derivatives to approximating integrals. Thus, similar to the derivative free case, we can now leverage the powerful framework of Gauss-Hermite quadrature \citep{quadratBook}.

Let 
\begin{equation}
\label{eq:Quadrature Scheme}
Q_m(f(\omega)) = \sum_{i=1}^{m}W_i^mf(\omega_i^m)
\end{equation}
denote the Gauss-Hermite quadrature scheme of order $m$ for the function $f$, where $W_i^m$ are its weights and $\omega_i^m$ its abscissas. Construct the $2m$ dimensional feature vector $\bm{\phi}(x)$ by adding for each $i$ the two components
\begin{equation}
\label{eq:Feature vector}
\left[ \sqrt{W_i^m} \cos( \omega_i^m \frac{\sqrt 2}{l} t_i ) \quad \sqrt{W_i^m}\sin(\omega_i^m \frac{\sqrt 2}{l} t_i) \right].
\end{equation}
Given this feature vector, we first observe that 
\begin{equation}
\bm{\phi}(t_i)^T\bm{\phi}(t_j)  = \sum_{i=1}^{m} W_i^m \cos( \omega_i^m \frac{\sqrt 2}{l} r )
= Q_m(\cos(\omega \frac{\sqrt 2}{l} r)),
\end{equation}
as desired when recovering the QFF approximation for $k(t_i, t_j)$ for calculating $\bm{C_\phi}$.

However, this feature expansion can also be differentiated w.r.t. $t_i$ to obtain
\begin{equation}
\label{eq:Deriv Feature vector}
\bm{\phi}(t_i)^{\prime} = \left[\begin{matrix} -\sqrt{W_i^m} \frac{\sqrt 2}{l} \omega_i^m  \sin( \omega_i^m \frac{\sqrt 2}{l} t_i ) \\ \hspace{8pt}\sqrt{W_i^m} \frac{\sqrt 2}{l} \omega_i^m \cos(\omega_i^m \frac{\sqrt 2}{l} t_i) \end{matrix}\right] _{i=1 \dots m}.
\end{equation}
 
Using trigonometric identities, it can be shown that this feature expansion indeed yields
\begin{align} \textstyle
\bm{\phi}(t_i)^{\prime T}\bm{\phi}(t_j) &= Q_m(-\frac{\sqrt 2}{l} \omega \sin(\omega r\frac{\sqrt 2}{l})),\\
\bm{\phi}(t_i)^T\bm{\phi}(t_j)^{\prime} &= Q_m(\frac{\sqrt 2}{l} \omega \sin(\omega r\frac{\sqrt 2}{l})), \\
\bm{\phi}(t_i)^{\prime T}\bm{\phi}(t_j)^{\prime} &= Q_m(\omega^2 \frac{2}{l^2} \cos(\omega r\frac{\sqrt 2}{l})).
\end{align}

Thus, the features given by \eqref{eq:Deriv Feature vector} represent the quadrature schemes for the integrals shown in Equations \eqref{eq:integrals1} and \eqref{eq:integrals2}. Thus, the kernel derivatives involved in calculating $\bm{A}$ and $\bm{D}$ can be approximated by
\begin{equation}
\frac{ \partial }{\partial x} k(x,y) \approx \bm{\phi}(x)^{\prime T}\bm{\phi}(y)
\end{equation}
and
\begin{equation}
\frac{ \partial^2 }{\partial x \partial y} k(x,y) \approx \bm{\phi}(x)^{\prime T}\bm{\phi}(y)^{\prime}.
\end{equation}

As we shall prove and empirically validate in the rest of this paper, this feature expansion is efficient in the sense that the approximation error decays \emph{exponentially} in the amount of features. This allows us to obtain accurate approximations of the kernel derivatives even for a small amount of \emph{deterministically} chosen features.

\subsection{Application to GPR with Derivatives}
To apply this approximation to GPR with derivatives, we make use of the matrix inversion lemma.
\begin{lemma}[Matrix Inversion Lemma]
	\label{lem:MatInvLemma}
	Let A and C be invertible matrices. Then for any matrices U, V of appropriate dimensions, it holds that $(A + U C V)^{-1} = A^{-1} - A^{-1} U (C^{-1} + V A^{-1} U)^{-1} V A^{-1}$.
\end{lemma}

In the context of standard GP regression \citep[e.g.][]{rahimi2008random, mutny2018efficient}, this can be used as follows. Let $\bm{\Phi} \in \mathbb{R}^{2m \times N}$ be such that its columns are given by the feature vectors $\bm{\bm{\phi}}(t_i)$ as defined in \eqref{eq:Feature vector} for a quadrature scheme of order m. After adding a small jitter $\lambda$ on the diagonal of $\bm{C}_{\bm{\phi}}$, we can use Lemma \ref{lem:MatInvLemma} to obtain
\begin{equation}
(\bm{C_\phi} + \lambda \bm{I})^{-1} \approx \frac{1}{\lambda}(\bm{{I }} - 
\bm{\Phi}^T(\bm{\Phi} \bm{\Phi}^T + \lambda \bm{\mathbb{I }})^{-1} \bm{\Phi}) \label{eq:standardApprox}
\end{equation}
Here, $\approx$ should be read as an approximation up to an exponentially decaying error in $M$, which we will quantify in Section \ref{sec:TheoryNew}.

Since all matrices required to calculate the quantities in Equations \eqref{eq:GPRD_post_state} and \eqref{eq:GPRD_post_deriv} allow for a direct feature expansion, the matrix inversion lemma can be directly applied to approximate these quantities. Since we work directly with the posterior mean and variance, we do not even need a jitter term due to the presence of the noise terms.

\subsection{SLEIPNIR}
\label{subsec:SLEIPNIR}
However, extending this concept to Equation \eqref{eq:riskDerivs} is slightly more tricky, as neither $\bm{D}$ nor $\bm{A}$ allow for a direct feature expansion. Nevertheless, it is still possible to derive a scalable approximation using the previously introduced feature expansions:

First, start by collecting the appropriate feature vectors from Equations \eqref{eq:Feature vector} and \eqref{eq:Deriv Feature vector} to write $\bm{C}^{\prime}_{\bm{\bm{\phi}}} \approx \bm{\Phi}^{\prime T}\bm{\Phi}$, ${}^\prime \bm{C}_{\bm{\bm{\phi}}} \approx \bm{\Phi}^T\bm{\Phi}^{\prime}$ and $\bm{C}^{\prime \prime}_{\bm{\bm{\phi}}} \approx \bm{\Phi}^{\prime T}\bm{\Phi}^{\prime}$. These approximations can then be inserted into the term given by Equation \eqref{eq:riskDerivs}. By applying Theorem \ref{lem:MatInvLemma} multiple times, it is possible to finally obtain
\begin{align}
\bm{z}^{T}(\bm{A} &+ \gamma \bm{I})^{-1}\bm{z} \approx \\
&\frac{1}{\gamma} \bm{z}^{T}( \bm{\mathbb{I }} - 
\bm{\Phi}^{\prime T}(\bm{\Phi}^{\prime} \bm{\Phi}^{\prime T} + \frac{\gamma}{\lambda} \bm{\Phi} \bm{\Phi}^T + \gamma \bm{\mathbb{I }})^{-1} \bm{\Phi}^{\prime}  )\bm{z}, \nonumber
\end{align}
where
\begin{equation*}
\bm{z} \coloneqq \bm{f}(\bm{\theta}, \bm{x}) - \bm{D}\bm{x} 
\approx \bm{f}(\bm{\theta}, \bm{x}) - \bm{\Phi}^{\prime T} (\bm{\Phi} \bm{\Phi}^T + \lambda \bm{\mathbb{I }})^{-1} \bm{\Phi} \bm{x}.
\end{equation*}

Combining these approximations with the one given by Equation \eqref{eq:standardApprox}, we obtain a computationally efficient scheme for calculating all objective functions in $\operatorname{ODIN}$. Furthermore, both the additional term $\log\det(\bm{A} + \gamma \bm{I})$ obtained when learning $\gamma$ and learning the hyperparameters via Equation \eqref{eq:marginalObs} can be scaled similarly using these approximations and directly applying the matrix inversion lemma.

Overall, this means that the original complexity of $\mathcal{O}(N^3)$ has been successfully reduced to $\mathcal{O}(N M^2 + M^3)$. For $M < N$, this is significantly accelerating $\operatorname{ODIN}$, which is why we name our approximation scheme $\operatorname{SLEIPNIR}$. The resulting algorithm will be referred to as $\operatorname{ODIN-S}$.

\section{Theoretical Results}
\label{sec:TheoryNew}
It is clear that this bound is especially appealing, since the error will exhibit exponential decay for $m > l^2$. In this section, we state that this behavior can be generalized to our approximations of the kernel derivatives. As we have demonstrated in the previous section, our feature approximation can be used to efficiently reduce the computational complexity of GPRD and $\operatorname{ODIN}$. In this section, we will provide an in-depth error analysis of this approximation, demonstrating the key theoretical result of this paper: The favorable properties of standard QFF carry over to the derivative case and can be efficiently deployed in practically relevant algorithms.

\subsection{Exponentially Decaying Kernel Approximation Errors}
Let us define 
\begin{equation}
\label{eq:Error term}
E_m:= \sqrt{\pi}\frac{1}{m^m}(\frac{e}{4l^2})^m
\end{equation}
In the context of standard GP regression, \citet{mutny2018efficient} have shown that
\begin{equation}
|k(t_i,t_j) - \bm{\phi}(t_i)^T\bm{\phi}(t_j) | \leq E_m
\end{equation} 

The results are summarized in Theorem \ref{thm:error bounds}, with a detailed proof given in the appendix in Section \ref{sec:AppendixKernelApproxProof}.

\begin{theorem} \label{thm:error bounds}
	Let $k(t_i,t_j)$ be defined as in Equation \eqref{eq:RBF} and consider the Gauss-Hermite quadrature scheme of order $m$ as described by Equation \eqref{eq:Quadrature Scheme}, defining $\bm{\phi}(t_i) $   and $\bm{\phi}(t_i)^{\prime}$ as in Equations \eqref{eq:Feature vector} and \eqref{eq:Deriv Feature vector}.
	Then, for $|r| = |t_i-t_j| \leq 1$, it holds that
	\begin{enumerate}[(i)]
		\item $|\frac{ \partial }{\partial t_i} k(t_i,t_j) - \bm{\phi}(t_i)^{\prime T}\bm{\phi}(t_j)|  \leq \frac{2e}{l^2} E_{m-2} $
		\item $| \frac{ \partial^2 }{\partial t_i \partial t_j} k(t_i,t_j) - \bm{\phi}(t_i)^{\prime T}\bm{\phi}(t_j)^{\prime} |   \leq \frac{2e}{l^4} E_{m-3}$
	\end{enumerate}
	where $E_m$ is defined as in Equation \eqref{eq:Error term}
\end{theorem}
Not that in the above Theorem, $|r| < 1$ can always be achieved by rescaling the data and adapting the lengthscale $l$ accordingly.

While the error is slightly larger for the derivative approximations than for the kernel itself, it is important to note that we are still getting the same exponential decaying behavior.

\subsection{GPR with Derivatives Bounds}
\label{sec:GPRD_Theory}

As we shall see in this section, this exponential decay carries over nicely to the case of GPR with derivatives. Define $e_{\tilde{\mu}}$, $e_{\tilde{\Sigma}}$, $e_{\tilde{\mu}'}$ and $e_{\tilde{\Sigma}'}$ as the absolute error between the feature approximations and the corresponding accurate quantities of the means and covariances of Equations \eqref{eq:GPRD_post_state} and \eqref{eq:GPRD_post_deriv}. For each $\tau \in \mathbb{R}$, define $e_{\text{tot}}$ as the maximum of these four errors. Using these definitions, we can show the exponential relation between feature approximation order $m$ and the corresponding approximation error, as summarized in Theorem \ref{thm:GPRD}, with proof in the appendix.

\begin{theorem}
	Let us consider an RBF kernel with hyperparameters $(\rho, l)$ and domain $[0,1]$. Define $c \coloneqq \min(\gamma, \sigma^2)$ and $R \coloneqq \max(||\bm{y}||_\infty, ||\bm{F}||_\infty)$. Let $C > 0$. Let us consider a QFF approximation scheme of order $m \geq 3+\max \left(\frac{e}{2l^2}, \log\left( \frac{270n^2\rho^3R}{l^8 c^2 C} \right) \right)$. 
	Then, it holds for all $\tau \in [0, 1]$ that $e_{\text{tot}} \leq C$.
	\label{thm:GPRD}
\end{theorem}

From this theorem, we can also observe the following fact: If we decrease the acceptable worst case performance $C$, we clearly need more features. However, due to the logarithm in the theorem, an \emph{exponential} decay in $C$ only leads to \emph{linear} growth in $m$, all other things being equal.

\subsection{SLEIPNIR Bounds}

Similarly, we can observe that when applying $\operatorname{SLEIPNIR}$ to $\operatorname{ODIN}$ for a fixed feature vector length $M$, we will always create a small error in the objective function. However, the deterministic nature of the QFF approximation still allows us to choose the amount of features in a way such that the approximation error will always be guaranteed to be smaller than a pre-chosen threshold. This result is summarized in Theorem \ref{theo:Consistency} and proven in the appendix, Section \ref{sec:AppendixRiskApproxProof}.

\begin{theorem}
	\label{theo:Consistency}
	Let $\mathcal{R}$ be the $\operatorname{ODIN}$-objective as defined in Equations \eqref{eq:riskPrior}-\eqref{eq:riskDerivs} and let $\tilde{\mathcal{R}}$ denote its counterpart obtained by approximating the matrices $\bm{C_{\phi}}$, $\bm{A}$ and $\bm{D}$ as described in Section \ref{subsec:SLEIPNIR}.
	Assume the parameters $\lambda$, $\gamma$, $\bm{\phi}=(\rho,l)$ and $N$ to be fixed. Suppose $N\geq 60$ and let $1 >\epsilon > 0$. Let $m$ denote the order of the quadrature scheme. Then
	\begin{equation}
	m\geq 10 + \max \lbrace \frac{e}{2l^2},\log_2(\frac{\rho^2 n^3}{\lambda ^ 2 \gamma l^4 \epsilon}) \rbrace
	\end{equation}
	implies
	\begin{equation}
	\frac{|R_{\lambda \gamma \phi}(x,\theta)-\tilde{R}_{\lambda \gamma \phi}(x,\theta)|}{R_{\lambda \gamma \phi}(x,\theta)} \leq \epsilon
	\end{equation}
	for any configuration of the variables $\bm{x}$ and $\bm{\theta}$.
\end{theorem}

Again, it should be noted that the threshold $\epsilon$ appears inside the logarithm. Thus, an \emph{exponential} decrease of the allowed error only requires a \emph{linear} increase in the amount of features.

In this theorem, the bound on $N$ is purely aesthetically motivated, to reduce the amount of terms. From a practical perspective however, one would probably want to use $\operatorname{ODIN}$ without any approximations anyways if only $<60$ observations are available. Furthermore, the bound on $\epsilon$ just requires that investigating an approximation scheme with more than 100\% relative error is of little interest.
\section{Empirical Validation}
\label{sec:ExperimentsNew}
In this section, we will provide empirical validation for all theoretical claims made in this paper. We will start by showing the exponentially decaying behavior of the kernel approximation error in Section \ref{sec:ExpKernelError}. In Sections \ref{sec:ExpGPRD} and \ref{sec:ExpODINSStandard}, we show that this exponentially decaying behavior carries over to both GPR with derivatives and $\operatorname{ODIN}$. Ultimately, we will conclude by showing how $\operatorname{ODIN}$ with $\operatorname{SLEIPNIR}$ can be used on a realistic dataset, demonstrating its pratical applicability in Section \ref{subsec:Quadro}.

Wherever applicable, we will use three standard benchmark systems from the GP-based gradient matching literature, namely the Lotka Volterra (LV) system \citep{lotka1932growth}, the Protein Transduction (PT) system \citep{vyshemirsky2007bayesian} and the Lorenz system \citep{lorenz1963deterministic}. While the LV system is quite easy to fit, both PT and Lorenz offer interesting challenges. The non-stationary dynamics of PT make it a formidable challenge for collocation methods. Classic literature \citep[e.g.][]{dondelinger2013ode, gorbach2017scalable, wenk2019odin} sidestep this issue by using a non-stationary sigmoid kernel. However, as we shall demonstrate in this section, this is not a problem in the case of densely observed trajectories, since we will be using the RBF kernel in all experiments. Finally, the Lorenz system is interesting due to its chaotic behavior. Chaotic systems are an interesting challenge for many parameter inference schemes due to the potentially high sensitivity to parameter changes and the presence of many local optima. For a full description of the experimental setup and all metrics used please refer to Section \ref{sec:AppendixExperiments}. For all experiments, we created our data-set using numerical integration of the ground truth parameters. We then added 25 different noise realizations to obtain 25 different data sets. This allows us to quantify robustness w.r.t. noise by showing median as well as 20\% and 80\% quantiles over these noise realizations for each experiment. In all experiments, we trained $\gamma$ and learned the kernel hyperparameters from the data using the scalable approximations described in the previous section. To compare with $\operatorname{ODIN-S}$, we chose to combine $\operatorname{ODIN}$ with the RFF and RFF-B feature expansions detailed in Section \ref{sec:Background}.

\subsection{Exponentially Decaying Kernel Approximation Errors}
\label{sec:ExpKernelError}

\begin{figure*}[!h]
	\centering
	\begin{subfigure}[t]{0.3\textwidth}
		\centering
		\includegraphics[width=\textwidth]{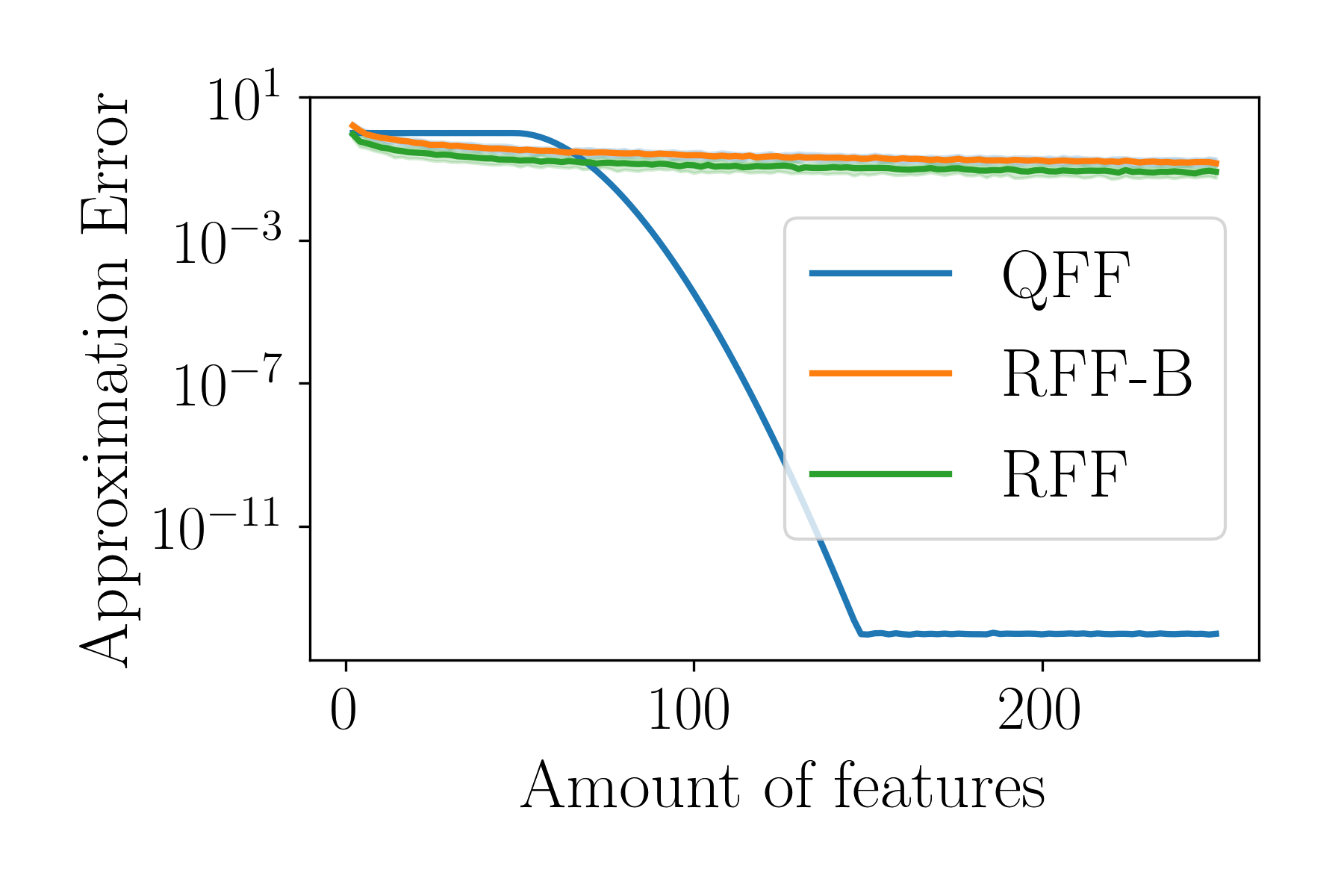}
		\vspace{-0.7cm}
		\caption{$k(r)$}
	\end{subfigure}
	\hfill
	\begin{subfigure}[t]{0.3\textwidth}
		\centering
		\includegraphics[width=\textwidth]{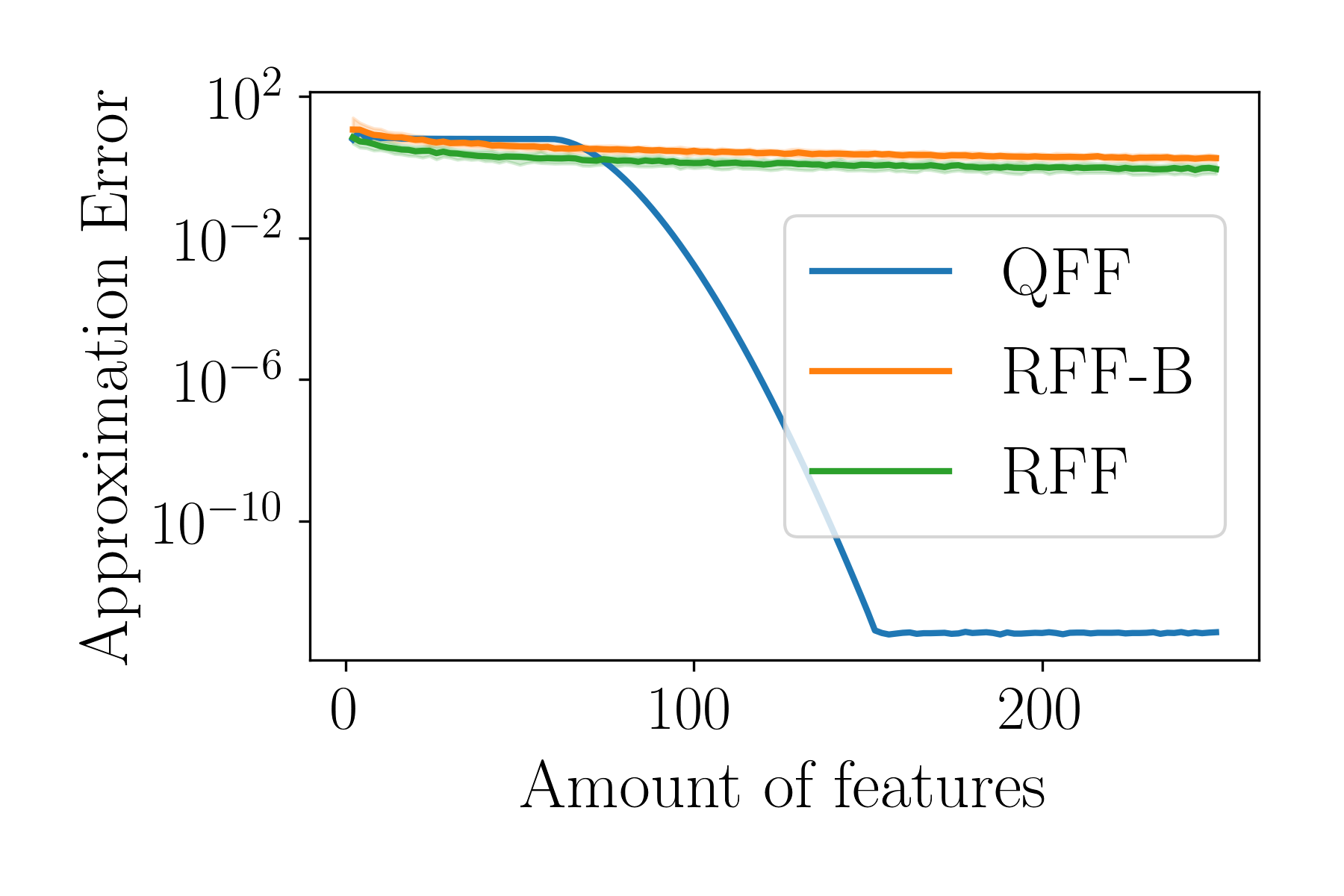}
		\vspace{-0.7cm}
		\caption{$k^{\prime}(r)$}
	\end{subfigure}
	\hfill
	\begin{subfigure}[t]{0.3\textwidth}
		\centering
		\includegraphics[width=\textwidth]{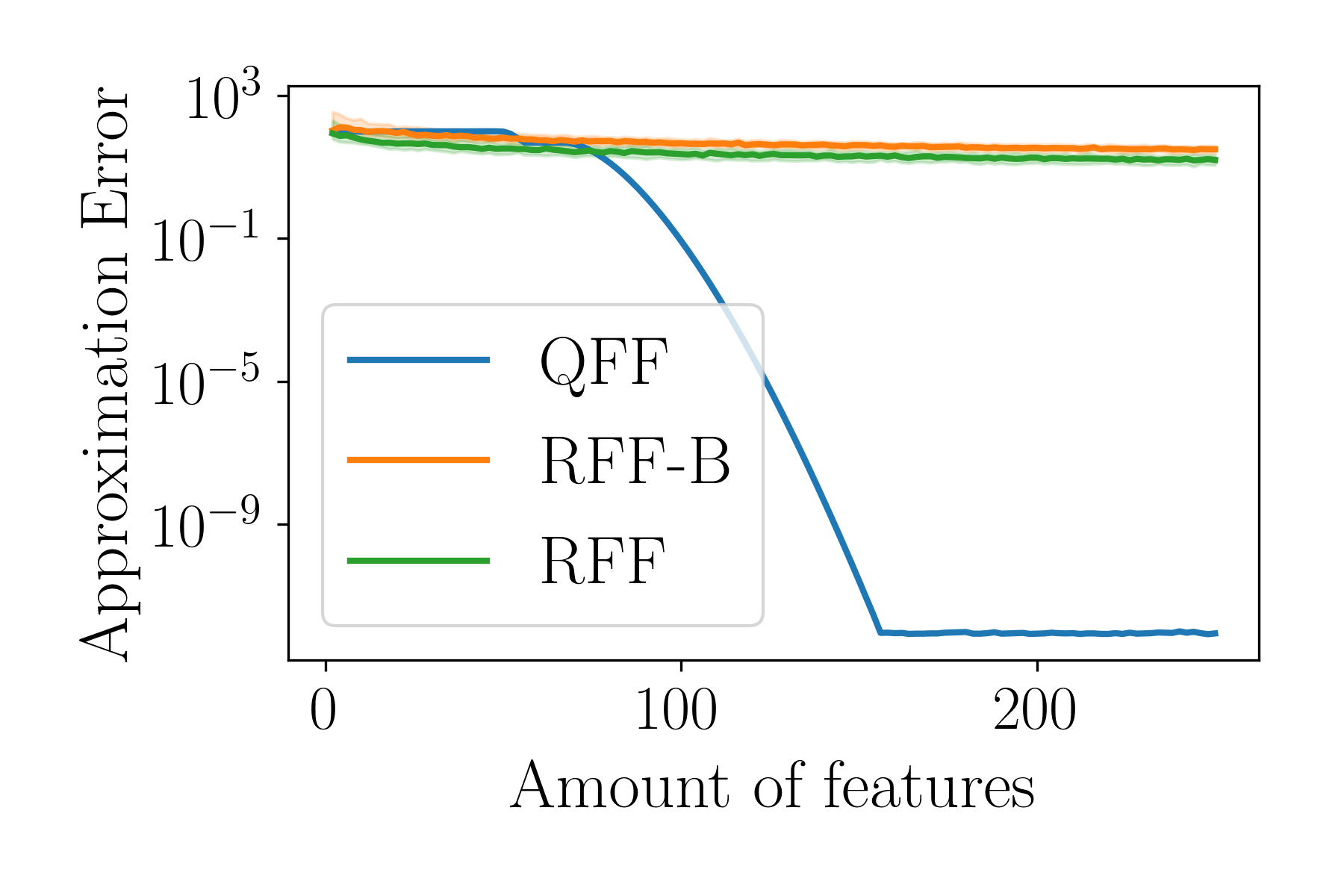}
		\vspace{-0.7cm}
		\caption{$k^{\prime \prime}(r)$}
	\end{subfigure}
	\vspace{-0.3cm}
	\caption{Comparing the maximum error of different feature expansions over $r \in [0, 1]$. For the random feature expansions, we show median as well as 12.5\% and 87.5\% quantiles over 100 random samples. Due to the exponential decay of the error of the QFF approximation, this stochasticity is barely visible. As given by the theoretical analysis, the error is a bit higher for the derivatives, but still decaying exponentially.}
	\label{fig:QFF_Approximation}
\end{figure*}

In Figure \ref{fig:QFF_Approximation}, we evaluate the maximum error of the feature approximations over an interval $r \in [0, 1]$. From Theorem \ref{thm:error bounds}, we would expect an exponential decay in the approximation error for both the kernel and the first and second order derivatives. This exponential decay is clearly visible and continues until we hit numerical boundaries. In Figure \ref{fig:QFF_Approximation}, the $l$ was chosen to be 0.1. However, as can be seen on the additional plots in the appendix (Sec. \ref{sec:AppendixKernelApproxPlots}), this behavior is robust across different lengthscales. Figure \ref{fig:QFF_Approximation} allows for the interesting observation that next to the exponential decay of QFF, the error of the RFF almost looks constant, even though it decays linearly.

\subsection{GPR with Derivatives}
\label{sec:ExpGPRD}

\begin{figure*}[!h]
	\centering
	\begin{subfigure}[t]{0.24\textwidth}
		\centering
		\includegraphics[width=\textwidth]{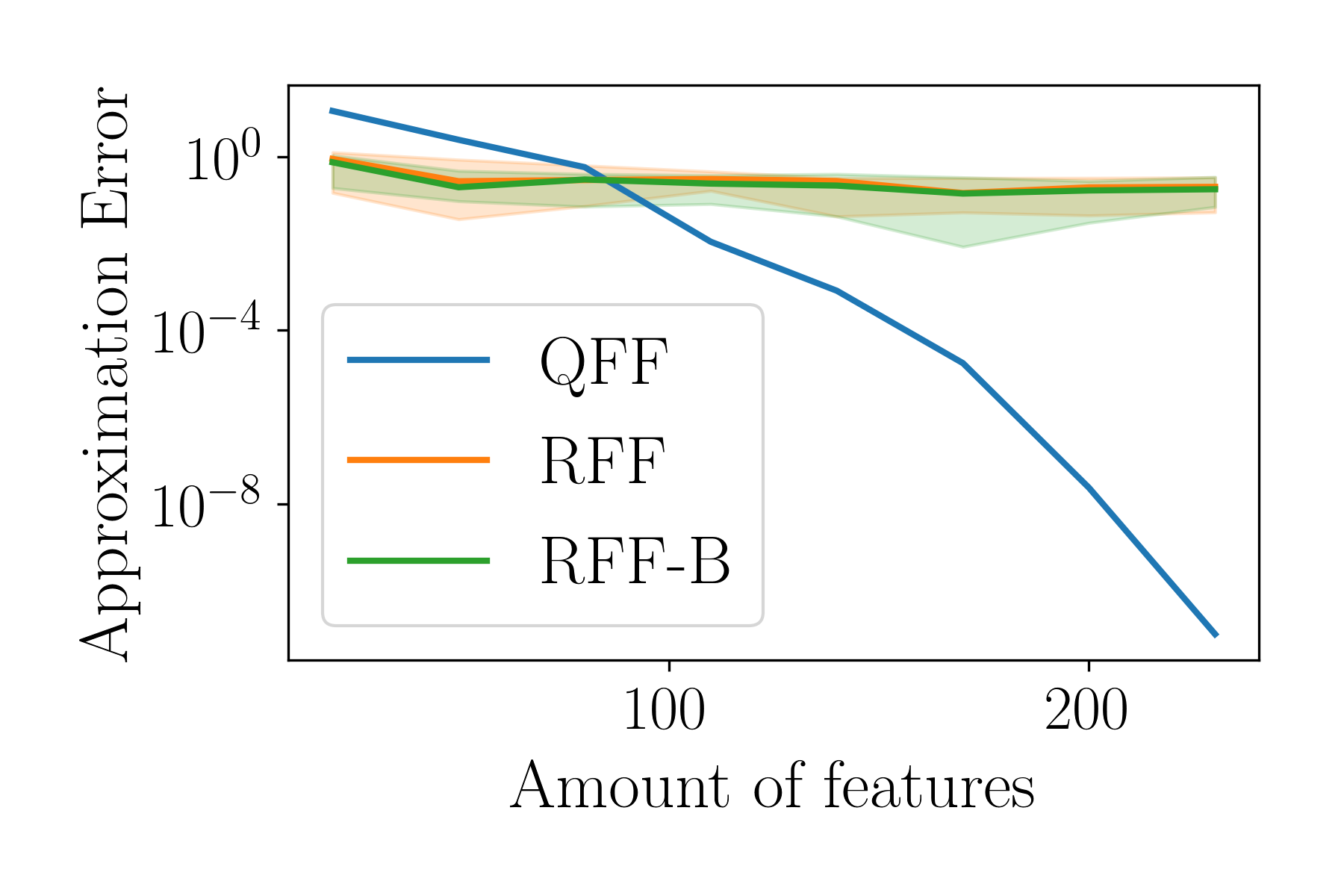}
		\vspace{-0.7cm}
		\caption{$\mu_0$}
	\end{subfigure}
	\hfill
	\begin{subfigure}[t]{0.24\textwidth}
		\centering
		\includegraphics[width=\textwidth]{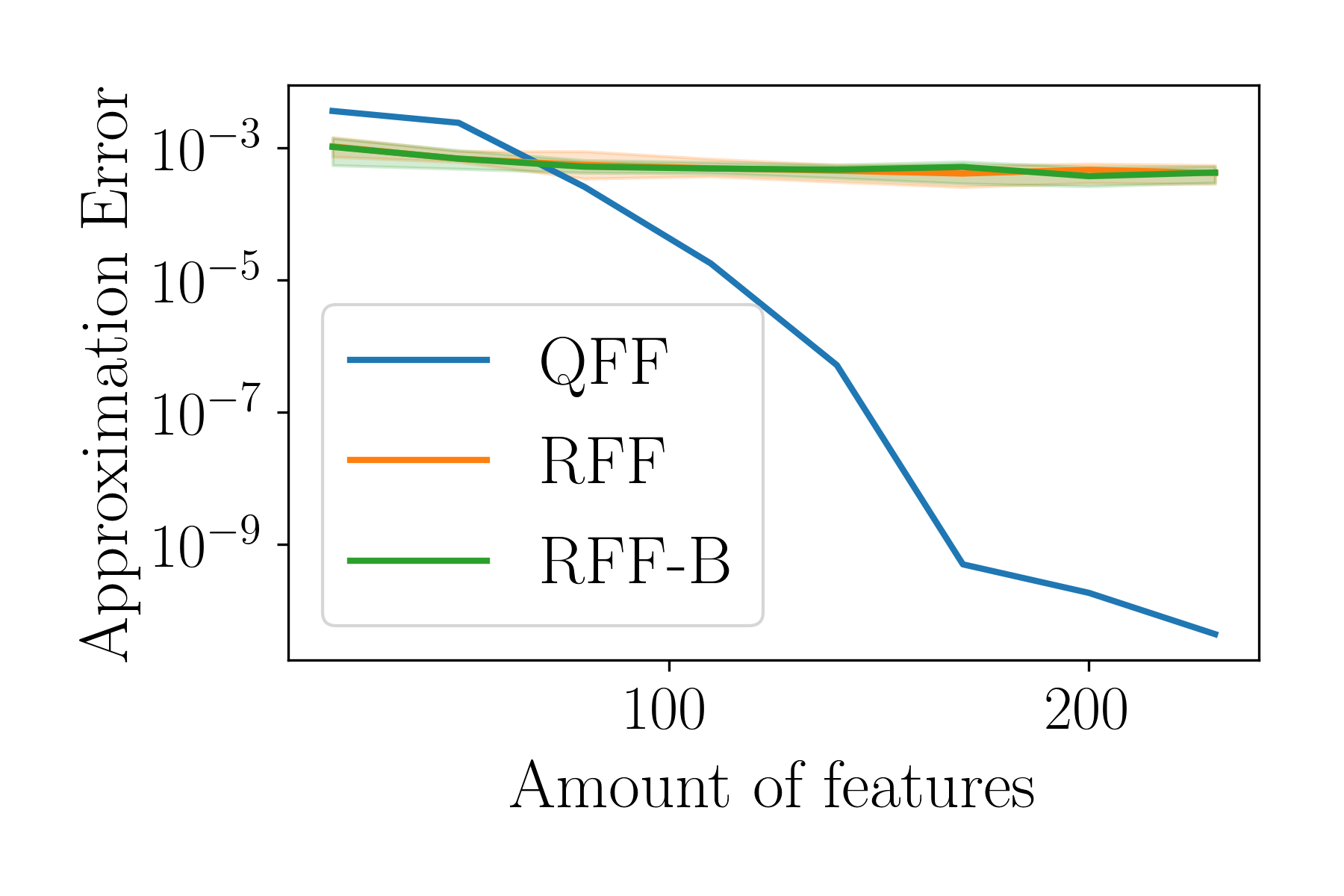}
		\vspace{-0.7cm}
		\caption{$\Sigma_0$}
	\end{subfigure}
	\hfill
	\begin{subfigure}[t]{0.24\textwidth}
		\centering
		\includegraphics[width=\textwidth]{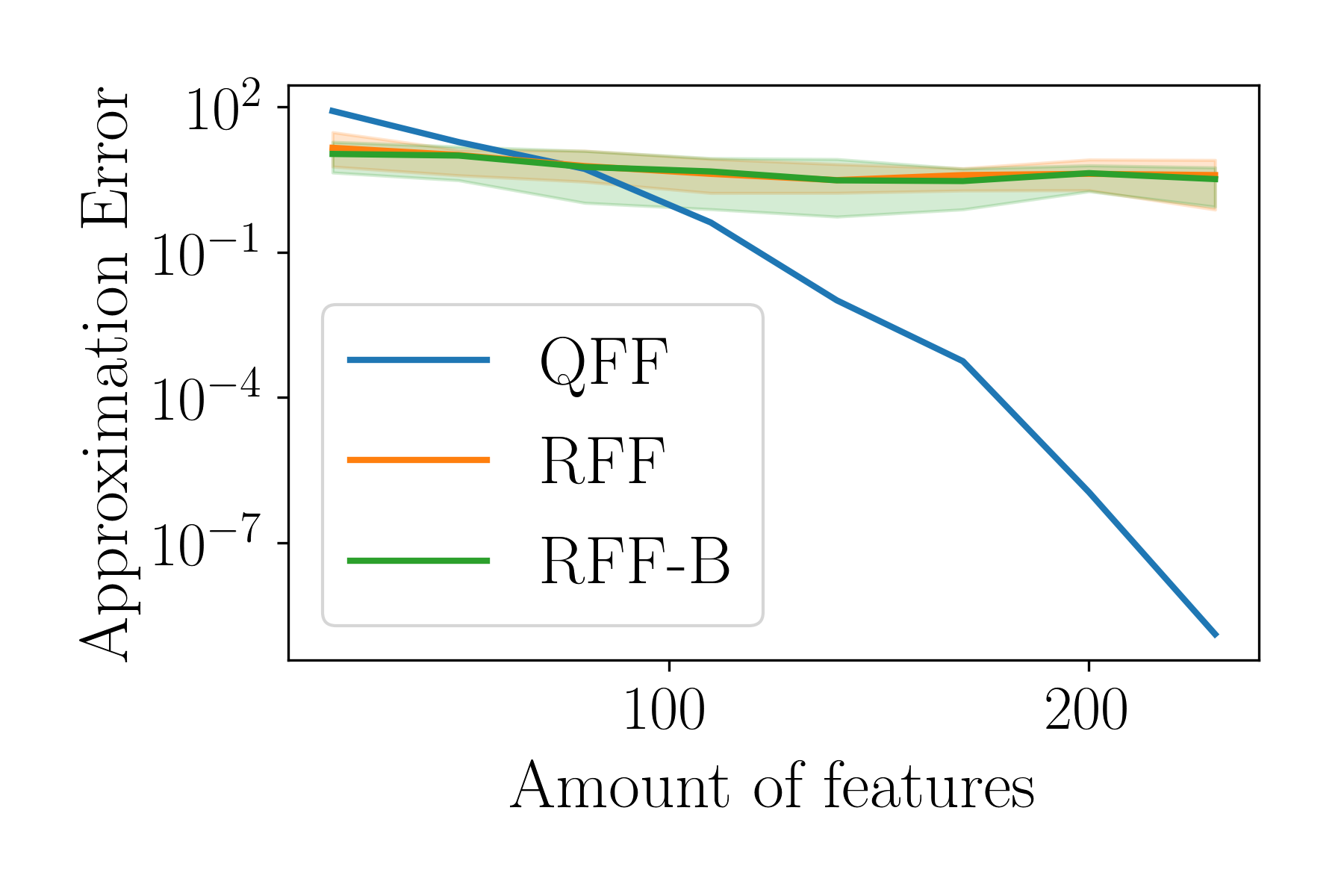}
		\vspace{-0.7cm}
		\caption{$\mu'_0$}
	\end{subfigure}
	\hfill
	\begin{subfigure}[t]{0.24\textwidth}
		\centering
		\includegraphics[width=\textwidth]{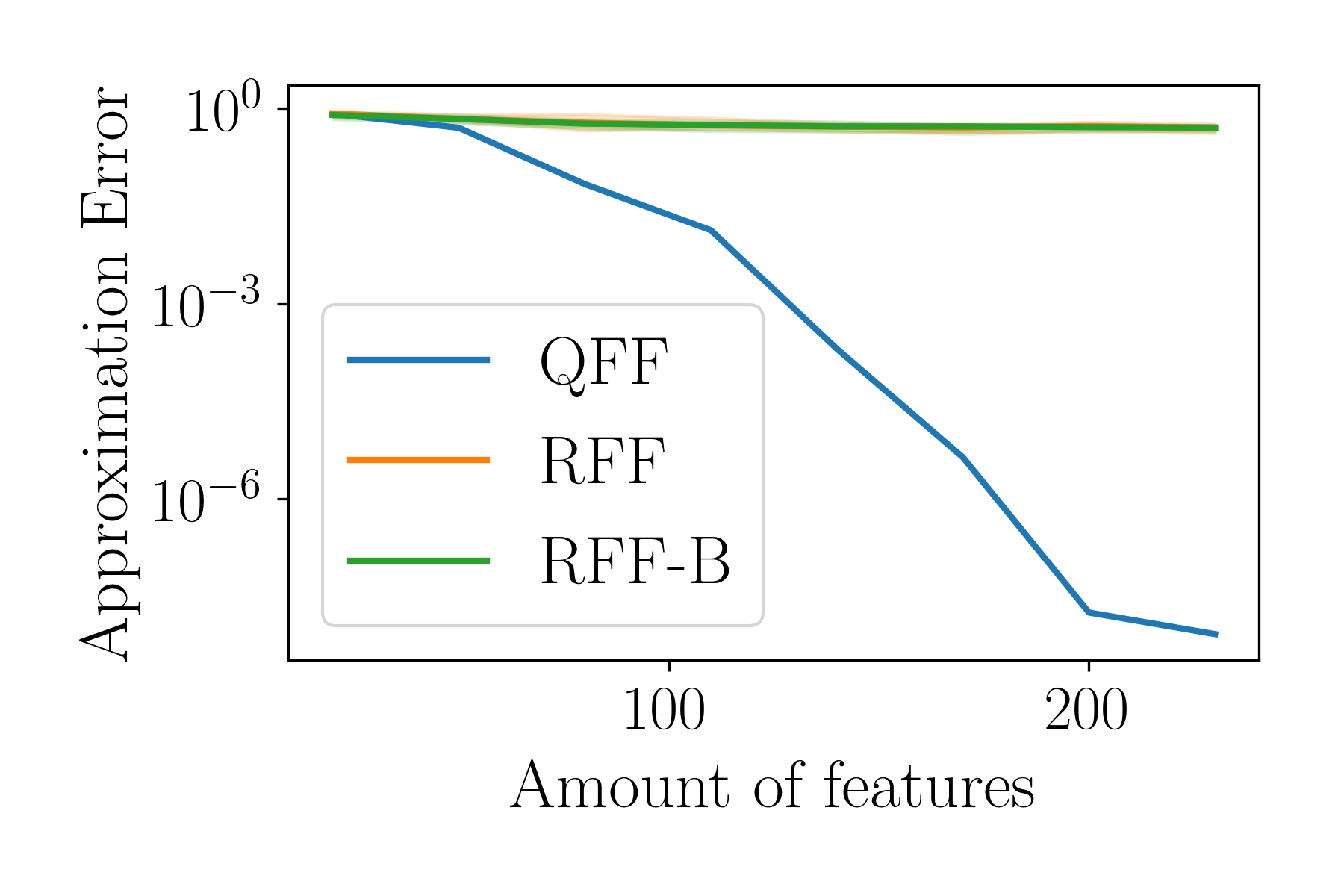}
		\vspace{-0.7cm}
		\caption{$\Sigma'_0$}
	\end{subfigure}
	\vspace{-0.3cm}
	\caption{Approximation error of the different feature approximations compared to the accurate GP, evaluated at $t=0.8$ for the Lorenz system with 1000 observations and an SNR of 100. For each feature, we show the median as well as the 12.5\% and 87.5\% quantiles over 10 independent noise realizations for the first state dimension.}
	\label{fig:GPRD_MAIN}
\end{figure*}

A similar behavior can be observed for the absolute error of the GP posterior we introduced in Section \ref{sec:GPRD_Intro}. As can be seen in Figure \ref{fig:GPRD_MAIN}, the exponential decay of the error predicted in Section \ref{sec:GPRD_Theory} also appears in practice. While we only show one state of one experiment, this behavior is consistent across experiments and noise settings, as can be seen by looking at the additional plots presented in the appendix, Section \ref{sec:AppendixGPRWithDerivEval}.

\subsection{ODIN-S on Standard Benchmark Systems}
\label{sec:ExpODINSStandard}

\begin{figure*}[h!]
	\centering
	\begin{subfigure}[t]{0.3\textwidth}
		\centering
		\includegraphics[width=\textwidth]{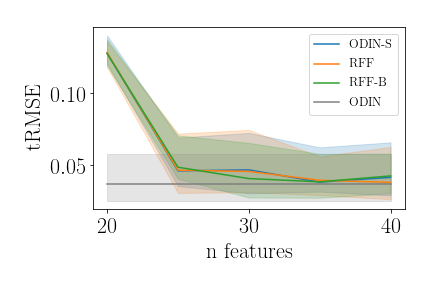}
		\vspace{-0.7cm}
		\caption{LV}
	\end{subfigure}
	\hfill
	\begin{subfigure}[t]{0.3\textwidth}
		\centering
		\includegraphics[width=\textwidth]{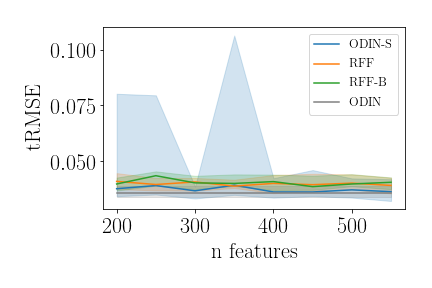}
		\vspace{-0.7cm}
		\caption{PT}
	\end{subfigure}
	\hfill
	\begin{subfigure}[t]{0.3\textwidth}
		\centering
		\includegraphics[width=\textwidth]{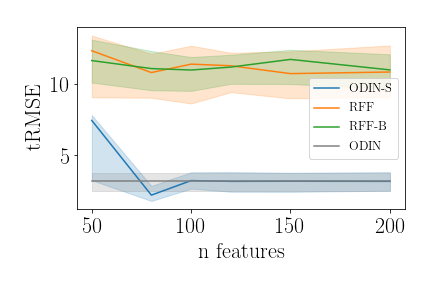}
		\vspace{-0.7cm}
		\caption{Lorenz}
	\end{subfigure}
	\\[-0.05cm]
	\begin{subfigure}[t]{0.3\textwidth}
		\centering
		\includegraphics[width=\textwidth]{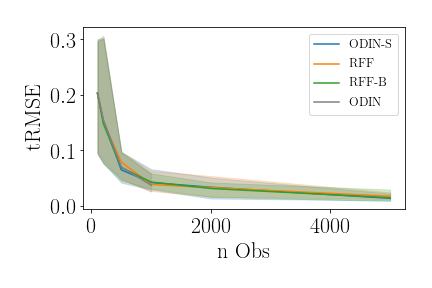}
	\end{subfigure}
	\hfill
	\begin{subfigure}[t]{0.3\textwidth}
		\centering
		\includegraphics[width=\textwidth]{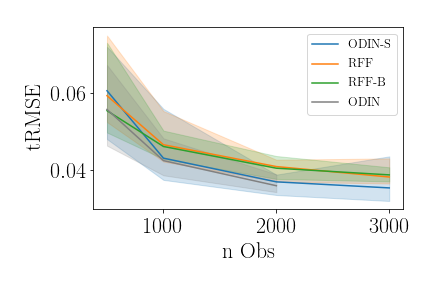}
	\end{subfigure}
	\hfill
	\begin{subfigure}[t]{0.3\textwidth}
		\centering
		\includegraphics[width=\textwidth]{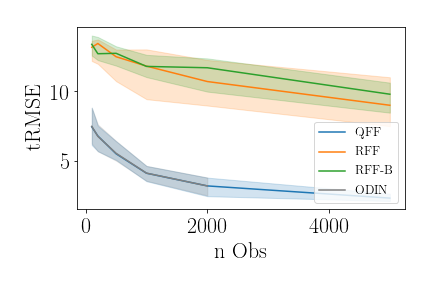}
	\end{subfigure}
	\vspace{-0.4cm}
	\caption{Comparing the tRMSE of the ODE parameters obtained via the three different approximation schemes. In the top row, we show how the results obtained by $\operatorname{ODIN-S}$ converge to the accurate results when increasing the amount of features. In the second row, we show the learning curves for a fixed feature vector length, demonstrating that increasing the amount of observations actually improves the parameter estimates and decreases the tRMSE.}
	\label{fig:standardBenchmarkResults}
\end{figure*}

In Figure \ref{fig:standardBenchmarkResults}, we compare the performance of $\operatorname{ODIN-S}$ against accurate $\operatorname{ODIN}$ as well as $\operatorname{ODIN}$ augmented with RFF and RFF-B on three standard benchmark systems. In the top row, we keep the total amount of observations fixed to 1000 for LV and 2000 for PT and Lorenz, while varying the length of the feature vector. In the bottom row, we keep the amount of features fixed to 40 for LV, 300 for PT and 150 for Lorenz. All the data was created using observation noise with $\sigma^2=0.1$ for LV, $\sigma^2=0.01$ for PT and a signal-to-noise ratio of 5 for Lorenz. Due to computational restrictions, it was not possible to evaluate accurate $\operatorname{ODIN}$ beyond what is shown in the plots. To provide an idea of the robustness of the evaluation, different noise, feature and observation settings are investigated in the appendix, Section \ref{sec:AppendixExpPlots}.

As expected by our theoretical analysis, the trajectory RMSE of the $\operatorname{SLEIPNIR}$-based $\operatorname{ODIN-S}$ eventually converges to the approximation-free $\operatorname{ODIN}$ in both median and quantiles if we increase the number of Fourier features. This clearly validates the main result of Theorem \ref{theo:Consistency}, expecting an exponentially small error.
Also, it is interesting to observe that the MCMC-based RFF and RFF-B seem to struggle to keep up with the approximation quality of $\operatorname{SLEIPNIR}$. While there is not much difference visible for the smooth and simple Lotka Volterra system, this is clearly visible for PT and especially striking for the chaotic Lorenz system. Complementing the results of Figure \ref{fig:QFF_Approximation} and Figure \ref{fig:GPRD_MAIN}, this further illustrates the power of the $\operatorname{SLEIPNIR}$ approximation, especially in this setting, where the kernel inputs are one-dimensional.

\begin{table}
	\captionof{table}{Comparing the run time of accurate $\operatorname{ODIN}$ against the run time of the feature approximations per iteration in miliseconds. For each setting, we show the median $\pm$ on standard deviation over 15 different iterations. As expected, the feature approximations lead to order of magnitude reductions in run time.}
	\label{tab:RunTime}
	\centering
	\begin{tabular}{lll}
		& accurate      & feature \\%\vspace{0.05cm} \\
		\textbf{LV}     & $973 \pm 42.5$  & $1.24 \pm 0.243$      \\
		\textbf{PT}     & $17900 \pm 668$ & $133 \pm 12.8$        \\
		\textbf{Lorenz} & $10700 \pm 466$ & $13.2 \pm 0.566$     
	\end{tabular}
\end{table}

In Table \ref{tab:RunTime}, we finally compare the run time per iteration of $\operatorname{ODIN}$ with $\operatorname{ODIN-S}$. Since the run time only depends on the amount of features and not on how the features were obtained, we omitted RFF and RFF-B. The run time was evaluated using the same amount of observation and amount of feature combinations as in Figure \ref{fig:standardBenchmarkResults}, namely $(2000, 150)$ for LV with $\sigma^2=0.1$, $(2000, 300)$ for PT with $\sigma^2=0.01$ and $(1000, 40)$ for Lorenz with an SNR of 5. While it should already be clear from theoretical analysis that $\operatorname{ODIN-S}$ scales linearly in the amount of observations $N$ and cubic in the length of the feature vector $M$, this was confirmed empirically as well. However, due to space restrictions, the plots have been moved to the appendix and are shown in Figure \ref{fig:AppendixTimeVsFeatures} and Figure \ref{fig:AppendixTimeVsObs}.

\subsection{Practical Applicability}
\label{subsec:Quadro}
\begin{figure*}[!h]
	\centering
	\begin{subfigure}[t]{0.32\textwidth}
		\centering
		\includegraphics[width=\textwidth]{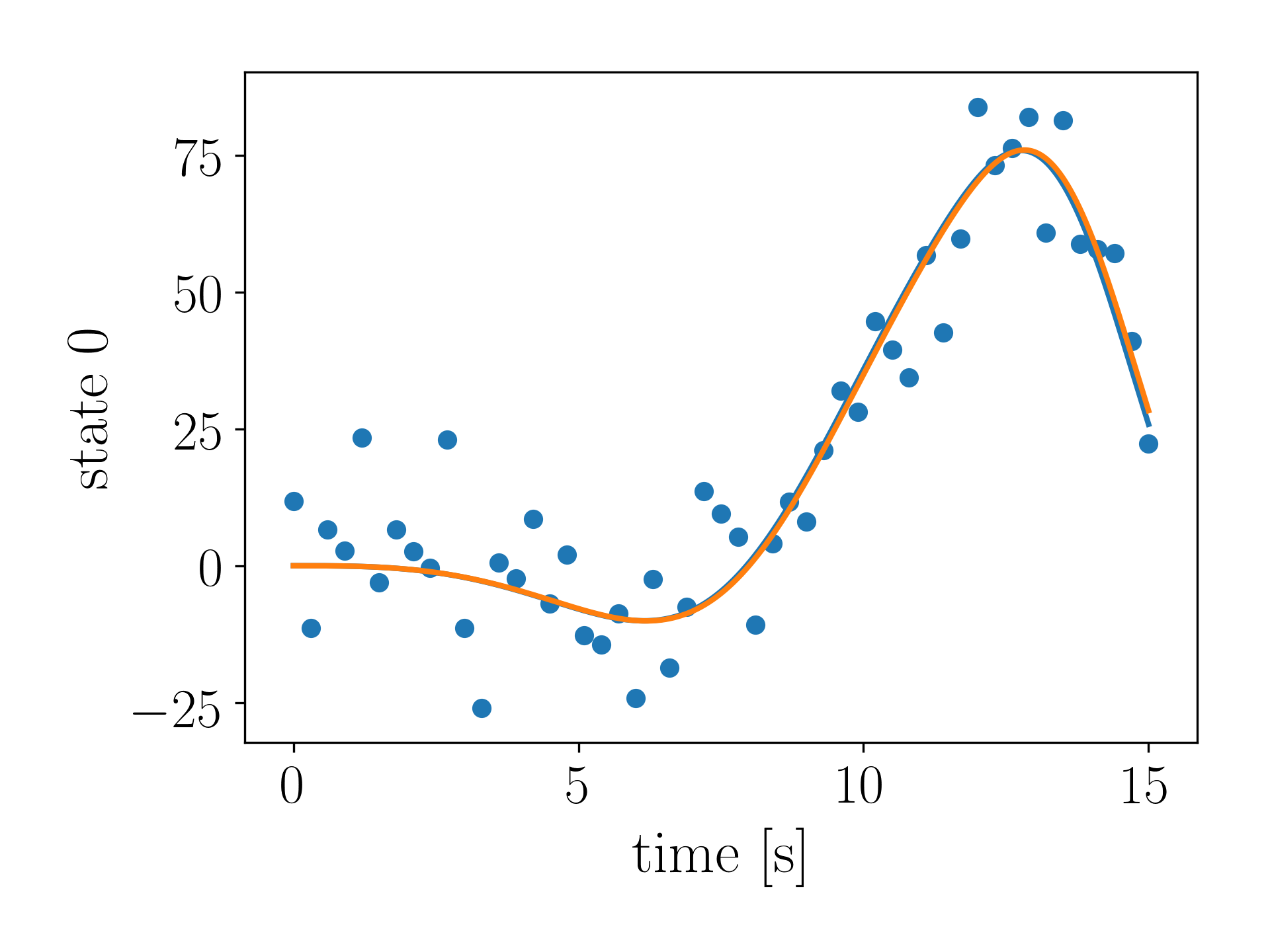}
	\end{subfigure}
	\hfill
	\begin{subfigure}[t]{0.32\textwidth}
		\centering
		\includegraphics[width=\textwidth]{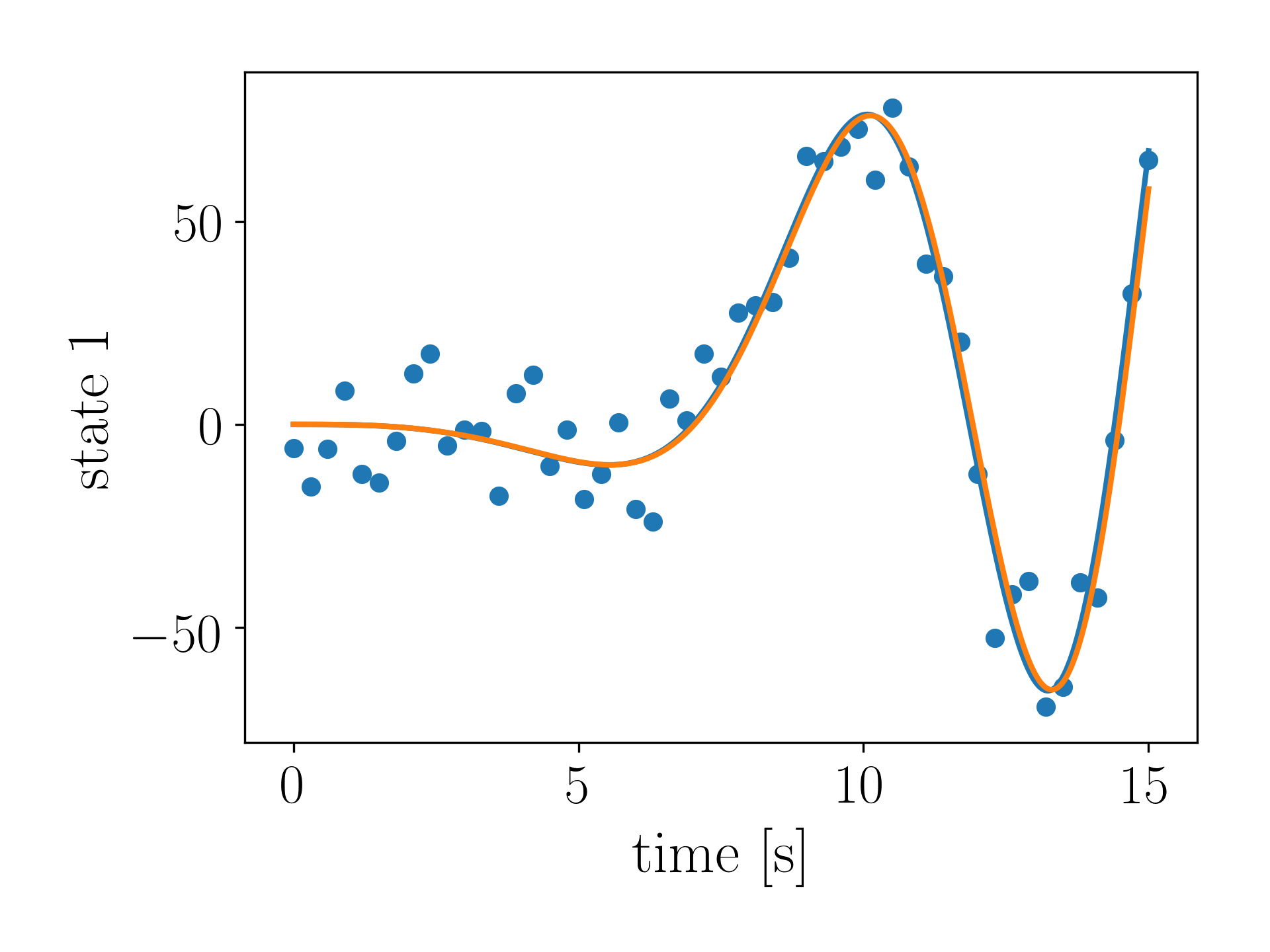}
	\end{subfigure}
	\hfill
	\begin{subfigure}[t]{0.32\textwidth}
		\centering
		\includegraphics[width=\textwidth]{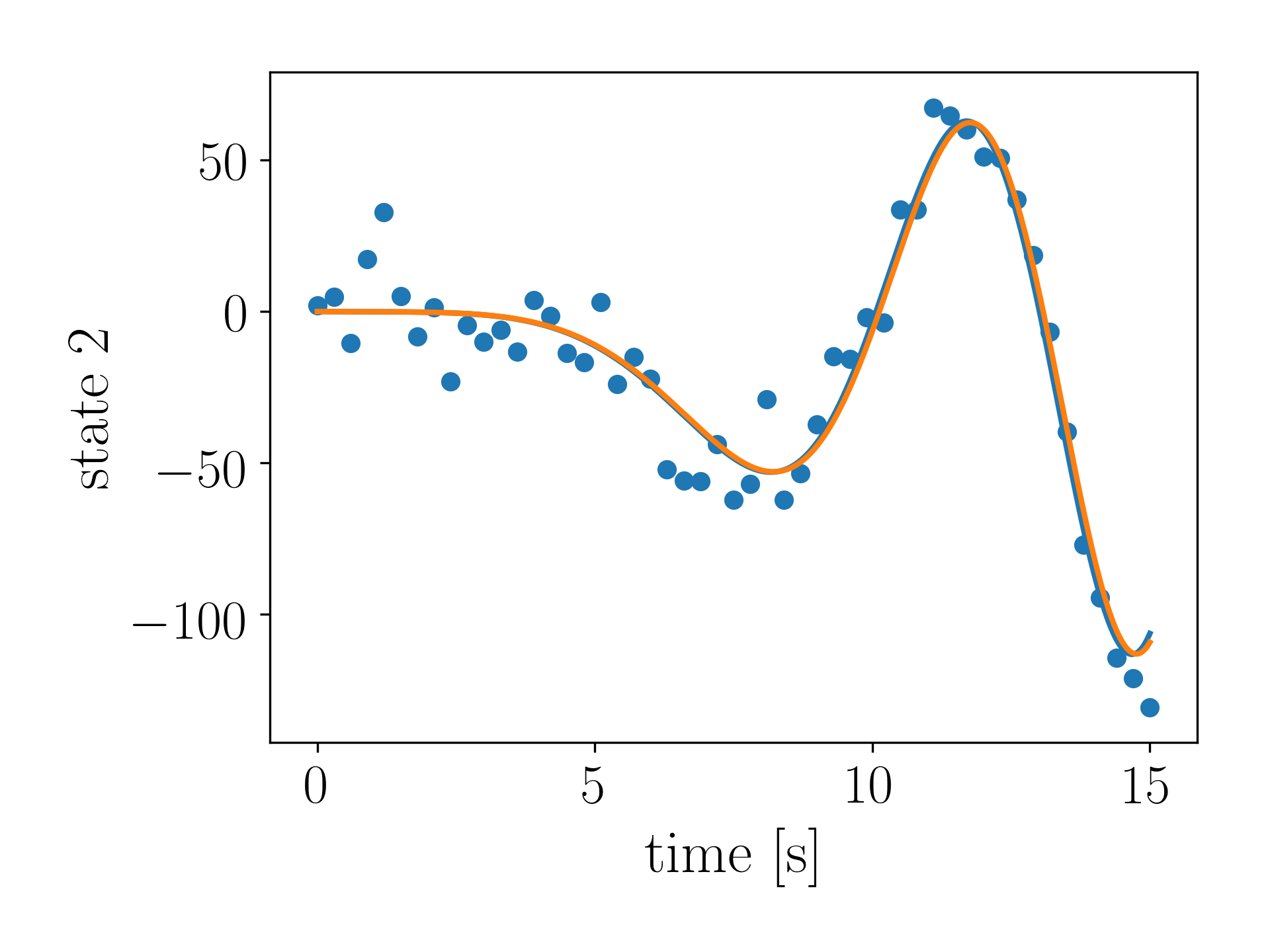}
	\end{subfigure}
	\vspace{-0.4cm}
	\caption{State trajectories of the first three states obtained by integrating the parameters inferred by $\operatorname{ODIN-S}$ (orange). The blue line represents the ground truth, while the blue dots show every 300-th observation for a signal-to-noise ratio of 10. States 5-12 have been moved to the appendix, Section \ref{subsec:QuadroStates}.}
	\vspace{-0.4cm}
	\label{fig:QuadroRMSE}
\end{figure*}

In a final experiment, we show that $\operatorname{ODIN-S}$ is able to scale to realistic data sets. For this, we introduce a 12-dimensional ODE system representing the dynamics of a 6DOF quadrocopter. We observe the system under Gaussian noise with SNR=10 over the time interval $t=[0, 15]$. We assume a sampling frequency of 1kHz, leading to 15'000 observations. We then run $\operatorname{ODIN-S}$ on a standard laptop (Lenovo Carbon X1) and obtain results in roughly 80min. Up to our knowledge, this is the first time that a system of such dimensions has been solved with a Gaussian process based parameter inference scheme, clearly demonstrating the power of our framework. The resulting trajectories are shown in Figure \ref{fig:QuadroRMSE}, including example observation points to visualize the noise level. The estimates of $\operatorname{ODIN-S}$ are so good that the ground truth is barely visible.

\section{Conclusion}
In this work, we introduced a new theoretical framework to scale up Gaussian process regression with derivative information, extending existing work based on quadrature Fourier features. We derived and proved deterministic, exponentially fast decaying error bounds for approximating the derivatives of an RBF kernel. We then combined these insights to create a computationally efficient approximation scheme for both standard GP regression with derivatives as well as the parameter inference scheme $\operatorname{ODIN}$. The theoretical analysis of this approximation yielded deterministic, non-asymptotic error bounds. In an extensive empirical evaluation, we then showed orders of magnitude improvements on the run time without sacrificing accuracy. In future work, we are excited to see how $\operatorname{SLEIPNIR}$ could be deployed in other areas such as scaling up Bayesian optimization with derivatives \citep[e.g.][]{wu2017bayesian} or probabilistic numerics \citep{hennig2015probabilistic}.

% Acknowledgements should go at the end, before appendices and references

\acks{This research was supported by the Max Planck ETH Center for Learning Systems. This projecthas received funding from the European Research Council(ERC) under the European Union’s Horizon 2020 researchand innovation programme grant agreement No 815943.}


\newpage

\clearpage
\bibliography{references}

\begin{thebibliography}{35}
\providecommand{\natexlab}[1]{#1}
\providecommand{\url}[1]{\texttt{#1}}
\expandafter\ifx\csname urlstyle\endcsname\relax
  \providecommand{\doi}[1]{doi: #1}\else
  \providecommand{\doi}{doi: \begingroup \urlstyle{rm}\Url}\fi

\bibitem[Abbati et~al.(2019)Abbati, Wenk, Osborne, Krause, Sch{\"o}lkopf, and
  Bauer]{abbati2019ares}
Gabriele Abbati, Philippe Wenk, Michael~A. Osborne, Andreas Krause, Bernhard
  Sch{\"o}lkopf, and Stefan Bauer.
\newblock {AR}e{S} and {M}a{RS} adversarial and {MMD}-minimizing regression for
  {SDE}s.
\newblock In Kamalika Chaudhuri and Ruslan Salakhutdinov, editors,
  \emph{Proceedings of the 36th International Conference on Machine Learning},
  volume~97 of \emph{Proceedings of Machine Learning Research}, pages 1--10,
  Long Beach, California, USA, 09--15 Jun 2019. PMLR.

\bibitem[Calderhead et~al.(2009)Calderhead, Girolami, and
  Lawrence]{calderhead2009accelerating}
Ben Calderhead, Mark Girolami, and Neil~D Lawrence.
\newblock Accelerating bayesian inference over nonlinear differential equations
  with gaussian processes.
\newblock In \emph{Advances in neural information processing systems}, pages
  217--224, 2009.

\bibitem[Cunningham et~al.(2008)Cunningham, Shenoy, and
  Sahani]{cunningham2008fast}
John~P Cunningham, Krishna~V Shenoy, and Maneesh Sahani.
\newblock Fast gaussian process methods for point process intensity estimation.
\newblock In \emph{Proceedings of the 25th international conference on Machine
  learning}, pages 192--199. ACM, 2008.

\bibitem[Dondelinger et~al.(2013)Dondelinger, Husmeier, Rogers, and
  Filippone]{dondelinger2013ode}
Frank Dondelinger, Dirk Husmeier, Simon Rogers, and Maurizio Filippone.
\newblock Ode parameter inference using adaptive gradient matching with
  gaussian processes.
\newblock In \emph{Artificial intelligence and statistics}, pages 216--228,
  2013.

\bibitem[Dony et~al.(2019)Dony, He, and Stumpf]{dony2019parametric}
Leander Dony, Fei He, and Michael~PH Stumpf.
\newblock Parametric and non-parametric gradient matching for network
  inference: a comparison.
\newblock \emph{BMC bioinformatics}, 20\penalty0 (1):\penalty0 52, 2019.

\bibitem[Eriksson et~al.(2018)Eriksson, Dong, Lee, Bindel, and
  Wilson]{eriksson2018scaling}
David Eriksson, Kun Dong, Eric Lee, David Bindel, and Andrew~G Wilson.
\newblock Scaling gaussian process regression with derivatives.
\newblock In \emph{Advances in Neural Information Processing Systems}, pages
  6867--6877, 2018.

\bibitem[Gorbach et~al.(2017)Gorbach, Bauer, and Buhmann]{gorbach2017scalable}
Nico~S Gorbach, Stefan Bauer, and Joachim~M Buhmann.
\newblock Scalable variational inference for dynamical systems.
\newblock In \emph{Advances in Neural Information Processing Systems}, pages
  4806--4815, 2017.

\bibitem[Hennig et~al.(2015)Hennig, Osborne, and
  Girolami]{hennig2015probabilistic}
Philipp Hennig, Michael~A Osborne, and Mark Girolami.
\newblock Probabilistic numerics and uncertainty in computations.
\newblock \emph{Proceedings of the Royal Society A: Mathematical, Physical and
  Engineering Sciences}, 471\penalty0 (2179):\penalty0 20150142, 2015.

\bibitem[Hensman et~al.(2017)Hensman, Durrande, Solin,
  et~al.]{hensman2017variational}
James Hensman, Nicolas Durrande, Arno Solin, et~al.
\newblock Variational fourier features for gaussian processes.
\newblock \emph{Journal of Machine Learning Research}, 18\penalty0
  (151):\penalty0 1--151, 2017.

\bibitem[Hildebrand(1987)]{quadratBook}
Francis~Begnaud Hildebrand.
\newblock \emph{Introduction to numerical analysis}.
\newblock Courier Corporation, 1987.

\bibitem[Lazaro-Gredilla et~al.(2010)Lazaro-Gredilla, Qui{\~n}onero-Candela,
  Rasmussen, and Figueiras-Vidal]{quia2010sparse}
Miguel Lazaro-Gredilla, Joaquin Qui{\~n}onero-Candela, Carl~Edward Rasmussen,
  and Anibal~R Figueiras-Vidal.
\newblock Sparse spectrum gaussian process regression.
\newblock \emph{Journal of Machine Learning Research}, 11\penalty0
  (Jun):\penalty0 1865--1881, 2010.

\bibitem[Lorenz(1963)]{lorenz1963deterministic}
Edward~N Lorenz.
\newblock Deterministic nonperiodic flow.
\newblock \emph{Journal of the atmospheric sciences}, 20\penalty0 (2):\penalty0
  130--141, 1963.

\bibitem[Lotka(1932)]{lotka1932growth}
Alfred~J Lotka.
\newblock The growth of mixed populations: two species competing for a common
  food supply.
\newblock \emph{Journal of the Washington Academy of Sciences}, 22\penalty0
  (16/17):\penalty0 461--469, 1932.

\bibitem[Macdonald and Husmeier(2015)]{macdonald2015gradient}
Benn Macdonald and Dirk Husmeier.
\newblock Gradient matching methods for computational inference in mechanistic
  models for systems biology: a review and comparative analysis.
\newblock \emph{Frontiers in bioengineering and biotechnology}, 3:\penalty0
  180, 2015.

\bibitem[Mockus(2012)]{mockus2012bayesian}
Jonas Mockus.
\newblock \emph{Bayesian approach to global optimization: theory and
  applications}, volume~37.
\newblock Springer Science \& Business Media, 2012.

\bibitem[Mutny and Krause(2018)]{mutny2018efficient}
Mojmir Mutny and Andreas Krause.
\newblock Efficient high dimensional bayesian optimization with additivity and
  quadrature fourier features.
\newblock In \emph{Advances in Neural Information Processing Systems}, pages
  9005--9016, 2018.

\bibitem[Qui{\~n}onero-Candela and Rasmussen(2005)]{quinonero2005unifying}
Joaquin Qui{\~n}onero-Candela and Carl~Edward Rasmussen.
\newblock A unifying view of sparse approximate gaussian process regression.
\newblock \emph{Journal of Machine Learning Research}, 6\penalty0
  (Dec):\penalty0 1939--1959, 2005.

\bibitem[Rahimi and Recht(2008)]{rahimi2008random}
Ali Rahimi and Benjamin Recht.
\newblock Random features for large-scale kernel machines.
\newblock In \emph{Advances in neural information processing systems}, pages
  1177--1184, 2008.

\bibitem[Rasmussen and Williams(2006)]{RasmussenBook}
CE. Rasmussen and CKI. Williams.
\newblock \emph{Gaussian Processes for Machine Learning}.
\newblock Adaptive Computation and Machine Learning. MIT Press, Cambridge, MA,
  USA, January 2006.

\bibitem[Rudin(1976)]{rudin1976principles}
W.~Rudin.
\newblock \emph{Principles of Mathematical Analysis}.
\newblock International series in pure and applied mathematics. McGraw-Hill,
  1976.
\newblock ISBN 9780070856134.

\bibitem[Sarkka et~al.(2013)Sarkka, Solin, and
  Hartikainen]{sarkka2013spatiotemporal}
Simo Sarkka, Arno Solin, and Jouni Hartikainen.
\newblock Spatiotemporal learning via infinite-dimensional bayesian filtering
  and smoothing: A look at gaussian process regression through kalman
  filtering.
\newblock \emph{IEEE Signal Processing Magazine}, 30\penalty0 (4):\penalty0
  51--61, 2013.

\bibitem[Seo et~al.(2000)Seo, Wallat, Graepel, and Obermayer]{seo2000gaussian}
Sambu Seo, Marko Wallat, Thore Graepel, and Klaus Obermayer.
\newblock Gaussian process regression: Active data selection and test point
  rejection.
\newblock In \emph{Mustererkennung 2000}, pages 27--34. Springer, 2000.

\bibitem[Snelson and Ghahramani(2006)]{snelson2006sparse}
Edward Snelson and Zoubin Ghahramani.
\newblock Sparse gaussian processes using pseudo-inputs.
\newblock In \emph{Advances in neural information processing systems}, pages
  1257--1264, 2006.

\bibitem[Solak et~al.(2003)Solak, Murray-Smith, Leithead, Leith, and
  Rasmussen]{solak2003derivative}
Ercan Solak, Roderick Murray-Smith, William~E Leithead, Douglas~J Leith, and
  Carl~E Rasmussen.
\newblock Derivative observations in gaussian process models of dynamic
  systems.
\newblock In \emph{Advances in neural information processing systems}, pages
  1057--1064, 2003.

\bibitem[Solin and S{\"a}rkk{\"a}()]{solin2014hilbert}
Arno Solin and Simo S{\"a}rkk{\"a}.
\newblock Hilbert space methods for reduced-rank gaussian process regression.
\newblock \emph{Statistics and Computing}, pages 1--28.

\bibitem[Solin et~al.(2018)Solin, Kok, Wahlstr{\"o}m, Sch{\"o}n, and
  S{\"a}rkk{\"a}]{solin2018modeling}
Arno Solin, Manon Kok, Niklas Wahlstr{\"o}m, Thomas~B Sch{\"o}n, and Simo
  S{\"a}rkk{\"a}.
\newblock Modeling and interpolation of the ambient magnetic field by gaussian
  processes.
\newblock \emph{IEEE Transactions on robotics}, 34\penalty0 (4):\penalty0
  1112--1127, 2018.

\bibitem[Szab{\'o} and Sriperumbudur(2018)]{szabo2018kernel}
Zolt{\'a}n Szab{\'o} and Bharath~K Sriperumbudur.
\newblock On kernel derivative approximation with random fourier features.
\newblock \emph{arXiv preprint arXiv:1810.05207}, 2018.

\bibitem[Titsias(2009)]{titsias2009variational}
Michalis Titsias.
\newblock Variational learning of inducing variables in sparse gaussian
  processes.
\newblock In \emph{Artificial Intelligence and Statistics}, pages 567--574,
  2009.

\bibitem[Vyshemirsky and Girolami(2007)]{vyshemirsky2007bayesian}
Vladislav Vyshemirsky and Mark~A Girolami.
\newblock Bayesian ranking of biochemical system models.
\newblock \emph{Bioinformatics}, 24\penalty0 (6):\penalty0 833--839, 2007.

\bibitem[Wenk et~al.(2018)Wenk, Gotovos, Bauer, Gorbach, Krause, and
  Buhmann]{wenk2018fast}
Philippe Wenk, Alkis Gotovos, Stefan Bauer, Nico Gorbach, Andreas Krause, and
  Joachim~M Buhmann.
\newblock Fast gaussian process based gradient matching for parameter
  identification in systems of nonlinear odes.
\newblock \emph{arXiv preprint arXiv:1804.04378}, 2018.

\bibitem[Wenk et~al.(2020)Wenk, Abbati, Bauer, Osborne, Krause, and
  Sch{\"o}lkopf]{wenk2019odin}
Philippe Wenk, Gabriele Abbati, Stefan Bauer, Michael~A Osborne, Andreas
  Krause, and Bernhard Sch{\"o}lkopf.
\newblock Odin: Ode-informed regression for parameter and state inference in
  time-continuous dynamical systems.
\newblock 2020.

\bibitem[Williams and Seeger(2001)]{williams2001using}
Christopher~KI Williams and Matthias Seeger.
\newblock Using the nystr{\"o}m method to speed up kernel machines.
\newblock In \emph{Advances in neural information processing systems}, pages
  682--688, 2001.

\bibitem[Wilson and Nickisch(2015)]{wilson2015kernel}
Andrew Wilson and Hannes Nickisch.
\newblock Kernel interpolation for scalable structured gaussian processes
  (kiss-gp).
\newblock In \emph{International Conference on Machine Learning}, pages
  1775--1784, 2015.

\bibitem[Wilson et~al.(2014)Wilson, Gilboa, Nehorai, and
  Cunningham]{wilson2014fast}
Andrew~G Wilson, Elad Gilboa, Arye Nehorai, and John~P Cunningham.
\newblock Fast kernel learning for multidimensional pattern extrapolation.
\newblock In \emph{Advances in Neural Information Processing Systems}, pages
  3626--3634, 2014.

\bibitem[Wu et~al.(2017)Wu, Poloczek, Wilson, and Frazier]{wu2017bayesian}
Jian Wu, Matthias Poloczek, Andrew~G Wilson, and Peter Frazier.
\newblock Bayesian optimization with gradients.
\newblock In \emph{Advances in Neural Information Processing Systems}, pages
  5267--5278, 2017.

\end{thebibliography}
\bibliographystyle{plainnat}

\newpage
\appendix
\onecolumn

\section{Kernel Approximation Error Bounds}
\label{sec:AppendixKernelApproxProof}
For any function $f$, let $I(f(\omega)) := \int_{-\infty}^{+\infty} e^{-\omega^2} f(\omega)d\omega $. \\
For any $r \in [0, 1]$, let $k(r) \coloneqq \sqrt{\pi} e^{-\frac{r^2}{2l^2}}= \int_{-\infty}^{+\infty} e^{-\omega^2}cos(\omega r\frac{\sqrt 2}{l})d\omega = I(cos(\omega r\frac{\sqrt 2}{l})) $. \\
Let $Q_m(f) = \sum_{i=1}^{m}W_i^mf(x_i^m)$ denote the Gauss-Hermite quadrature scheme of order $m$ and function $f$ with weights $W_i^m \geq 0$ and abscissas $x_i^m$ and let $S^m\coloneqq\lbrace x_1^m,x_2^m,...,x_m^m \rbrace$ denote the set of these abscissas.\\
From Gauss-Hermite quadrature, we know: If $f$ is a polynomial of degree $2m-1$ at most, then  $Q_m(f(\omega)) = I(f(\omega)) $, i.e the quadrature scheme exactly computes the integral.

Let $H_m(x)$ be the Hermite polynomial of order $m$ and $h_m(x) \coloneqq \frac{H_m(x)}{2^m}$ be its normalized version.\\
Since $I ( H_i(\omega)H_j(\omega)) = r_{ij} 2^j j! \sqrt{\pi} $, we know that $I ( h_m(\omega)^2 ) = \sqrt{\pi} \frac{m!}{2^m}$.

Let $E_m \coloneqq \sqrt{\pi}\frac{1}{m^m}(\frac{e}{4l^2})^m$.\\
Using this definition, we can restate the error bounds derived by \citet{mutny2018efficient} as
\begin{equation}
|I(\cos(\omega r\frac{\sqrt 2}{l})) - Q_m(  \cos(\omega r\frac{\sqrt 2}{l})) | \leq E_m
\end{equation}
and our bounds from Theorem $\ref{thm:error bounds}$ as
\begin{align}
	\frac{\sqrt 2}{l}|I( \omega \sin(\omega r\frac{\sqrt 2}{l})) - Q_m( \omega \sin(\omega r\frac{\sqrt 2}{l})) | &\leq 8(m-1)E_{m-1} \leq \frac{2e}{l^2} E_{m-2}, \\
	\frac{2}{l^2}| I(  \omega^2  \cos(\omega r\frac{\sqrt 2}{l})) - Q_m( \omega^2  \cos(\omega r\frac{\sqrt 2}{l})) | &\leq \frac{4}{l^2} (m-1)E_{m-2} \leq \frac{2e}{l^4} E_{m-3}.
\end{align}

Similar to normal quadrature, our proof technique is based on choosing polynomial $q$ of degree $2m-1$ whose values and derivatives agree with the original function $f$ and its derivatives at a specific set of points. We can then define the remainder $s \coloneqq f - q$. If we now approximate the inegral of the function $f$ by the integral of the polynomial $q$, bounding the approximation error is equivalent to bounding the integral of the remainder $r$.

We thus need two important components. After showing the existence of polynomials of a required degree that agree with $f$ at specific points, we need some results regarding the residuals of polynomial approximations to functions. These components are demonstrated in the next two sections, which will then be combined in the final proof in Section \ref{section:Derivation of bounds}.

\subsection{Polynomial approximation residuals}
\label{section:Polynomial approximation residuals}
First, let us restate the following two well known Lemmata \citep[see e.g.][]{quadratBook}:
\begin{lemma}
	\label{lem:first}
	Let $f$ be a real function $n$ times continuously differentiable and $q_{n}(x)$ a polynomial of degree $n-1$ that agrees with $f$ at the set of distinct points $S=\lbrace x_1,x_2, ... ,x_n \rbrace$. Let $\pi(x) = \prod_{i=1}^{n}(x-x_i)$. Then $\forall x \in \mathbb{R}$ we have 
	\begin{equation}
	f(x) -q_{n}(x) = \frac{f^{(n)}(\xi)}{n!}\pi(x)
	\end{equation}
	for some $\xi = \xi(x) \in I$, where $I$ is the interval limited by the smallest and largest of the numbers $x_1,x_2,...,x_n $ and $x$.
%	\begin{proof}
%		If $x \in S$ the property holds. Otherwise, fix $\bar{x}$. We define $F(x)= f(x) -q_{n}(x) - K\pi(x)$, where $K$ is chosen such that $F(\bar{x})=0$. We see that $F$ has $n+1$ distinct roots at $I$, so by Rolle's theorem $F^{\prime} $ has $n$ distinct roots in $I$,$F^{\prime\prime} $ has $n-1$ distinct roots in $I$ ..., $F^{(n)}$ has one root $\xi \in I$. At this point we have $$ 0=F^{(n)}(\xi) = f^{(n)}(\xi) - n!K \Leftrightarrow K=\frac{f^{(n)}(\xi)}{n!}. $$
%		So $F(\bar{x})=0 = f(\bar{x}) -q_{n}(\bar{x}) - \frac{f^{(n)}(\xi)}{n!}\pi(\bar{x}) \Leftrightarrow f(\bar{x}) -q_{n}(\bar{x}) = \frac{f^{(n)}(\xi)}{n!}\pi(\bar{x})$. And since $\bar{x}$ was chosen arbritrarily the result holds for all $x$.
%	\end{proof}
\end{lemma}

\begin{lemma}\label{second}
	Let $f$ be a real function $2n$ times continuously differentiable and $q_{2n}(x)$ a polynomial of degree $2n-1$ that agrees with $f$ at the set of distinct points $S=\lbrace x_1,x_2, ... ,x_n \rbrace$ and its derivative also agrees with $f^{\prime}$ at $S$. Let $\pi(x) = \prod_{i=1}^{n}(x-x_i)^2$. Then $\forall x \in \mathbb{R}$ we have 
	\begin{equation}
	f(x) -q_{2n}(x) = \frac{f^{(2n)}(\xi)}{(2n)!}\pi(x)
	\end{equation}
	for some $\xi = \xi(x) \in I$, where $I$ is the interval limited by the smallest and largest of the numbers $x_1,x_2,...,x_n $ and $x$.
%	\begin{proof}
%		If $x \in S$ the property holds. Otherwise, fix $\bar{x}$. We define $F(x)= f(x) -q_{2n}(x) - K\pi(x)$, where $K$ is chosen such that $F(\bar{x})=0$. We see that $F$ has $n+1$ distinct roots at $I$, so by Rolle's theorem $F^{\prime} $ has $n$ distinct roots in $I$ which are different from the points of $S$ (because $F^{\prime} $ vanishes at intermediate points). Moreover, $F^{\prime} $ vanishes at all the points in $S$ so in total $F^{\prime} $ has at least $2n$ distinct roots in $I$. Consequently, by Rolle's theorem again, $F^{\prime\prime} $ has $2n-1$ distinct roots in $I$ ..., $F^{(2n)}$ has one root $\xi \in I$. At this point we have 
%		\begin{equation}
%		0=F^{(2n)}(\xi) = f^{(2n)}(\xi) - (2n)!K \Leftrightarrow K=\frac{f^{(2n)}(\xi)}{(2n)!}.
%		\end{equation}
%		So $F(\bar{x})=0 = f(\bar{x}) -q_{2n}(\bar{x}) - \frac{f^{(2n)}(\xi)}{(2n)!}\pi(\bar{x}) \Leftrightarrow f(\bar{x}) -q_{2n}(\bar{x}) = \frac{f^{(2n)}(\xi)}{(2n)!}\pi(\bar{x})$. And since $\bar{x}$ was chosen arbritrarily the result holds for all $x$.
%	\end{proof}
\end{lemma}

For our proof, we will need the following two extensions:

\begin{lemma}\label{third}
	Let $f$ be a real function $2n+1$ times continuously differentiable and $q_{2n+1}(x)$ a polynomial of degree $2n$ that agrees with $f$ at the set of distinct points $S=\lbrace x_1,x_2, ... ,x_n,x_{*} \rbrace$ and its derivative also agrees with $f^{\prime}$ at 
	$S\setminus\lbrace x_{*} \rbrace$. Let $\pi(x) = (x-x_{*})\prod_{i=1}^{n}(x-x_i)^2$. Then $\forall x \in \mathbb{R}$ we have 
	\begin{equation}
	f(x) -q_{2n+1}(x) = \frac{f^{(2n+1)}(\xi)}{(2n+1)!}\pi(x)
	\end{equation} 
	for some $\xi = \xi(x) \in I$, where $I$ is the interval limited by the smallest and largest of the numbers $x_1,x_2,...,x_n,x_{*} $ and $x$.
	\begin{proof}
		If $x \in S$ the property holds. Otherwise, fix $\bar{x}$. We define $F(x)= f(x) -q_{2n+1}(x) - K\pi(x)$, where $K$ is chosen such that $F(\bar{x})=0$. We see that $F$ has $n+2$ distinct roots at $I$, so by Rolle's theorem $F^{\prime} $ has $n+1$ distinct roots in $I$ which are different from the points of $S$ (because $F^{\prime} $ vanishes at intermediate points). Moreover, $F^{\prime} $ vanishes at all the points in $S\setminus \lbrace x_{*} \rbrace $ so in total $F^{\prime} $ has at least $2n+1$ distinct roots in $I$. Consequently, by Rolle's theorem again, $F^{\prime\prime} $ has $2n$ distinct roots in $I$, ..., $F^{(2n+1)}$ has one root $\xi \in I$. At this point we have 
		\begin{equation}
		0=F^{(2n+1)}(\xi) = f^{(2n+1)}(\xi) - (2n+1)!K \Leftrightarrow K=\frac{f^{(2n+1)}(\xi)}{(2n+1)!}.
		\end{equation}
		
		So $F(\bar{x})=0 = f(\bar{x}) -q_{2n+1}(\bar{x}) - \frac{f^{(2n+1)}(\xi)}{(2n+1)!}\pi(\bar{x})^2 \Leftrightarrow f(\bar{x}) -q_{2n+1}(\bar{x}) = \frac{f^{(2n+1)}(\xi)}{(2n+1)!}\pi(\bar{x})^2$. And since $\bar{x}$ was chosen arbritrarily the result holds for all $x$.
	\end{proof}
\end{lemma}

\begin{lemma}\label{fourth}
	Let $f$ be a real function $2n+1$ times continuously differentiable and $q_{2n+1}(x)$ a polynomial of degree $2n$ that agrees with $f$ at the set of distinct points $S=\lbrace x_1,x_2, ... ,x_n \rbrace$, its derivative also agrees with $f^{\prime}$ at  $S$ and its second derivative agrees with the second derivative of $f$ at $x_1$. Let $\pi(x) = (x-x_1)\prod_{i=1}^{n}(x-x_i)^2$. Then $\forall x \in \mathbb{R}$ we have 
	\begin{equation}
	f(x) -q_{2n+1}(x) = \frac{f^{(2n+1)}(\xi)}{(2n+1)!}\pi(x)
	\end{equation} 
	for some $\xi = \xi(x) \in I$, where $I$ is the interval limited by the smallest and largest of the numbers $x_1,x_2,...,x_n $ and $x$.
	\begin{proof}
		If $x \in S$ the property holds. Otherwise, fix $\bar{x}$. We define $F(x)= f(x) -q_{2n+1}(x) - K\pi(x)$, where $K$ is chosen such that $F(\bar{x})=0$. We see that $F$ has $n+1$ distinct roots at $I$, so by Rolle's theorem $F^{\prime} $ has $n$ distinct roots in $I$ which are different from the points of $S$ (because $F^{\prime} $ vanishes at intermediate points). Moreover, $F^{\prime} $ vanishes at all the points in $S$ so in total $F^{\prime} $ has at least $2n$ distinct roots in $I$. Consequently, by Rolle's theorem again, $F^{\prime\prime} $ has $2n-1$ distinct roots in $I$ which are different from the points of $S$ (because $F^{\prime \prime} $ vanishes at intermediate points) and also $F^{\prime \prime} $ vanishes at $x_1$, so in total $F^{\prime \prime} $ vanishes at $2n$ points (at least),  ..., $F^{(2n+1)}$ has one root $\xi \in I$. At this point we have 
		\begin{equation}
		0=F^{(2n+1)}(\xi) = f^{(2n+1)}(\xi) - (2n+1)!K \Leftrightarrow K=\frac{f^{(2n+1)}(\xi)}{(2n+1)!}.
		\end{equation}
		
		So $F(\bar{x})=0 = f(\bar{x}) -q_{2n+1}(\bar{x}) - \frac{f^{(2n+1)}(\xi)}{(2n+1)!}\pi(\bar{x})^2 \Leftrightarrow f(\bar{x}) -q_{2n+1}(\bar{x}) = \frac{f^{(2n+1)}(\xi)}{(2n+1)!}\pi(\bar{x})^2$. And since $\bar{x}$ was chosen arbritrarily the result holds for all $x$.
	\end{proof}
\end{lemma}

\subsection{Existence of polynomials that agree with function at certain points}
\label{section:Existence of polynomials}
Lemma \ref{lem:first} - \ref{fourth} assume the existence of polynomials with certain properties. In the following, we will show that such polynomials actually exist and how they can be constructed.

Let $S = \lbrace x_1,x_2, \cdots x_n  \rbrace $ be a set of $n$ distinct points in $\mathbb{R}$ and define $\pi(x) = \prod_{i=1}^n(x-x_i)$ and for $i=1,..., n$, $l_i(x) = \frac{\pi(x)}{(x-x_i)\pi^{\prime}(x_i) }$. It holds that $l_i(x_j) = r_{ij}$ and that $deg(l_i)=n-1$. Moreover, we define for $i=1,..., n$, $h_i(x)=l_i^2(x)(-2l_i^{\prime}(x_i)(x-x_i) +1 )$ and $\bar{h}_i(x)=l_i^2(x)(x-x_i)$. It holds that $h_i(x_j) = r_{ij}$ , $h_i^{\prime}(x_j)=0$, $\bar{h}_i(x_j)=0$ and $\bar{h}^{\prime}_i(x_j)=r_{ij}$ and $deg(h_i)=deg(\bar{h}_i)=2n-1$ for $i=1,..., n$. Thus, we can directly state the following four Lemmata
\begin{lemma}\label{lem:existence of pol simple}
	Let $f$ be a real function and define $y_1(x) = \sum_{i=1}^n f(x_i)l_i(x)$. Then $y_1$ is a polynomial with degree $n-1$ that agrees with $f$ at $S$.
\end{lemma}

\begin{lemma}\label{lem:existence of pol with der}
	Let $f$ be a real differential function and define $y_2(x) = \sum_{i=1}^n f(x_i)h_i(x) + \sum_{i=1}^n f^{\prime}(x_i)\bar{h}_i(x)$. Then $y_2$ is a polynomial with degree $2n-1$ that agrees with $f$ at $S$ and its derivative also agrees with the derivative in our proofs  of $f$ at $S$.
\end{lemma}

\begin{lemma}\label{lem:existence of pol with der minus point}
	Consider a point $x_0 \not\in S$. Let $f$ be a real differential function and define $y_3(x) = y_2(x) + \frac{f(x_0) - y_2(x_0)}{\pi^2(x_0)} \pi^2(x) $ . Then $y_3$ is a polynomial with degree $2n$ that agrees with $f$ at $S\cup \lbrace x_0 \rbrace$ and its derivative also agrees with the derivative of $f$ at $S$.
\end{lemma}

\begin{lemma}\label{lem:existence of pol with second der}
	Consider the point $x_1 \in S$. Let $f$ be a real two times differential function and define $y_4(x) = y_2(x) + \frac{f^{\prime \prime}(x_1) - y_2^{\prime \prime}(x_1)}{2\pi^{\prime 2}(x_1)} \pi^2(x) $ . Then $y_4$ is a polynomial with degree $2n$ that agrees with $f$ at $S$, its derivative also agrees with the derivative of $f$ at $S$ and its second derivative agrees with the second derivative of $f$ at $x_1$.
\end{lemma}

\subsection{Derivation of bounds}
\label{section:Derivation of bounds}
In this section, we prove one of the two main theoretical results of our paper, namely the bounds on the error of our feature approximations (Theorem \ref{thm:error bounds}). We will start with Theorem \ref{thm:QuadratureBasics}, restating a well known result from Gauss-Hermite quadrature \citep{quadratBook}. This theorem will inspire the proofs for Theorem \ref{thm:GMFirstDeriv} and \ref{thm:GMSecondDerivative}, where we prove similar bounds for the more complex cases of first and second order derivatives.

\begin{theorem}
	\label{thm:QuadratureBasics}
	Consider the function $cos(\frac{\sqrt{2}}{l}r\omega)$. If we approximate the integral $I(cos(\frac{\sqrt{2}}{l}r\omega) = \int_{-\infty}^{+\infty} e^{-\omega^2}cos(\omega r\frac{\sqrt 2}{l})d\omega $ by $Q_m(cos(\frac{\sqrt{2}}{l}r\omega)$, the Gauss-Hermite quadrature scheme of order $m$, we have
	\begin{equation}
	|I(cos(\frac{\sqrt{2}}{l}r\omega)) - Q_m(cos(\frac{\sqrt{2}}{l}r\omega)) | \leq E_m.
	\end{equation}
	\begin{proof}
		Let $y_{2m}(\omega)$ be a polynomial of degree $2m-1$ that agrees with $cos(\frac{\sqrt{2}}{l}r\omega)$ at $S^m$  and its derivate also agrees with the derivative of $cos(\frac{\sqrt{2}}{l}r\omega)$ at $S^m$ (we know that such a polynomial exists by Lemma \ref{lem:existence of pol with der} ) . By Lemma \ref{second} we have for every $\omega$ that
		
		\begin{equation}
		cos(\frac{\sqrt{2}}{l}r\omega) -y_{2m}(\omega) = (\frac{\sqrt{2}r}{l})^{2m}\frac{cos(\xi)}{(2m)!}h_m^2(\omega)
		\end{equation}
		
		and since $r \leq 1 $ and $|cos(\xi)| \leq 1 $ we get
		
		\begin{equation}
		|cos(\frac{\sqrt{2}}{l}r\omega) -y_{2m}(\omega)| \leq (\frac{\sqrt{2}}{l})^{2m} \frac{h_m^2(\omega)}{(2m)!}.
		\end{equation}
		
		Moreover, $Q_m(cos(\frac{\sqrt{2}}{l}r\omega)) = Q_m(y_{2m}(\omega)) $ (because $cos(\frac{\sqrt{2}}{l}r\omega)$ and $y_{2m}(\omega)$ agree at $S^m$ ) and $ Q_m(y_{2m}(\omega)) = I(y_{2m}(\omega))$  (because $y_{2m}$ has degree less than $2m$) so $ Q_m(cos(\frac{\sqrt{2}}{l}r\omega)) = I(y_{2m}(\omega))$  and 
		
		\begin{align}
		&|I(cos(\frac{\sqrt{2}}{l}r\omega)) - Q_m(cos(\frac{\sqrt{2}}{l}r\omega)) | = |I(cos(\frac{\sqrt{2}}{l}r\omega)) -I(y_{2m}(\omega)) | \leq \\&   I(|cos(\omega r\frac{\sqrt 2}{l}) - y_{2m}(\omega)| ) \leq I (\frac{\sqrt{2}}{l})^{2m}\frac{h_m^2(\omega)}{(2m)!} ) = \\ & (\frac{\sqrt{2}}{l})^{2m} \frac{m!\sqrt{\pi}}{2^m(2m!)} \leq  E_m.
		\end{align}
	\end{proof}
\end{theorem}

In the preceding proof, the basic idea was to approximate the function with a polynomial (of degree less than $2m$) that agrees with the function at a specific set of points. This gives us a remainder that can be relatively efficiently bounded, leading to a tight error bound. In the following proofs, the main challenge lies in finding the right  approximating polynomial (of degree less than $2m$) that yields an easy to handle and efficiently bounded remainder. Once such a polynomial is constructed, we can use the following idea: Let $f(\omega)$ be the function to be approximated by the approximating polynomial $p(\omega)$ and assume that the remainder can be absolutely bounded by the polynomial $s(\omega)$. Then
\begin{align}
&|I(f(\omega)) - Q_m(f(\omega))| = |I(f(\omega)) - I(p(\omega)) + Q_m(p(\omega)) - Q_m(f(\omega))| \leq \\
&|I(f(\omega)) - I(p(\omega))| + |Q_m(p(\omega)) - Q_m(f(\omega))| \leq \\
&I(|f(\omega) - p(\omega) |) + Q_m(|f(\omega) - p(\omega) |) \leq  I(s(\omega)) + Q_m(s(\omega)).
\end{align}

If $s(\omega)$ has degree less than $2m$ so that $I(s(\omega)) = Q_m(s(\omega))$), then the final bound is $2I(s(\omega))$. Following standard practice, $s(\omega)$ will be chosen as the square of a polynomial. This is motivated by the following observation:

Consider $s(\omega)$ to be the square of a polynomial of degree $n$. W.l.o.g assume $s(\omega)$ to be monic (has leading coefficient $1$). Then $s(\omega) = (h_n(\omega) + q(\omega) )^2 $ with $deg(q)<n$. Consequently, $I(s(\omega)) = I(h^2_n(\omega)) + I(q^2(\omega)) + 2I(h_n(\omega)  q(\omega)) = I(h^2_n(\omega)) + I(q^2(\omega)) \geq I(h^2_n(\omega))$. This means that $s(\omega) = h^2_n(\omega)$ minimizes $I(s(\omega))$ and suggests that in our proofs, the approximating polynomial should agree with the function at the set $S^n$, so that the remainder  $s(\omega)$ is of the form $h^2_n(\omega)$ and gives us good values for $I(s(\omega))$.

We now have all the necessary tools to state and prove Theorem \ref{thm:error bounds}. We split the theorem into two parts: Theorem \ref{thm:GMFirstDeriv} restates the claim of Theorem \ref{thm:error bounds} for the first order derivative, while Theorem \ref{thm:GMSecondDerivative} restates the claim for the second order derivative.
\begin{theorem}
	\label{thm:GMFirstDeriv}
	Consider the function $\frac{\sqrt 2}{l}\omega \sin(\omega r\frac{\sqrt 2}{l})$. If we approximate the integral $I(\frac{\sqrt 2}{l}\omega \sin(\omega r\frac{\sqrt 2}{l})) = \int_{-\infty}^{+\infty} e^{-\omega^2}\frac{\sqrt 2}{l}\omega \sin(\omega r\frac{\sqrt 2}{l})d\omega $ by $Q_m(\frac{\sqrt 2}{l}\omega \sin(\omega r\frac{\sqrt 2}{l})$, the Gauss-Hermite quadrature scheme of order $m$, we have
	\begin{equation}
	\frac{\sqrt 2}{l} | I(\omega \sin(\omega r\frac{\sqrt 2}{l})) - Q_m(\omega \sin(\omega r\frac{\sqrt 2}{l})) | \leq 8(m-1)E_{m-1}.
	\end{equation}
	\begin{proof}
		
		Depending on wether $m$ is odd or even we have:\\
		Suppose $m$ is even. Then $0\in S^{m-1} $. Let $y_{2m-3}(\omega)$ be a polynomial of degree $2m-4$ that agrees with $sin(\frac{\sqrt{2}}{l}r\omega)$ at $S^{m-1}$ and its derivate also agrees with the derivative of $sin(\frac{\sqrt{2}}{l}r\omega)$ at $S^{m-1}\setminus \lbrace 0 \rbrace $ (we know that such a polynomial exists by Lemma \ref{lem:existence of pol with der minus point} ). By Lemma \ref{third} we have for every $\omega$ that 
		\begin{equation}
		\omega (sin(\frac{\sqrt{2}}{l}r\omega) -y_{2m-3}(\omega)) = (\frac{\sqrt{2}r}{l})^{2m-3}\frac{cos(\xi)}{(2m-3)!}h_{m-1}^2(\omega) 
		\end{equation}
		so since $r \leq 1 $ and $|cos(\xi)| \leq 1 $ we get
		\begin{equation}
		|\omega sin(\frac{\sqrt{2}}{l}r\omega) -\omega y_{2m-3}(\omega)| \leq (\frac{\sqrt{2}}{l})^{2m-3} \frac{h_{m-1}^2(\omega)}{(2m-3)!}.
		\end{equation}
		Consequently, since the weights of a quadrature scheme $W_i^m$ are positive and $Q_m(h_{m-1}^2(\omega)) = I(h_{m-1}^2(\omega)) $ (because $h^2_{m-1}$ has degree less than $2m$), using the above relation we get  
		\begin{align}
		&|Q_m(\omega y_{2m-3}(\omega) ) - Q_m(\omega \sin(\omega r\frac{\sqrt 2}{l})| = |Q_m(\omega y_{2m-3}(\omega)  - \omega \sin(\omega r\frac{\sqrt 2}{l}))| \leq \\ &  Q_m(|\omega y_{2m-3}(\omega)  - \omega \sin(\omega r\frac{\sqrt 2}{l})|) \leq Q_m((\frac{\sqrt{2}}{l})^{2m-3} \frac{h_{m-1}^2(\omega)}{(2m-3)!}) = \\ & I((\frac{\sqrt{2}}{l})^{2m-3} \frac{h_{m-1}^2(\omega)}{(2m-3)!}) :=R
		\end{align}
		
		where 
		\begin{equation}
		R= (\frac{\sqrt 2}{l})^{2m-3} \frac{(m-1)!\sqrt{\pi}}{2^{m-1}(2m-3)!}.
		\end{equation}
		
		Similarly, 
		\begin{align}
		&|I(\omega y_{2m-3}(\omega) ) - I(\omega \sin(\omega r\frac{\sqrt 2}{l})| = |I(\omega y_{2m-3}(\omega)  - \omega \sin(\omega r\frac{\sqrt 2}{l}))| \leq \\ &  I(|\omega y_{2m-3}(\omega)  - \omega \sin(\omega r\frac{\sqrt 2}{l})|) \leq I((\frac{\sqrt{2}}{l})^{2m-3} \frac{h_{m-1}^2(\omega)}{(2m-3)!}) =R.
		\end{align}
		
		Finally, using that $Q_m(\omega y_{2m-3}(\omega) ) = I(\omega y_{2m-3}(\omega) )$ (because $\omega y_{2m-3}(\omega)$ has degree less than $2m$) we get
		
		\begin{align}
		& \frac{\sqrt 2}{l} | I(\omega \sin(\omega r\frac{\sqrt 2}{l})) - Q_m(\omega \sin(\omega r\frac{\sqrt 2}{l})) | =\\
		&\frac{\sqrt 2}{l} | I(\omega \sin(\omega r\frac{\sqrt 2}{l})) - I(\omega y_{2m-3}(\omega) ) + Q_m(\omega y_{2m-3}(\omega) )  - Q_m(\omega \sin(\omega r\frac{\sqrt 2}{l})) | \leq 
		\\ &\frac{\sqrt 2}{l} (| I(\omega \sin(\omega r\frac{\sqrt 2}{l})) - I(\omega y_{2m-3}(\omega) )| + |Q_m(\omega y_{2m-3}(\omega) )  - Q_m(\omega \sin(\omega r\frac{\sqrt 2}{l})) |)
		\leq \\ & \frac{\sqrt{2}}{l}2R  =2 (\frac{\sqrt 2}{l})^{2m-2} \frac{(m-1)!\sqrt{\pi}}{2^{m-1}(2m-3)!} \leq 4(m-1)E_{m-1}.
		\end{align}
		
		Suppose $m$ is odd. Then $0\in S^{m-2} $. Let $y_{2m-3}(\omega)$ be a polynomial of degree $2m-4$ that agrees with $sin(\frac{\sqrt{2}}{l}r\omega)$ at $S^{m-2}$ and its derivate also agrees with the derivative of $sin(\frac{\sqrt{2}}{l}r\omega)$ at $S^{m-2}$ and its second derivative agrees at $0$ with with the second derivative of $sin(\frac{\sqrt{2}}{l}r\omega)$ (we know that such a polynomial exists by Lemma \ref{lem:existence of pol with second der} ). By Lemma \ref{fourth} we have for every $\omega$ that 
		\begin{equation}
		\omega (sin(\frac{\sqrt{2}}{l}r\omega) -y_{2m-3}(\omega)) = (\frac{\sqrt{2}r}{l})^{2m-3}\frac{cos(\xi)}{(2m-3)!}\omega^2 h_{m-2}^2(\omega)
		\end{equation}   
		so since $r \leq 1 $ and $|cos(\xi)| \leq 1 $ we get 
		
		\begin{equation}
		|\omega \sin(\frac{\sqrt{2}}{l}r\omega) -\omega y_{2m-3}(\omega)| \leq (\frac{\sqrt{2}}{l})^{2m-3} \frac{\omega^2 h_{m-2}^2(\omega)}{(2m-3)!}.
		\end{equation}

		Consequently, since the weights of a quadrature scheme $W_i^m$ are positive and $Q_m(\omega^2h_{m-2}^2(\omega)) = I(\omega^2h_{m-2}^2(\omega)) $ (because $\omega^2h^2_{m-2}$ has degree less than $2m$), using the above relation we get  
		\begin{align}
		&|Q_m(\omega y_{2m-3}(\omega) ) - Q_m(\omega \sin(\omega r\frac{\sqrt 2}{l})| = |Q_m(\omega y_{2m-3}(\omega)  - \omega \sin(\omega r\frac{\sqrt 2}{l}))| \leq \\ &  Q_m(|\omega y_{2m-3}(\omega)  - \omega \sin(\omega r\frac{\sqrt 2}{l})|) \leq Q_m((\frac{\sqrt{2}}{l})^{2m-3} \frac{\omega^2 h_{m-2}^2(\omega)}{(2m-3)!}) = \\ & I((\frac{\sqrt{2}}{l})^{2m-3} \frac{\omega^2 h_{m-2}^2(\omega)}{(2m-3)!}) :=R
		\end{align}

		where $$R= (\frac{\sqrt 2}{l})^{2m-3} \frac{I( \omega^2 h_{m-2}^2(\omega) )}{(2m-3)!}.  $$
		We will now use the identity $$xh_n(x) = h_{n+1}(x) + \frac{n}{2}h_{n-1}(x) $$
		and that $I(h_{n+1}(\omega) h_{n-1}(\omega)) =0$ (normality) so by squaring we have
		\begin{equation}
		I( \omega^2 h_{m-2}^2(\omega) ) = I( h_{m-1}^2(\omega) ) + (\frac{m-2}{2})^2 I ( h_{m-3}^2(\omega) ) \leq 2\frac{(m-1)!\sqrt{\pi}}{2^{m-1}}.
		\end{equation}
		
		So 
		\begin{equation}
		R \leq 2(\frac{\sqrt 2}{l})^{2m-3} \frac{(m-1)!\sqrt{\pi}}{2^{m-1}(2m-3)!}.
		\end{equation}
		
		Similarly, 
		\begin{align}
		&|I(\omega y_{2m-3}(\omega) ) - I(\omega \sin(\omega r\frac{\sqrt 2}{l})| = |I(\omega y_{2m-3}(\omega)  - \omega \sin(\omega r\frac{\sqrt 2}{l}))| \leq \\ &  I(|\omega y_{2m-3}(\omega)  - \omega \sin(\omega r\frac{\sqrt 2}{l})|) \leq I((\frac{\sqrt{2}}{l})^{2m-3} \frac{\omega^2 h_{m-2}^2(\omega)}{(2m-3)!}) =R.
		\end{align}
		
		Finally, using that $Q_m(\omega y_{2m-3}(\omega) ) = I(\omega y_{2m-3}(\omega) )$ (because $\omega y_{2m-3}(\omega)$ has degree less than $2m$) we get
		
		\begin{align}
		&\frac{\sqrt 2}{l} | I (\omega \sin(\omega r\frac{\sqrt 2}{l}) ) - Q_m(\omega \sin(\omega r\frac{\sqrt 2}{l})) | = \\
		&\frac{\sqrt 2}{l} | I(\omega \sin(\omega r\frac{\sqrt 2}{l})) - I(\omega y_{2m-3}(\omega) ) + Q_m(\omega y_{2m-3}(\omega) )  - Q_m(\omega \sin(\omega r\frac{\sqrt 2}{l})) | \leq 
		\\ &\frac{\sqrt 2}{l} (| I(\omega \sin(\omega r\frac{\sqrt 2}{l})) - I(\omega y_{2m-3}(\omega) )| + |Q_m(\omega y_{2m-3}(\omega) )  - Q_m(\omega \sin(\omega r\frac{\sqrt 2}{l})) |)
		\leq \\ &  \frac{\sqrt{2}}{l}2R  =4 (\frac{\sqrt 2}{l})^{2m-2} \frac{(m-1)!\sqrt{\pi}}{2^{m-1}(2m-3)!} \leq  8(m-1)E_{m-1}.
		\end{align}
		
	\end{proof}
\end{theorem}

\begin{theorem}
	\label{thm:GMSecondDerivative}
	Consider the function $\frac{ 2}{l^2} \omega^2 \cos(\omega r\frac{\sqrt 2}{l})$. If we approximate the integral $I(\frac{ 2}{l^2} \omega^2 \cos(\omega r\frac{\sqrt 2}{l})) = \int_{-\infty}^{+\infty} e^{-\omega^2}\frac{ 2}{l^2} \omega^2 \cos(\omega r\frac{\sqrt 2}{l})d\omega $ by $Q_m(\frac{ 2}{l^2} \omega^2 \cos(\omega r\frac{\sqrt 2}{l}))$, the Gauss-Hermite quadrature scheme of order $m$, we have
	\begin{equation}
	\frac{2}{l^2} | I(\omega^2 \cos(\omega r\frac{\sqrt 2}{l})  - Q_m(\omega^2 \cos(\omega r\frac{\sqrt 2}{l}) | \leq \frac{4}{l^2} (m-1)E_{m-2}.
	\end{equation}
	
	\begin{proof}
		
		Depending on wether $m$ is odd or even we have:\\
		Suppose $m$ is even. Then $0\in S^{m-1} $. Let $y_{2m-4}(\omega)$ be a polynomial of degree $2m-5$ that agrees with $\cos(\frac{\sqrt{2}}{l}r\omega)$ at $S^{m-1}\setminus \lbrace 0 \rbrace $ and its derivate also agrees with the derivative of $\cos(\frac{\sqrt{2}}{l}r\omega)$ at $S^{m-1}\setminus \lbrace 0 \rbrace $ (we know that such a polynomial exists by Lemma \ref{lem:existence of pol with der minus point} ). By Lemma \ref{third} we have for every $\omega$ that 
		\begin{equation}
		\omega^2 (\cos(\frac{\sqrt{2}}{l}r\omega) -y_{2m-4}(\omega)) = (\frac{\sqrt{2}r}{l})^{2m-4}\frac{\cos(\xi)}{(2m-4)!}h_{m-1}^2(\omega) 
		\end{equation}
		so since $r \leq 1 $ and $|cos(\xi)| \leq 1 $ we get
		\begin{equation}
		|\omega^2 \cos(\frac{\sqrt{2}}{l}r\omega) -\omega^2 y_{2m-4}(\omega)| \leq (\frac{\sqrt{2}}{l})^{2m-4} \frac{h_{m-1}^2(\omega)}{(2m-4)!}.
		\end{equation}
		Consequently, since the weights of a quadrature scheme $W_i^m$ are positive and $Q_m(h_{m-1}^2(\omega)) = I(h_{m-1}^2(\omega)) $ (because $h^2_{m-1}$ has degree less than $2m$), using the above relation we get  
		\begin{align}
		&|Q_m(\omega^2 y_{2m-4}(\omega) ) - Q_m(\omega^2 \cos(\omega r\frac{\sqrt 2}{l})| = |Q_m(\omega^2 y_{2m-4}(\omega)  - \omega^2 \cos(\omega r\frac{\sqrt 2}{l}))| \leq \\ &  Q_m(|\omega^2 y_{2m-4}(\omega)  - \omega^2 \cos(\omega r\frac{\sqrt 2}{l})|) \leq Q_m((\frac{\sqrt{2}}{l})^{2m-4} \frac{h_{m-1}^2(\omega)}{(2m-4)!}) = \\ & I((\frac{\sqrt{2}}{l})^{2m-4} \frac{h_{m-1}^2(\omega)}{(2m-4)!}) :=R
		\end{align}
		
		where 
		\begin{equation}
		R= (\frac{\sqrt 2}{l})^{2m-4} \frac{(m-1)!\sqrt{\pi}}{2^{m-1}(2m-4)!}.
		\end{equation}
		Similarly,
		\begin{align}
		&|I(\omega^2 y_{2m-4}(\omega) ) - I(\omega^2 \cos(\omega r\frac{\sqrt 2}{l})| = |I(\omega^2 y_{2m-4}(\omega)  - \omega^2 \cos(\omega r\frac{\sqrt 2}{l}))| \leq \\ &  I(|\omega^2 y_{2m-4}(\omega)  - \omega^2 \cos(\omega r\frac{\sqrt 2}{l})|) \leq  I((\frac{\sqrt{2}}{l})^{2m-4} \frac{h_{m-1}^2(\omega)}{(2m-4)!}) =R.
		\end{align}
		
		Finally using that $Q_m(\omega^2 y_{2m-4}(\omega) ) = I(\omega^2 y_{2m-4}(\omega) )$ (because $\omega^2 y_{2m-4}(\omega)$ has degree less than $2m$) and following exactly the same procedure as in the previous proof we get
		
		\begin{align}
		&\frac{2}{l^2} | I(\omega^2 \cos(\omega r\frac{\sqrt 2}{l})  - Q_m(\omega^2 \cos(\omega r\frac{\sqrt 2}{l}) |  \leq  \frac{2}{l^2}2R  =2 (\frac{\sqrt 2}{l})^{2m-2} \frac{(m-1)!\sqrt{\pi}}{2^{m-1}(2m-4)!} \leq  \frac{2}{l^2} (m-1)E_{m-2}.
		\end{align}

		Suppose $m$ is odd. Let $y_{2m-4}(\omega)$ be a polynomial of degree $2m-5$ that agrees with $\cos(\frac{\sqrt{2}}{l}r\omega)$ at $S^{m-2}$ and its derivate also agrees with the derivative of $\cos(\frac{\sqrt{2}}{l}r\omega)$ at $S^{m-2}$ (we know that such a polynomial exists by Lemma \ref{lem:existence of pol with der} ). By Lemma \ref{second} we have for every $\omega$ that
		
		\begin{equation}
		\omega^2 (\cos(\frac{\sqrt{2}}{l}r\omega) -y_{2m-4}(\omega)) = (\frac{\sqrt{2}r}{l})^{2m-4}\frac{cos(\xi)}{(2m-4)!}\omega^2 h_{m-2}^2(\omega) 
		\end{equation}
		so since $r \leq 1 $ and $|cos(\xi)| \leq 1 $ we get
		\begin{equation}
		|\omega^2 \cos(\frac{\sqrt{2}}{l}r\omega) -\omega^2 y_{2m-4}(\omega)| \leq (\frac{\sqrt{2}}{l})^{2m-4} \frac{\omega^2 h_{m-2}^2(\omega)}{(2m-4)!}.
		\end{equation}

		Consequently, since the weights of a quadrature scheme $W_i^m$ are positive and $Q_m(\omega^2h_{m-2}^2(\omega)) = I(\omega^2h_{m-2}^2(\omega)) $ (because $\omega^2h^2_{m-2}$ has degree less than $2m$), using the above relation we get  
		
		\begin{align}
		&|Q_m(\omega^2 y_{2m-4}(\omega) ) - Q_m(\omega^2 \cos(\omega r\frac{\sqrt 2}{l})| = |Q_m(\omega^2 y_{2m-4}(\omega)  - \omega^2 \cos(\omega r\frac{\sqrt 2}{l}))| \leq \\ &  Q_m(|\omega^2 y_{2m-4}(\omega)  - \omega^2 \cos(\omega r\frac{\sqrt 2}{l})|) \leq Q_m((\frac{\sqrt{2}}{l})^{2m-4} \frac{\omega^2 h_{m-2}^2(\omega)}{(2m-4)!}) = \\ & I((\frac{\sqrt{2}}{l})^{2m-4} \frac{\omega^2 h_{m-2}^2(\omega)}{(2m-4)!}) :=R
		\end{align}
		
		where $$R= (\frac{\sqrt 2}{l})^{2m-4} \frac{I( \omega^2 h_{m-2}^2(\omega) )}{(2m-4)!} . $$
		We will now use the identity $$xh_n(x) = h_{n+1}(x) + \frac{n}{2}h_{n-1}(x) $$
		and that $I(h_{n+1}(\omega) h_{n-1}(\omega)) =0$ (normality) so by squaring we have
		\begin{equation}
		I( \omega^2 h_{m-2}^2(\omega) ) = I( h_{m-1}^2(\omega) ) + (\frac{m-2}{2})^2 I ( h_{m-3}^2(\omega) ) \leq 2\frac{(m-1)!\sqrt{\pi}}{2^{m-1}}.
		\end{equation}
		
		So 
		\begin{equation}
		R \leq 2(\frac{\sqrt 2}{l})^{2m-4} \frac{(m-1)!\sqrt{\pi}}{2^{m-1}(2m-4)!}.
		\end{equation}
		
		Similarly,
		\begin{align}
		&|I(\omega^2 y_{2m-4}(\omega) ) - I(\omega^2 \cos(\omega r\frac{\sqrt 2}{l})| = |I(\omega^2 y_{2m-4}(\omega)  - \omega^2 \cos(\omega r\frac{\sqrt 2}{l}))| \leq \\ &  I(|\omega^2 y_{2m-4}(\omega)  - \omega^2 \cos(\omega r\frac{\sqrt 2}{l})|) \leq  I((\frac{\sqrt{2}}{l})^{2m-4} \frac{\omega^2 h_{m-2}^2(\omega)}{(2m-4)!}) =R.
		\end{align}
		
		Finally, using that $Q_m(\omega^2 y_{2m-4}(\omega) ) = I(\omega^2 y_{2m-4}(\omega) )$ (because $\omega^2 y_{2m-4}(\omega)$ has degree less than $2m$) and following exactly the same procedure as in the previous proof we get
		
		\begin{align}
		&\frac{2}{l^2} | I(\omega^2 \cos(\omega r\frac{\sqrt 2}{l})  - Q_m(\omega^2 \cos(\omega r\frac{\sqrt 2}{l}) |  \leq  \frac{2}{l^2}2R  =4 (\frac{\sqrt 2}{l})^{2m-2} \frac{(m-1)!\sqrt{\pi}}{2^{m-1}(2m-4)!} \leq  \frac{4}{l^2} (m-1)E_{m-2}.
		\end{align}
		
	\end{proof}
\end{theorem}

\newpage

\section{Kernel Approximation Additional Plots}
\label{sec:AppendixKernelApproxPlots}
\begin{figure}[!h]
	\centering
	\begin{subfigure}[t]{0.325\textwidth}
		\centering
		\includegraphics[width=\textwidth]{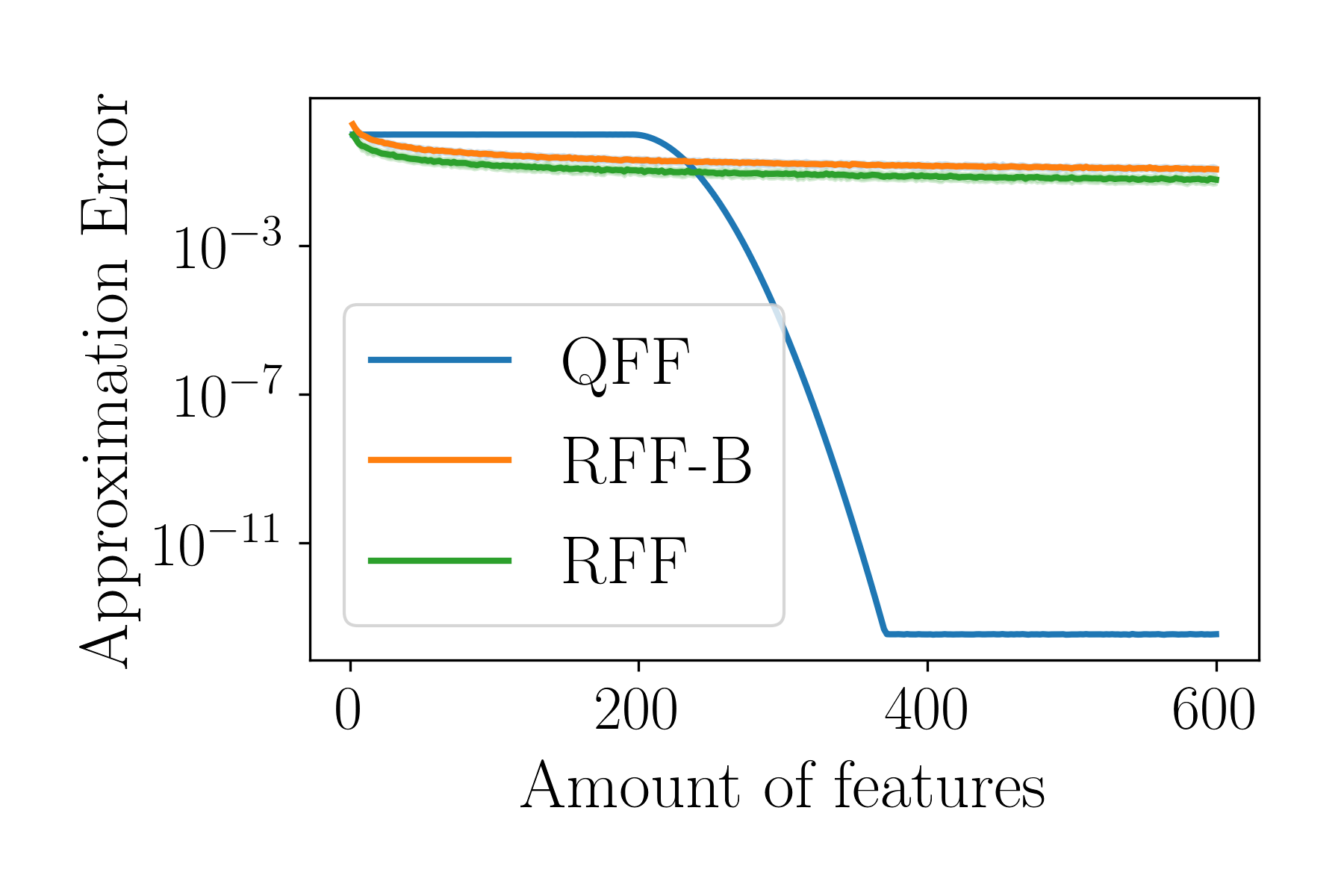}
		\caption{$k(r)$}
	\end{subfigure}
	\hfill
	\begin{subfigure}[t]{0.325\textwidth}
		\centering
		\includegraphics[width=\textwidth]{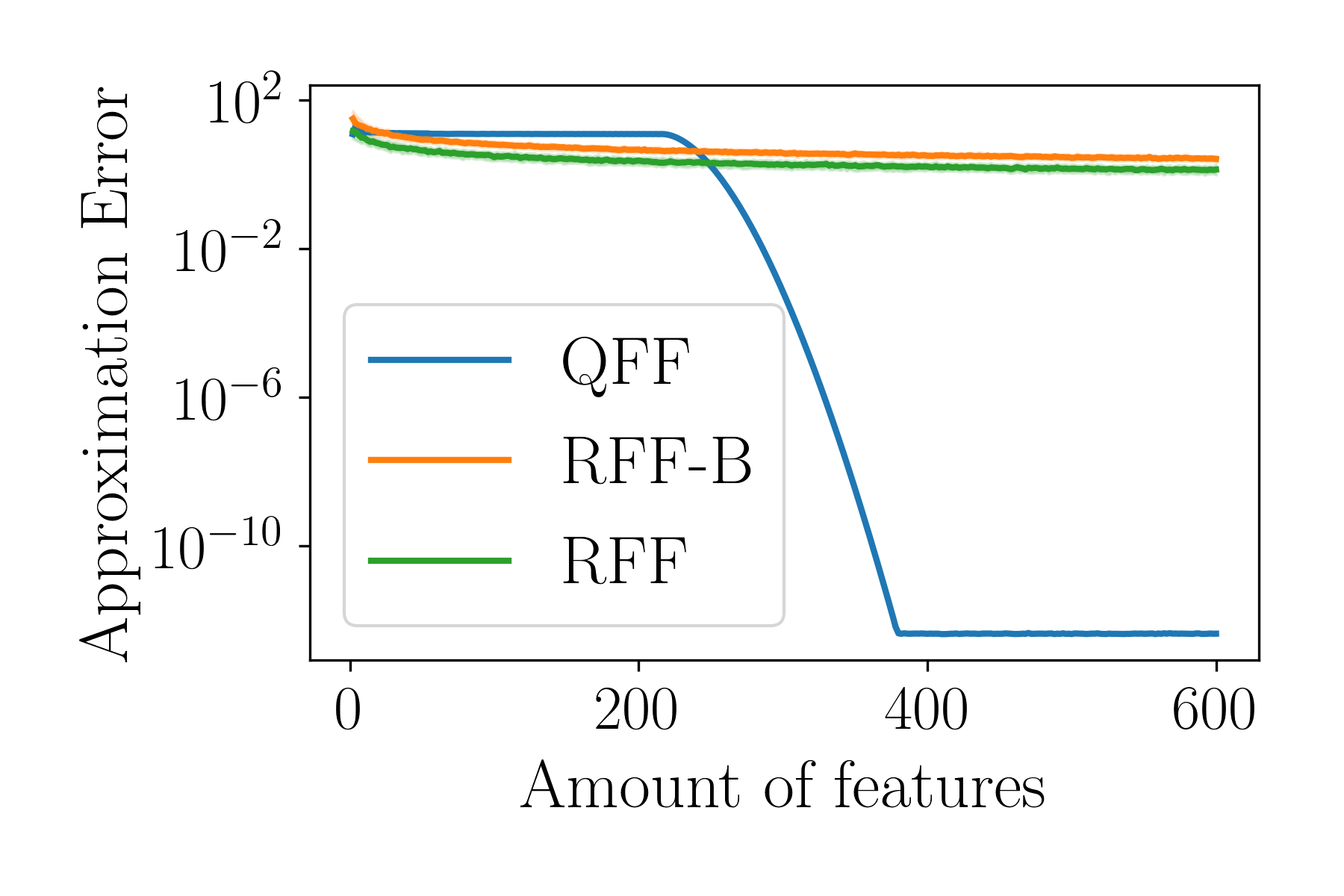}
		\caption{$k^{\prime}(r)$}
	\end{subfigure}
	\hfill
	\begin{subfigure}[t]{0.325\textwidth}
		\centering
		\includegraphics[width=\textwidth]{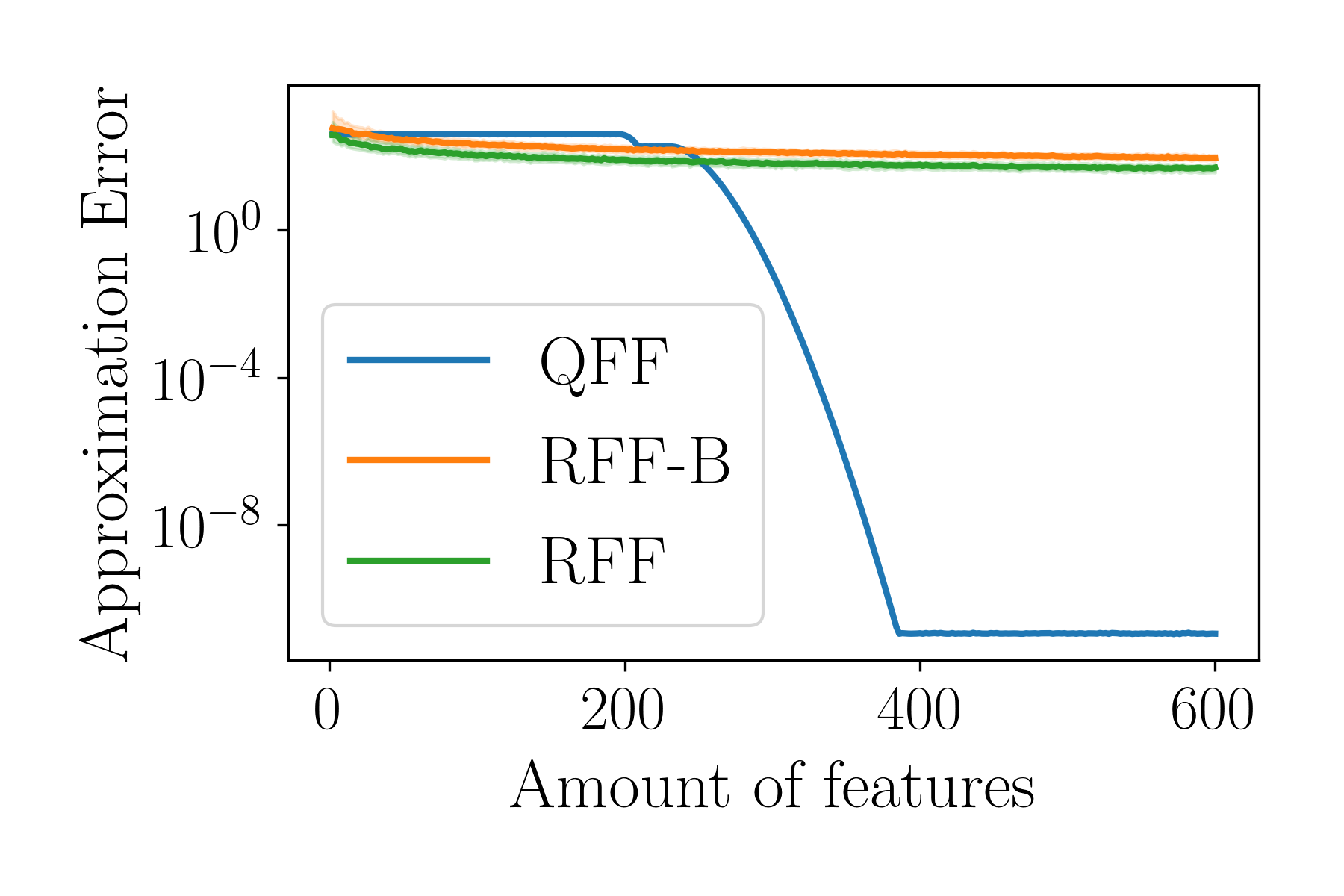}
		\caption{$k^{\prime \prime}(r)$}
	\end{subfigure}
	\caption{Comparing the maximum error of different feature expansions over $r \in [0, 1]$. For the random feature expansions, we show median as well as 12.5\% and 87.5\% quantiles over 100 random samples. Due to the exponential decay of the error of the QFF approximation, this stochasticity is barely visible. As given by the theoretical analysis, the error is a bit higher for the derivatives, but still decaying exponentially. In this plot, we set $l=0.05$.}

\end{figure}

\begin{figure}[!h]
	\centering
	\begin{subfigure}[t]{0.325\textwidth}
		\centering
		\includegraphics[width=\textwidth]{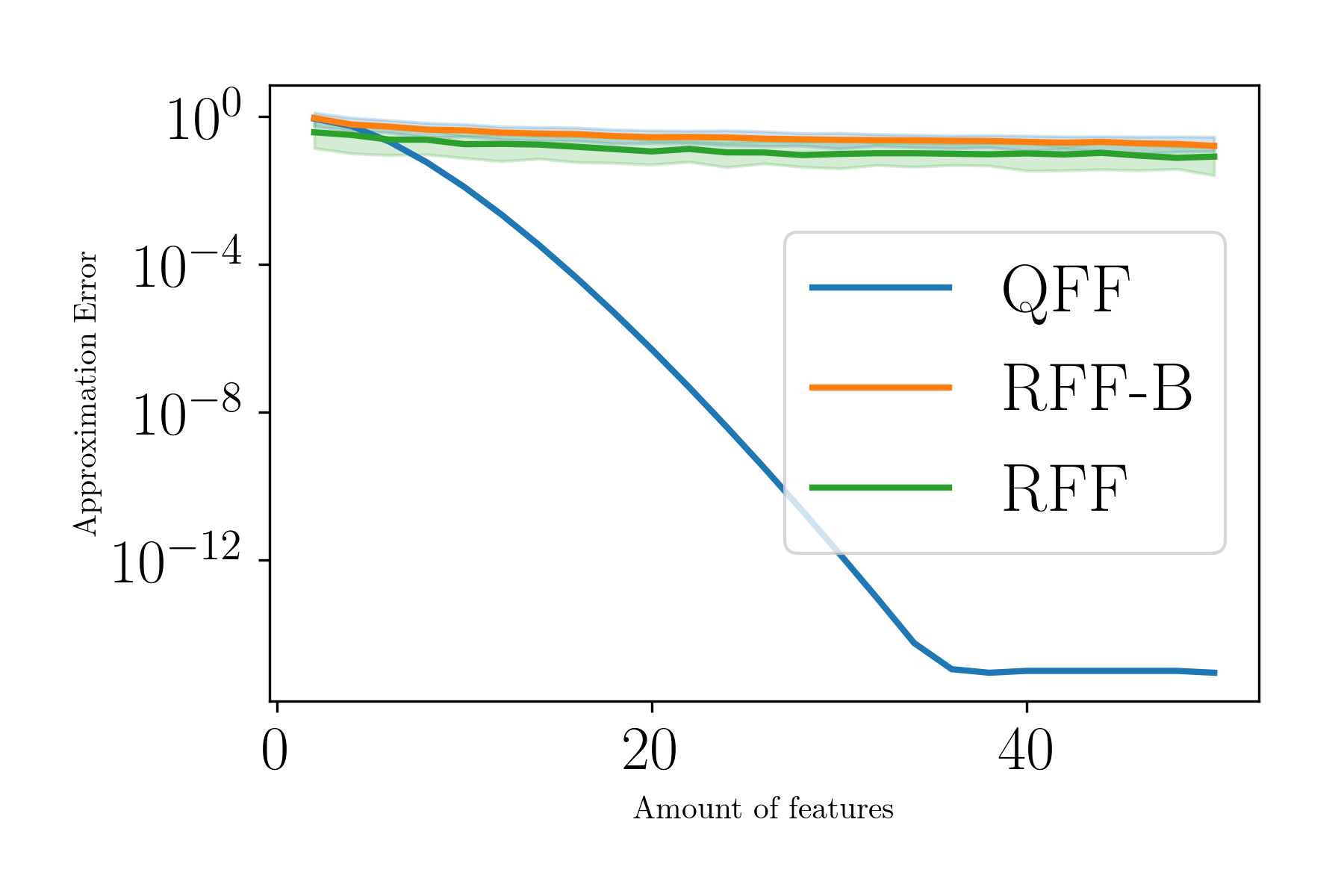}
		\caption{$k(r)$}
	\end{subfigure}
	\hfill
	\begin{subfigure}[t]{0.325\textwidth}
		\centering
		\includegraphics[width=\textwidth]{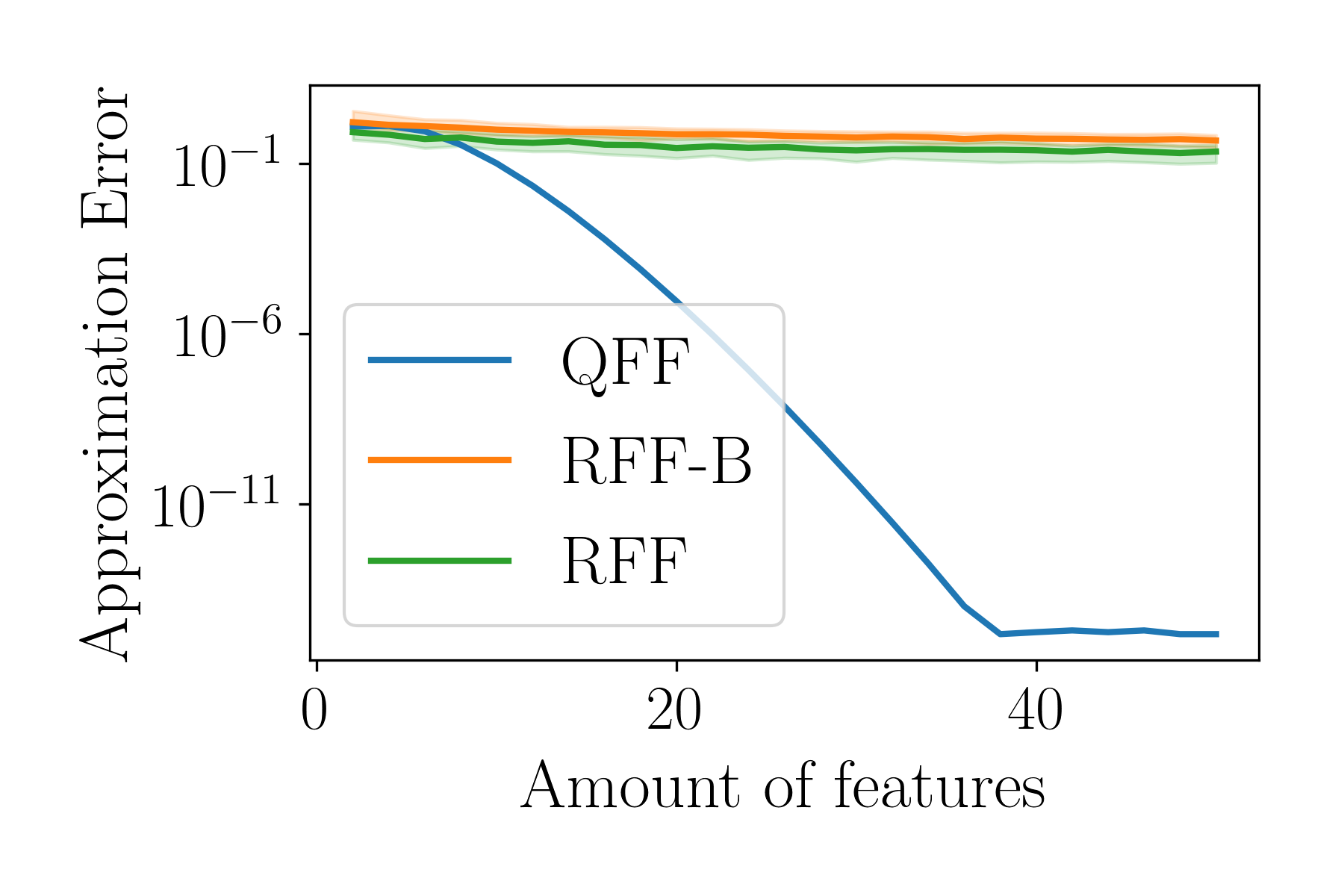}
		\caption{$k^{\prime}(r)$}
	\end{subfigure}
	\hfill
	\begin{subfigure}[t]{0.325\textwidth}
		\centering
		\includegraphics[width=\textwidth]{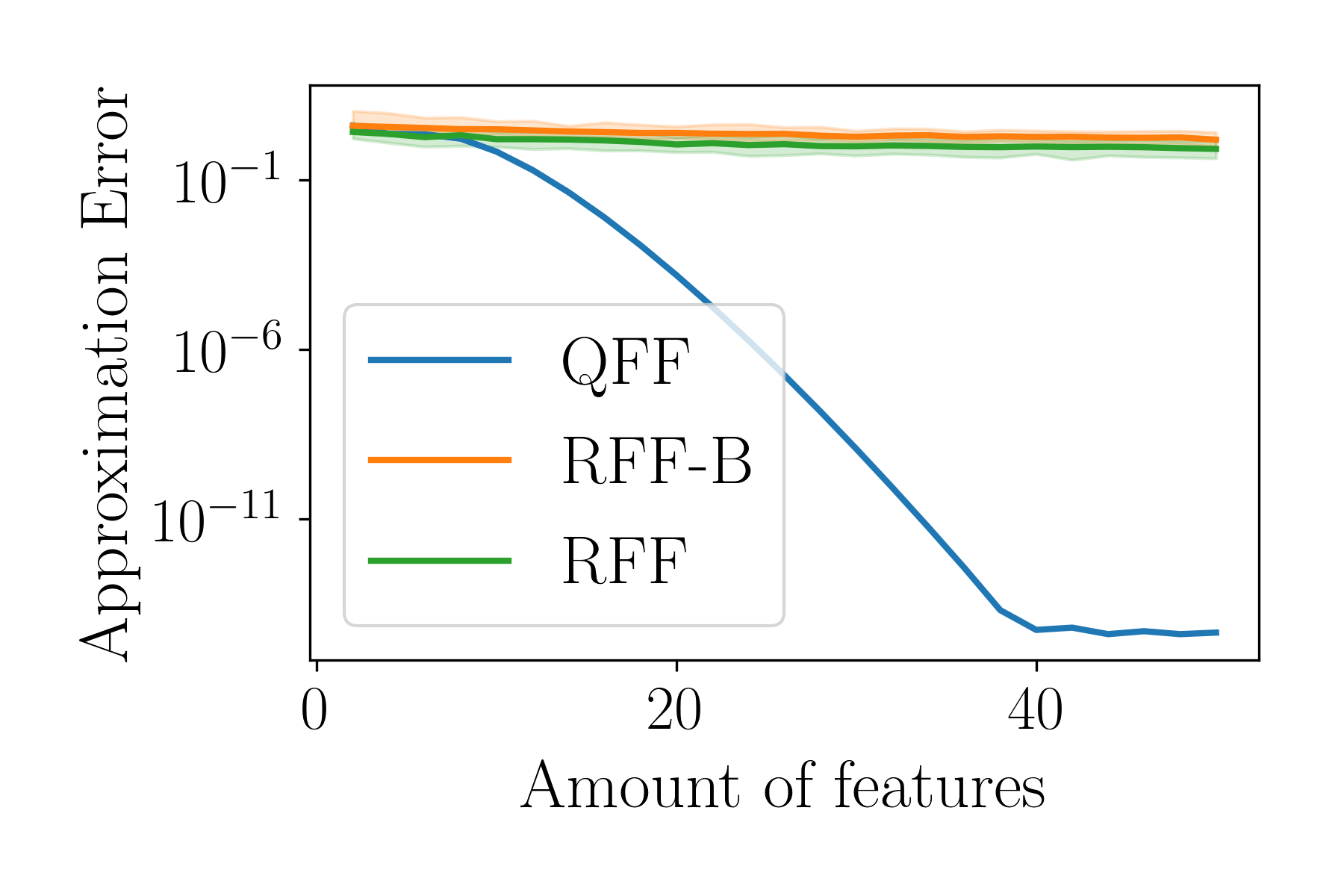}
		\caption{$k^{\prime \prime}(r)$}
	\end{subfigure}
	\caption{Comparing the maximum error of different feature expansions over $r \in [0, 1]$. For the random feature expansions, we show median as well as 12.5\% and 87.5\% quantiles over 100 random samples. Due to the exponential decay of the error of the QFF approximation, this stochasticity is barely visible. As given by the theoretical analysis, the error is a bit higher for the derivatives, but still decaying exponentially. In this plot, we set $l=0.5$.}
\end{figure}

\newpage

\section{GP Regression with Derivatives}
\label{sec:AppendixGPRWithDerivs}
In this section, we will introduce the necessary notation and the proof for our theoretical results on GP regression with derivatives.

\subsection{Notation}
\label{subsec:AppendixNotation}
Consider the problem of Gaussian Process regression, using zero mean prior and the RBF kernel function $k_{\bm{\phi}}(x,y) := \rho e^{-\frac{(x-y)^2}{2l^2}}$ for some fixed hyperparameters $\bm{\phi}=(\rho,l)$, which denote the variance and the lengthscale. Suppose we are given at $n$ observation points $\bm{t}:=\bm{t}_n=(t_1,\dots,t_n)$ the n dimensional (column) vectors $\bm{y}:= \bm{x} + \bm{\epsilon}_{\sigma^2}$ and $\bm{F}:= \bm{\dot{x}} + \bm{\epsilon}_{\gamma}$ of noisy state and noisy state derivative observations. Our goal is to infer at the observation point $T$ the values of state $x:=x(T)$  and state derivative $\dot{x}:=\dot{x}(T)$.

We repeat some of the definitions already given in this paper and this appendix as well as add a few new ones relevant for this section only:

Let $\bm{k}_{\bm{\phi}}(\bm{t},T) $ denote the $n$ dimensional kernel (column) vector, i.e.
\begin{equation}
\bm{k}_{\bm{\phi}}(\bm{t},T)_{i} := k_{\bm{\phi}}(t_i,T).
\end{equation}
Let $'k_{\bm{\phi}}(x,y)$  denote the partial derivative of $k_{\bm{\phi}}$ w.r.t. its first argument, i.e.
\begin{equation}
'k_{\bm{\phi}}(x,y) := \frac{\partial}{\partial a} k_{\bm{\phi}}(a, b) \rvert_{a=x, b=y}.
\end{equation}
Let $'\bm{k}_{\bm{\phi}}(\bm{t},T)$ denote the $n$ dimensional kernel derivative (column) vector, i.e.
\begin{equation}
'\bm{k}_{\bm{\phi}}(\bm{t},T)_{i} := \hspace{2pt}'\hspace{-1pt}k_{\bm{\phi}}(t_i,T).
\end{equation}
Let $k_{\bm{\phi}}'(x,y)$ denote the partial derivative of $k_{\bm{\phi}}$ w.r.t. its second argument, i.e.
\begin{equation}
k_{\bm{\phi}}'(x,y) := \frac{\partial}{\partial b} k_{\bm{\phi}}(a, b) \rvert_{a=x, b=y}.
\end{equation}
Let $\bm{k}_{\bm{\phi}}'(\bm{t},T)$ denote the $n$ dimensional kernel derivative (column) vector, i.e. 
\begin{equation}
\bm{k}_{\bm{\phi}}'(\bm{t},T)_{i} := k_{\bm{\phi}}' (t_i,T).
\end{equation}
Let $k_{\bm{\phi}}''(x,y)$ denote the mixed partial derivative of $k_{\bm{\phi}}$, i.e.
\begin{equation}
k_{\bm{\phi}}''(x,y) := \frac{\partial^2}{\partial a\partial b} k_{\bm{\phi}}(a, b) \rvert_{a=x, b=y}.
\end{equation}
Let $\bm{k}_{\bm{\phi}}''(\bm{t},T)$ denote the $n$ dimensional kernel derivative (column) vector, i.e.
\begin{equation}
\bm{k}_{\bm{\phi}}''(\bm{t},T)_{i} := k_{\bm{\phi}}'' (t_i,T).
\end{equation}
Let $\bm{C}_{\bm{\phi}} $ denote the $n\times n$ covariance kernel matrix, whose elements are given by
\begin{equation}
\left[\bm{C}_{\bm{\phi}} \right]_{i,j} := k_{\bm{\phi}}(t_i, t_j).
\end{equation}
Let $'\bm{C}_{\bm{\phi}}$ denote the kernel derivative matrix, whose elements are given by
\begin{equation}
\left['\bm{C}_{\bm{\phi}} \right]_{i,j} := \frac{\partial}{\partial a} k_{\bm{\phi}}(a, b) \rvert_{a=t_i, b=t_j}.
\end{equation}
Let $\bm{C}_{\bm{\phi}}' $ denote the kernel derivative matrix, whose elements are given by
\begin{equation}
\left[\bm{C}_{\bm{\phi}}'  \right]_{i,j} := \frac{\partial}{\partial b} k_{\bm{\phi}}(a, b) \rvert_{a=t_i, b=t_j}.
\end{equation}
Let $\bm{C}_{\bm{\phi}}''$ denote the mixed kernel derivative matrix, whose elements are given by
\begin{equation}
\left[\bm{C}_{\bm{\phi}}'' \right]_{i,j} := \frac{\partial^2}{\partial a \partial b} k_{\bm{\phi}}(a, b) \rvert_{a=t_i, b=t_j}.
\end{equation}
Let $\bm{\hat{K}}_{\bm{\phi}}$ denote the sum of the $2n\times 2n$ block matrix with the covariance matrix and its derivatives plus the diagonal noise matrix, i.e.
\begin{equation}
\bm{\hat{K}}_{\bm{\phi}} := \left( \begin{array}{cc}
\bm{C}_{\bm{\phi}} & \bm{C}_{\bm{\phi}}'  \\
'\bm{C}_{\bm{\phi}} &  \bm{C}_{\bm{\phi}}''
\end{array} \right)
+ \left( \begin{array}{cc}
\sigma^2 \mathbb{I}_n& \bm{0}  \\
\bm{0}  & \gamma \mathbb{I}_n
\end{array} \right).
\end{equation}
Let $\bm{\hat{k}}_{\phi}(\bm{t},T)$ denote the $2n$ dimensional (column) vector, which is a concatenation of $\bm{k}_{\bm{\phi}}(\bm{t},T)$ and $'\bm{k}_{\bm{\phi}}(\bm{t},T)$, i.e.
\begin{equation}
\bm{\hat{k}}_{\phi}(\bm{t},T):= \left( \begin{array}{c}
\bm{k}_{\bm{\phi}}(\bm{t},T)   \\
'\bm{k}_{\bm{\phi}}(\bm{t},T)
\end{array} \right).
\end{equation}
Let $\bm{\hat{k}}_{\phi}'(\bm{t},T)$ denote the $2n$ dimensional (column) vector, which is a concatenation of $\bm{k}_{\bm{\phi}}'(\bm{t},T)$ and $\bm{k}_{\bm{\phi}}''(\bm{t},T)$, i.e. 
\begin{equation}
\bm{\hat{k}}_{\phi}'(\bm{t},T):= \left( \begin{array}{c}
\bm{k}_{\bm{\phi}}'(\bm{t},T)   \\
\bm{k}_{\bm{\phi}}''(\bm{t},T)
\end{array} \right).
\end{equation}

Finally, we are able to write down the formulas for the scalar predictive mean and covariance at a new point $T$. Here, we let $\mu$ denote the mean of the state, $\mu'$ denote the mean of the derivative, $\Sigma$ denote the variance of the state and $\Sigma'$ the variance of the derivative prediction. They are given by
\begin{align}
\mu(T) &= \bm{\hat{k}}_{\phi}(\bm{t},T)^T \bm{\hat{K}}_{\bm{\phi}}^{-1} \left(\begin{array}{c}
\bm{y}\\
\bm{F} \end{array} \right), \\
\mu'(T) &= \bm{\hat{k}}_{\phi}'(\bm{t},T)^T \bm{\hat{K}}_{\bm{\phi}}^{-1} \left(\begin{array}{c}
\bm{y}\\
\bm{F} \end{array} \right),\\
\Sigma(T) &= k(T,T) - \bm{\hat{k}}_{\phi}(\bm{t},T)^T \bm{\hat{K}}_{\bm{\phi}}^{-1} \bm{\hat{k}}_{\phi}(\bm{t},T),\\
\Sigma'(T) &= k''(T,T) - \bm{\hat{k}}_{\phi}'(\bm{t},T)^T \bm{\hat{K}}_{\bm{\phi}}^{-1} \bm{\hat{k}}_{\phi}'(\bm{t},T).
\end{align}

\subsection{Proof of Theorem \ref{thm:GPRD}}
In this Section, we will prove the main theorem regarding the approximation error of GP regression with derivatives. For ease of reference, we will restate it as Theorem \ref{thm:GPRDAppendix}. Throughout this section, we will use $\tilde{a}$ to denote the feature approximation of any scalar, vector or matrix $a$.
\begin{theorem}
	Let us consider an RBF kernel with hyperparameters $(\rho, l)$ and domain $[0,1]$. Define $e_{\tilde{\mu}}$, $e_{\tilde{\Sigma}}$, $e_{\tilde{\mu}'}$ and $e_{\tilde{\Sigma}'}$ as the absolute error between the feature approximations and the corresponding accurate quantities of the means and covariances of Equations \eqref{eq:GPRD_post_state} and \eqref{eq:GPRD_post_deriv}. For each $\tau \in [0,1]$, define $e_{\text{tot}}$ as the maximum of these four errors. Define $c \coloneqq \min(\gamma, \sigma^2)$ and $R \coloneqq \max(||\bm{y}||_\infty, ||\bm{F}||_\infty)$. Let $C > 0$. Let us consider a QFF approximation scheme of order $m \geq 3+\max \left(\frac{e}{2l^2}, \log\left( \frac{270n^2\rho^3R}{l^8 c^2 C} \right) \right)$. 
	Then, it holds for all $\tau \in [0,1]$ that $e_{\text{tot}} \leq C$.
	\label{thm:GPRDAppendix}
\end{theorem}
\begin{proof}
	Suppose that we apply a QFF approximation scheme of order $m$ for the functions $k_{\bm{\phi}},'k_{\bm{\phi}},k_{\bm{\phi}}',k_{\bm{\phi}}''$, which gives us a deterministic and uniform (over their domain) approximation guarantee of absolute error less than $\epsilon:=\epsilon_{\phi}(m)$ (for any of them). W.l.o.g we can assume that the domain of these functions is $[0,1]^2$, $l \leq 1 $ and $\rho \geq 1$ (so also $0\leq T,t_1,\dots t_n \leq 1$). Moreover, we assume that $|\bm{y}|_{max}, |\bm{F}|_{max} \leq R$, for some positive constant $R$.
	
	Let $\bm{E}_1$ be the error (matrix) when approximating $\bm{\hat{K}}_{\bm{\phi}}$, i.e.
	\begin{equation}
	\bm{E}_1 := \bm{\hat{K}}_{\bm{\phi}} -  \bm{\tilde{ \hat{K}  }}_{\bm{\phi}} .
	\end{equation}
	Let $\bm{E}_2$be the error (matrix) when approximating $\bm{\hat{K}}_{\bm{\phi}}^{-1}$, i.e.
	\begin{equation}
	\bm{E}_2 := \bm{\hat{K}}_{\bm{\phi}}^{-1} -  \bm{\tilde{ \hat{K}}  }_{\bm{\phi}^{-1} }.
	\end{equation}
	Let $\bm{e}_1$ be the error (vector) when approximating $\bm{\hat{k}}_{\phi}(\bm{t},T)$, i.e.
	\begin{equation}
	\bm{e}_1 := \bm{\hat{k}}_{\phi}(\bm{t},T) -  \bm{ \tilde{\hat{k} }}_{\phi}(\bm{t},T).
	\end{equation}
	Let $\bm{e}_2$ be the error (vector) when approximating $\bm{\hat{k}}_{\phi}(\bm{t},T)$, i.e.
	\begin{equation}
	\bm{e}_2 := \bm{\hat{ k }}_{\phi}'(\bm{t},T) - \bm{\tilde{ \hat{ k }}_{\phi}  }'(\bm{t},T).
	\end{equation}
	
	Given these assumptions and definitions, we can introduce the following bounds:
	\begin{align}
	||\bm{e}_1|| &\leq \sqrt{2n}\epsilon,\\
	||\bm{e}_2|| &\leq \sqrt{2n}\epsilon,\\
	\sigma_1(\bm{E}_1) &\leq 2n\epsilon,\\
	\left|\left| \left(\begin{array}{c}
	\bm{y}\\
	\bm{F}
	\end{array} \right) \right|\right| &\leq \sqrt{2n}R.
	\end{align}
	Using the fact that $k_{\bm{\phi}} \leq \rho$, $'k_{\bm{\phi}} \leq \frac{\rho}{l}$ and $k_{\bm{\phi}}'' \leq \frac{2\rho}{l^2}$, we can also get
	\begin{align}
	\left|\left|\bm{\hat{k}}_{\phi}(\bm{t},T)\right|\right| &\leq \sqrt{2n}\frac{\rho}{l},\\
	\left|\left|\bm{\hat{k}}_{\phi}'(\bm{t},T) \right|\right|& \leq \sqrt{2n}\frac{2\rho}{l^2}.
	\end{align}
	We know that the matrix $\left( \begin{array}{cc}
	\bm{C}_{\bm{\phi}} & \bm{C}_{\bm{\phi}}'  \\
	'\bm{C}_{\bm{\phi}} &  \bm{C}_{\bm{\phi}}''
	\end{array} \right)$ and its QFF approximation are symmetric positive semi definite and thus, their smallest singular value is non negative. Also, $ \left( \begin{array}{cc}
	\sigma^2 \mathbb{I}_n& \bm{0}  \\
	\bm{0}  & \gamma \mathbb{I}_n
	\end{array} \right)$ has smallest singular value $c$. Thus, both $\bm{\hat{K}}$ and $ \bm{ \tilde{\hat{K}}}$ are symmetric, positive definite matrices and we have
	\begin{align}
	\sigma_1( \bm{\hat{K}}_{\bm{\phi}}^{-1} ) &\leq \frac{1}{c}, \\
	\sigma_1(   \bm{ \tilde{\hat{K}  }}_{\bm{\phi}}^{-1}    ) &\leq \frac{1}{c}.
	\end{align}
	Finally, we can use the Woodbury identity to get $\bm{E}_2 = - \bm{\hat{K}}_{\bm{\phi}}^{-1}   \bm{E}_1 \bm{\tilde{  \hat{K}  }}_{\bm{\phi}^{-1} }$, justifying
	\begin{equation}
	\sigma_1 ( \bm{E}_2 ) \leq \frac{2n}{c^2} \epsilon.
	\end{equation}
	We now have all the building blocks needed to provide a bound for $e_{\tilde{\mu}}$:
	Using all the above notations, assumptions and results we get
	\begin{align}
	|\epsilon_{\mu}| &= \left|\bm{\hat{k}}_{\phi}(\bm{t},T)^T \bm{\hat{K}}_{\bm{\phi}}^{-1} \left(\begin{array}{c}
	\bm{y}\\
	\bm{F}
	\end{array} \right) -
	(\bm{\hat{k}}_{\phi}(\bm{t},T)-\bm{e}_1)^T (\bm{\hat{K}}_{\bm{\phi}}^{-1} - \bm{E}_1) \left(\begin{array}{c}
	\bm{y}\\
	\bm{F}
	\end{array} \right)\right|\\
    &=
	\left|(\bm{e}_1^T\bm{\hat{K}}_{\bm{\phi}}^{-1} + \bm{\hat{k}}_{\phi}(\bm{t},T)^T\bm{E}_1 -\bm{e}_1^T\bm{E}_1 )\left(\begin{array}{c}
	\bm{y}\\
	\bm{F}
	\end{array} \right)\right| \\
	& \leq \sqrt{2n}R \left(\frac{\sqrt{2n}}{c}\epsilon +  \frac{\rho}{l}\sqrt{2n}2n\epsilon + \sqrt{2n}2n \epsilon^2\right) \\
	&\leq 10 \frac{n^2R\rho}{lc} \epsilon.
	\end{align}
	Similarly, we get
	\begin{align}
	|\epsilon_{\mu'}|&= 
	\left|(\bm{e}_2^T\bm{\hat{K}}_{\bm{\phi}}^{-1} + \bm{\hat{k}}_{\phi}'(\bm{t},T)^T\bm{E}_1 -\bm{e}_2^T\bm{E}_1 )\left(\begin{array}{c}
	\bm{y}\\
	\bm{F}
	\end{array} \right)\right|\\
	&\leq \sqrt{2n}R \left(\frac{\sqrt{2n}}{c}\epsilon +  \frac{\rho}{l^2}\sqrt{2n}4n\epsilon + \sqrt{2n}2n \epsilon^2\right)\\
	&\leq 14 \frac{n^2R\rho}{l^2c} \epsilon.
	\end{align}
	Moreover,
	\begin{align}
	|\epsilon_{\Sigma}| &= | k(T,T) - \bm{\hat{k}}_{\phi}(\bm{t},T)^T \bm{\hat{K}}_{\bm{\phi}}^{-1} \bm{\hat{k}}_{\phi}(\bm{t},T) -  \tilde{k}(T,T) \\
	 &+ (\bm{\hat{k}}_{\phi}(\bm{t},T) - \bm{e}_1)^T (\bm{\hat{K}}_{\bm{\phi}}^{-1} - \bm{E}_1) (\bm{\hat{k}}_{\phi}(\bm{t},T) - \bm{e}_1)| \\
	 &\leq \epsilon + |-2\bm{e}_1^T \bm{\tilde{ \hat{K}  }}_{\bm{\phi}^{-1} } \bm{\hat{k}}_{\phi}(\bm{t},T) + \bm{e}_1^T \bm{\tilde{ \hat{K}  }}_{\bm{\phi}}^{-1} \bm{e}_1 - \bm{\hat{k}}_{\phi}(\bm{t},T)^T\bm{E}_2\bm{\hat{k}}_{\phi}(\bm{t},T)| \\
	 &\leq \epsilon + \frac{4n\rho}{lc}\epsilon +\frac{2n}{c}\epsilon^2 + \frac{4n^2\rho^2}{l^2c^2}\epsilon \\
	 &\leq \frac{14n^2\rho^2}{l^2c^2}\epsilon
	\end{align}
	and
	\begin{align}
	|\epsilon_{\Sigma'}| &\leq \epsilon + |-2\bm{e}_2^T \bm{\tilde{ \hat{K}  }}_{\bm{\phi}}^{-1}  \bm{\hat{k}}_{\phi}' (\bm{t},T) + \bm{e}_2^T \bm{\tilde{ \hat{K}  }}_{\bm{\phi}}^{-1} \bm{e}_2 - \bm{\hat{k}}_{\phi}'  (\bm{t},T)^T\bm{E}_2\bm{\hat{k}}_{\phi}'(\bm{t},T)|\\
	&\leq \epsilon + \frac{8n\rho}{l^2c}\epsilon + \frac{2n}{c}\epsilon^2 + \frac{16n^2\rho^2}{l^4c^2}\epsilon\\
	&\leq \frac{27n^2\rho^2}{l^4c^2}\epsilon.
	\end{align}
	
	To summarize all of these upper bounds, we can observe that
	\begin{equation}
	|e_{\text{tot}}| \leq \frac{27n^2\rho^2R}{l^4c^2}\epsilon.
	\end{equation}
	Let us now fix a constant $0<C<1$ and define $M \coloneqq m-3$. Let us choose M such that
	\begin{equation}
	M \geq \max(\frac{e}{2l^2},\log(\frac{270n^2\rho^3R}{l^8c^2C})).
	\end{equation}
	With this choice, it holds that
	\begin{alignat}{2}
	& (\frac{e}{4l^2M})^M &&\leq \frac{l^8c^2}{270n^2\rho^3R}C\\
	\implies& \frac{2e\sqrt{\pi}\rho}{l^4} (\frac{e}{4l^2M})^M &&\leq \frac{l^4c^2}{27n^2\rho^2R}C\\
	\implies& \epsilon &&\leq \frac{l^4c^2}{27n^2\rho^2R}C\\
	\iff& \frac{27n^2\rho^2R}{l^4c^2}\epsilon &&\leq C,\\
	\end{alignat}
	which concludes the proof of this theorem.
\end{proof}

The above bound could be reformulated as $m \geq 12+ \max(\frac{e}{2l^2},\log(\frac{n^2\rho^3R}{l^8c^2C})) $. Moreover, we assumed that $R\geq 1 $ and that $c\leq1$. If any of these conditions is not met, then the bounds are still valid if we substitute these quantities by $1$ (the same holds also for $\rho$ and $l$). Finally, we implicitly assumed that $\epsilon$, the uniform upper bound of the approximation for $k_{\bm{\phi}},'k_{\bm{\phi}},k_{\bm{\phi}}',k_{\bm{\phi}}''$ is smaller than 1. This happens when $m \geq 3+ \max(\frac{e}{2l^2},\log( \frac{10\rho}{l^4} )) $ , a condition which is met if $m \geq 3+ \max(\frac{e}{2l^2},\log(\frac{270n^2\rho^3R}{l^8c^2C})) $.

\newpage

\section{Additional Empirical Evaluation GPR}
\label{sec:AppendixGPRWithDerivEval}
\subsection{Lotka Volterra}

\begin{figure}[!h]
	\centering
	\begin{subfigure}[t]{0.24\textwidth}
		\centering
		\includegraphics[width=\textwidth]{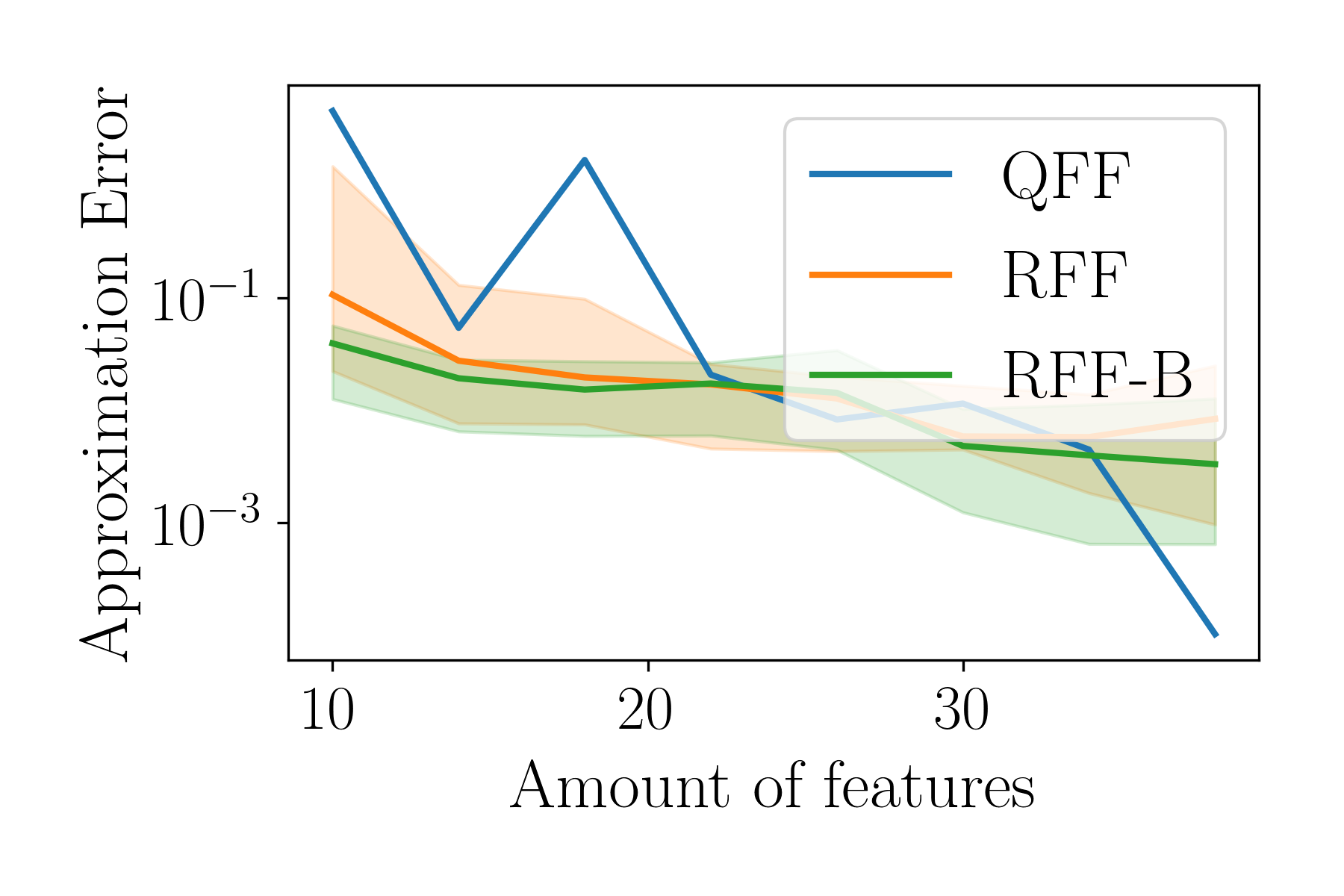}
		\caption{$\mu_0$}
	\end{subfigure}
	\hfill
	\begin{subfigure}[t]{0.24\textwidth}
	\centering
	\includegraphics[width=\textwidth]{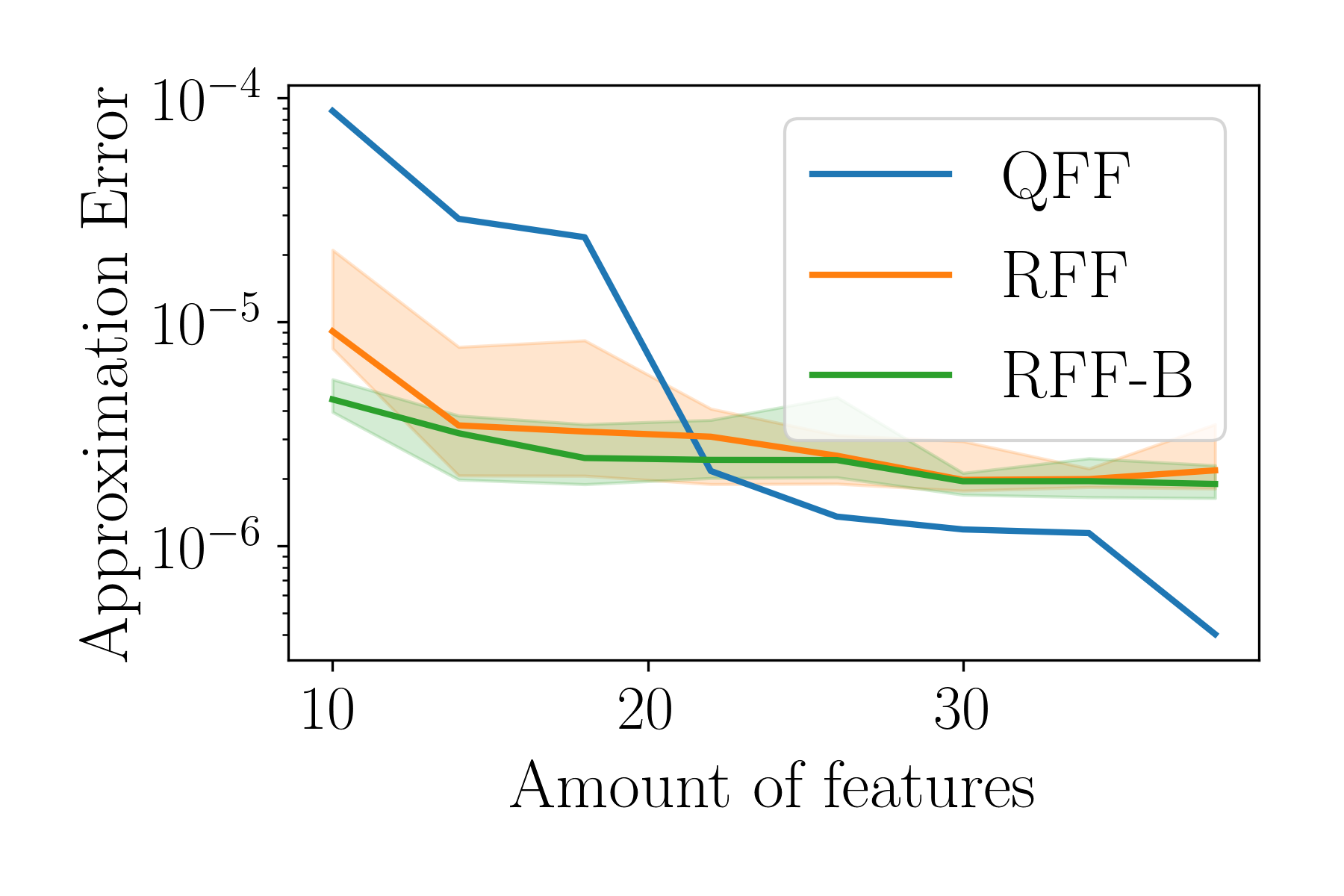}
	\caption{$\Sigma_0$}
	\end{subfigure}
	\hfill
	\begin{subfigure}[t]{0.24\textwidth}
	\centering
	\includegraphics[width=\textwidth]{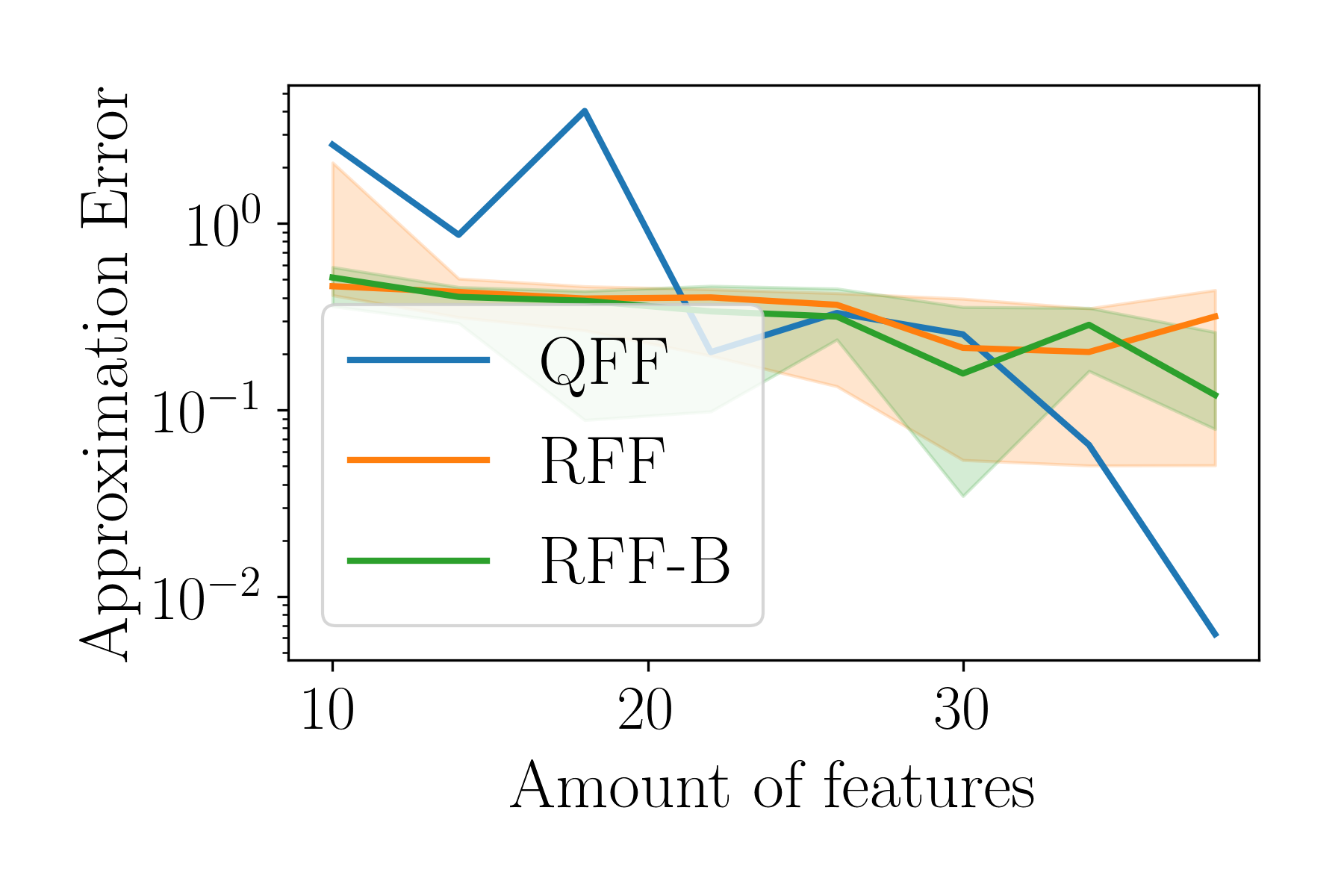}
	\caption{$\mu'_0$}
	\end{subfigure}
	\hfill
	\begin{subfigure}[t]{0.24\textwidth}
	\centering
	\includegraphics[width=\textwidth]{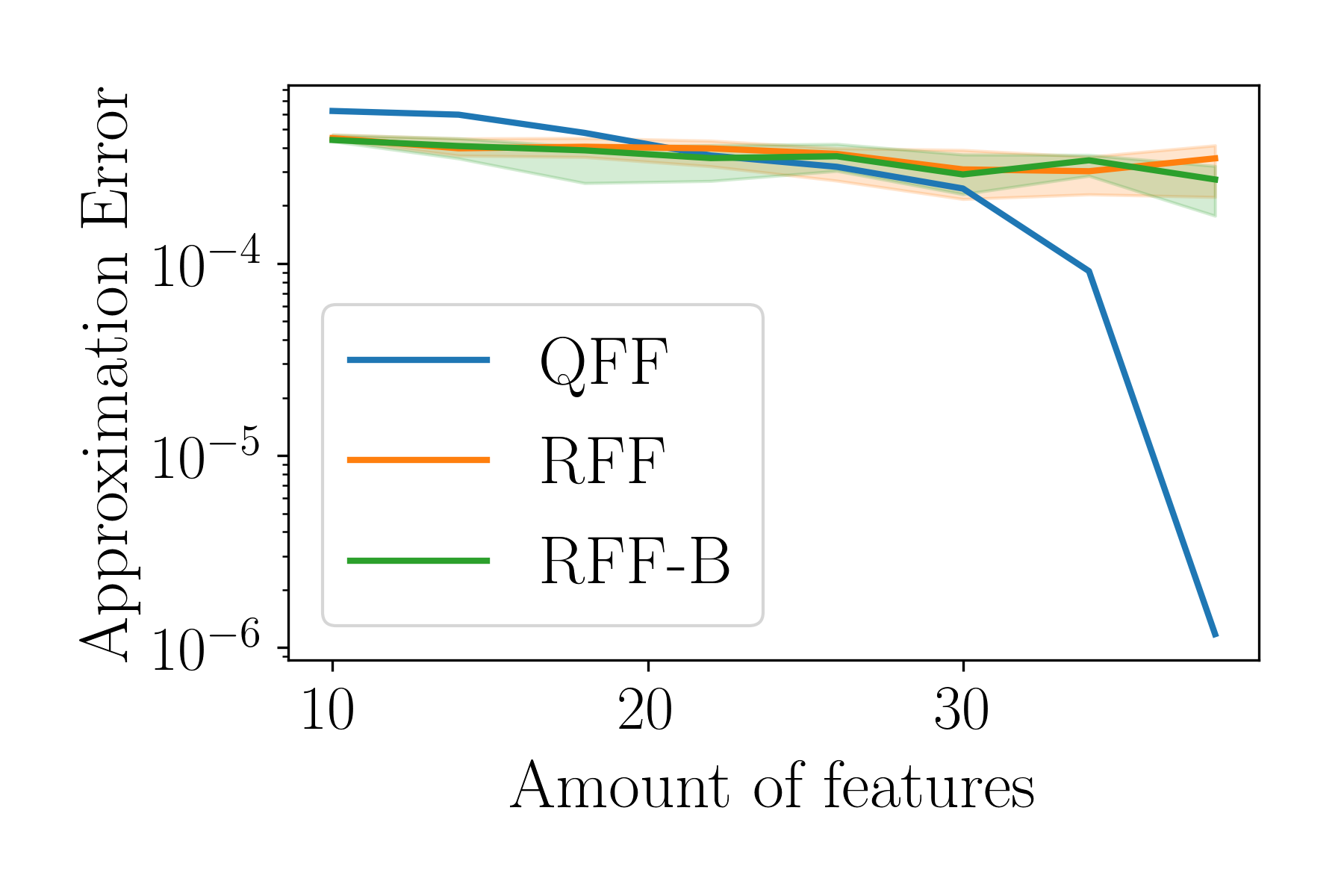}
	\caption{$\Sigma'_0$}
	\end{subfigure}\\
	\begin{subfigure}[t]{0.24\textwidth}
	\centering
	\includegraphics[width=\textwidth]{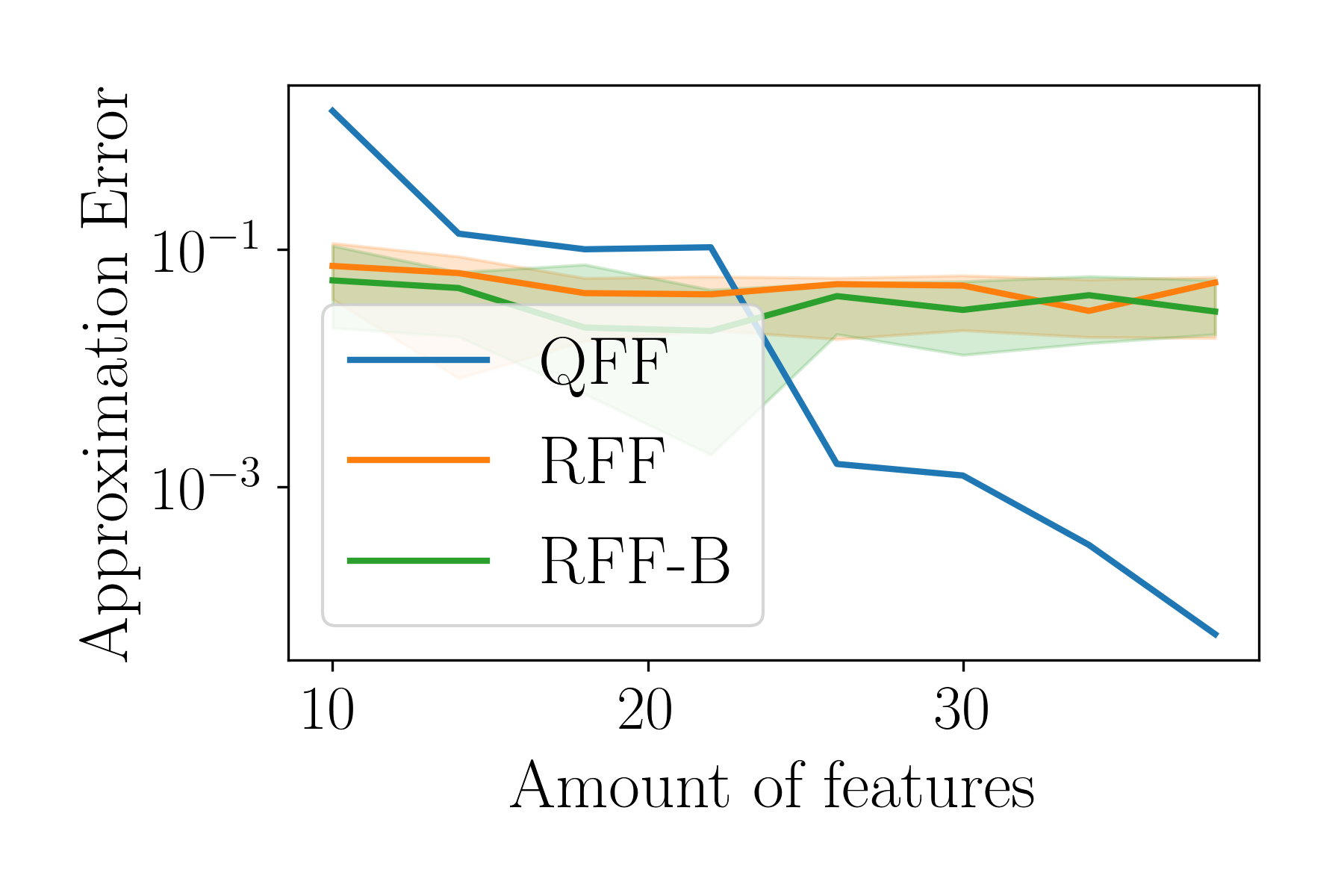}
	\caption{$\mu_1$}
	\end{subfigure}
	\hfill
	\begin{subfigure}[t]{0.24\textwidth}
		\centering
		\includegraphics[width=\textwidth]{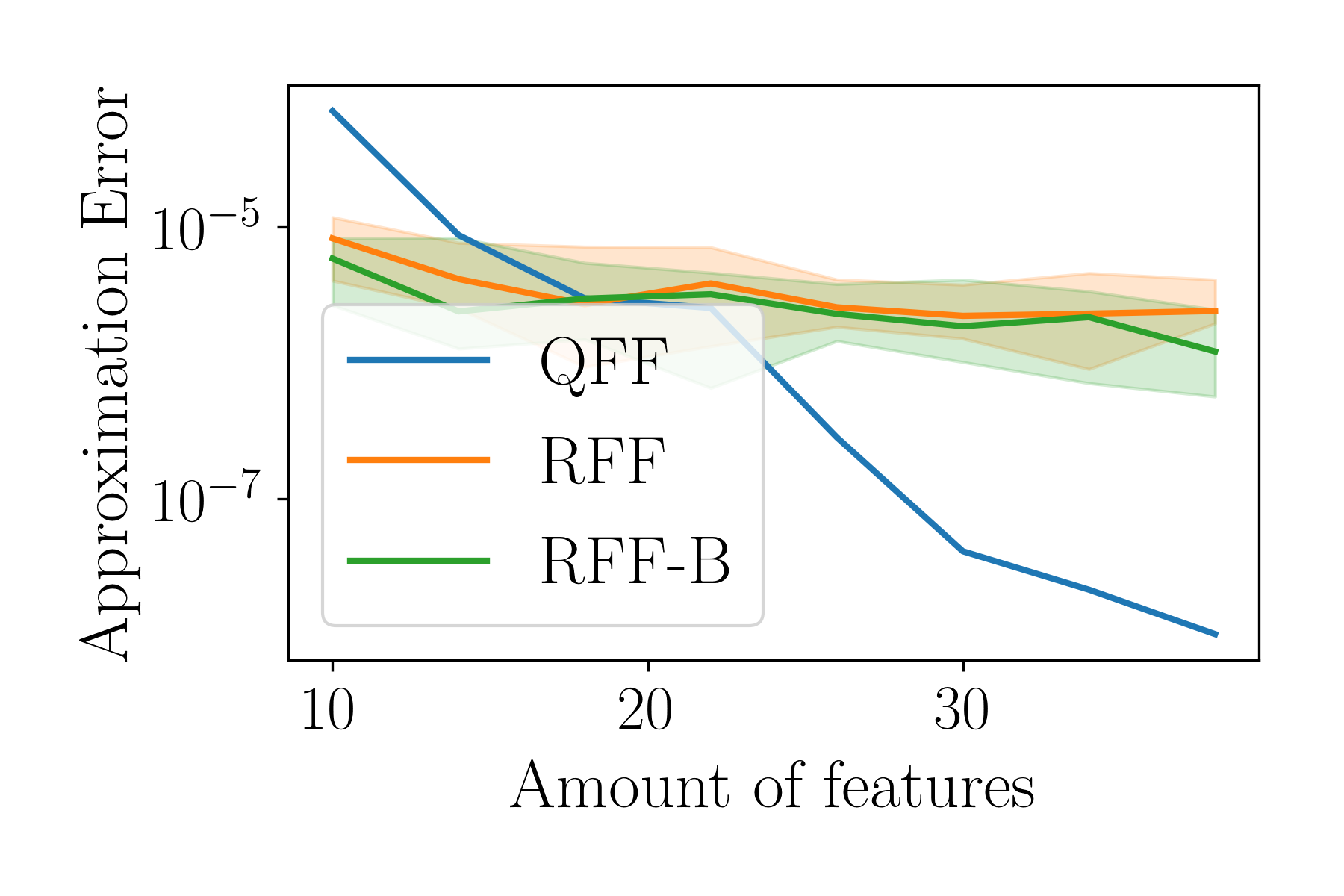}
		\caption{$\Sigma_1$}
	\end{subfigure}
	\hfill
	\begin{subfigure}[t]{0.24\textwidth}
		\centering
		\includegraphics[width=\textwidth]{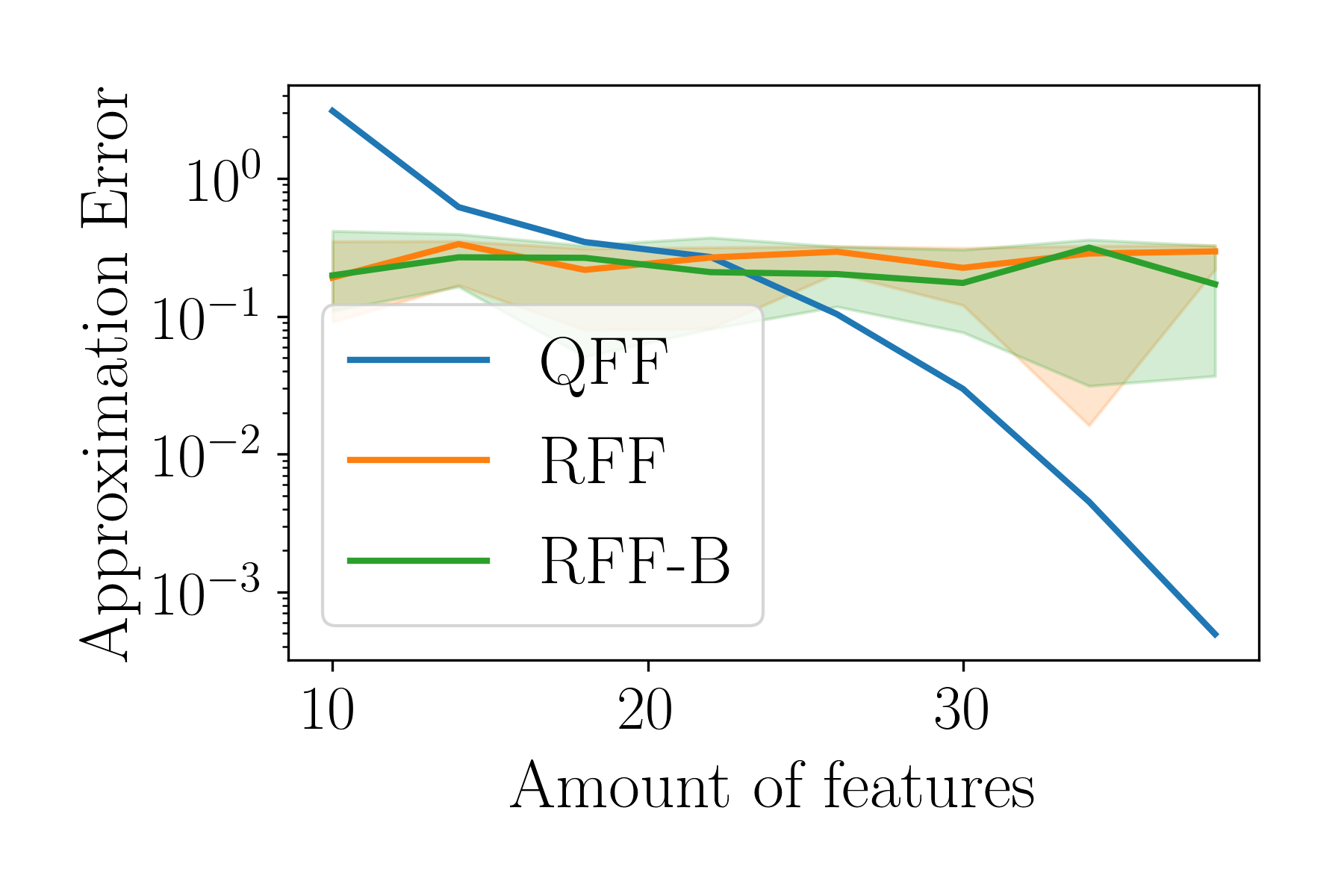}
		\caption{$\mu'_1$}
	\end{subfigure}
	\hfill
	\begin{subfigure}[t]{0.24\textwidth}
		\centering
		\includegraphics[width=\textwidth]{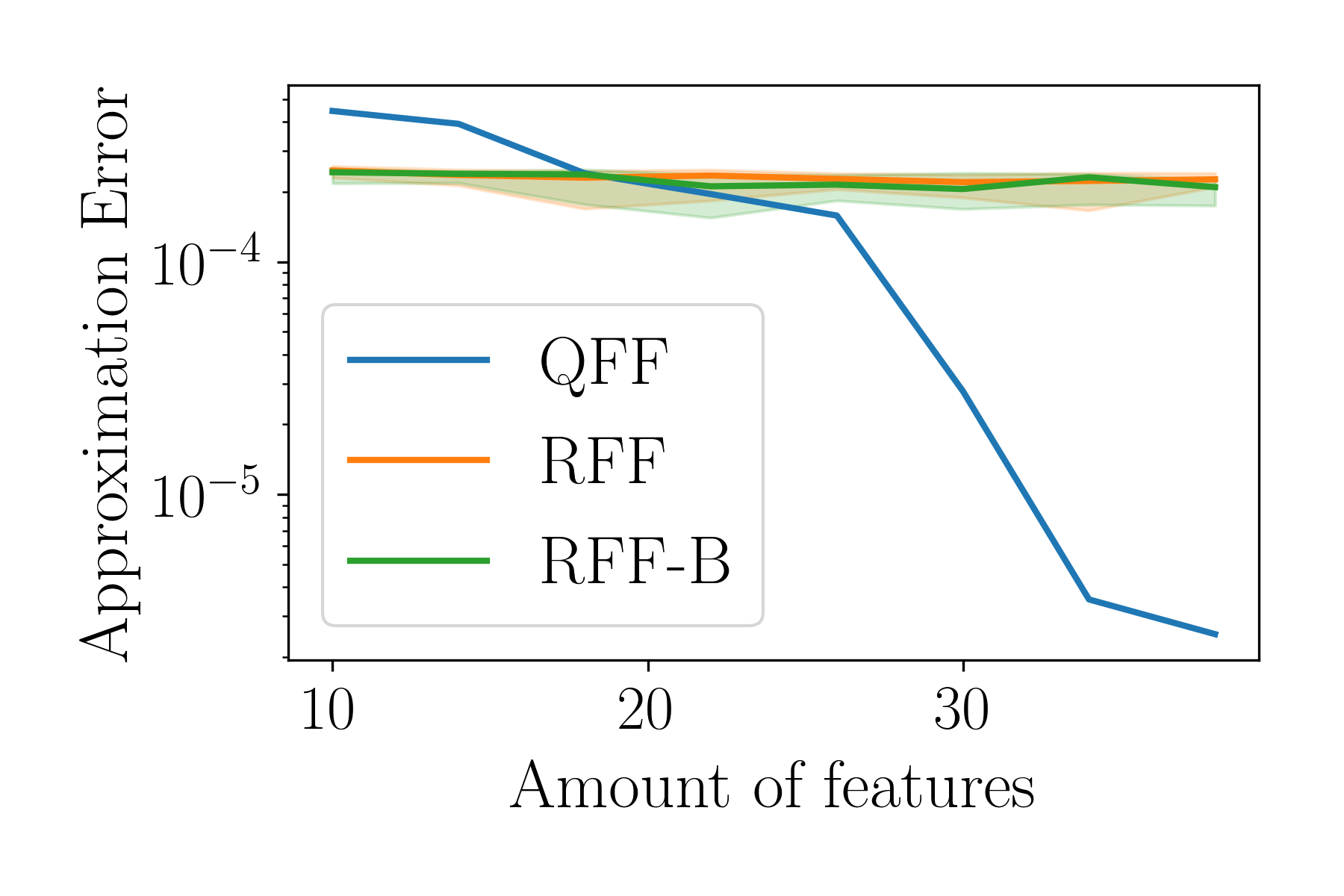}
		\caption{$\Sigma'_1$}
	\end{subfigure}
	\caption{Approximation error of the different feature approximations compared to the accurate GP, evaluated at $t=1.75$ for the Lotka Volterra system with 1000 observations and $\sigma^2=0.1$. For each feature, we show the median as well as the 12.5\% and 87.5\% quantiles over 10 independent noise realizations, separately for each state dimension.}
\end{figure}

\begin{figure}[!h]
	\centering
	\begin{subfigure}[t]{0.24\textwidth}
		\centering
		\includegraphics[width=\textwidth]{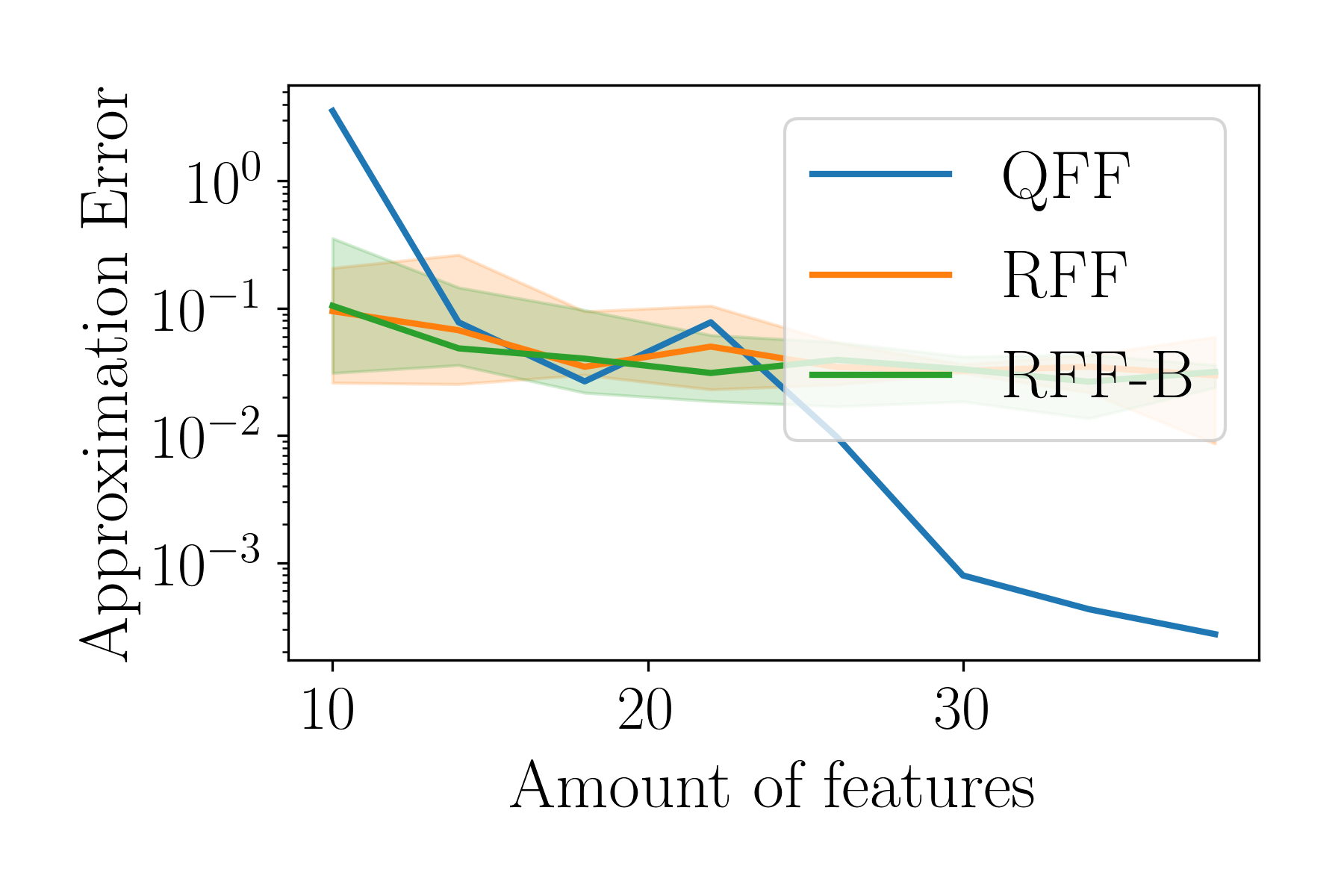}
		\caption{$\mu_0$}
	\end{subfigure}
	\hfill
	\begin{subfigure}[t]{0.24\textwidth}
		\centering
		\includegraphics[width=\textwidth]{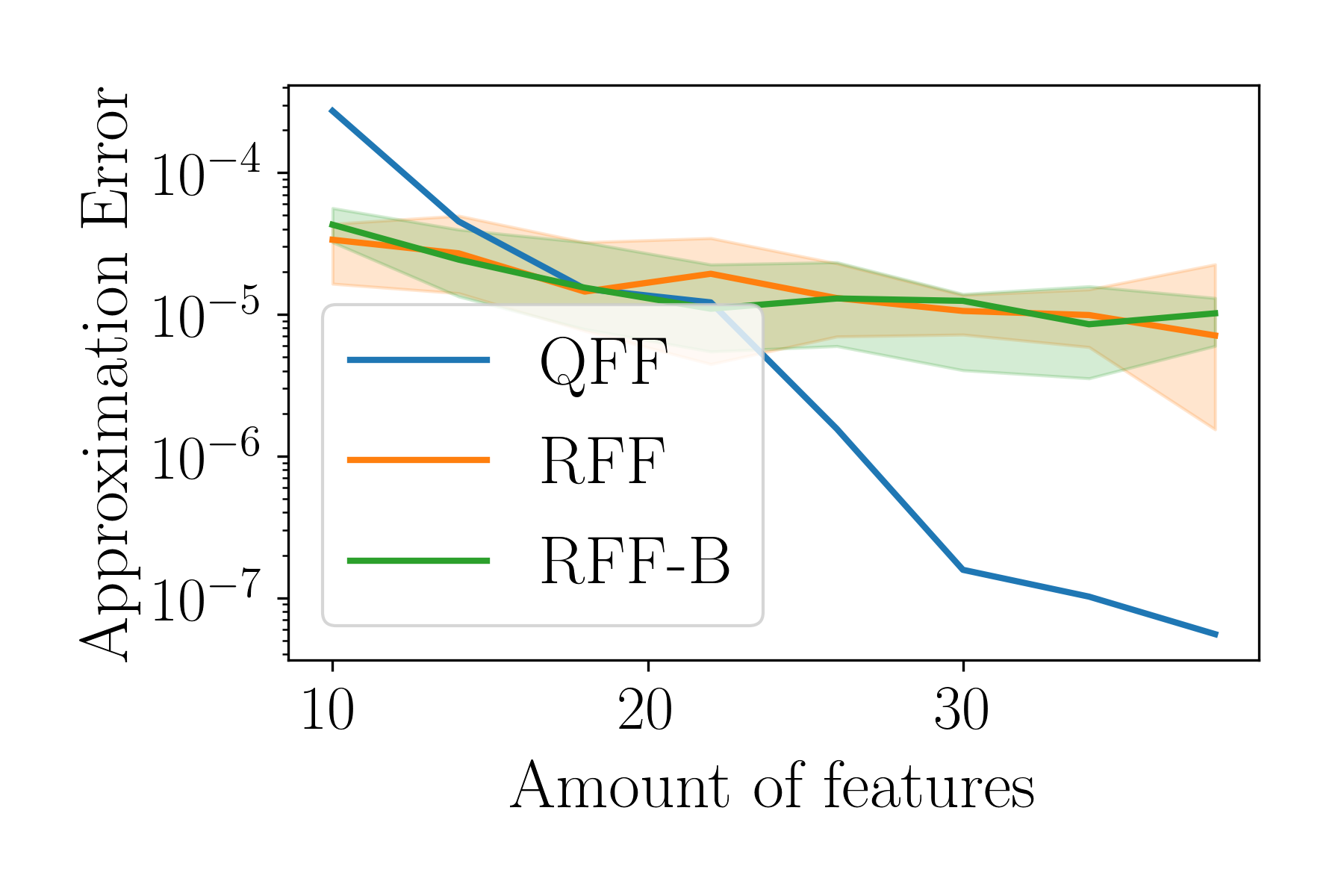}
		\caption{$\Sigma_0$}
	\end{subfigure}
	\hfill
	\begin{subfigure}[t]{0.24\textwidth}
		\centering
		\includegraphics[width=\textwidth]{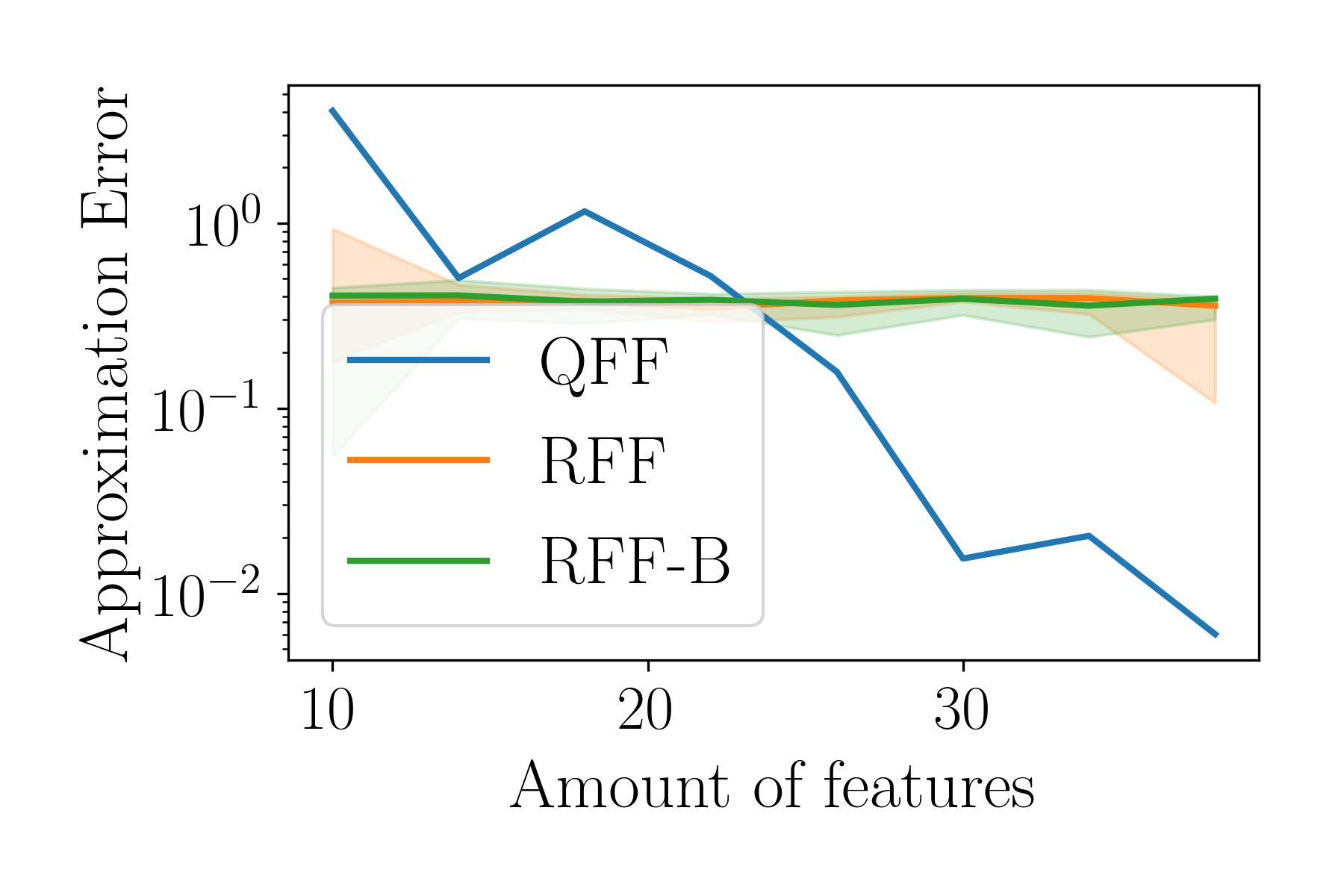}
		\caption{$\mu'_0$}
	\end{subfigure}
	\hfill
	\begin{subfigure}[t]{0.24\textwidth}
		\centering
		\includegraphics[width=\textwidth]{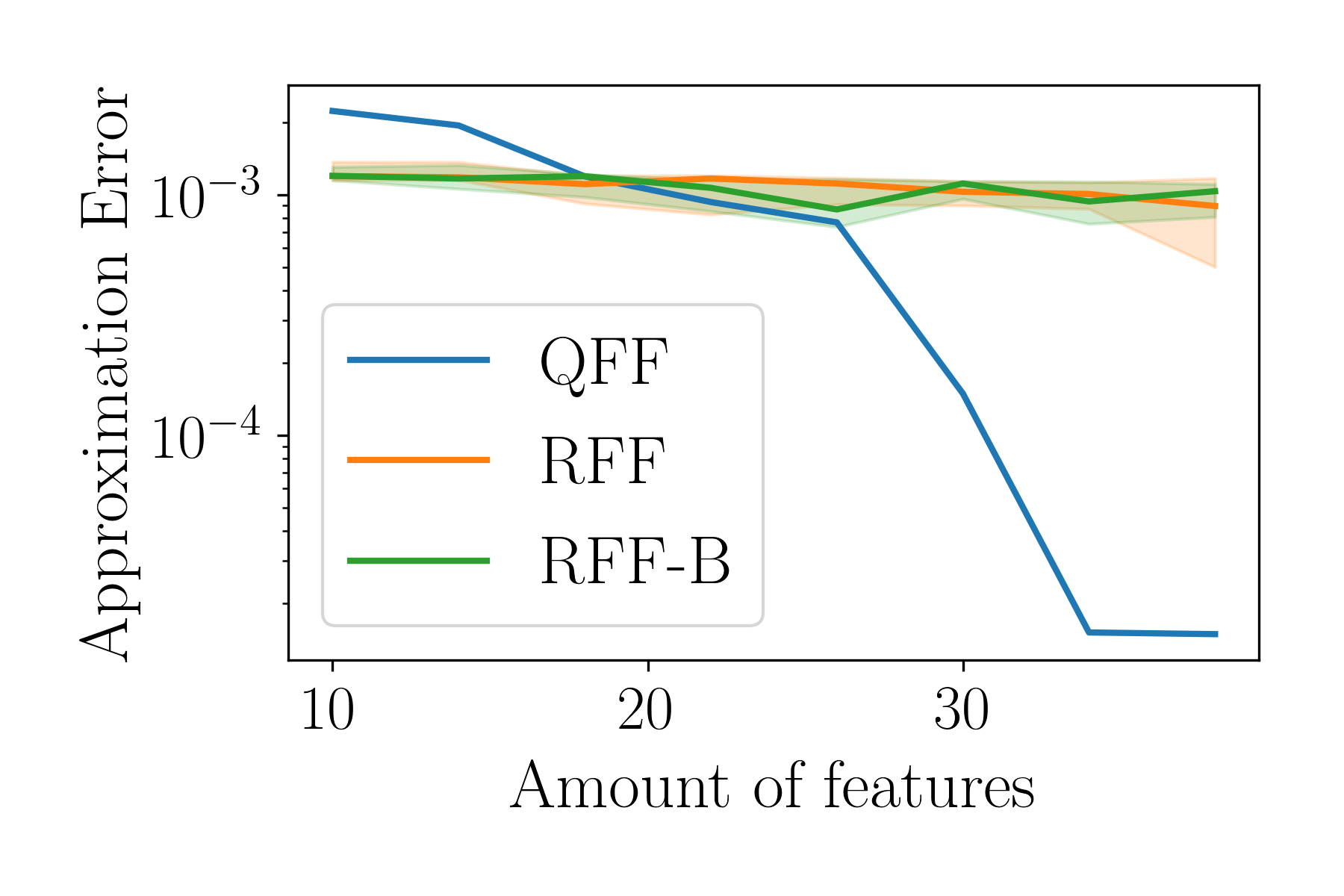}
		\caption{$\Sigma'_0$}
	\end{subfigure}\\
	\begin{subfigure}[t]{0.24\textwidth}
		\centering
		\includegraphics[width=\textwidth]{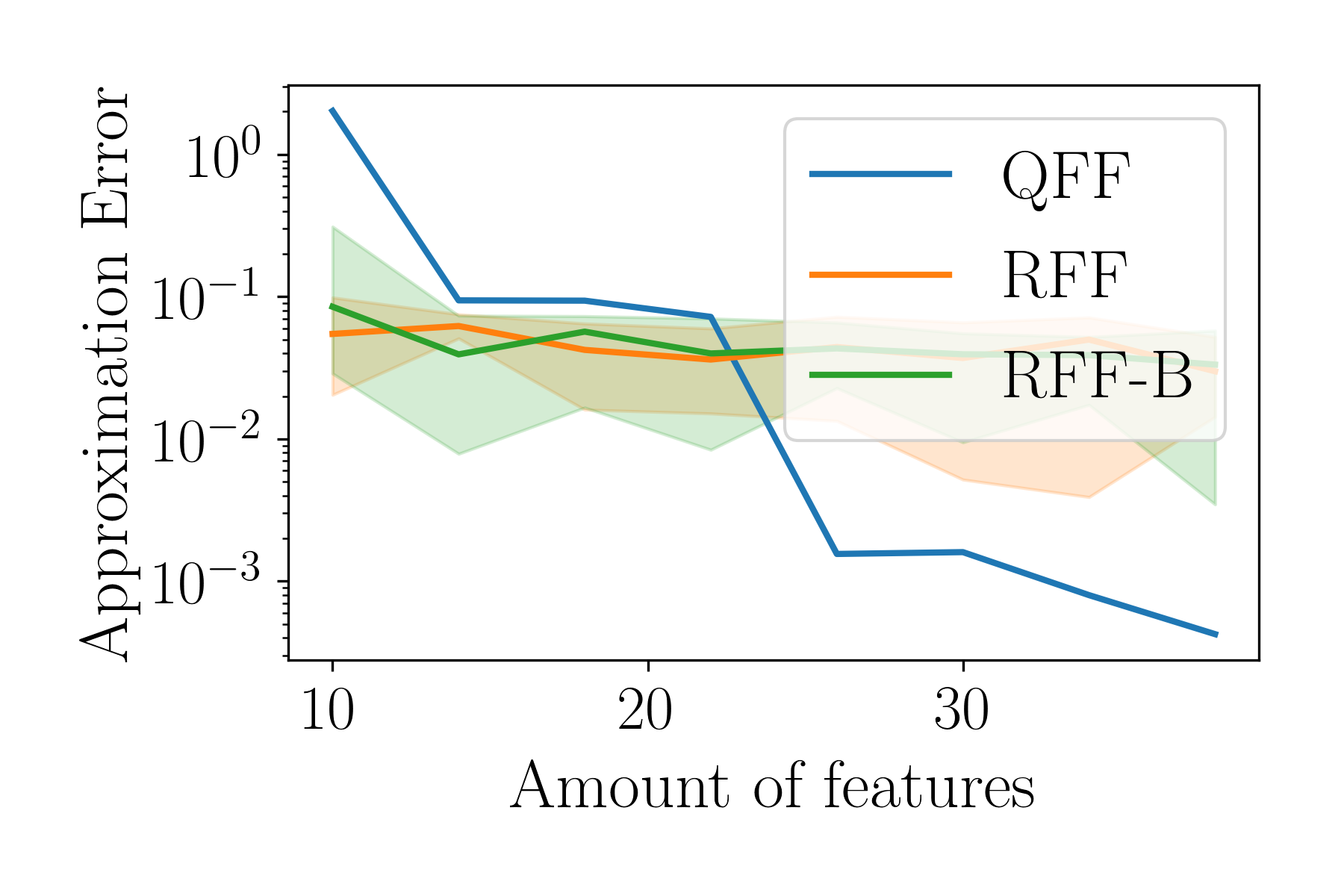}
		\caption{$\mu_1$}
	\end{subfigure}
	\hfill
	\begin{subfigure}[t]{0.24\textwidth}
		\centering
		\includegraphics[width=\textwidth]{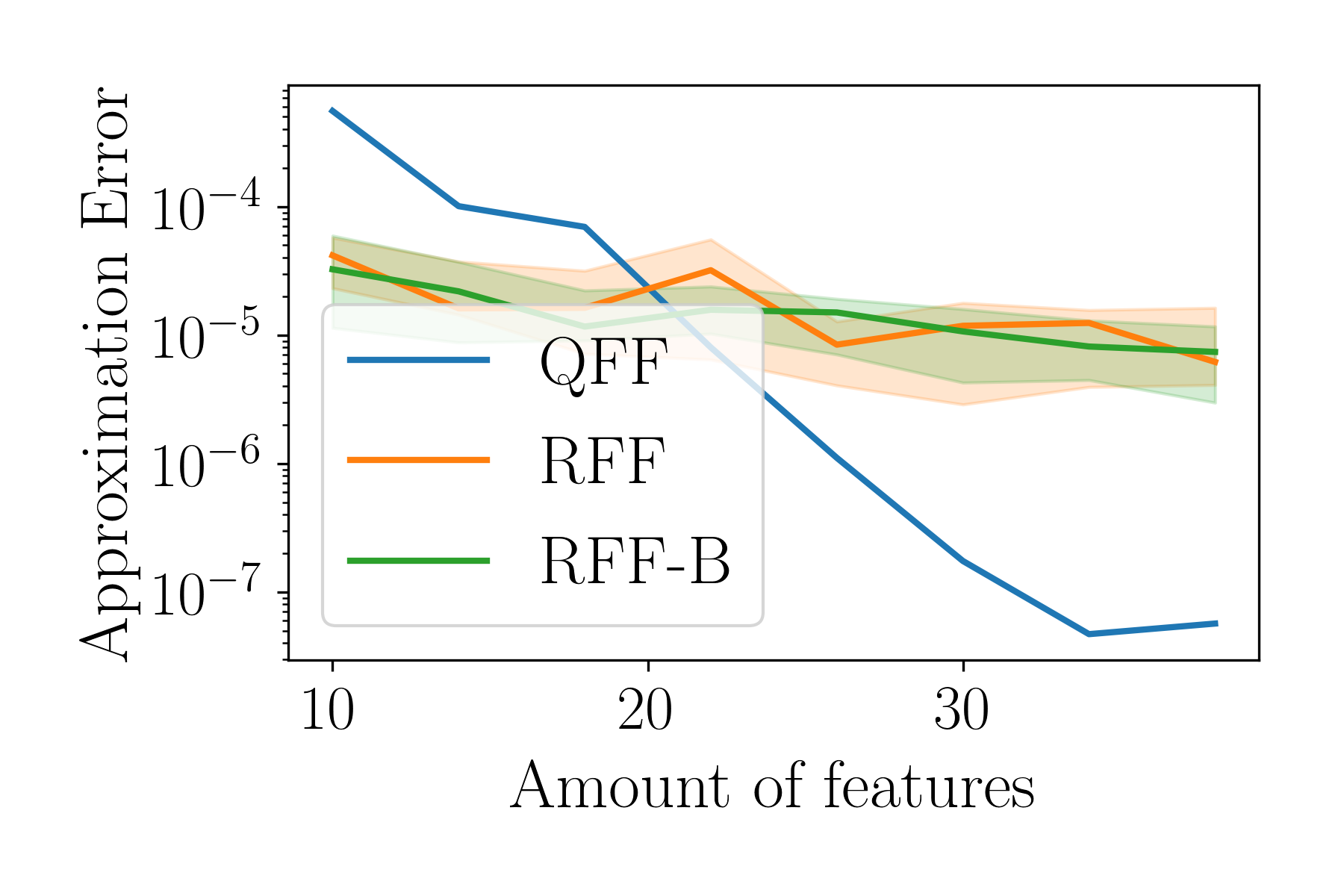}
		\caption{$\Sigma_1$}
	\end{subfigure}
	\hfill
	\begin{subfigure}[t]{0.24\textwidth}
		\centering
		\includegraphics[width=\textwidth]{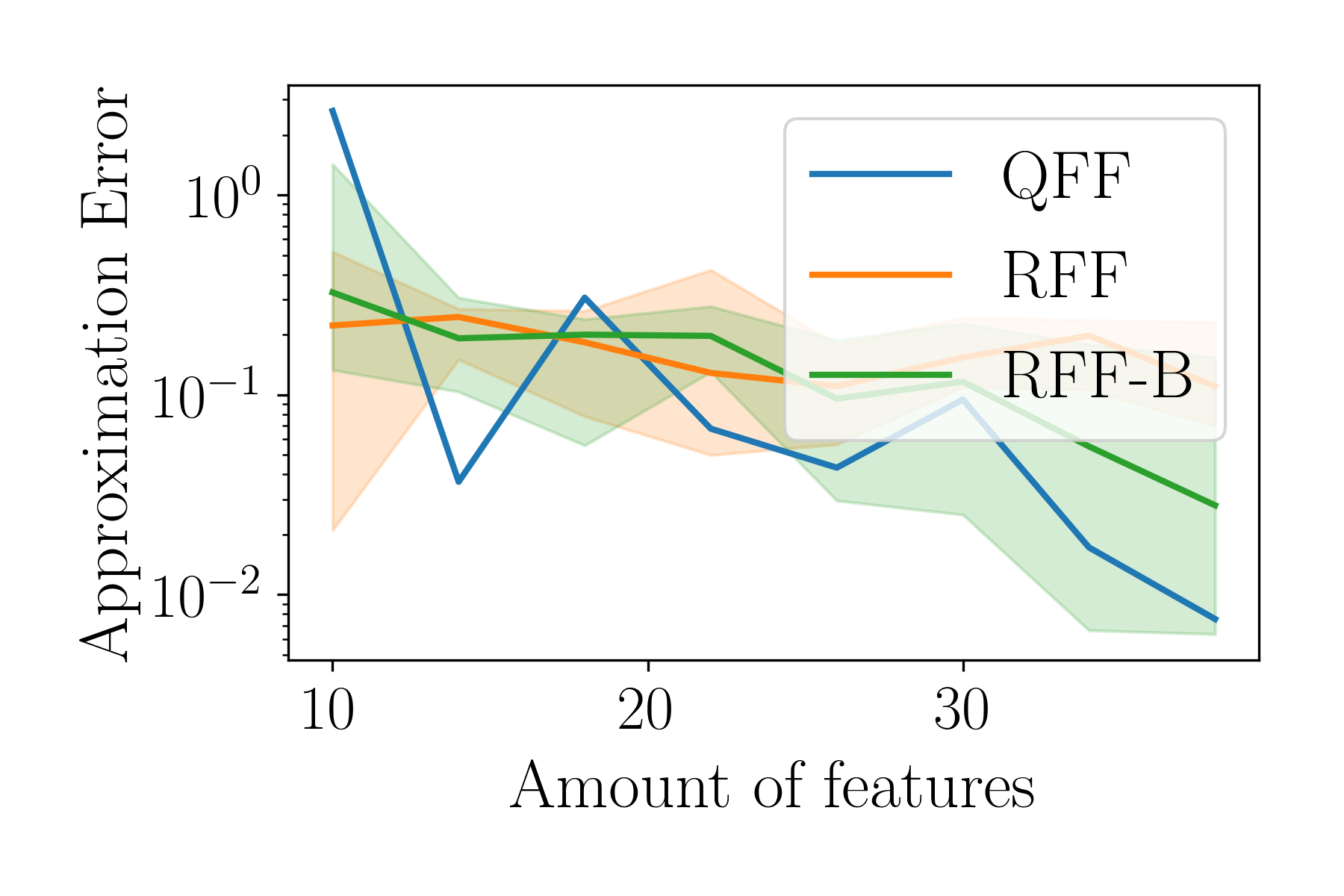}
		\caption{$\mu'_1$}
	\end{subfigure}
	\hfill
	\begin{subfigure}[t]{0.24\textwidth}
		\centering
		\includegraphics[width=\textwidth]{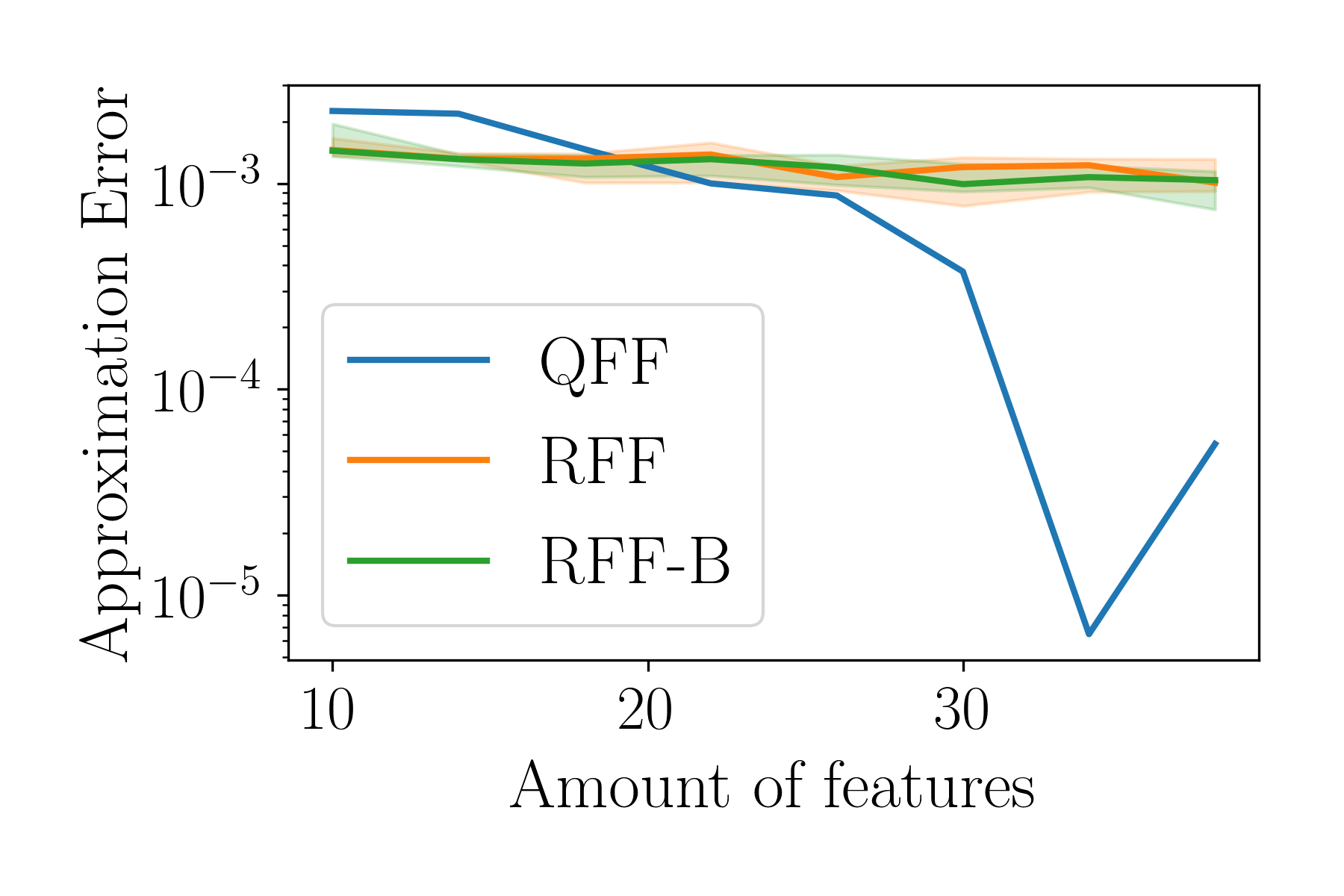}
		\caption{$\Sigma'_1$}
	\end{subfigure}
	\caption{Approximation error of the different feature approximations compared to the accurate GP, evaluated at $t=1.75$ for the Lotka Volterra system with 1000 observations and $\sigma^2=0.5$. For each feature, we show the median as well as the 12.5\% and 87.5\% quantiles over 10 independent noise realizations, separately for each state dimension.}
\end{figure}

\newpage

\subsection{Protein Transduction}
\begin{figure}[!h]
	\centering
	\begin{subfigure}[t]{0.24\textwidth}
		\centering
		\includegraphics[width=\textwidth]{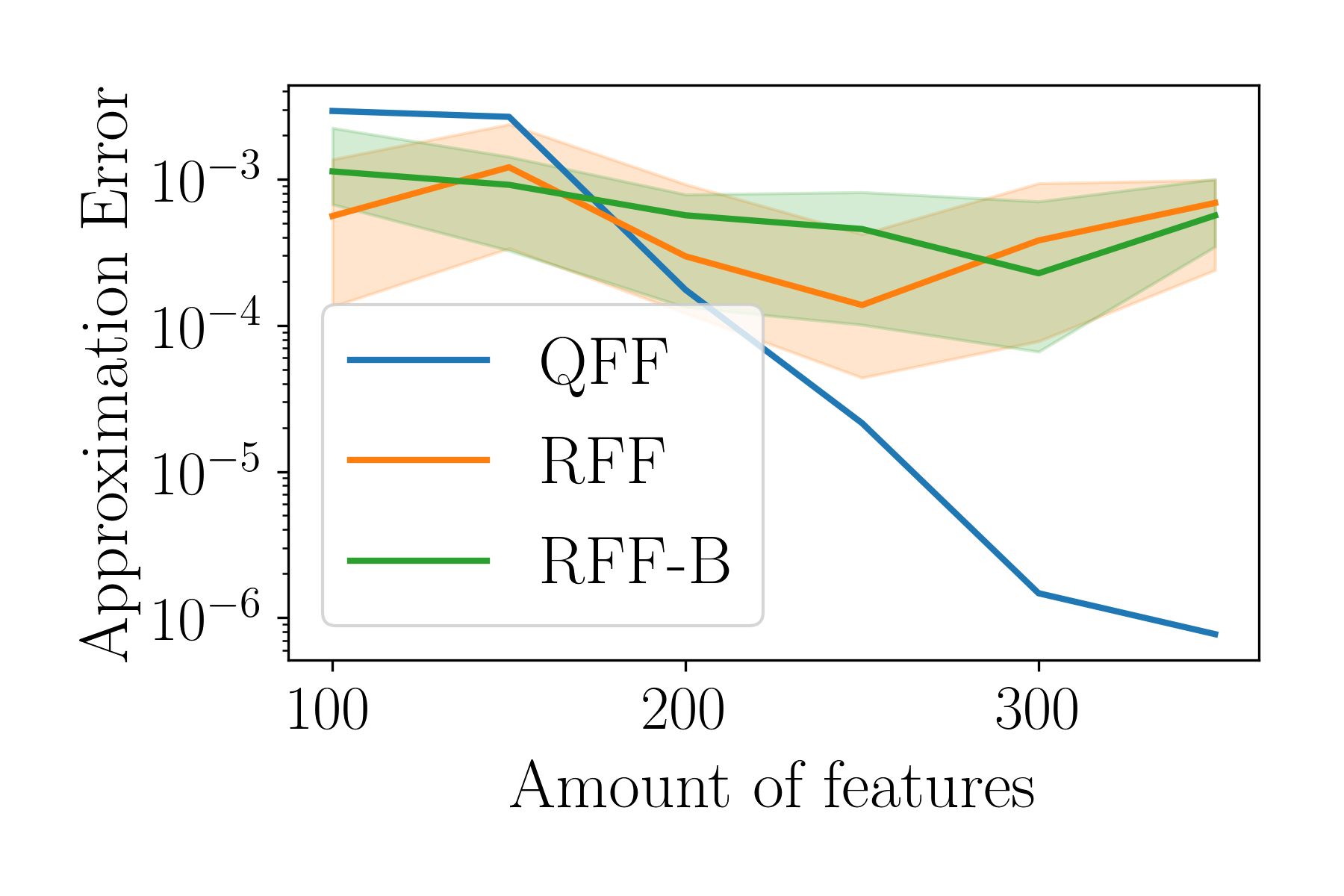}
		\caption{$\mu_1$}
	\end{subfigure}
	\hfill
	\begin{subfigure}[t]{0.24\textwidth}
		\centering
		\includegraphics[width=\textwidth]{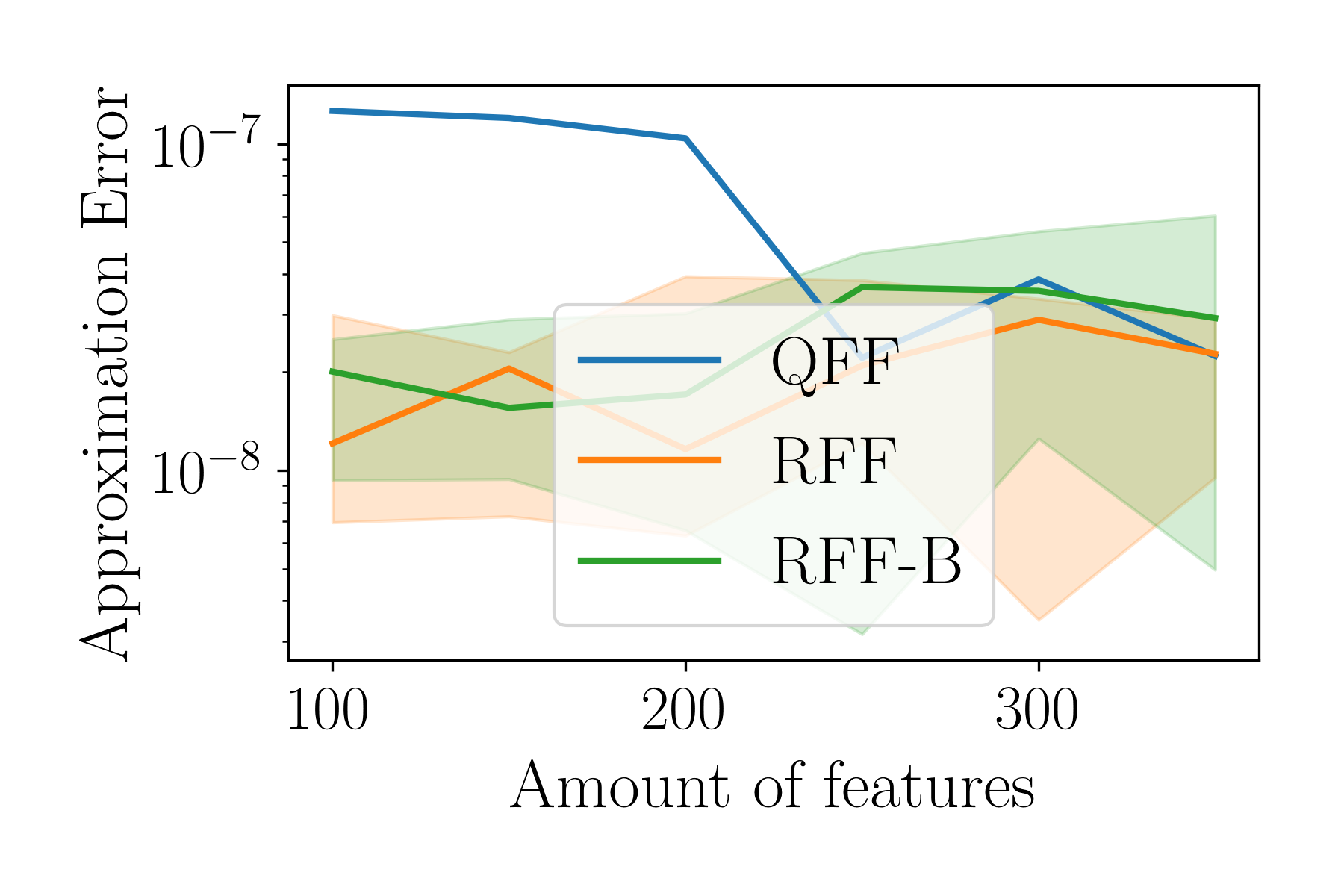}
		\caption{$\Sigma_1$}
	\end{subfigure}
	\hfill
	\begin{subfigure}[t]{0.24\textwidth}
		\centering
		\includegraphics[width=\textwidth]{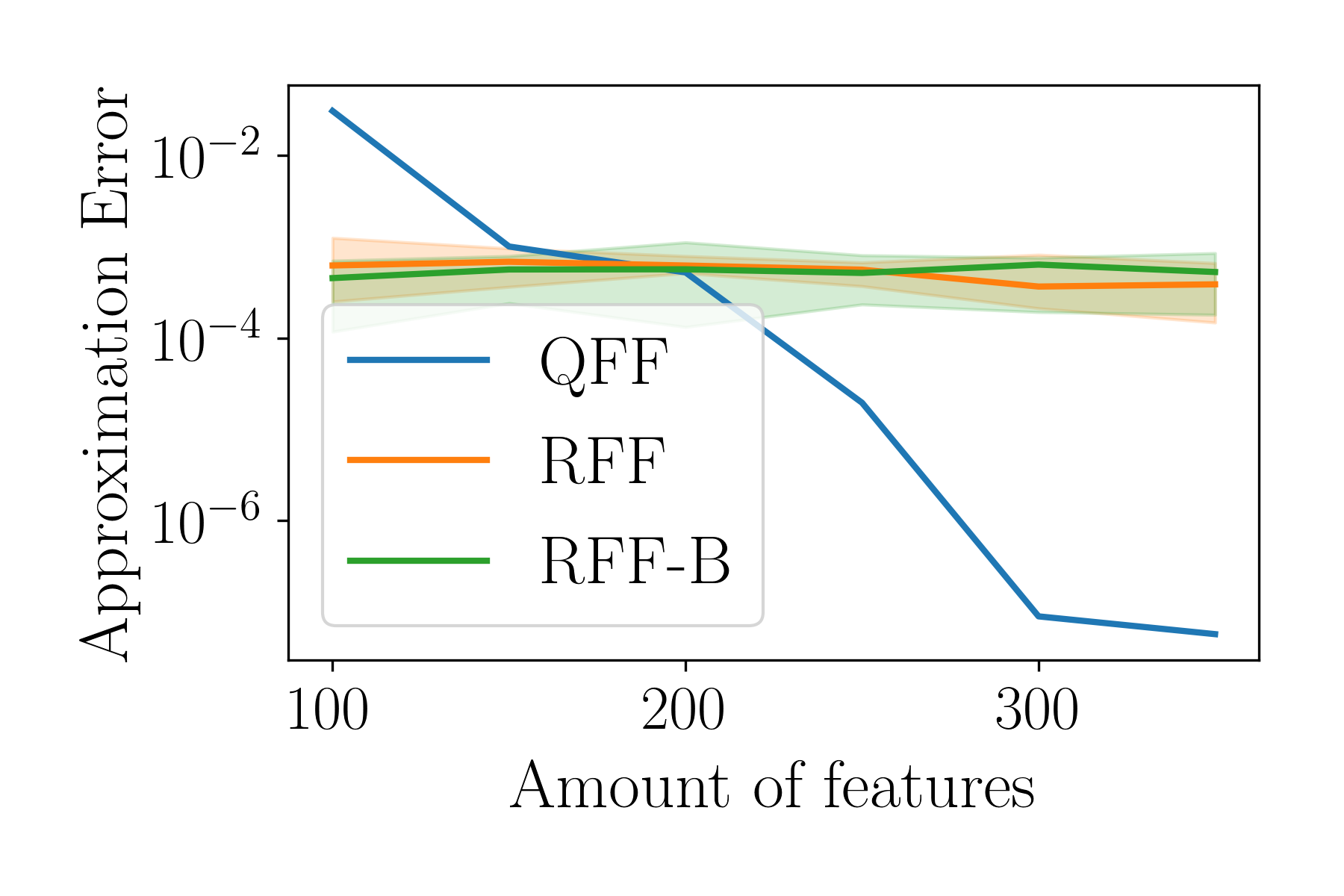}
		\caption{$\mu'_1$}
	\end{subfigure}
	\hfill
	\begin{subfigure}[t]{0.24\textwidth}
		\centering
		\includegraphics[width=\textwidth]{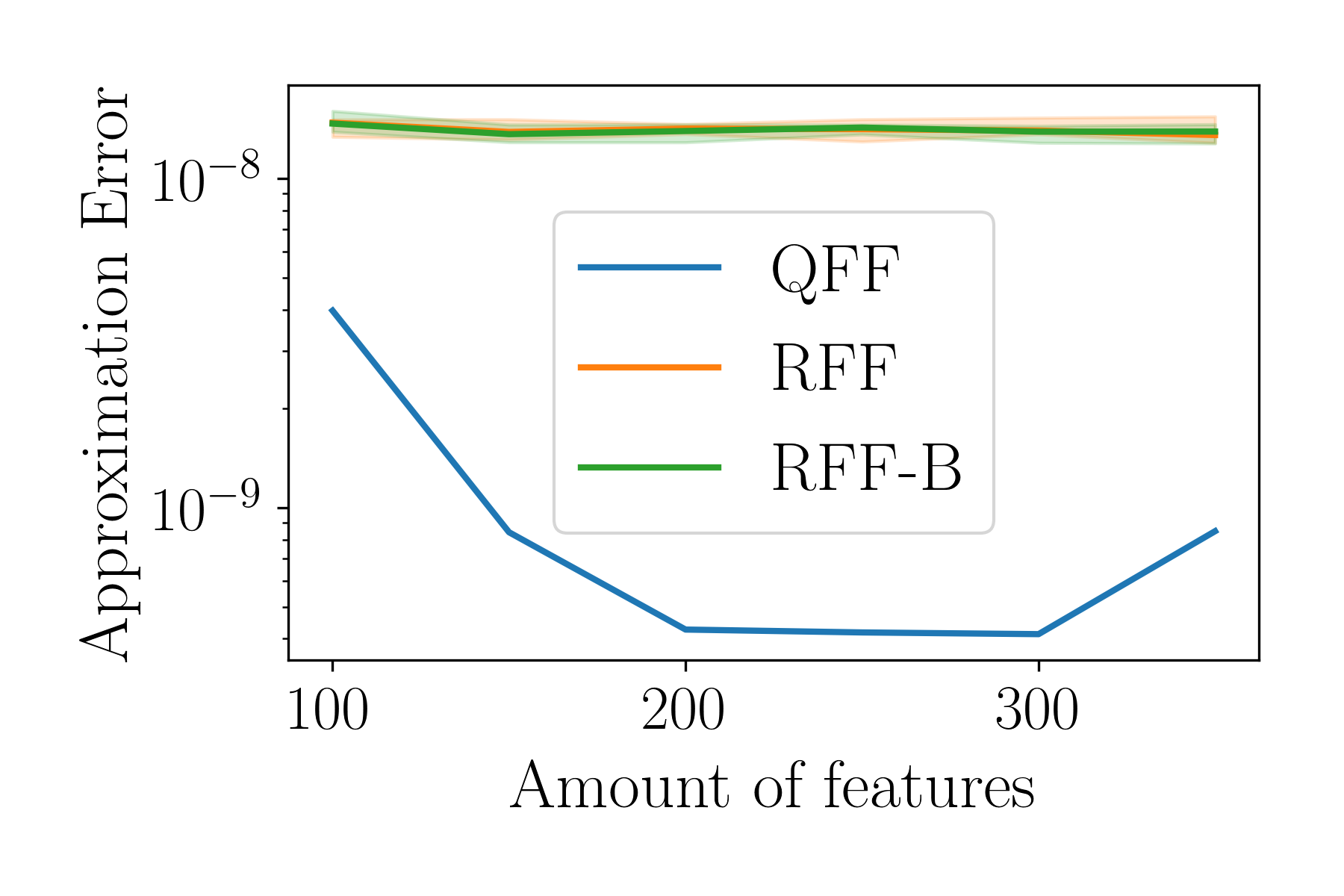}
		\caption{$\Sigma'_1$}
	\end{subfigure}\\
	\begin{subfigure}[t]{0.24\textwidth}
		\centering
		\includegraphics[width=\textwidth]{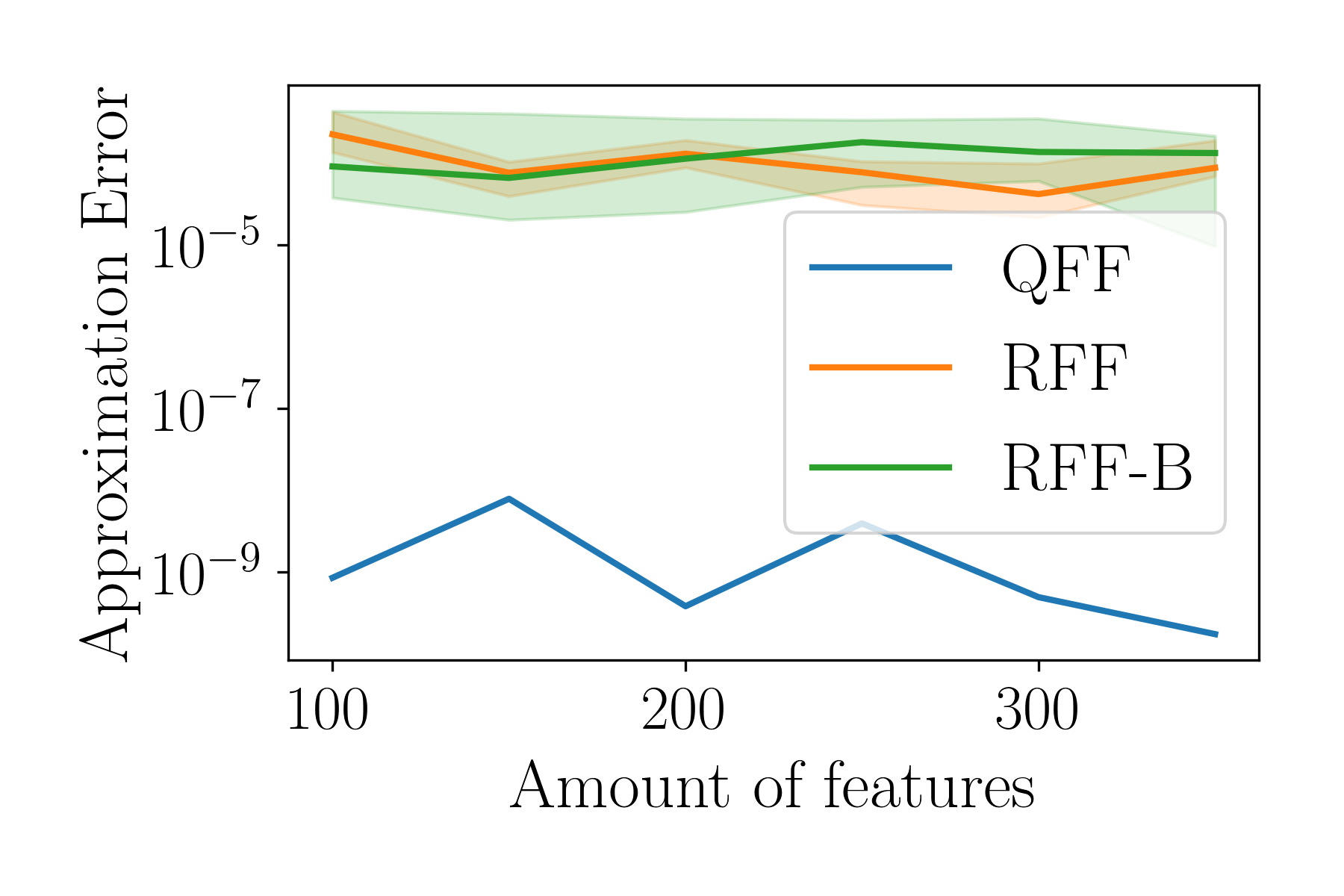}
		\caption{$\mu_2$}
	\end{subfigure}
	\hfill
	\begin{subfigure}[t]{0.24\textwidth}
		\centering
		\includegraphics[width=\textwidth]{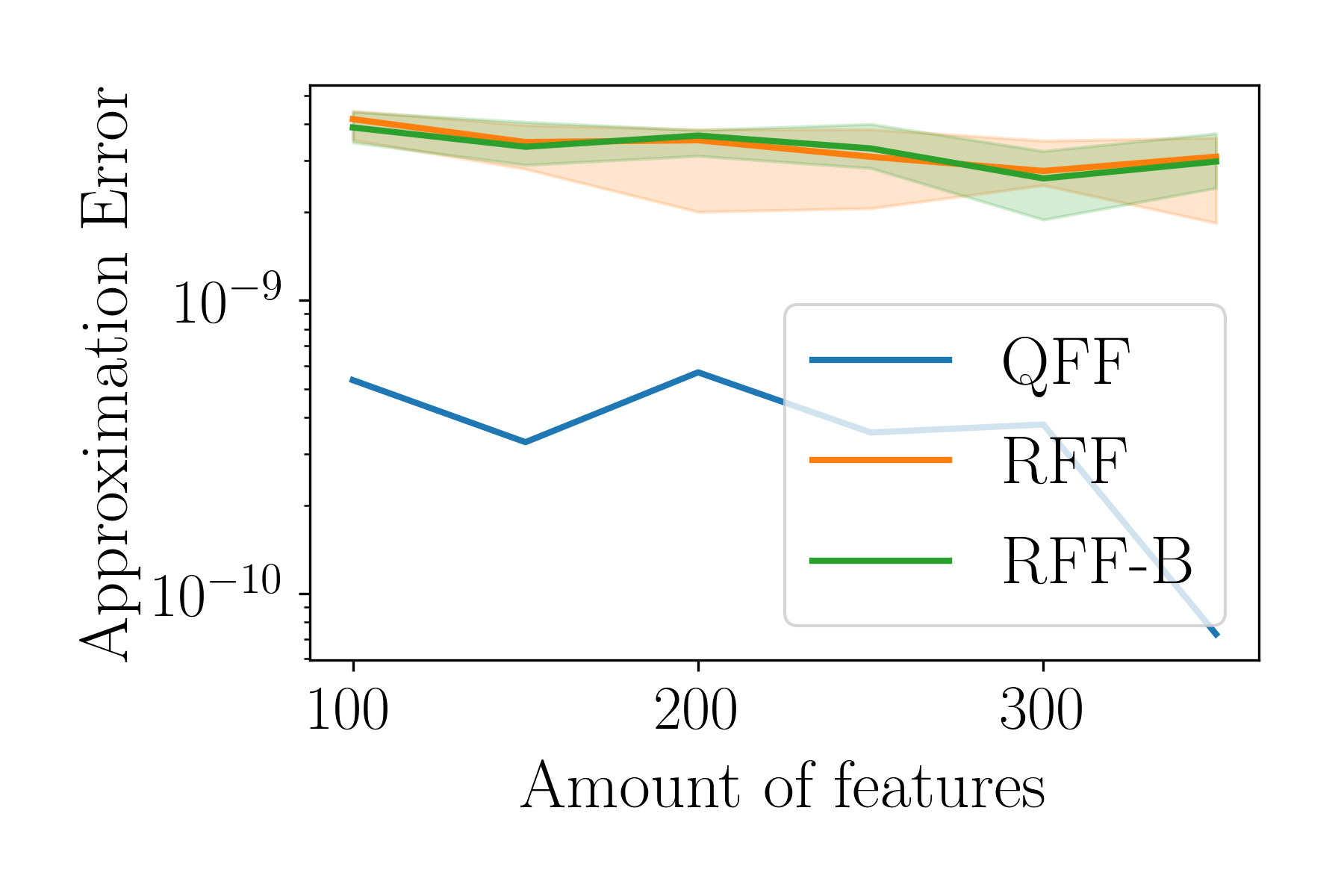}
		\caption{$\Sigma_2$}
	\end{subfigure}
	\hfill
	\begin{subfigure}[t]{0.24\textwidth}
		\centering
		\includegraphics[width=\textwidth]{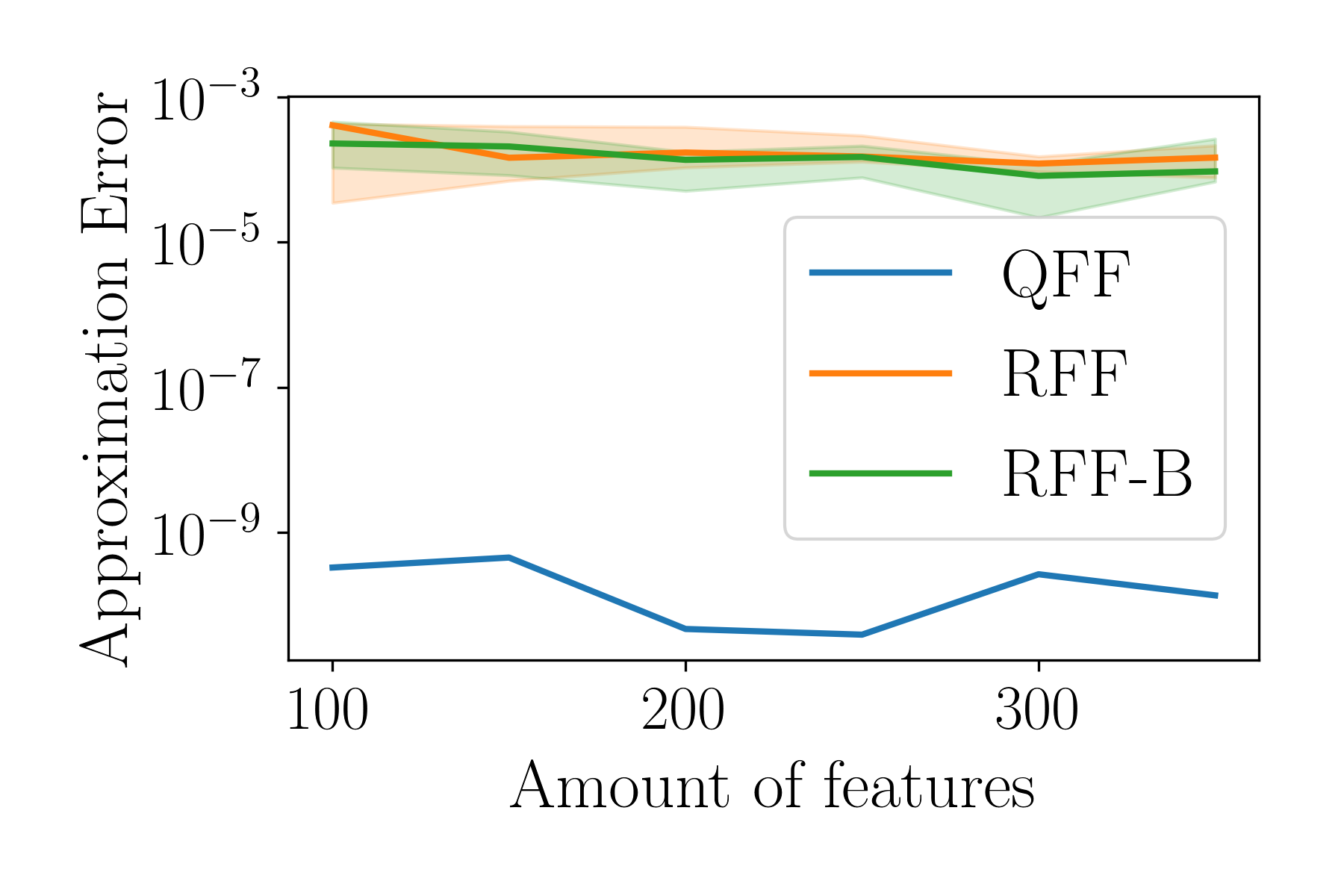}
		\caption{$\mu'_2$}
	\end{subfigure}
	\hfill
	\begin{subfigure}[t]{0.24\textwidth}
		\centering
		\includegraphics[width=\textwidth]{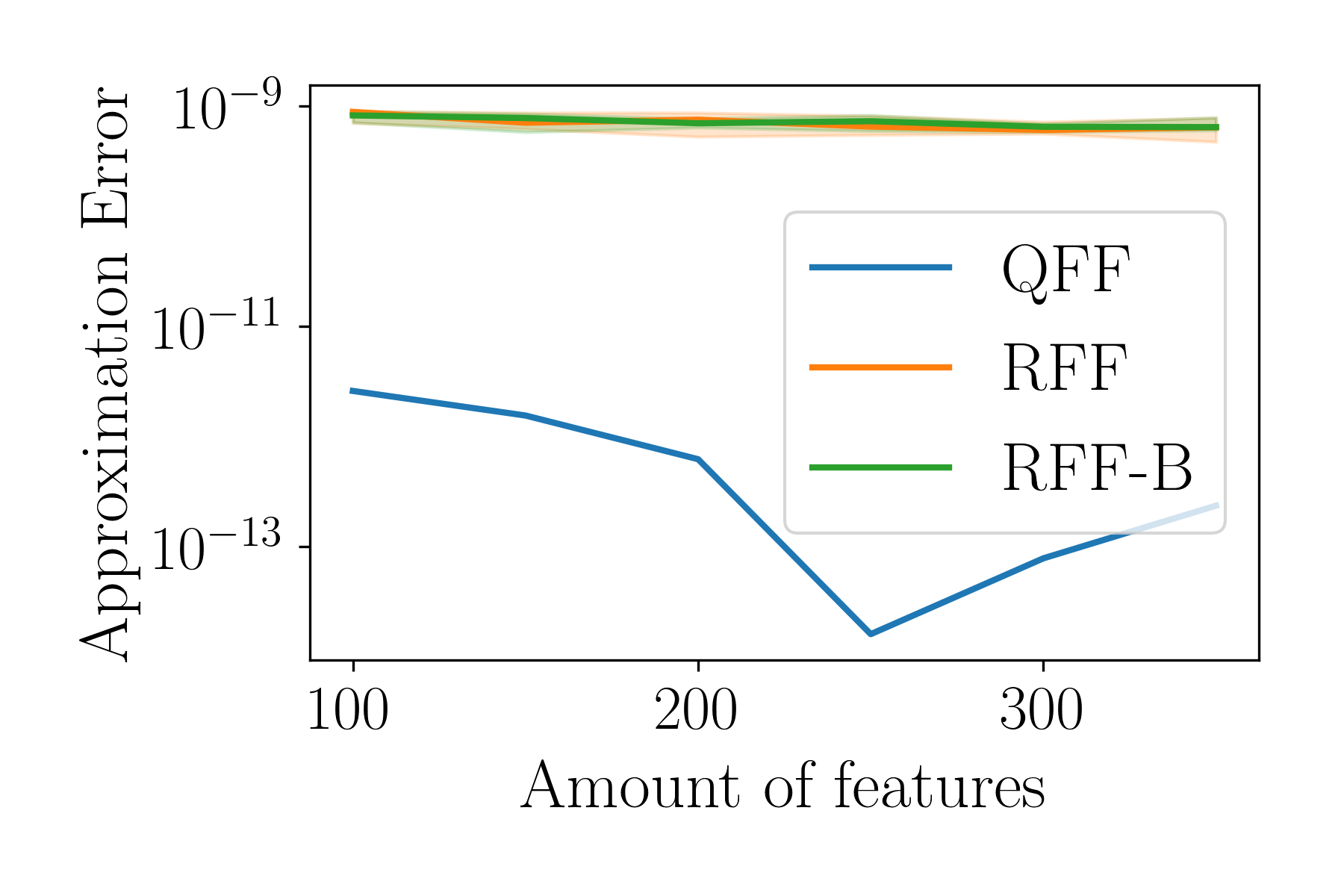}
		\caption{$\Sigma'_2$}
	\end{subfigure}\\
	\begin{subfigure}[t]{0.24\textwidth}
	\centering
	\includegraphics[width=\textwidth]{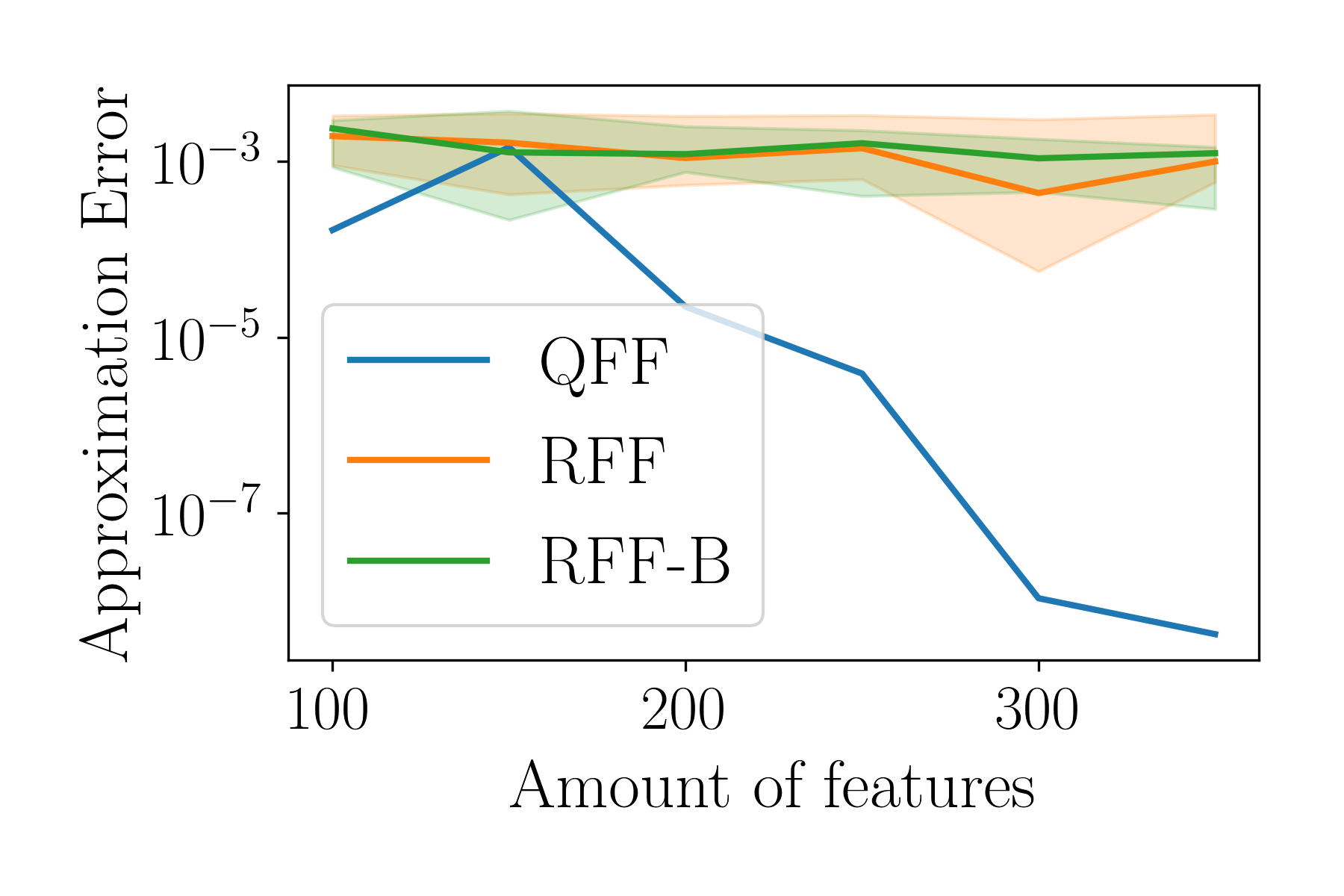}
	\caption{$\mu_3$}
	\end{subfigure}
	\hfill
	\begin{subfigure}[t]{0.24\textwidth}
	\centering
	\includegraphics[width=\textwidth]{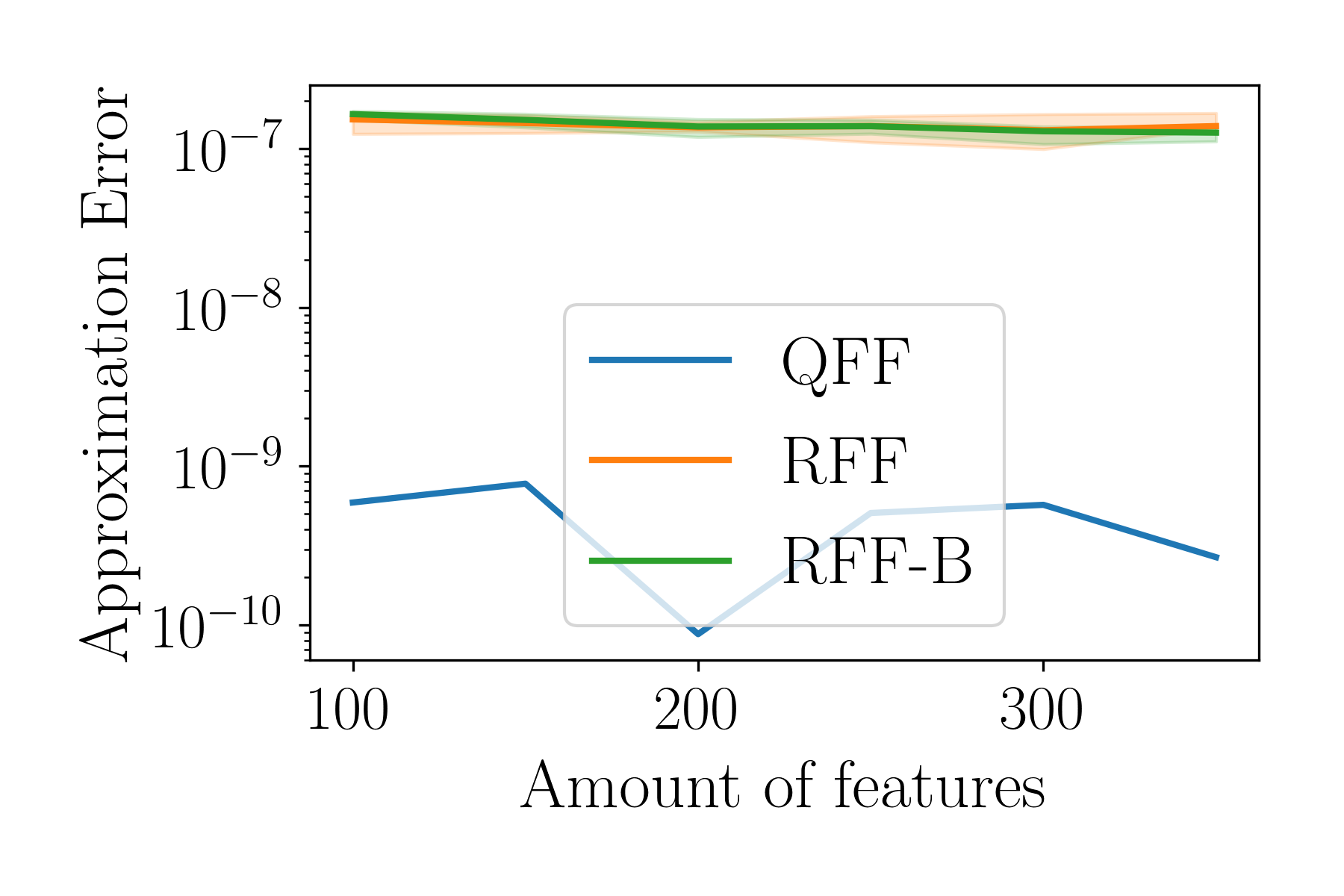}
	\caption{$\Sigma_3$}
	\end{subfigure}
	\hfill
	\begin{subfigure}[t]{0.24\textwidth}
	\centering
	\includegraphics[width=\textwidth]{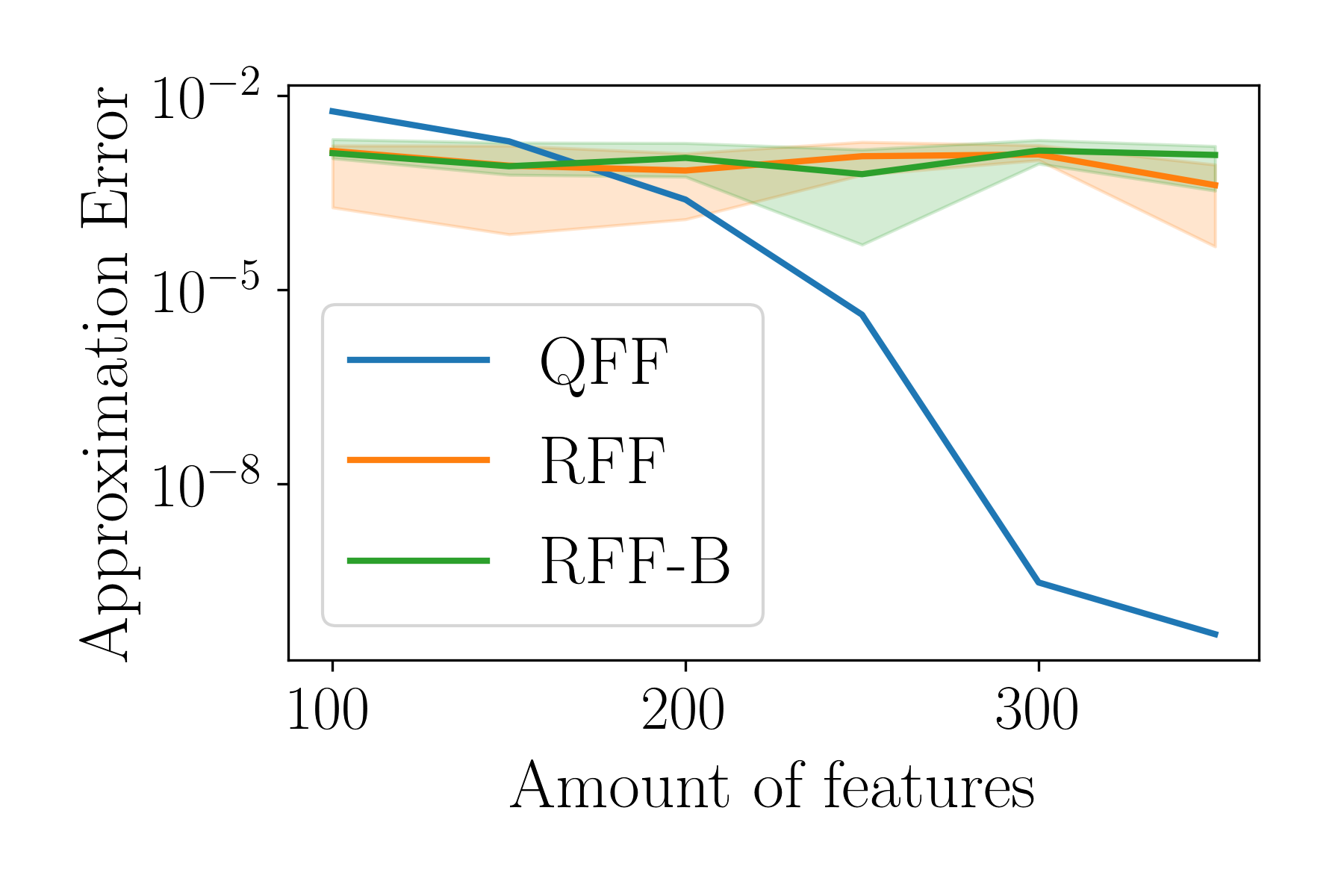}
	\caption{$\mu'3$}
	\end{subfigure}
	\hfill
	\begin{subfigure}[t]{0.24\textwidth}
	\centering
	\includegraphics[width=\textwidth]{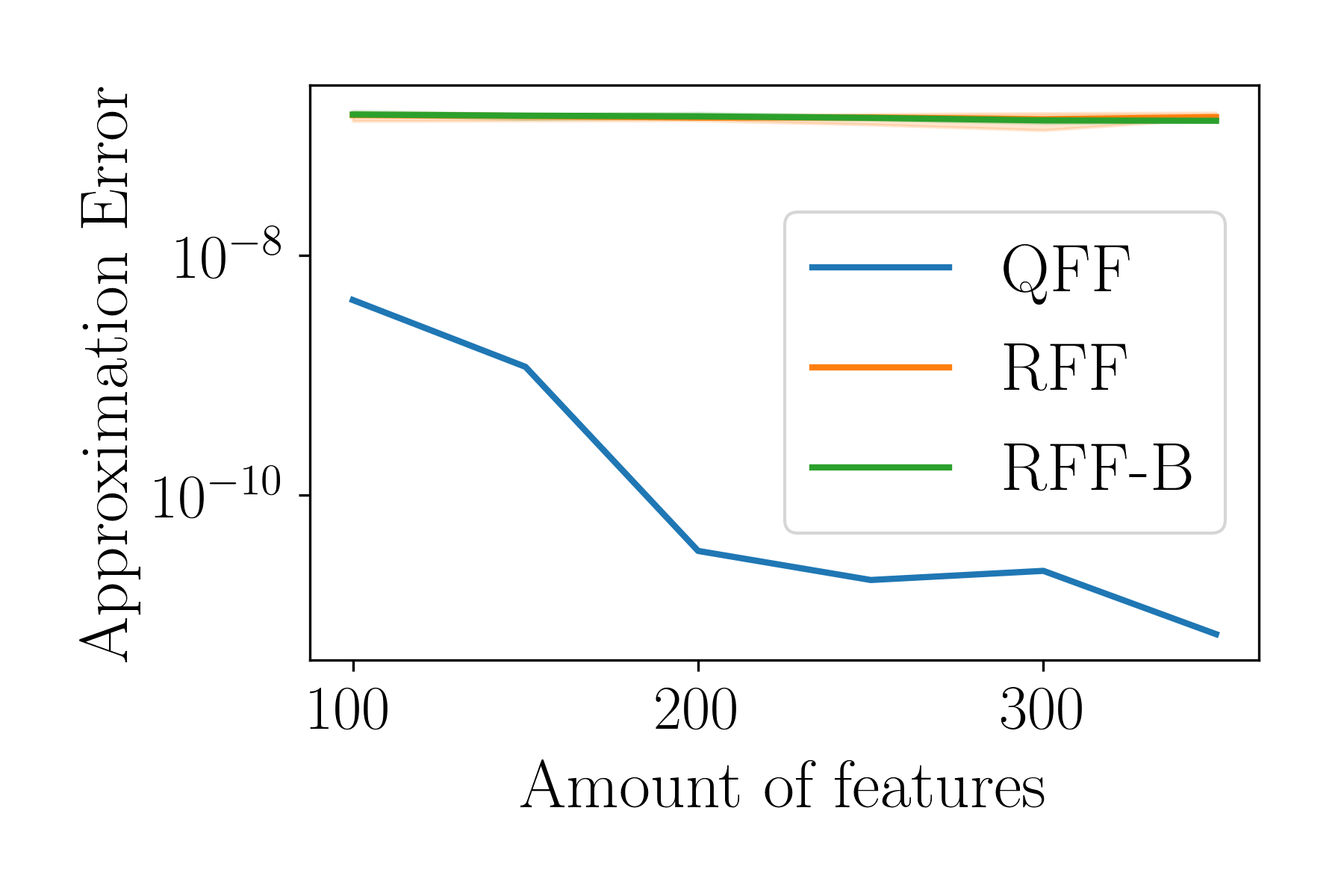}
	\caption{$\Sigma'_3$}
	\end{subfigure}\\
	\begin{subfigure}[t]{0.24\textwidth}
	\centering
	\includegraphics[width=\textwidth]{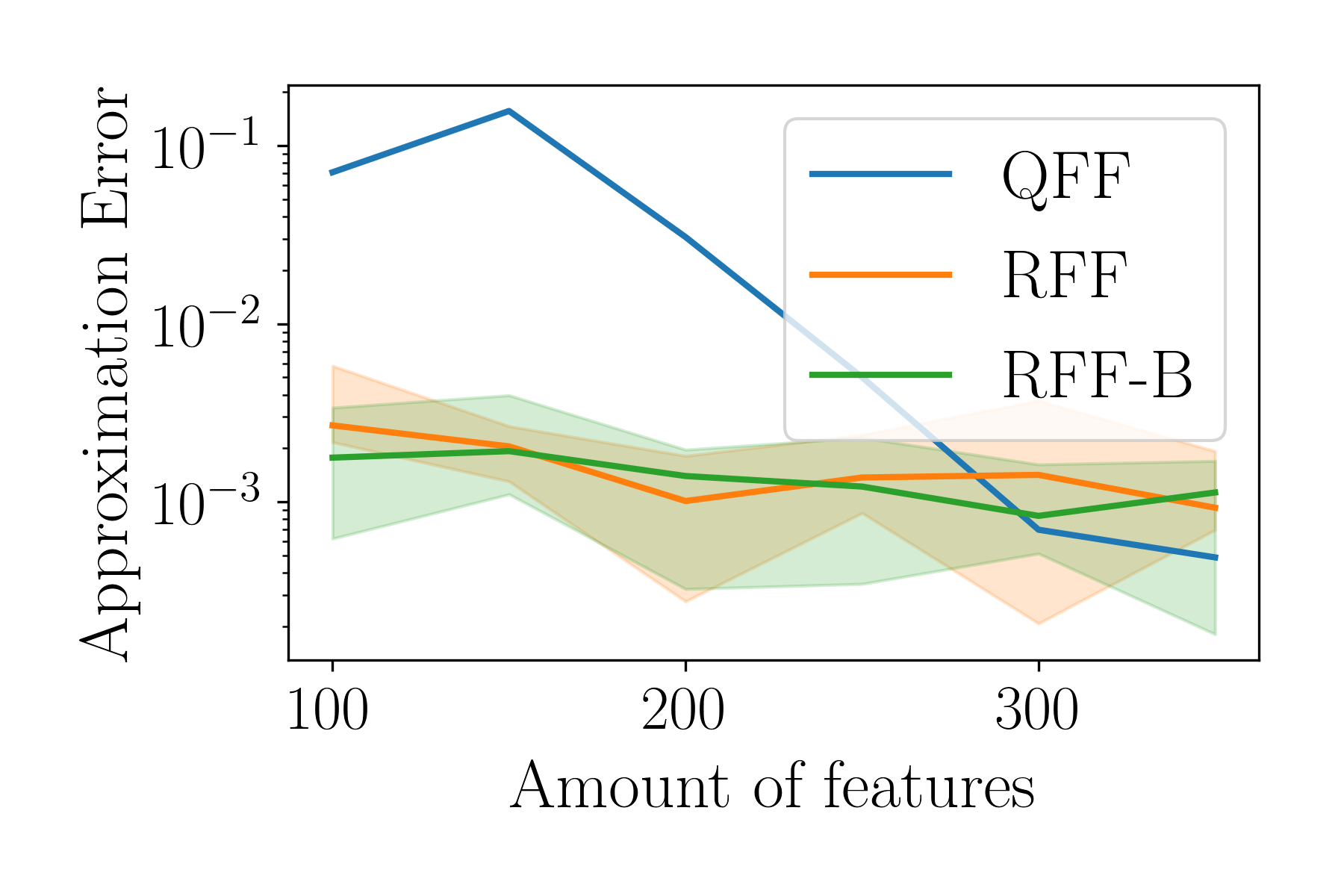}
	\caption{$\mu_4$}
	\end{subfigure}
	\hfill
	\begin{subfigure}[t]{0.24\textwidth}
	\centering
	\includegraphics[width=\textwidth]{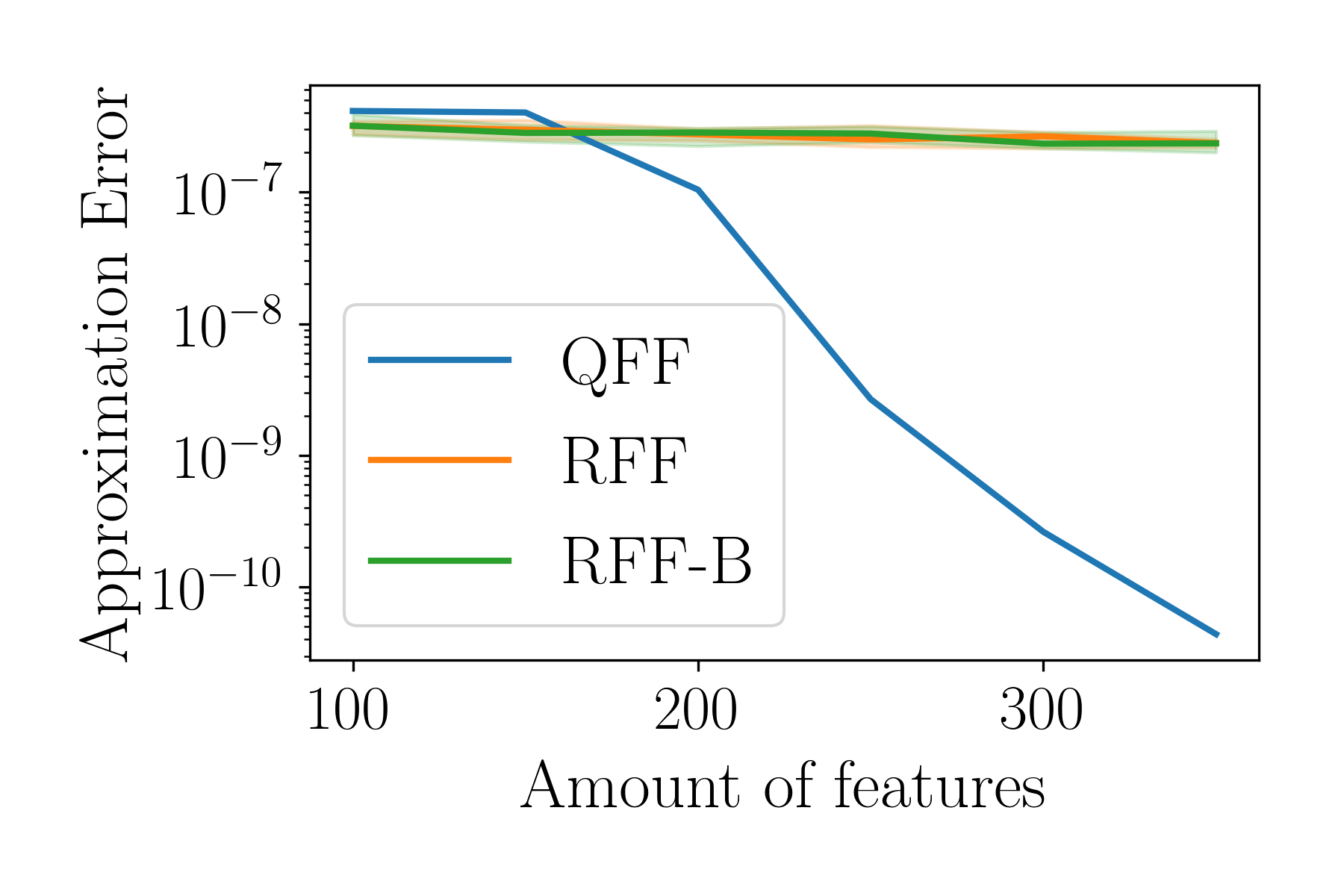}
	\caption{$\Sigma_4$}
	\end{subfigure}
	\hfill
	\begin{subfigure}[t]{0.24\textwidth}
	\centering
	\includegraphics[width=\textwidth]{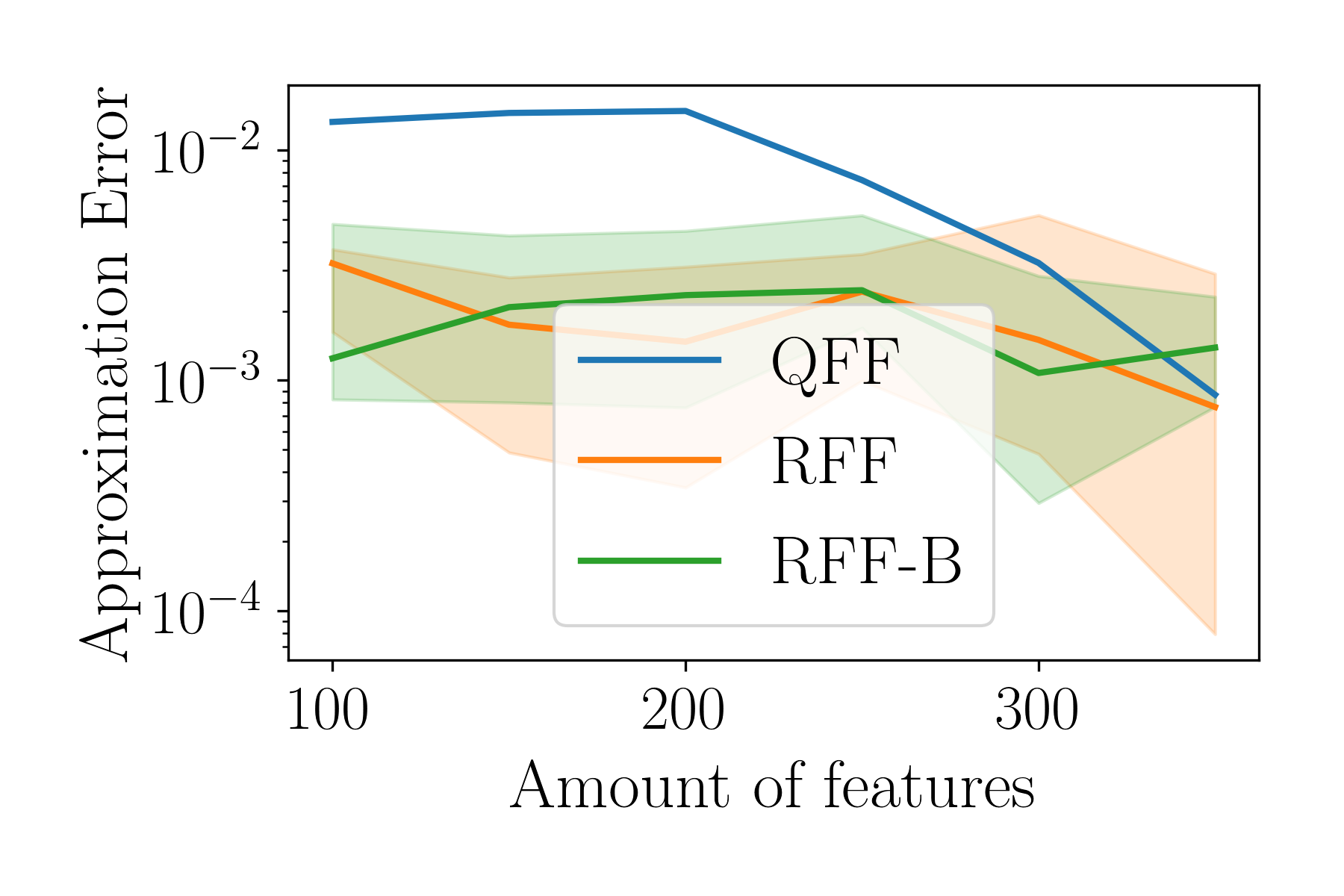}
	\caption{$\mu'_4$}
	\end{subfigure}
	\hfill
	\begin{subfigure}[t]{0.24\textwidth}
	\centering
	\includegraphics[width=\textwidth]{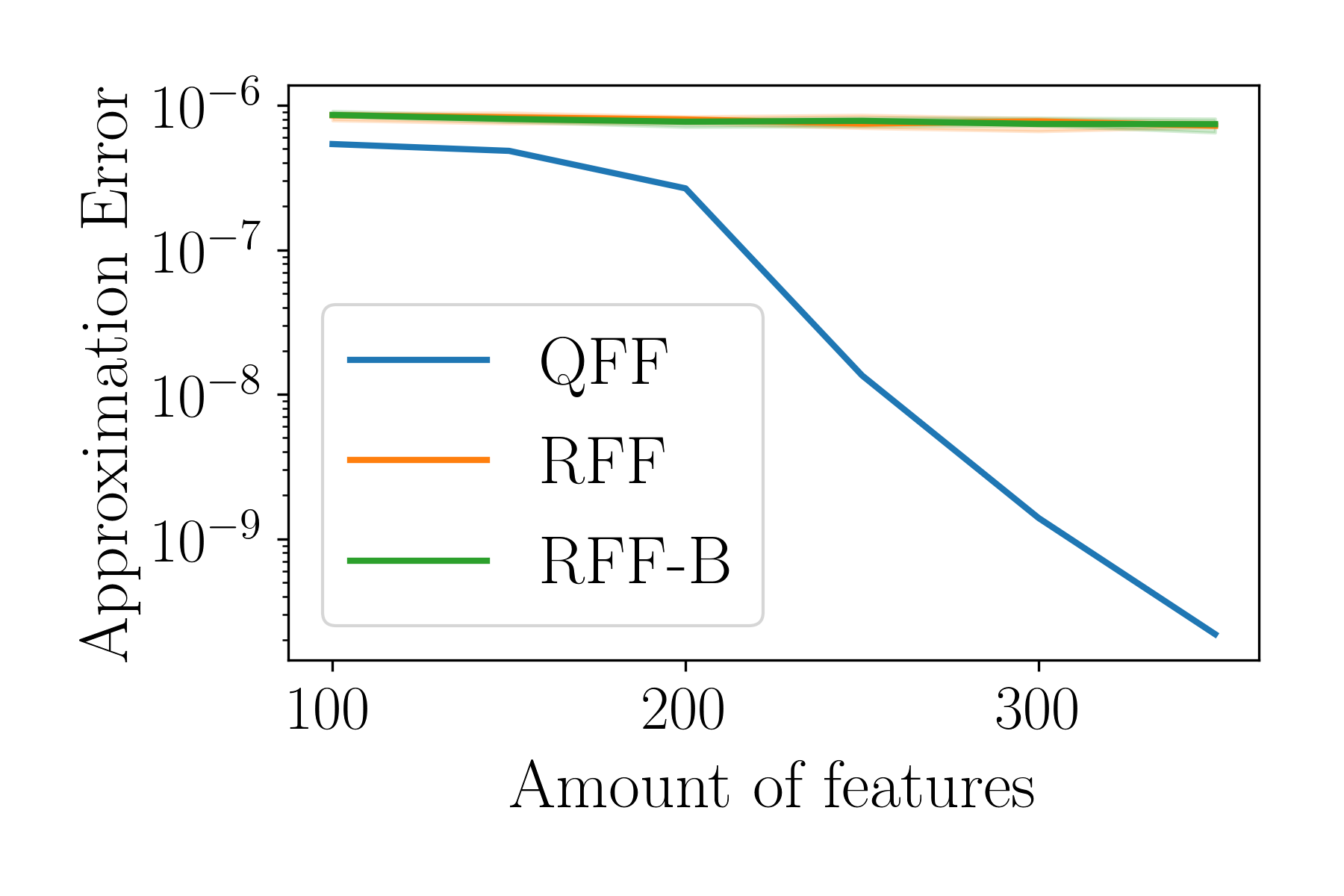}
	\caption{$\Sigma'_4$}
	\end{subfigure}\\
	\begin{subfigure}[t]{0.24\textwidth}
	\centering
	\includegraphics[width=\textwidth]{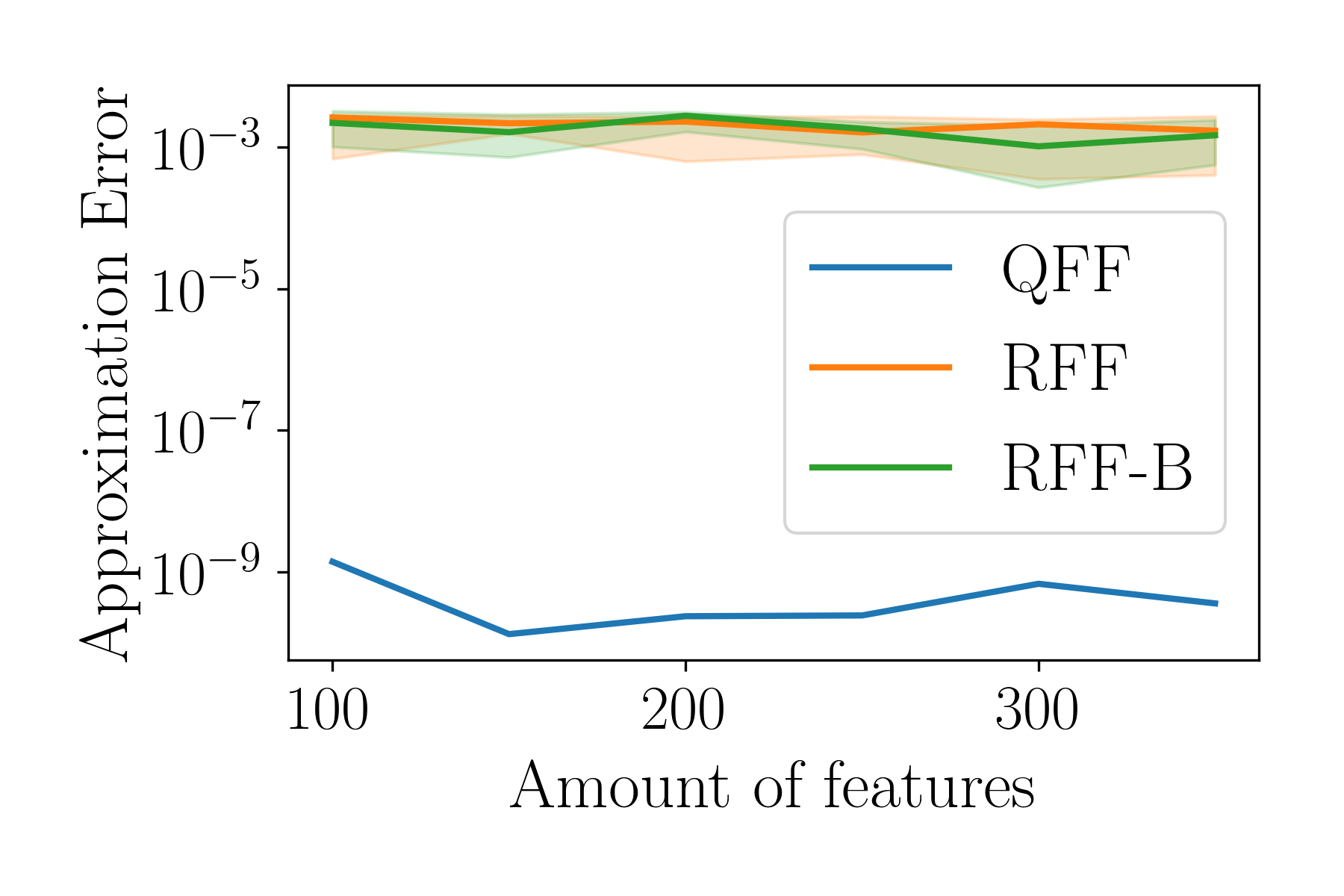}
	\caption{$\mu_5$}
	\end{subfigure}
	\hfill
	\begin{subfigure}[t]{0.24\textwidth}
	\centering
	\includegraphics[width=\textwidth]{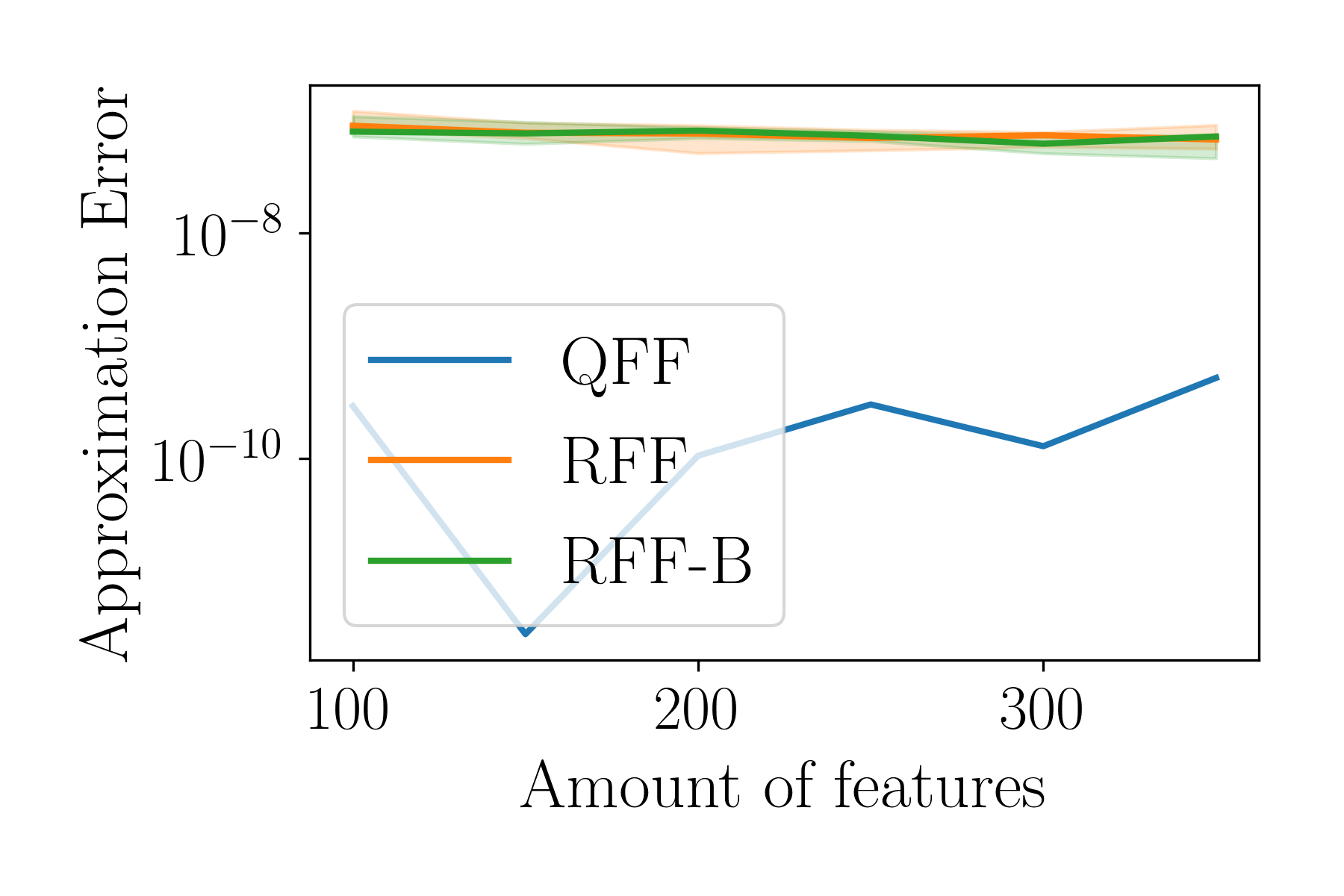}
	\caption{$\Sigma_5$}
	\end{subfigure}
	\hfill
	\begin{subfigure}[t]{0.24\textwidth}
	\centering
	\includegraphics[width=\textwidth]{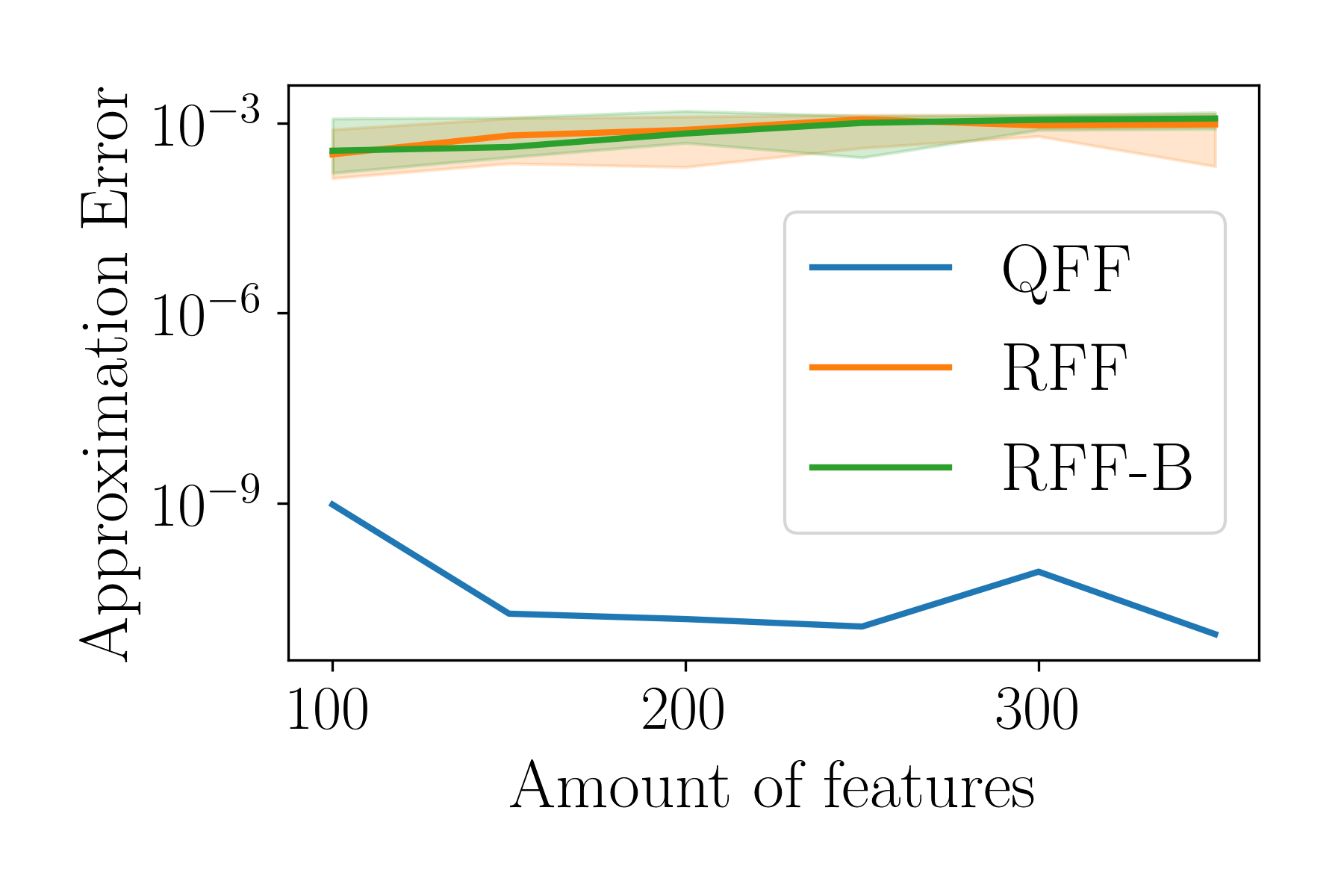}
	\caption{$\mu'_5$}
	\end{subfigure}
	\hfill
	\begin{subfigure}[t]{0.24\textwidth}
	\centering
	\includegraphics[width=\textwidth]{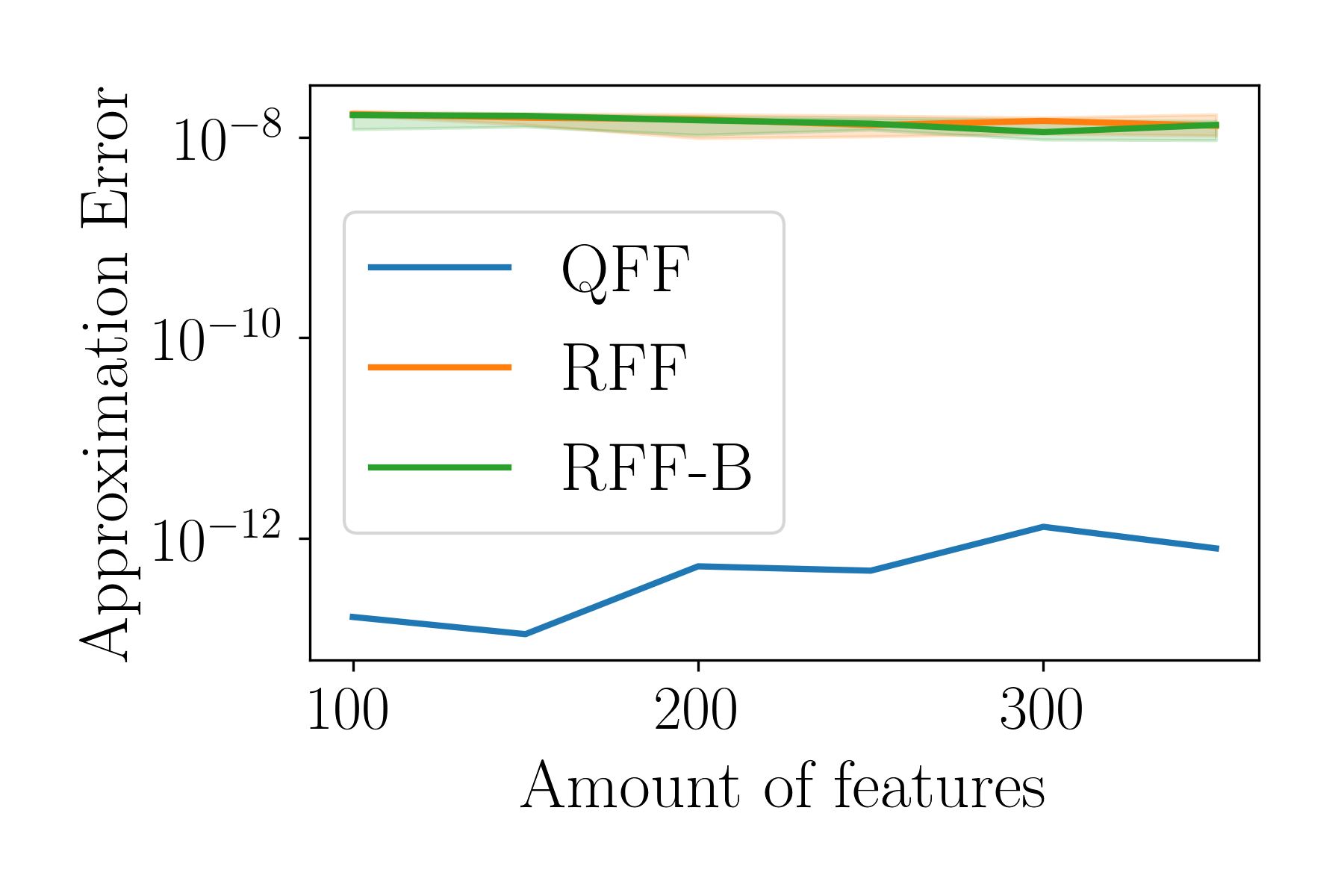}
	\caption{$\Sigma'_5$}
	\end{subfigure}
	\caption{Approximation error of the different feature approximations compared to the accurate GP, evaluated at $t=30$ for the Protein Transduction system with 1000 observations and $\sigma^2=0.0001$. For each feature, we show the median as well as the 12.5\% and 87.5\% quantiles over 10 independent noise realizations, separately for each state dimension.}
\end{figure}

\begin{figure}[!h]
	\centering
	\begin{subfigure}[t]{0.24\textwidth}
		\centering
		\includegraphics[width=\textwidth]{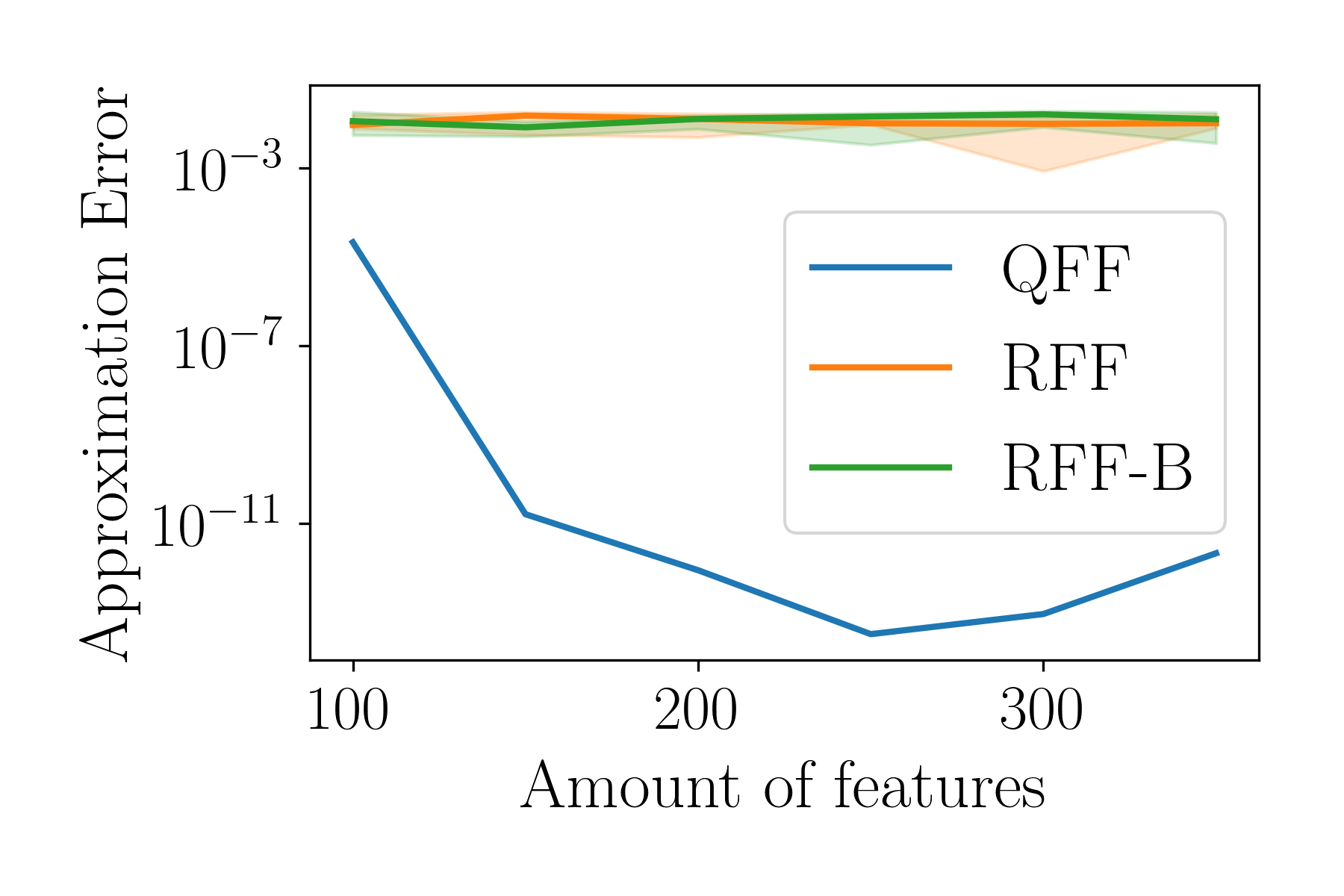}
		\caption{$\mu_1$}
	\end{subfigure}
	\hfill
	\begin{subfigure}[t]{0.24\textwidth}
		\centering
		\includegraphics[width=\textwidth]{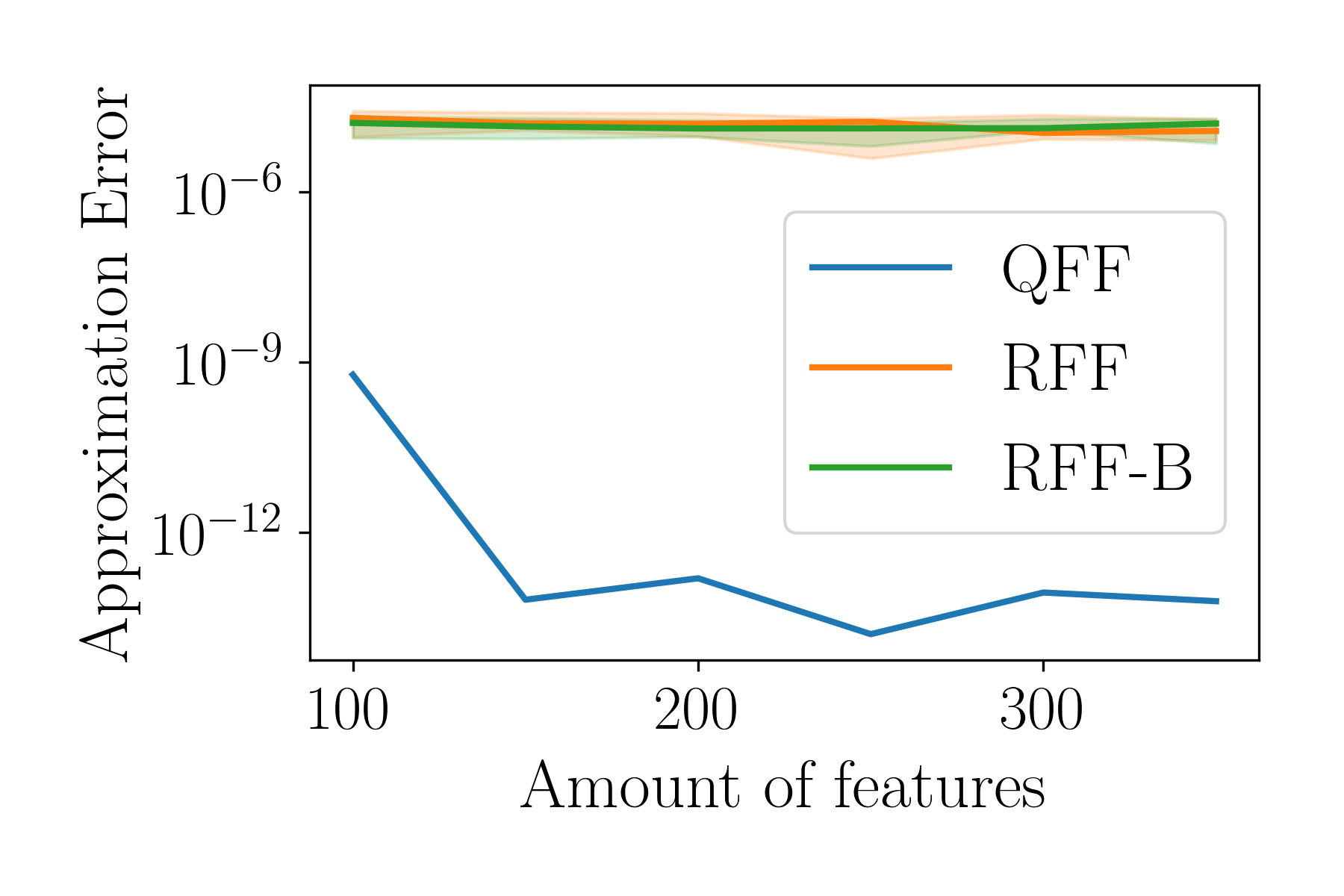}
		\caption{$\Sigma_1$}
	\end{subfigure}
	\hfill
	\begin{subfigure}[t]{0.24\textwidth}
		\centering
		\includegraphics[width=\textwidth]{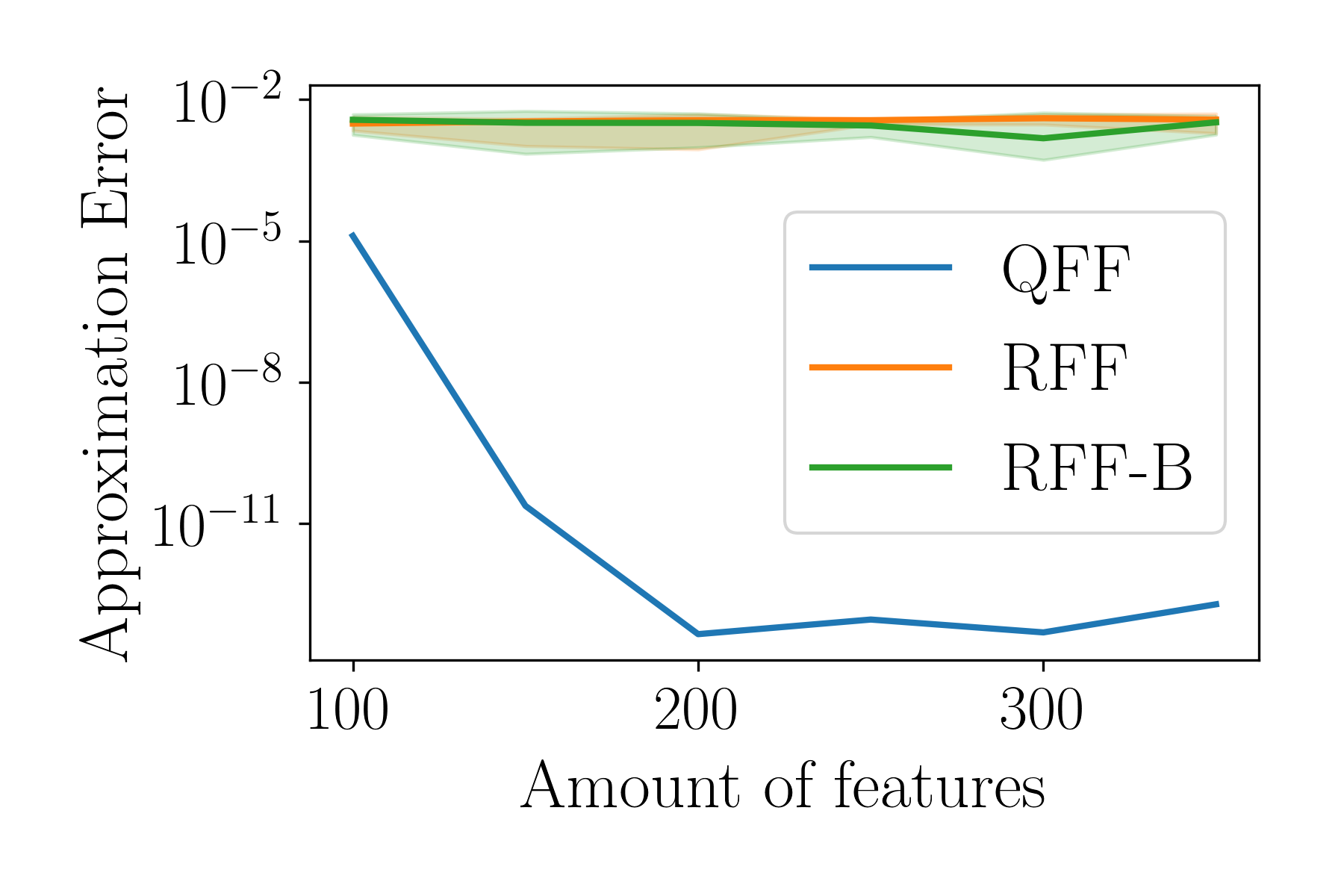}
		\caption{$\mu'_1$}
	\end{subfigure}
	\hfill
	\begin{subfigure}[t]{0.24\textwidth}
		\centering
		\includegraphics[width=\textwidth]{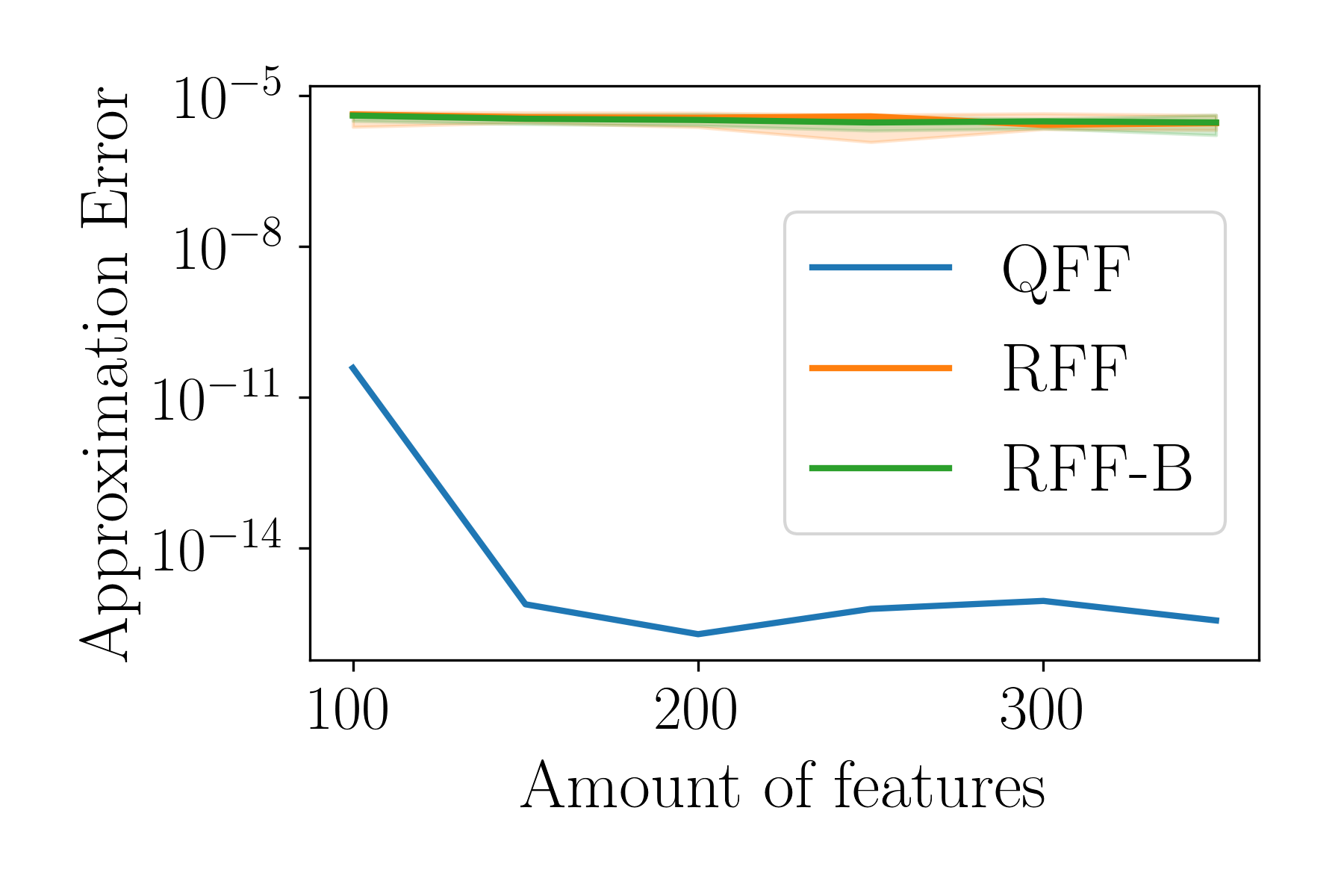}
		\caption{$\Sigma'_1$}
	\end{subfigure}\\
	\begin{subfigure}[t]{0.24\textwidth}
		\centering
		\includegraphics[width=\textwidth]{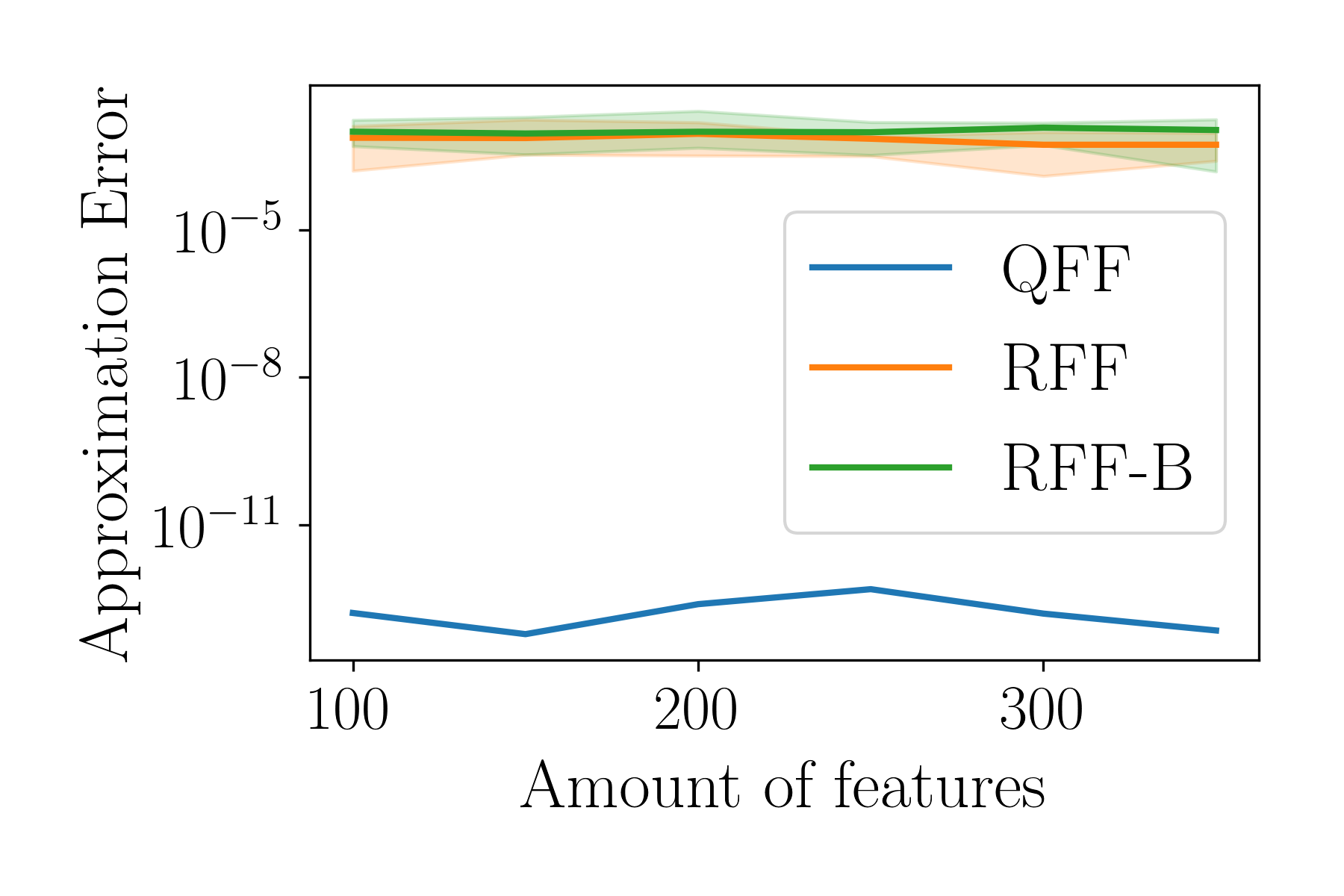}
		\caption{$\mu_2$}
	\end{subfigure}
	\hfill
	\begin{subfigure}[t]{0.24\textwidth}
		\centering
		\includegraphics[width=\textwidth]{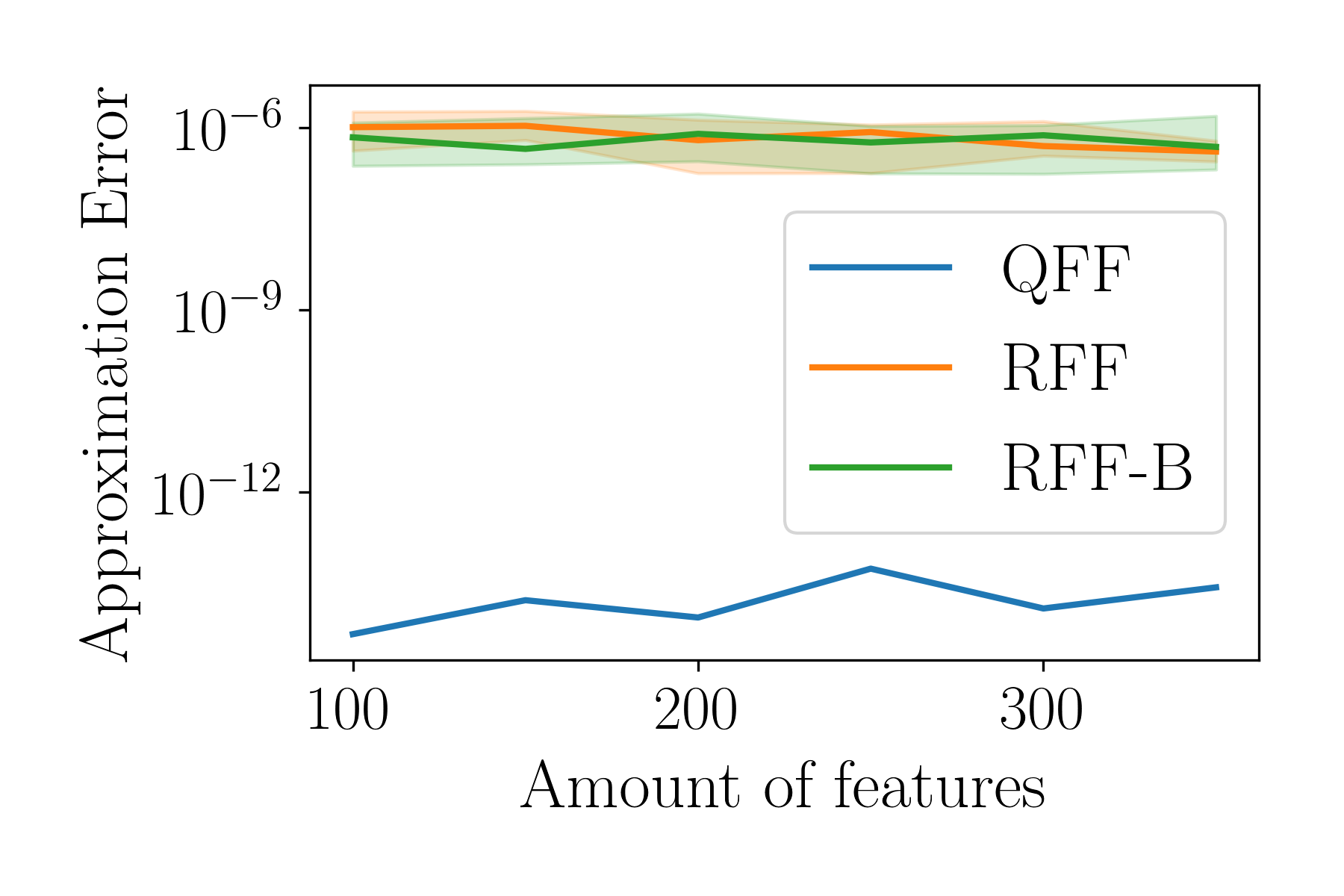}
		\caption{$\Sigma_2$}
	\end{subfigure}
	\hfill
	\begin{subfigure}[t]{0.24\textwidth}
		\centering
		\includegraphics[width=\textwidth]{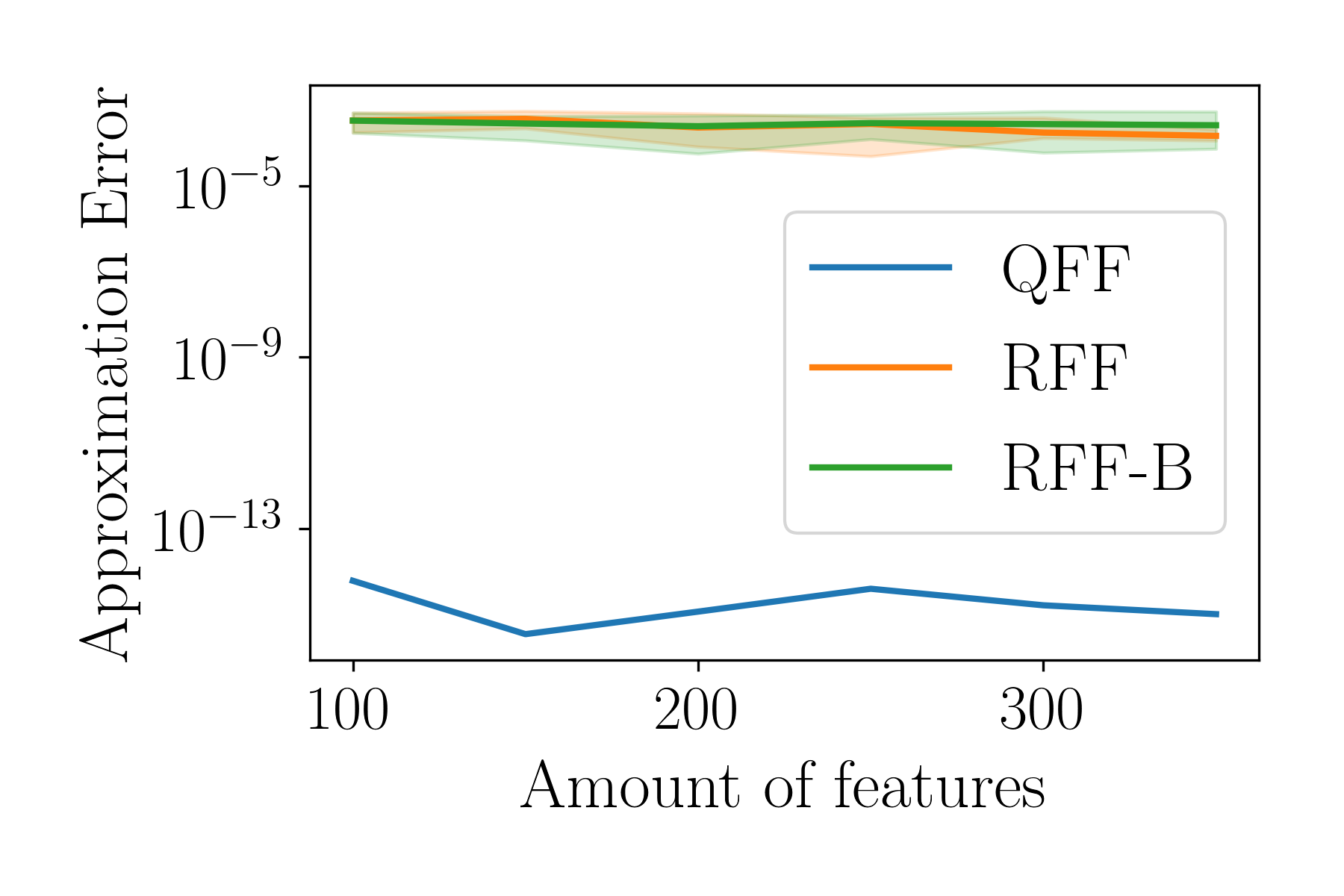}
		\caption{$\mu'_2$}
	\end{subfigure}
	\hfill
	\begin{subfigure}[t]{0.24\textwidth}
		\centering
		\includegraphics[width=\textwidth]{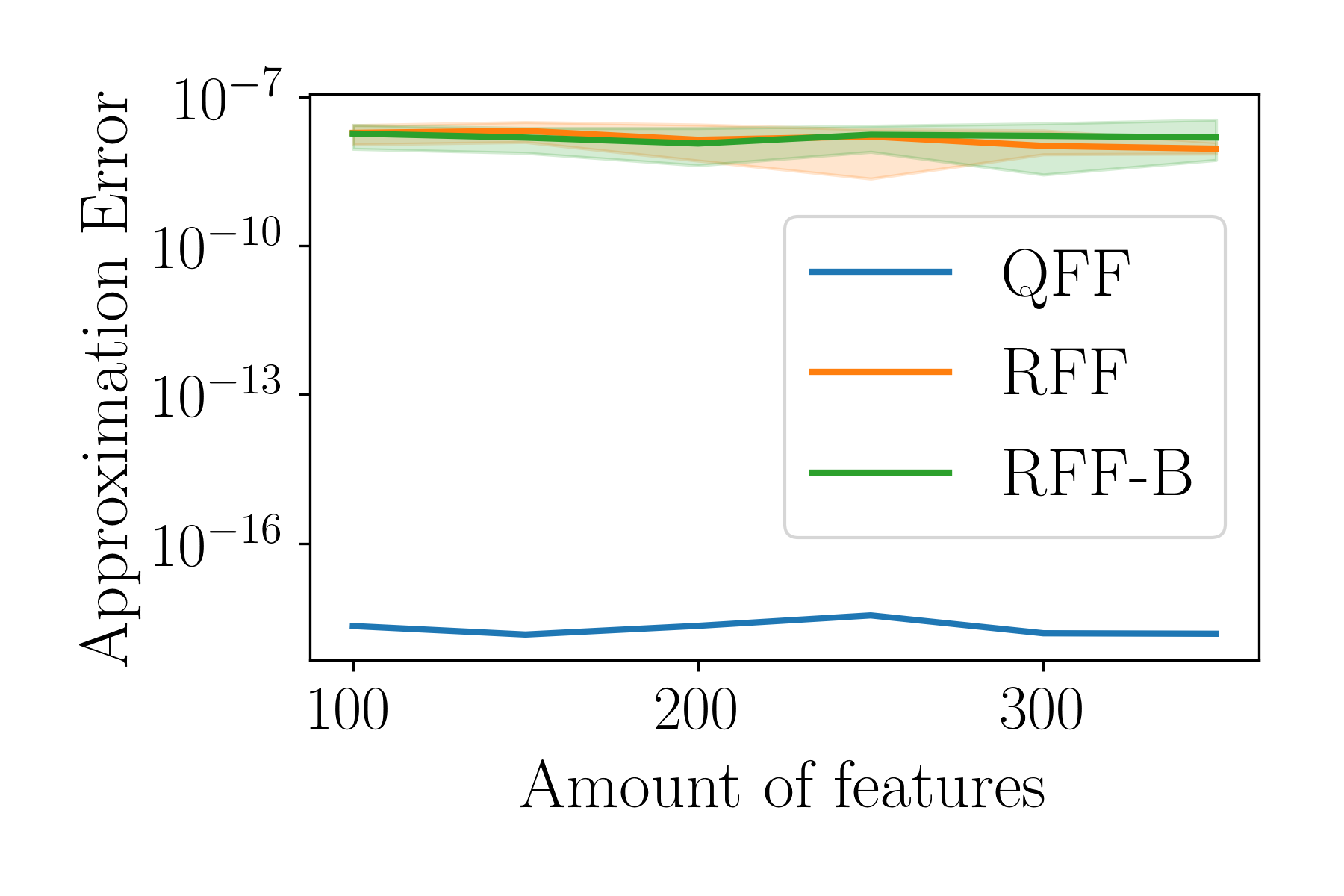}
		\caption{$\Sigma'_2$}
	\end{subfigure}\\
	\begin{subfigure}[t]{0.24\textwidth}
		\centering
		\includegraphics[width=\textwidth]{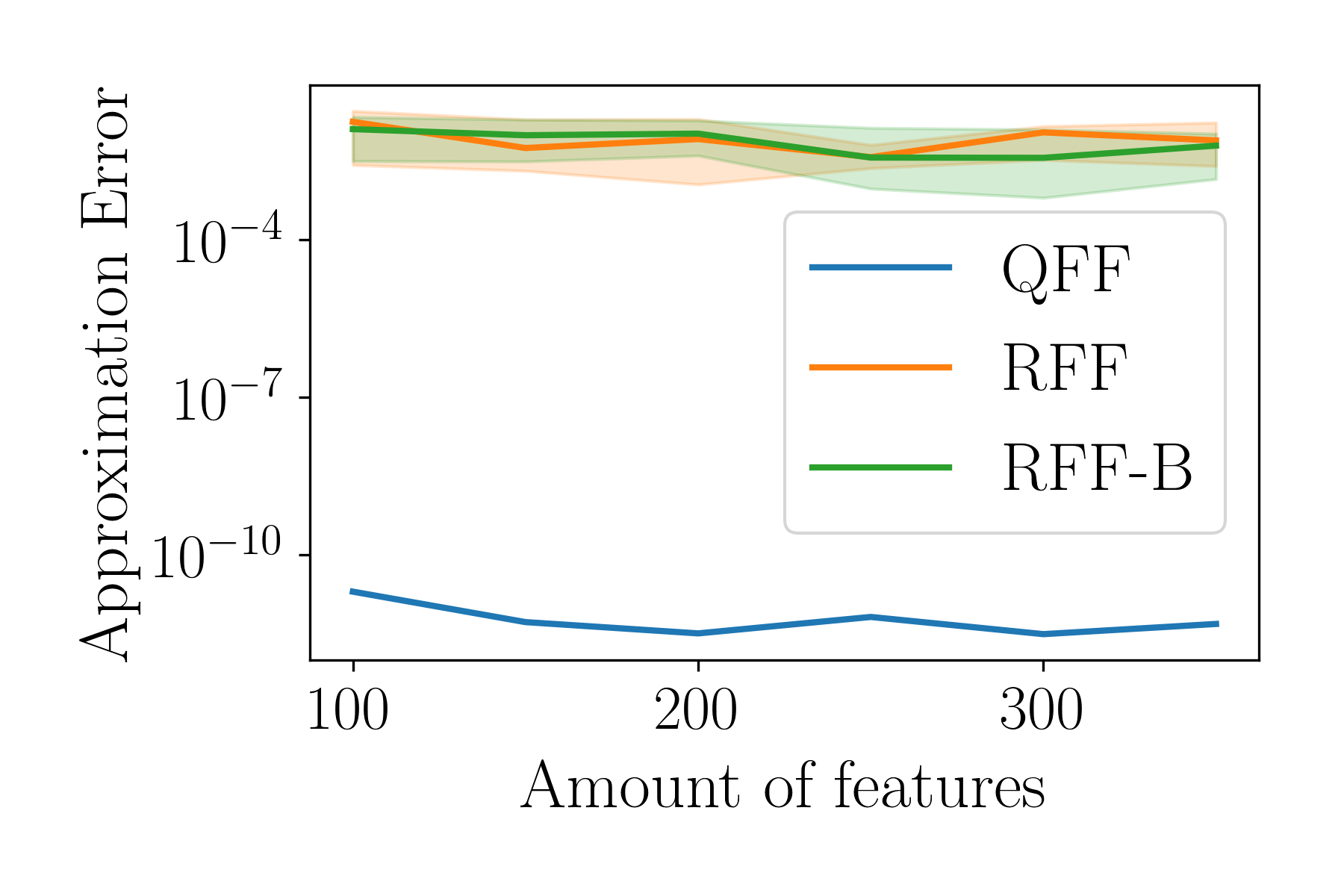}
		\caption{$\mu_3$}
	\end{subfigure}
	\hfill
	\begin{subfigure}[t]{0.24\textwidth}
		\centering
		\includegraphics[width=\textwidth]{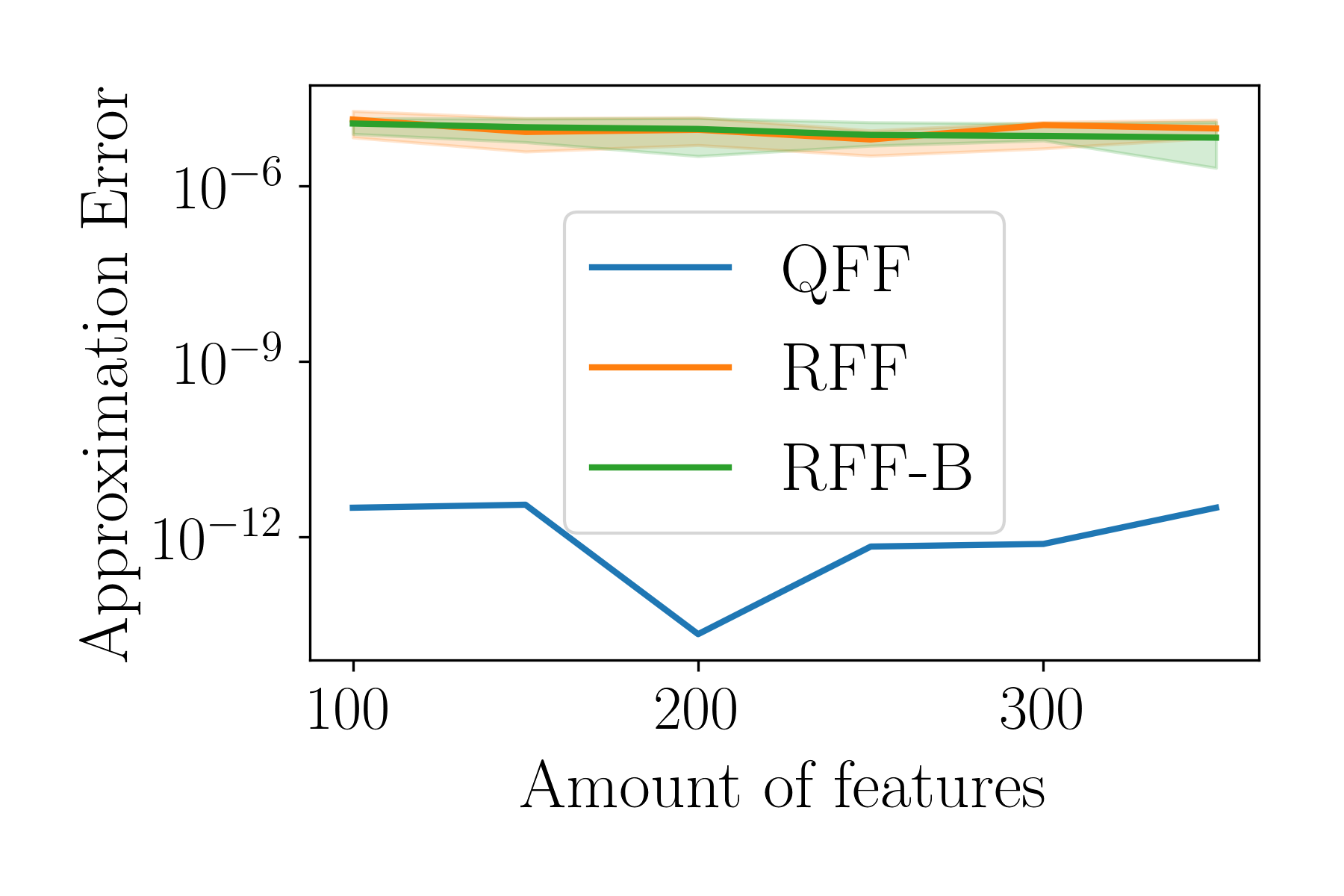}
		\caption{$\Sigma_3$}
	\end{subfigure}
	\hfill
	\begin{subfigure}[t]{0.24\textwidth}
		\centering
		\includegraphics[width=\textwidth]{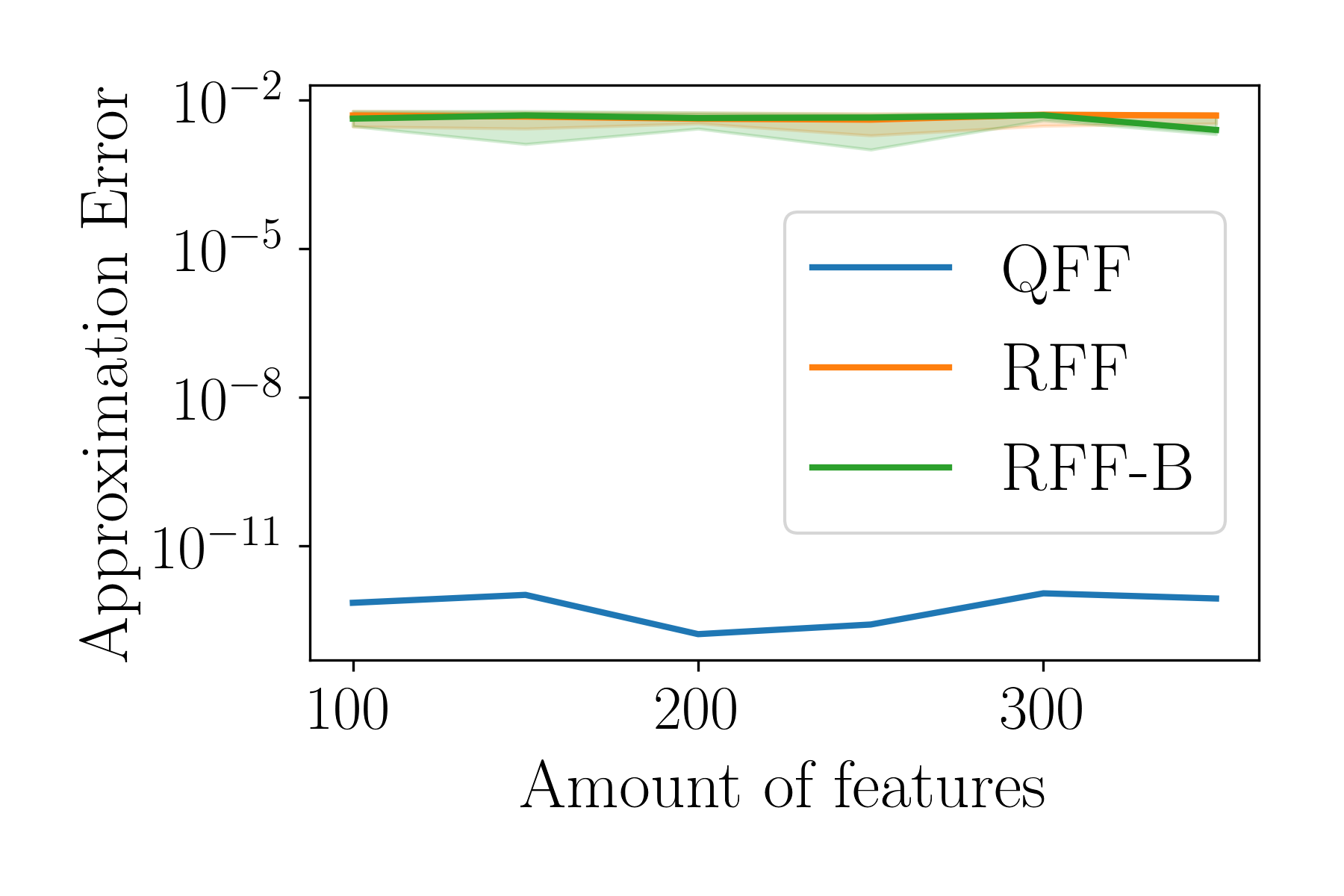}
		\caption{$\mu'3$}
	\end{subfigure}
	\hfill
	\begin{subfigure}[t]{0.24\textwidth}
		\centering
		\includegraphics[width=\textwidth]{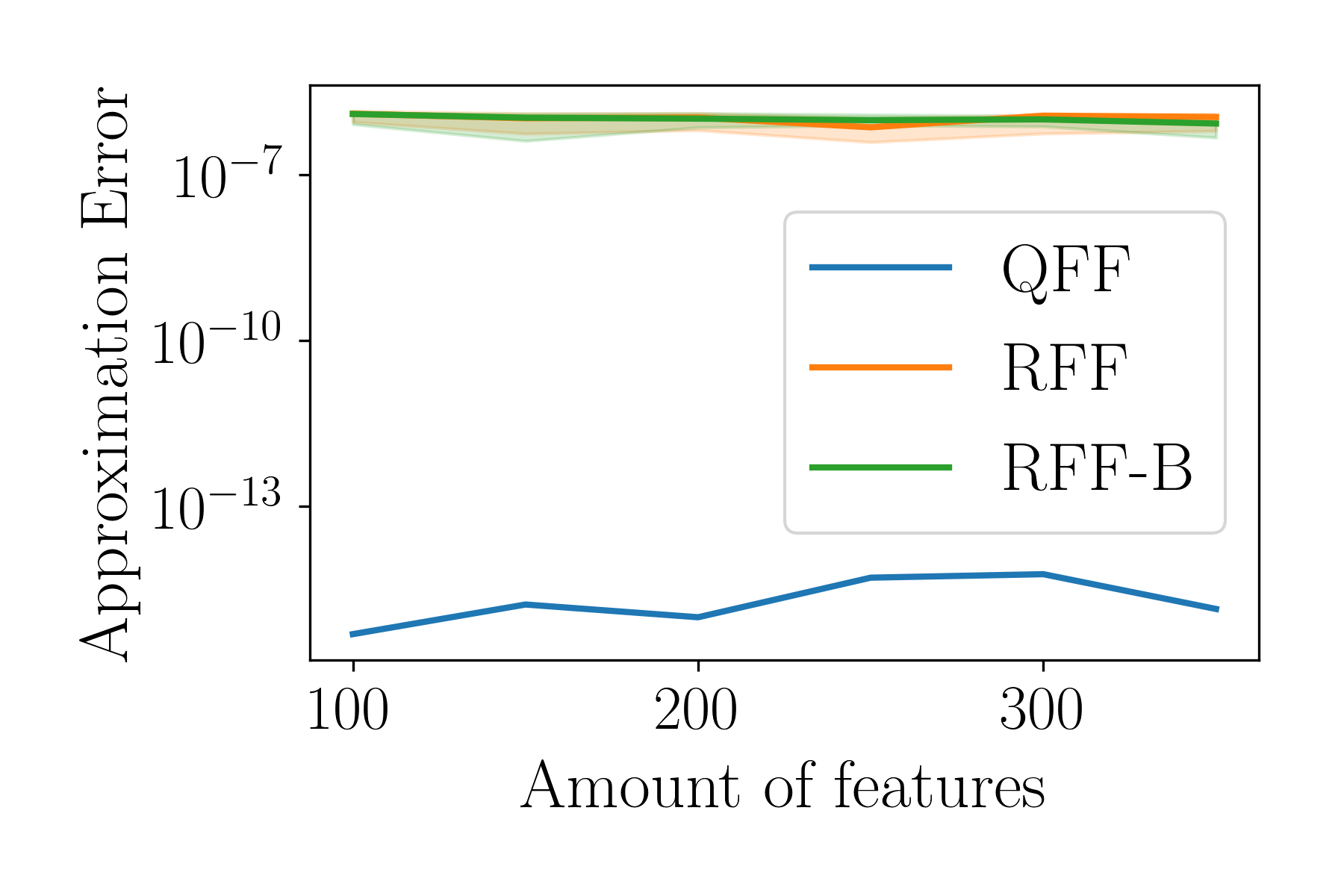}
		\caption{$\Sigma'_3$}
	\end{subfigure}\\
	\begin{subfigure}[t]{0.24\textwidth}
		\centering
		\includegraphics[width=\textwidth]{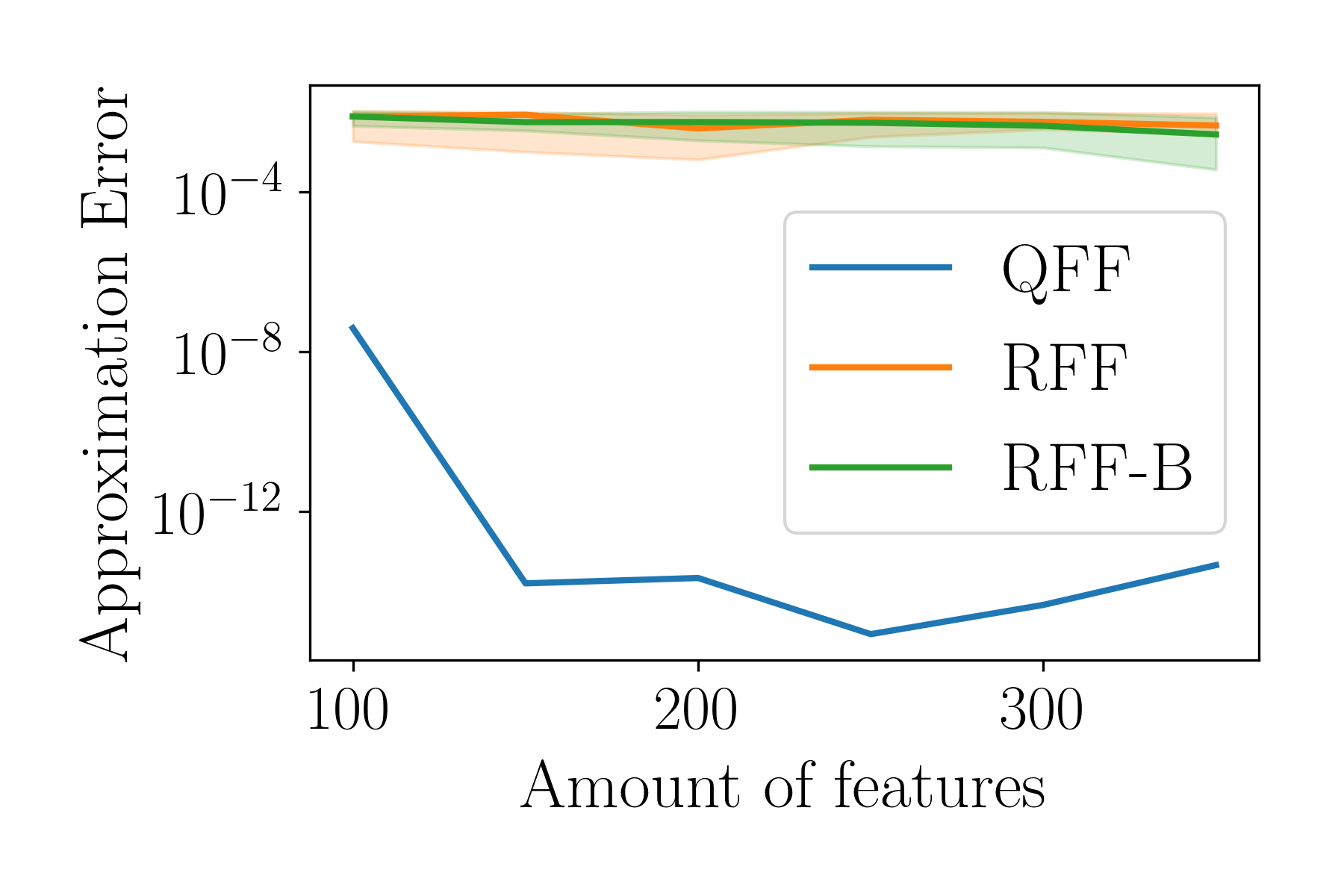}
		\caption{$\mu_4$}
	\end{subfigure}
	\hfill
	\begin{subfigure}[t]{0.24\textwidth}
		\centering
		\includegraphics[width=\textwidth]{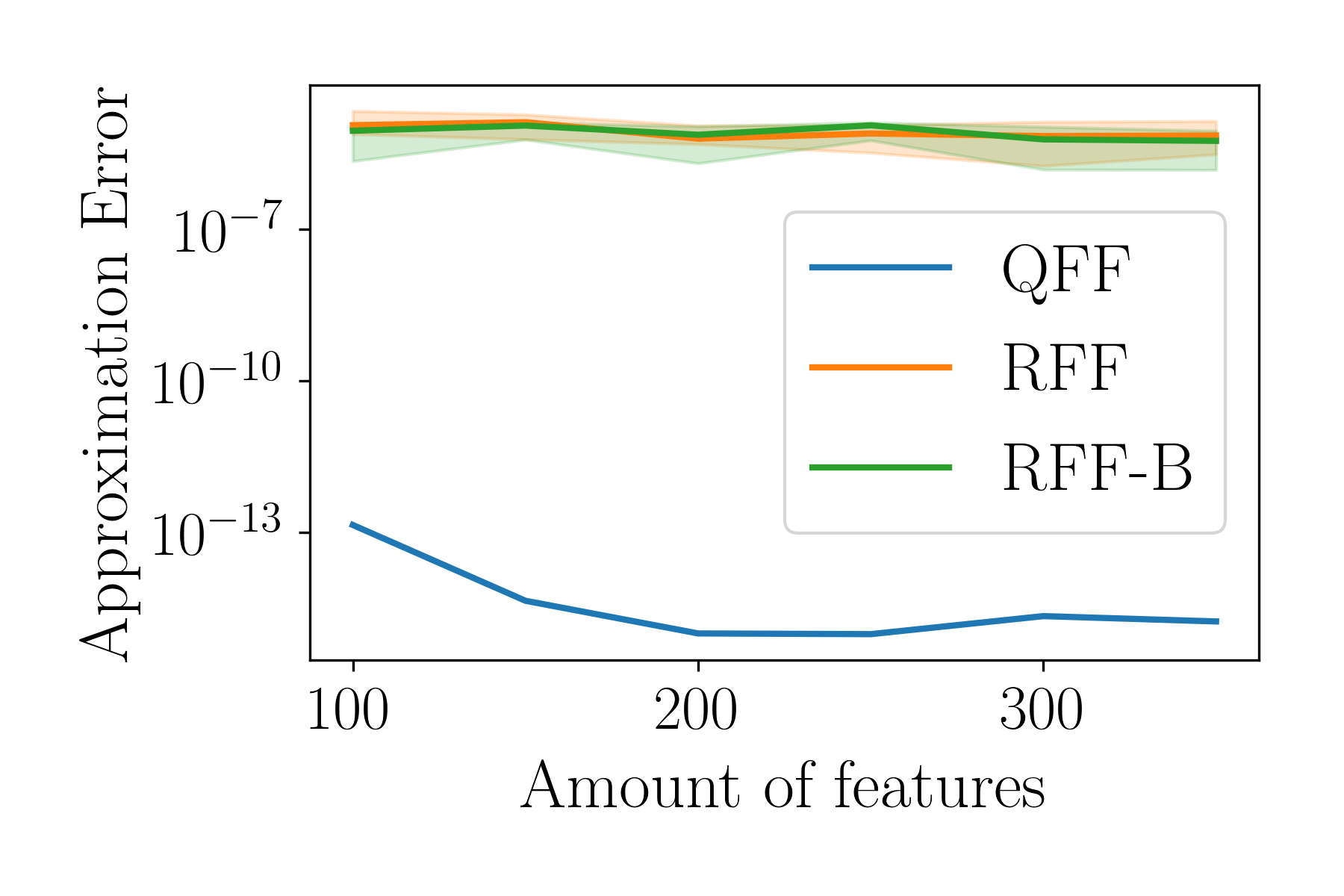}
		\caption{$\Sigma_4$}
	\end{subfigure}
	\hfill
	\begin{subfigure}[t]{0.24\textwidth}
		\centering
		\includegraphics[width=\textwidth]{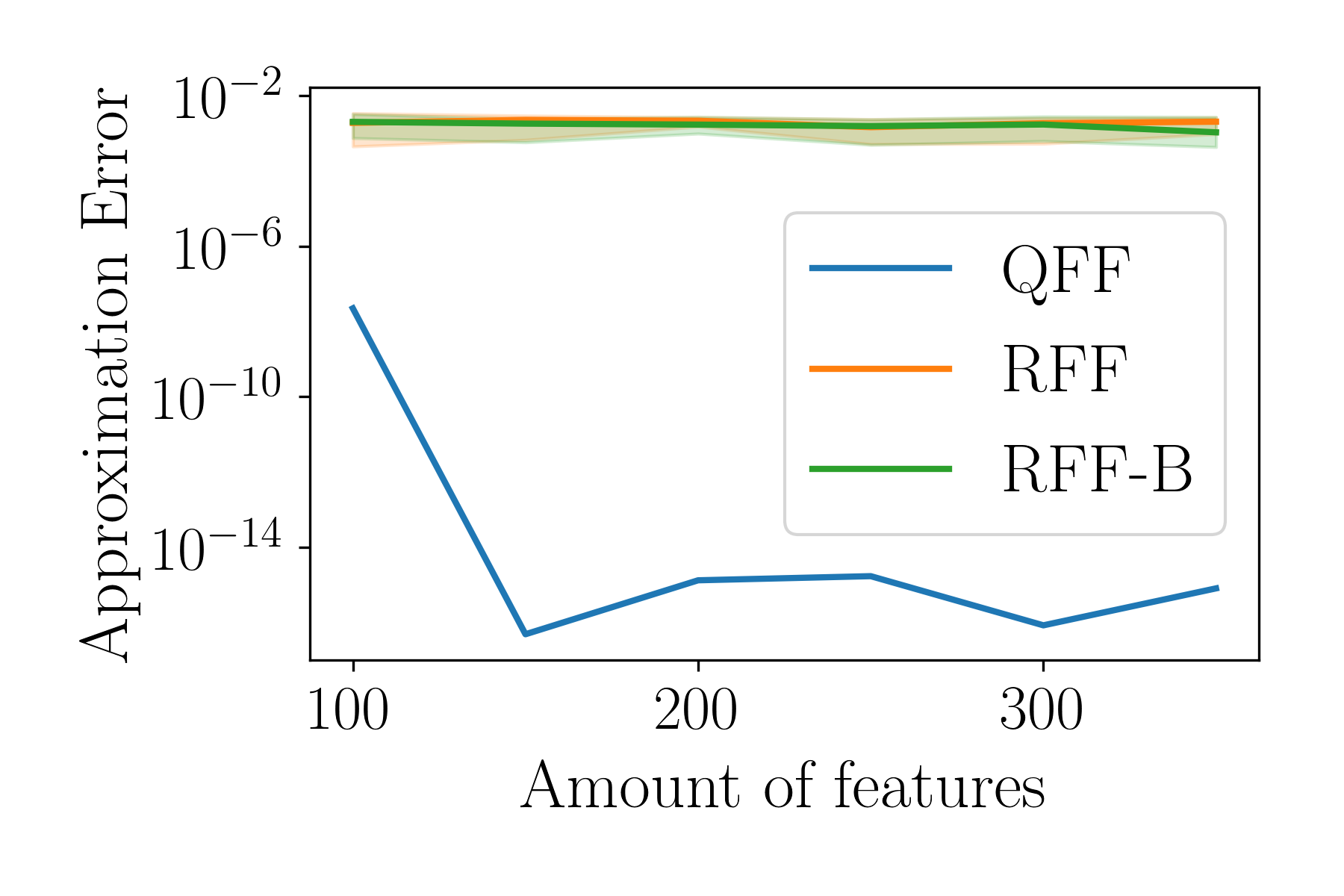}
		\caption{$\mu'_4$}
	\end{subfigure}
	\hfill
	\begin{subfigure}[t]{0.24\textwidth}
		\centering
		\includegraphics[width=\textwidth]{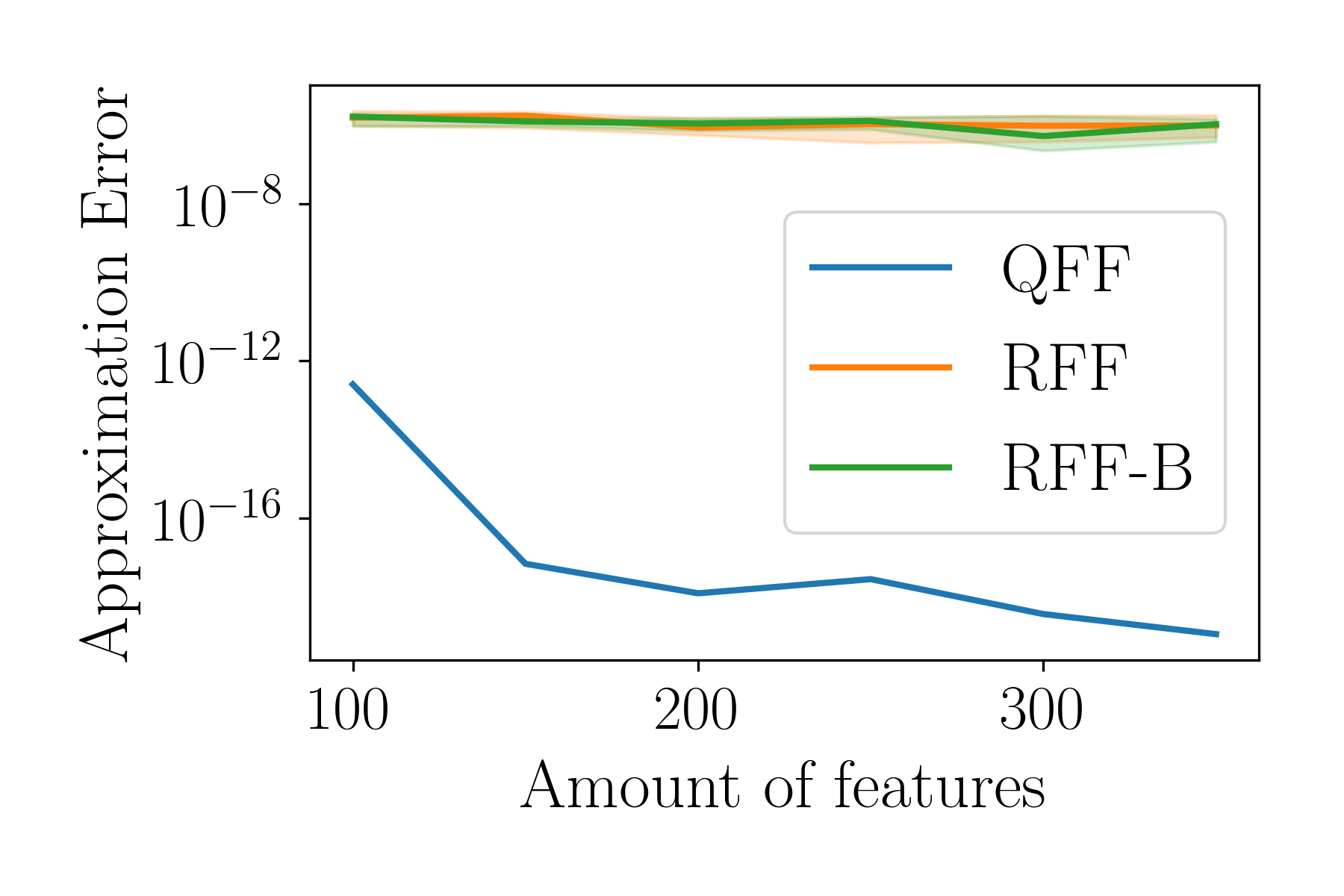}
		\caption{$\Sigma'_4$}
	\end{subfigure}\\
	\begin{subfigure}[t]{0.24\textwidth}
		\centering
		\includegraphics[width=\textwidth]{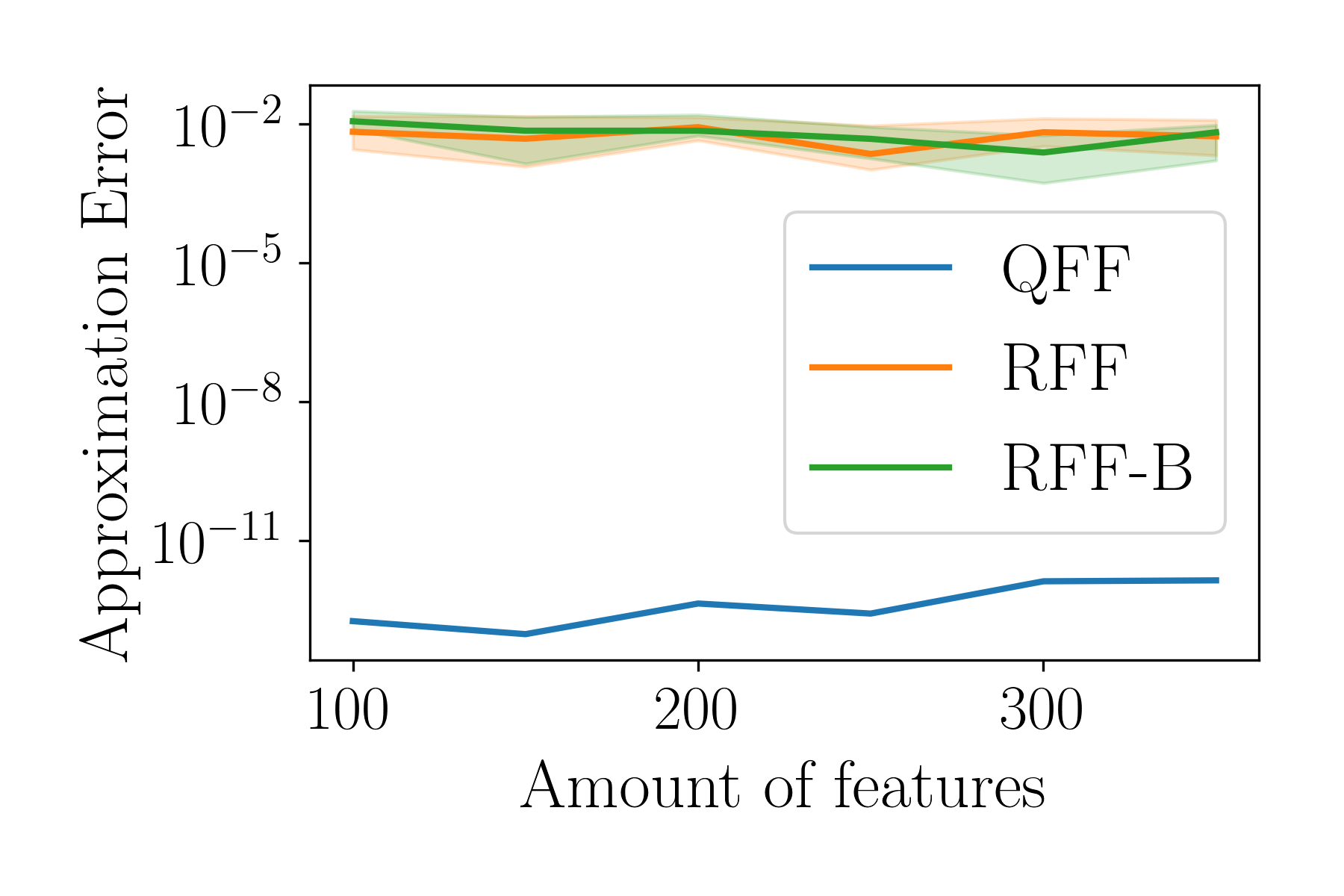}
		\caption{$\mu_5$}
	\end{subfigure}
	\hfill
	\begin{subfigure}[t]{0.24\textwidth}
		\centering
		\includegraphics[width=\textwidth]{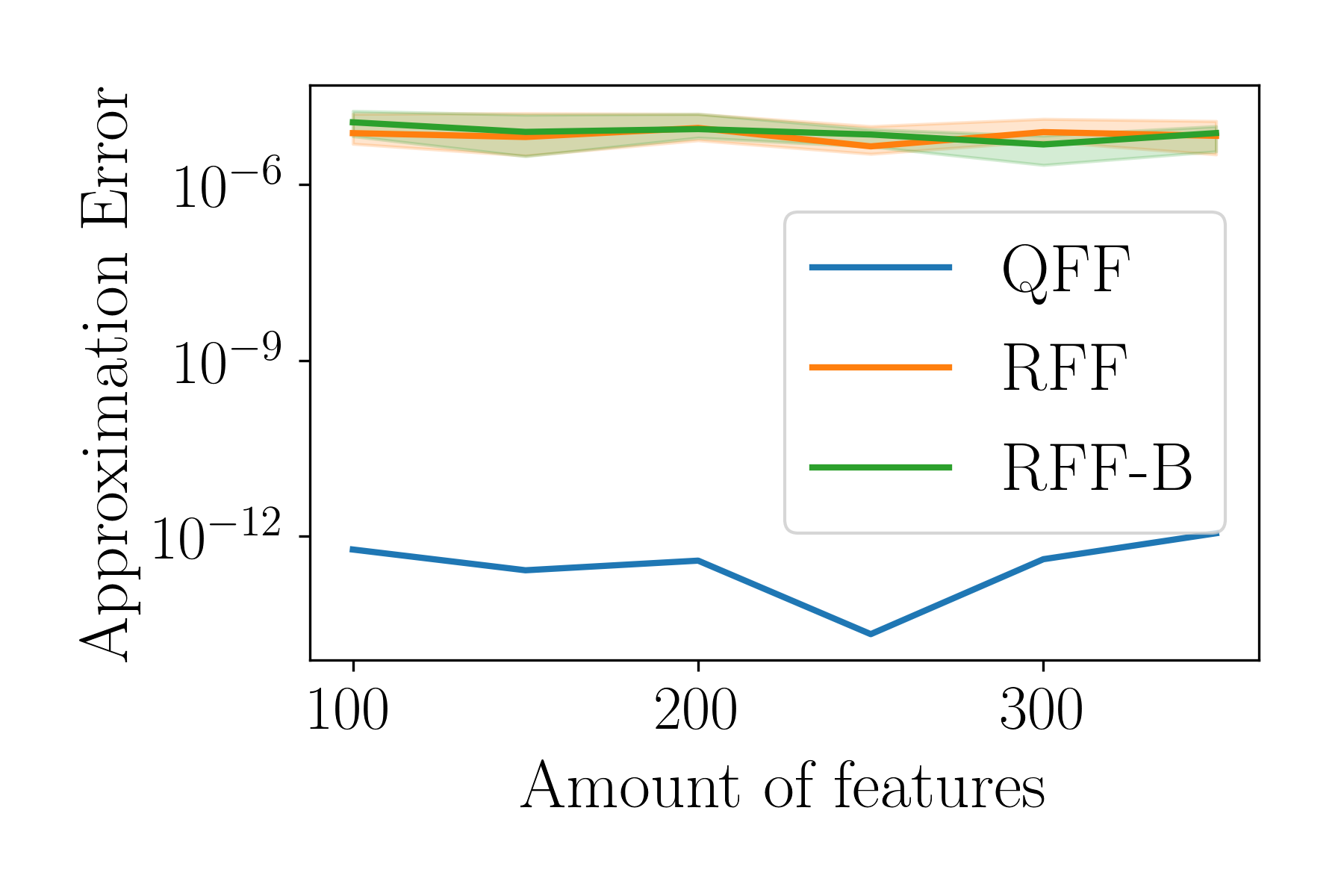}
		\caption{$\Sigma_5$}
	\end{subfigure}
	\hfill
	\begin{subfigure}[t]{0.24\textwidth}
		\centering
		\includegraphics[width=\textwidth]{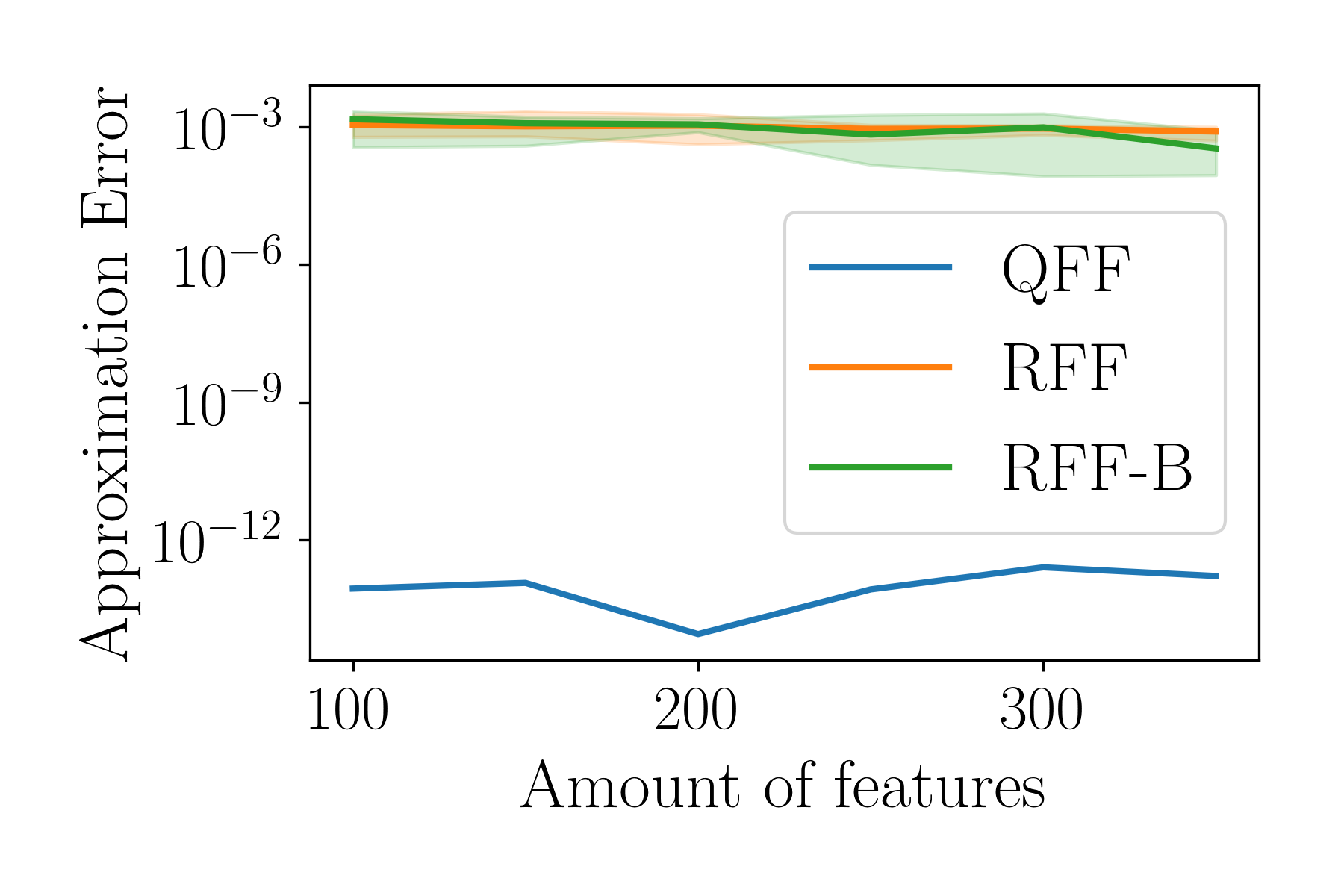}
		\caption{$\mu'_5$}
	\end{subfigure}
	\hfill
	\begin{subfigure}[t]{0.24\textwidth}
		\centering
		\includegraphics[width=\textwidth]{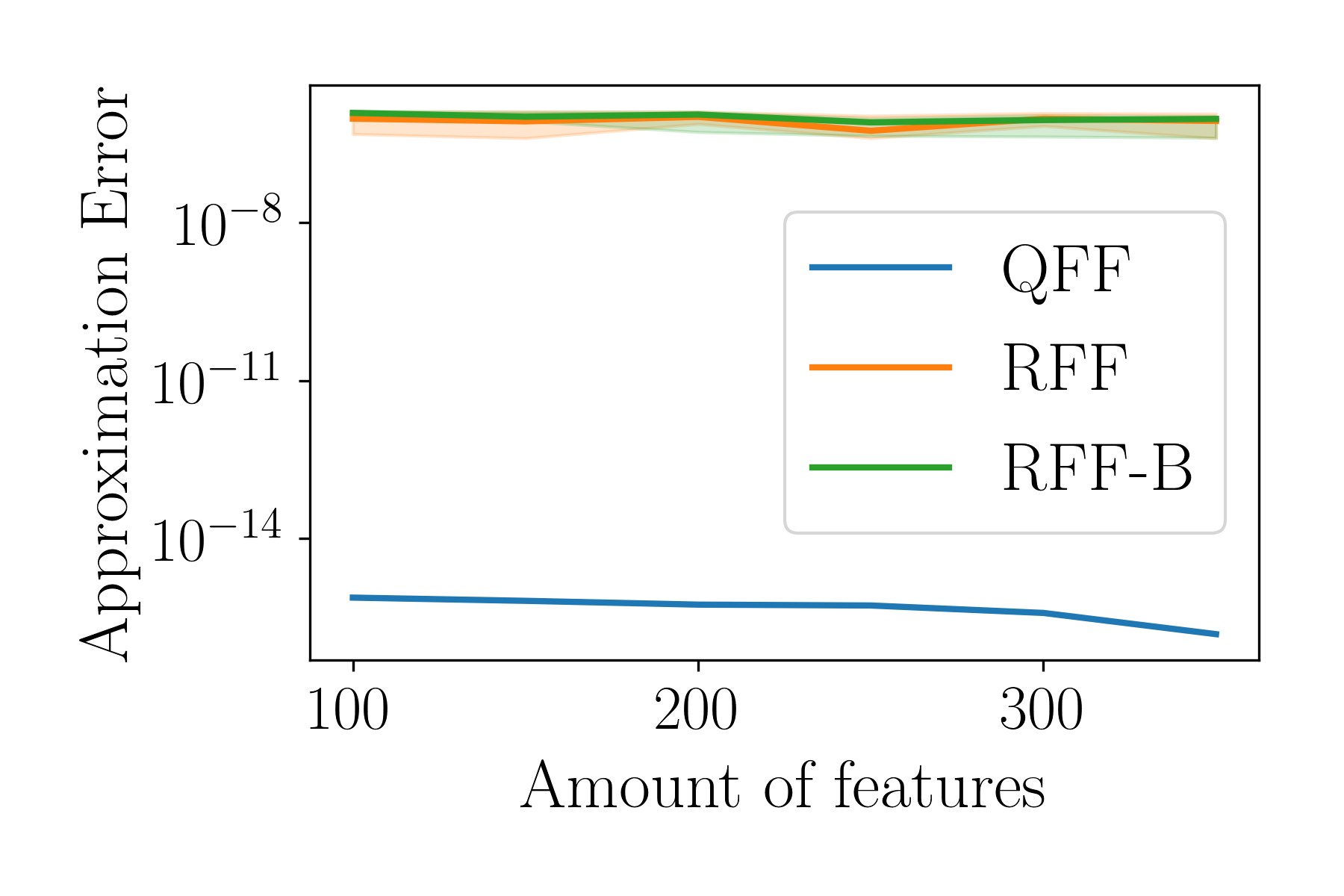}
		\caption{$\Sigma'_5$}
	\end{subfigure}
	\caption{Approximation error of the different feature approximations compared to the accurate GP, evaluated at $t=30$ for the Protein Transduction system with 1000 observations and $\sigma^2=0.01$. For each feature, we show the median as well as the 12.5\% and 87.5\% quantiles over 10 independent noise realizations, separately for each state dimension.}
\end{figure}

\clearpage
\newpage

\subsection{Lorenz}

\begin{figure}[!h]
	\centering
	\begin{subfigure}[t]{0.24\textwidth}
		\centering
		\includegraphics[width=\textwidth]{graphs/PostApprox/Lorenz/100_0_mu.png}
		\caption{$\mu_0$}
	\end{subfigure}
	\hfill
	\begin{subfigure}[t]{0.24\textwidth}
		\centering
		\includegraphics[width=\textwidth]{graphs/PostApprox/Lorenz/100_0_var.png}
		\caption{$\Sigma_0$}
	\end{subfigure}
	\hfill
	\begin{subfigure}[t]{0.24\textwidth}
		\centering
		\includegraphics[width=\textwidth]{graphs/PostApprox/Lorenz/100_0_muD.png}
		\caption{$\mu'_0$}
	\end{subfigure}
	\hfill
	\begin{subfigure}[t]{0.24\textwidth}
		\centering
		\includegraphics[width=\textwidth]{graphs/PostApprox/Lorenz/100_0_varD.png}
		\caption{$\Sigma'_0$}
	\end{subfigure}\\
	\begin{subfigure}[t]{0.24\textwidth}
		\centering
		\includegraphics[width=\textwidth]{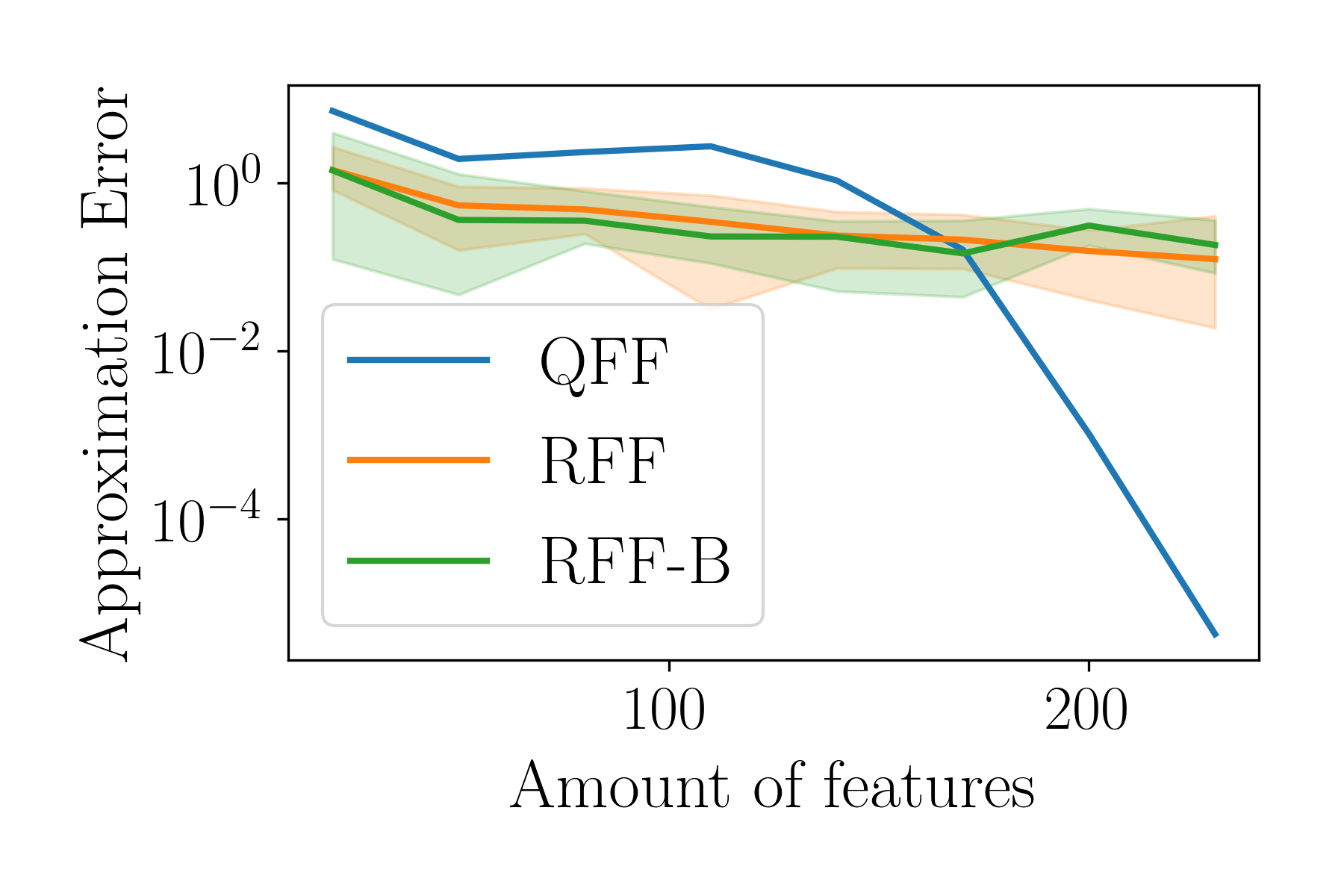}
		\caption{$\mu_1$}
	\end{subfigure}
	\hfill
	\begin{subfigure}[t]{0.24\textwidth}
		\centering
		\includegraphics[width=\textwidth]{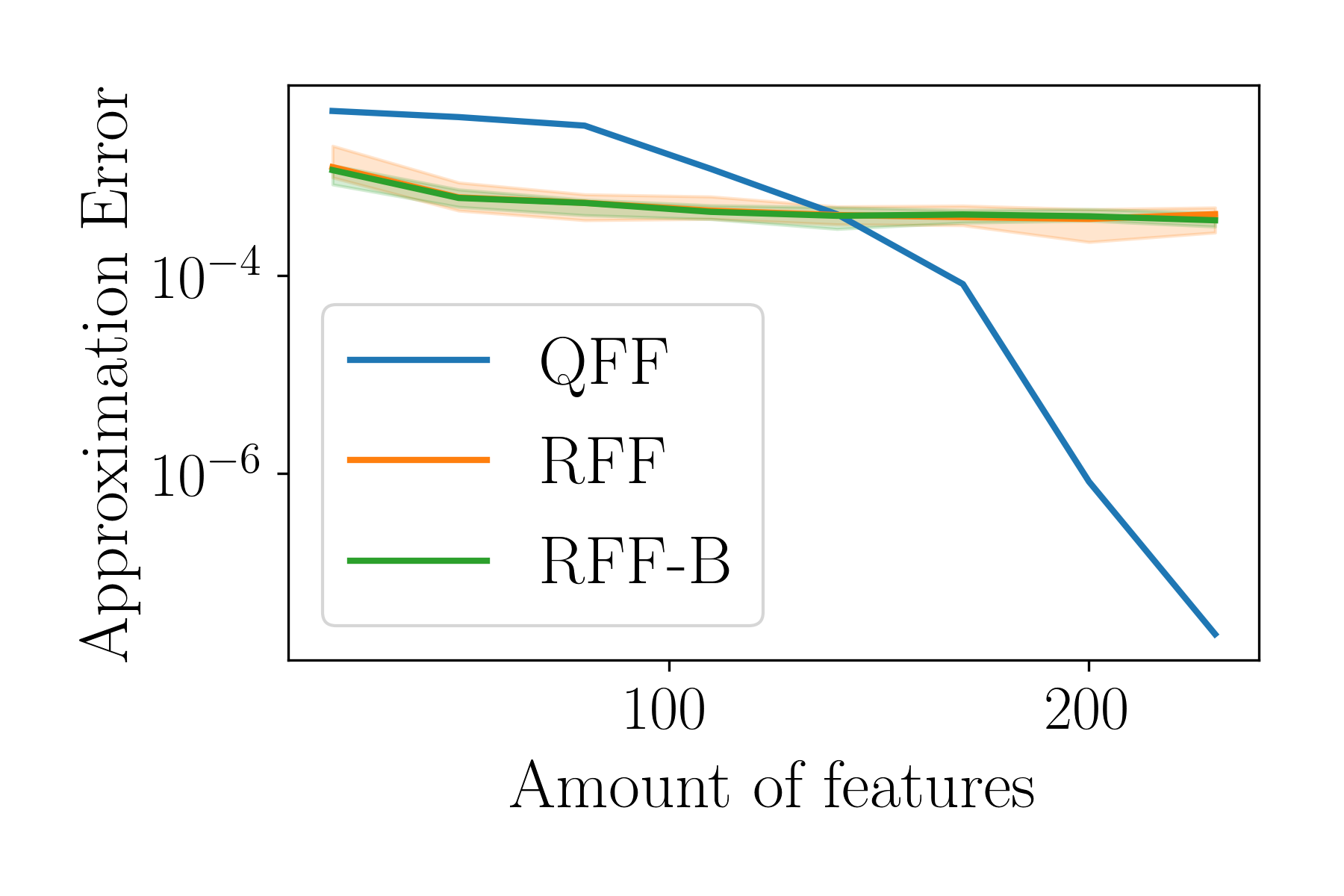}
		\caption{$\Sigma_1$}
	\end{subfigure}
	\hfill
	\begin{subfigure}[t]{0.24\textwidth}
		\centering
		\includegraphics[width=\textwidth]{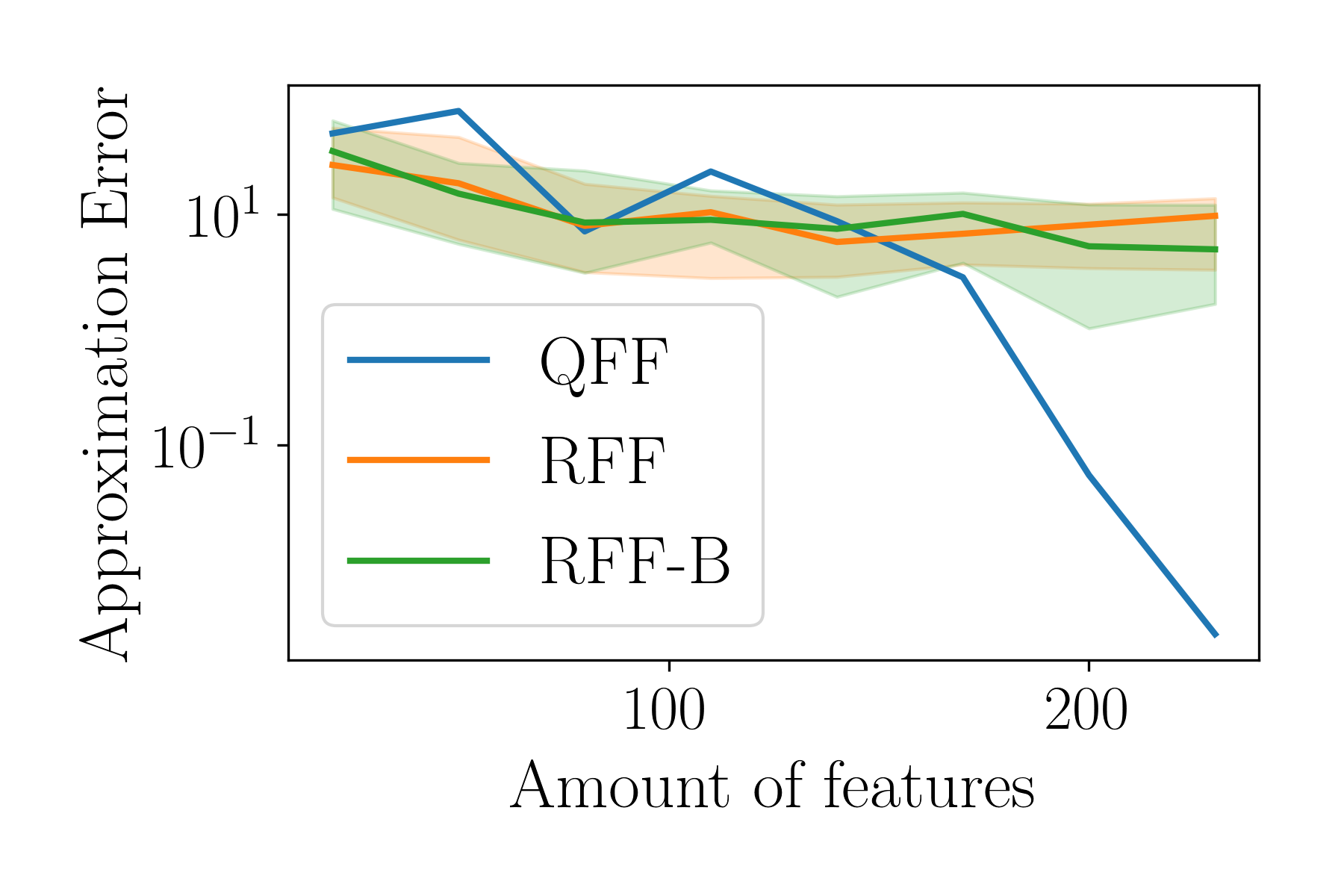}
		\caption{$\mu'_1$}
	\end{subfigure}
	\hfill
	\begin{subfigure}[t]{0.24\textwidth}
		\centering
		\includegraphics[width=\textwidth]{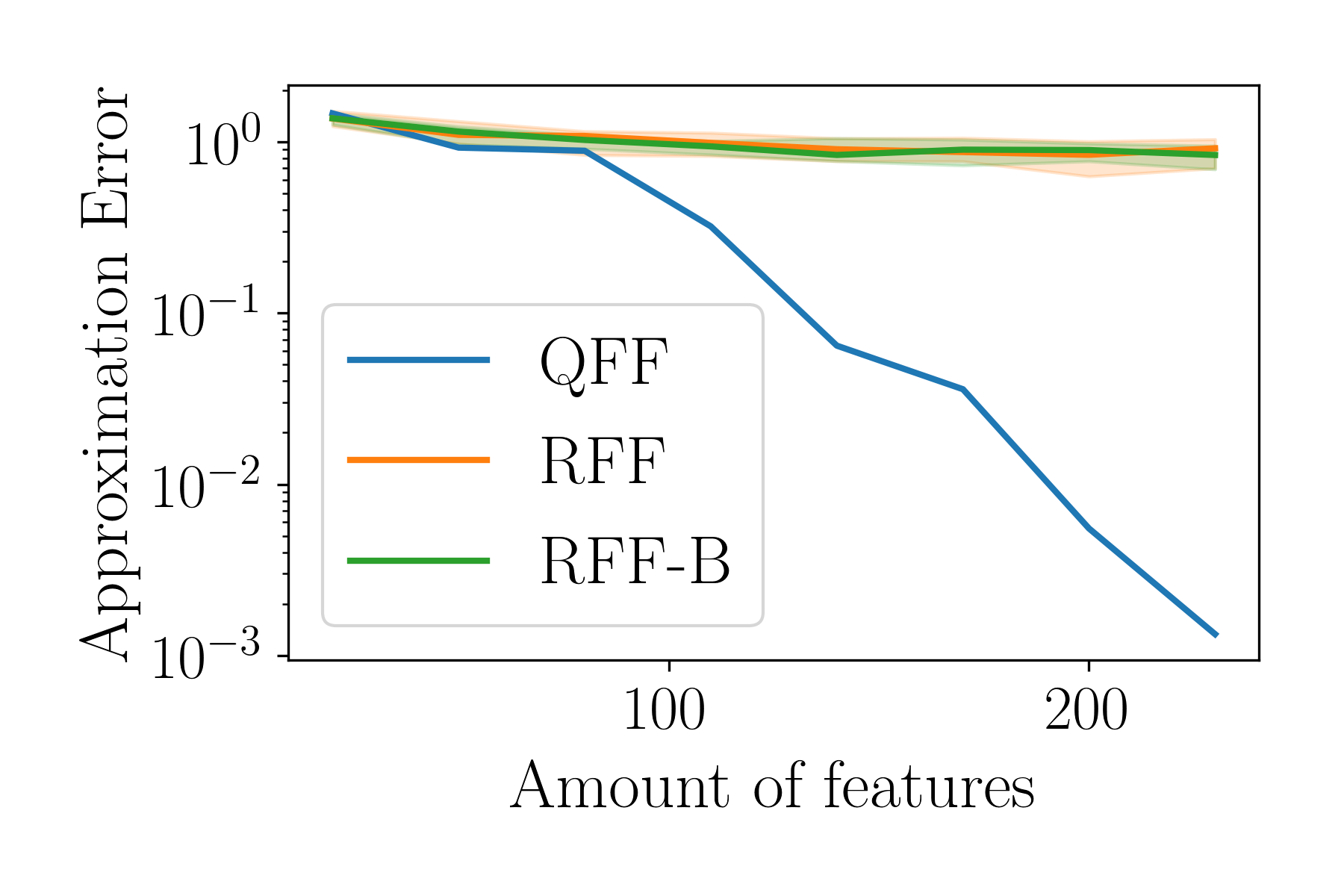}
		\caption{$\Sigma'_1$}
	\end{subfigure}
	\caption{Approximation error of the different feature approximations compared to the accurate GP, evaluated at $t=0.8$ for the Lorenz system with 1000 observations and an SNR of 100. For each feature, we show the median as well as the 12.5\% and 87.5\% quantiles over 10 independent noise realizations, separately for each state dimension.}
\end{figure}

\begin{figure}[!h]
	\centering
	\begin{subfigure}[t]{0.24\textwidth}
		\centering
		\includegraphics[width=\textwidth]{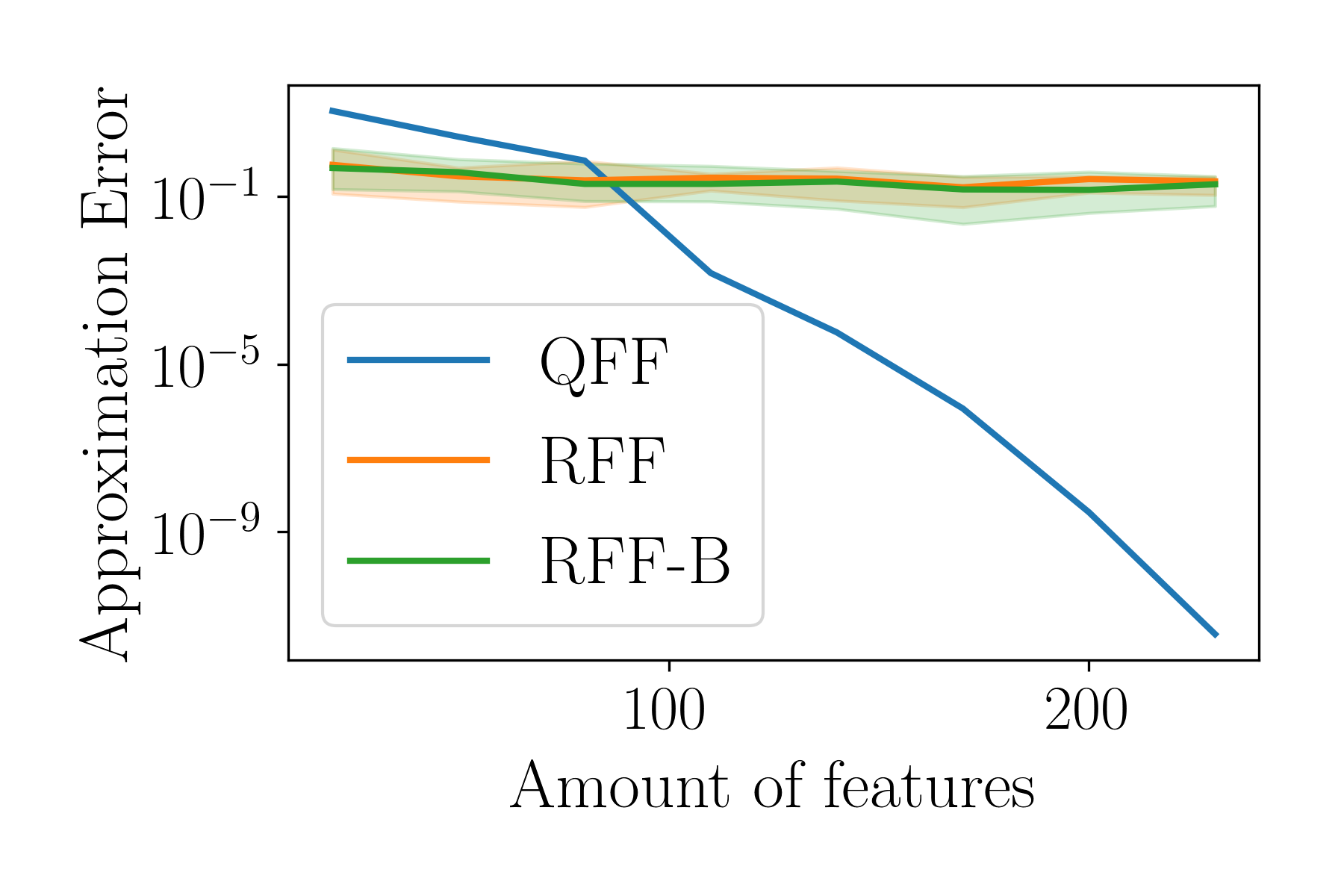}
		\caption{$\mu_0$}
	\end{subfigure}
	\hfill
	\begin{subfigure}[t]{0.24\textwidth}
		\centering
		\includegraphics[width=\textwidth]{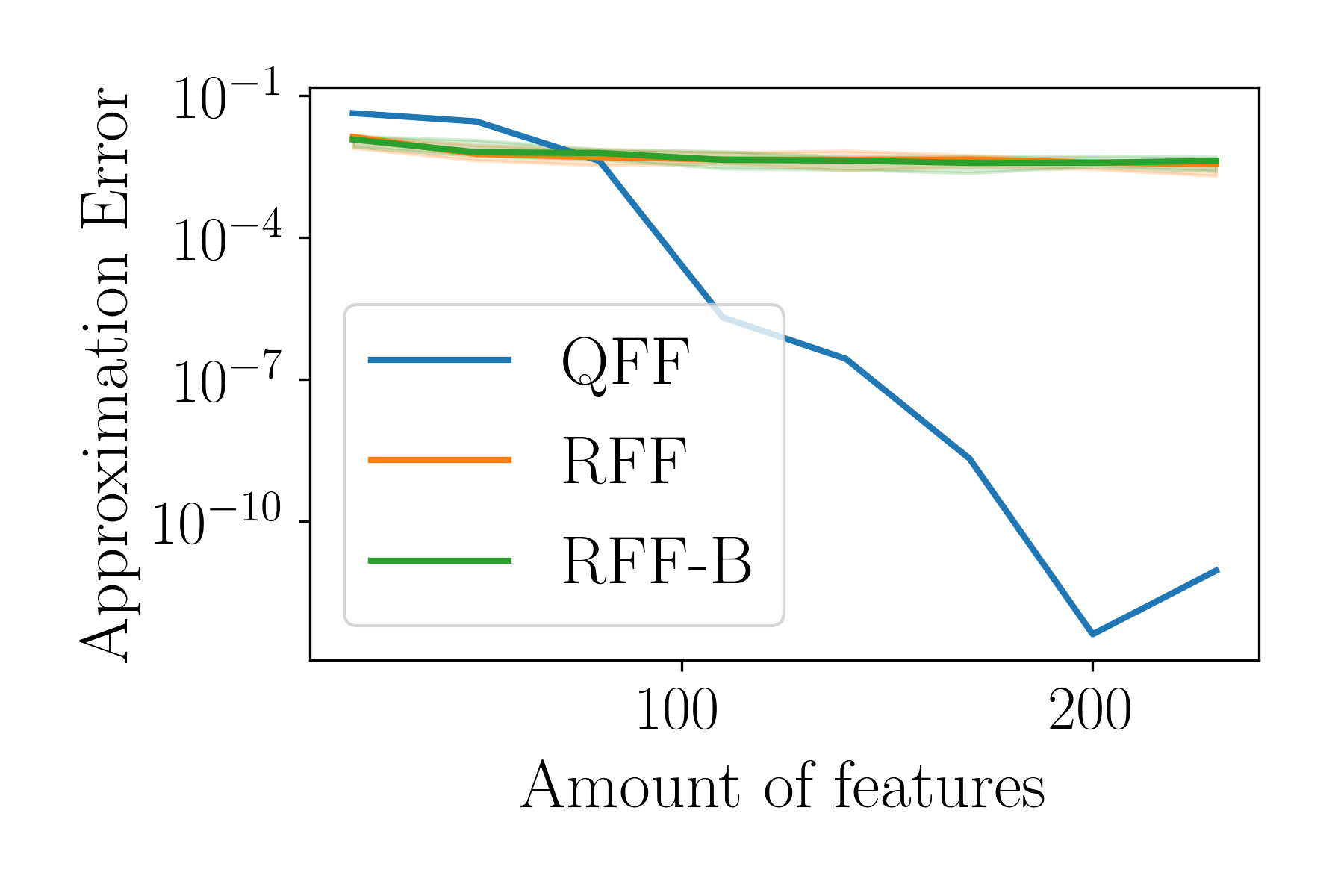}
		\caption{$\Sigma_0$}
	\end{subfigure}
	\hfill
	\begin{subfigure}[t]{0.24\textwidth}
		\centering
		\includegraphics[width=\textwidth]{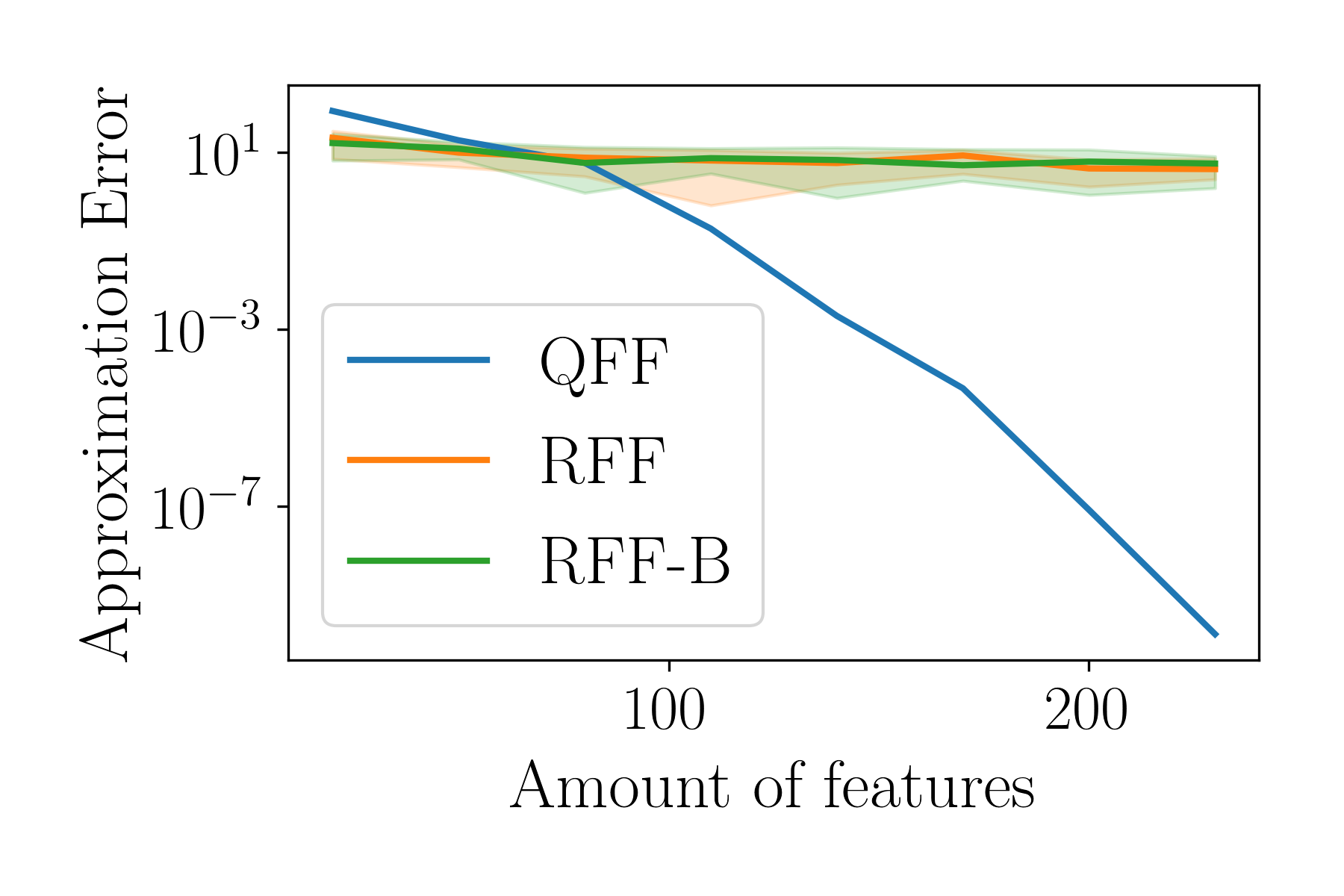}
		\caption{$\mu'_0$}
	\end{subfigure}
	\hfill
	\begin{subfigure}[t]{0.24\textwidth}
		\centering
		\includegraphics[width=\textwidth]{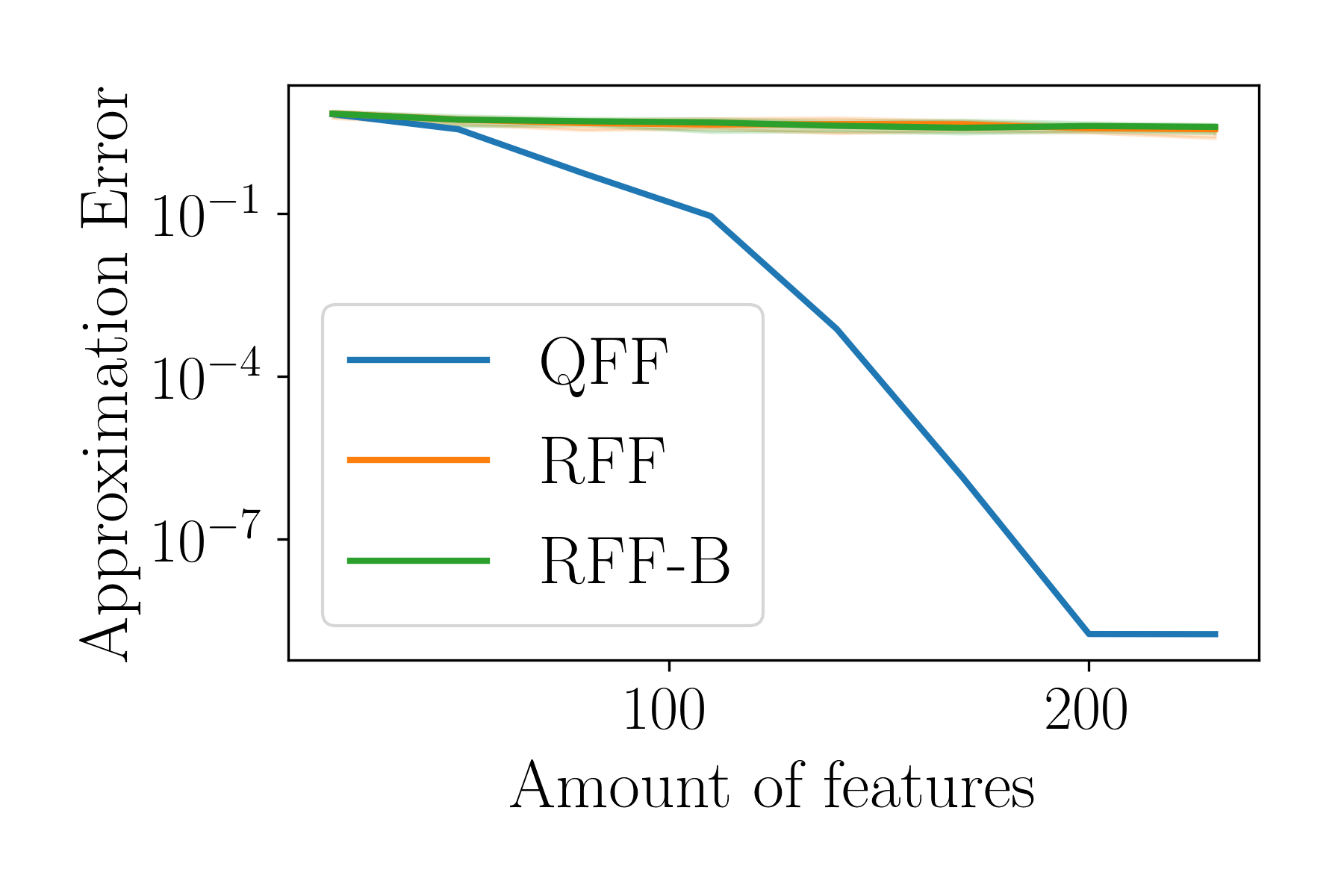}
		\caption{$\Sigma'_0$}
	\end{subfigure}\\
	\begin{subfigure}[t]{0.24\textwidth}
		\centering
		\includegraphics[width=\textwidth]{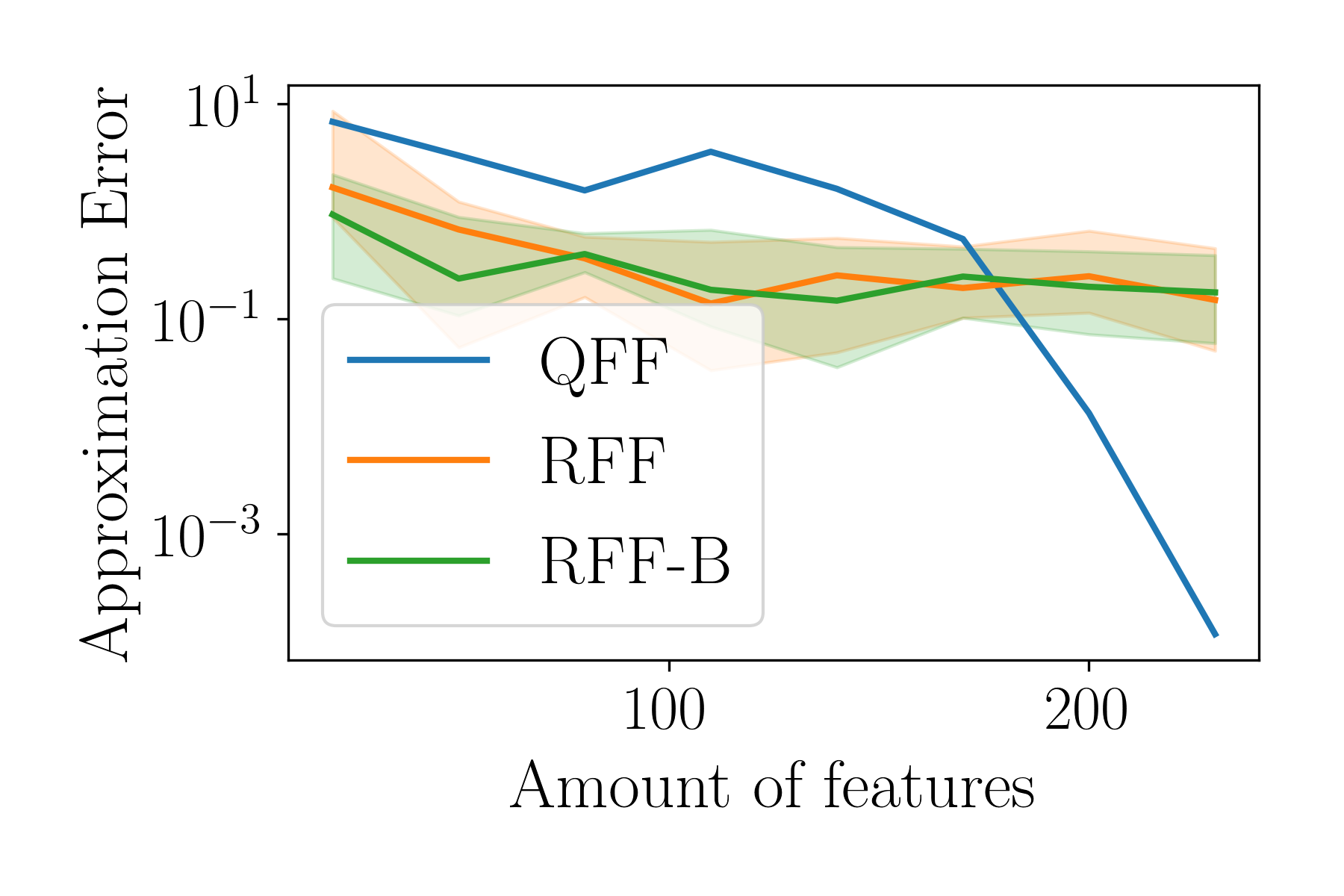}
		\caption{$\mu_1$}
	\end{subfigure}
	\hfill
	\begin{subfigure}[t]{0.24\textwidth}
		\centering
		\includegraphics[width=\textwidth]{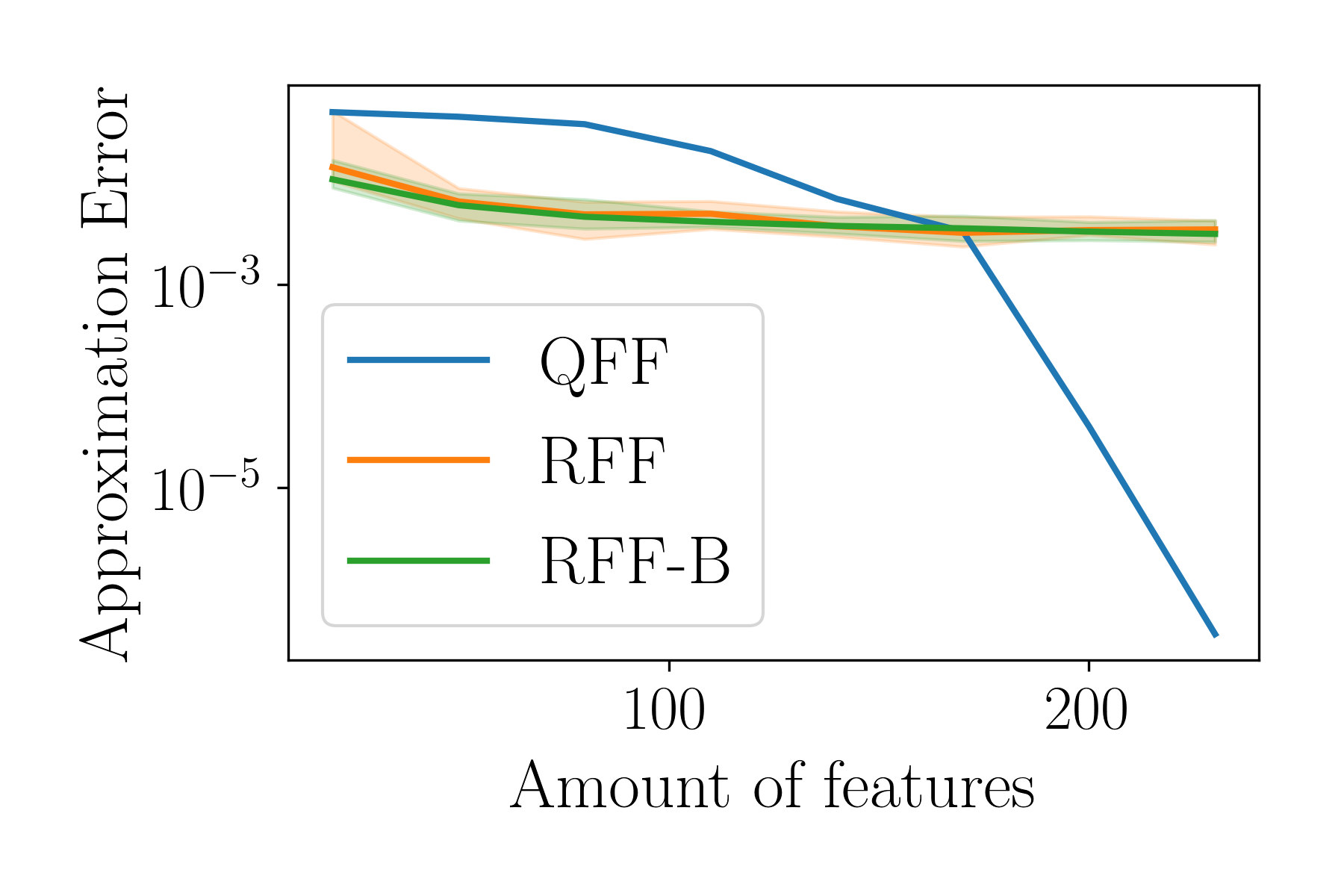}
		\caption{$\Sigma_1$}
	\end{subfigure}
	\hfill
	\begin{subfigure}[t]{0.24\textwidth}
		\centering
		\includegraphics[width=\textwidth]{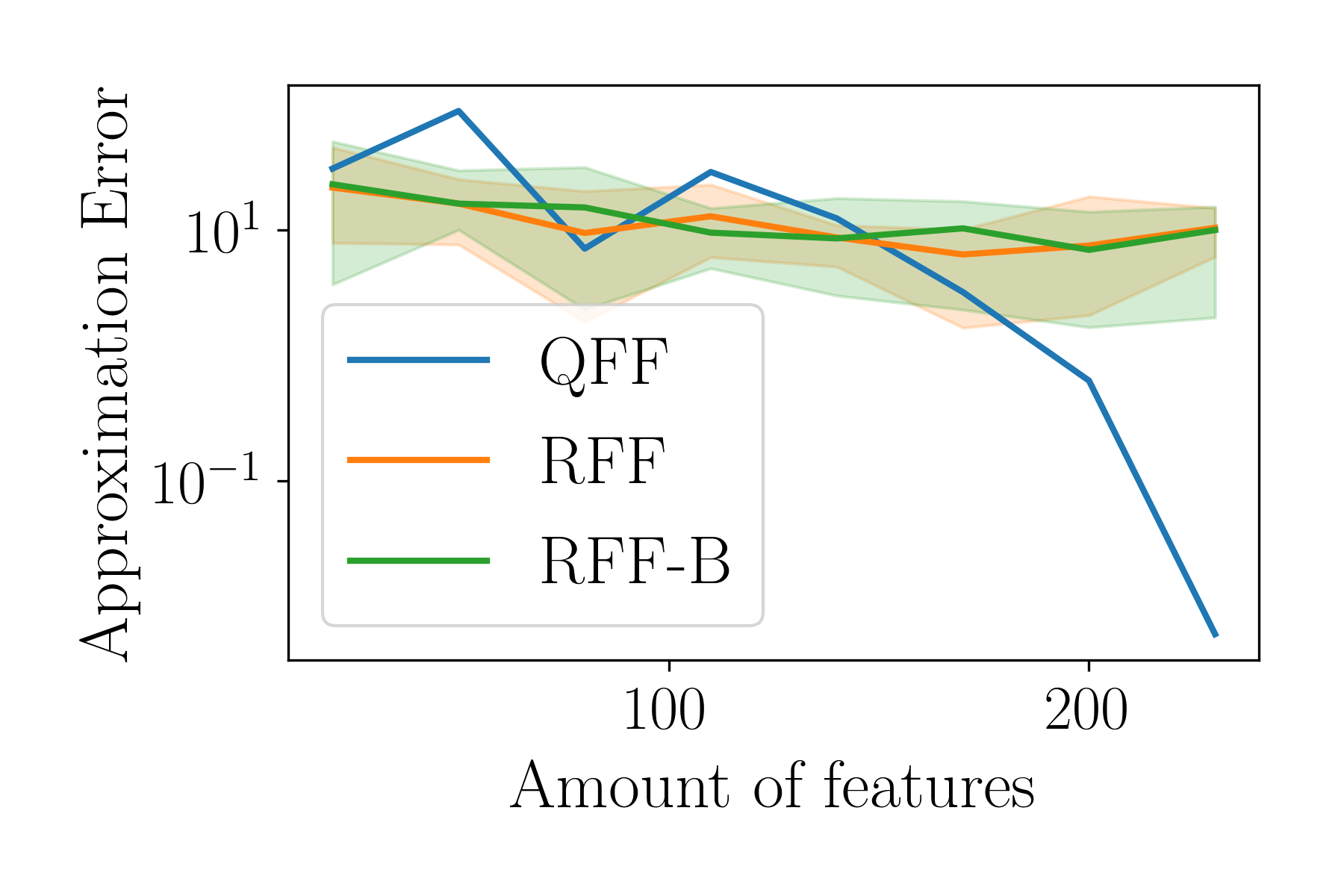}
		\caption{$\mu'_1$}
	\end{subfigure}
	\hfill
	\begin{subfigure}[t]{0.24\textwidth}
		\centering
		\includegraphics[width=\textwidth]{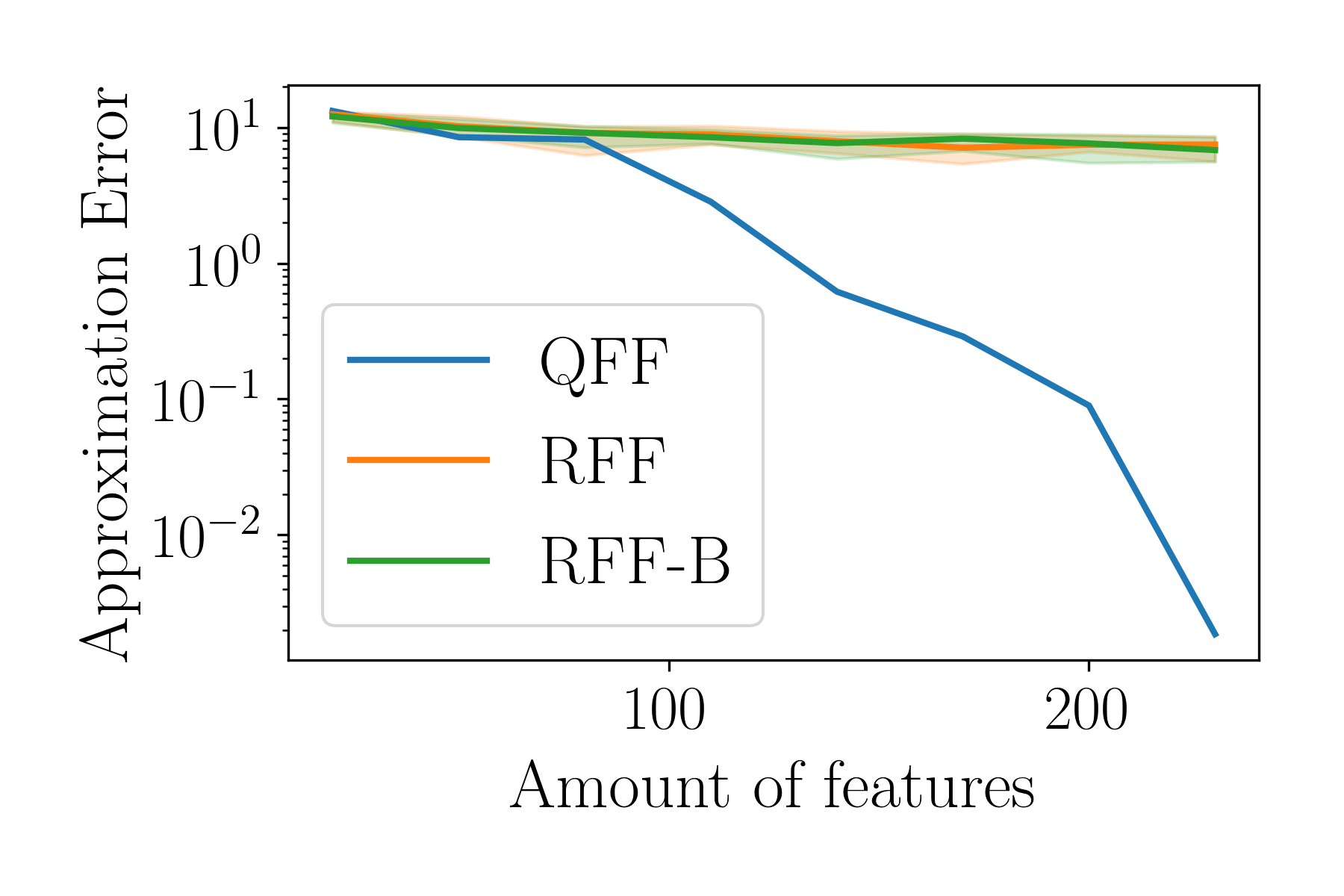}
		\caption{$\Sigma'_1$}
	\end{subfigure}
	\caption{Approximation error of the different feature approximations compared to the accurate GP, evaluated at $t=0.8$ for the Lorenz system with 1000 observations and an SNR of 10. For each feature, we show the median as well as the 12.5\% and 87.5\% quantiles over 10 independent noise realizations, separately for each state dimension.}
\end{figure}

\begin{figure}[!h]
	\centering
	\begin{subfigure}[t]{0.24\textwidth}
		\centering
		\includegraphics[width=\textwidth]{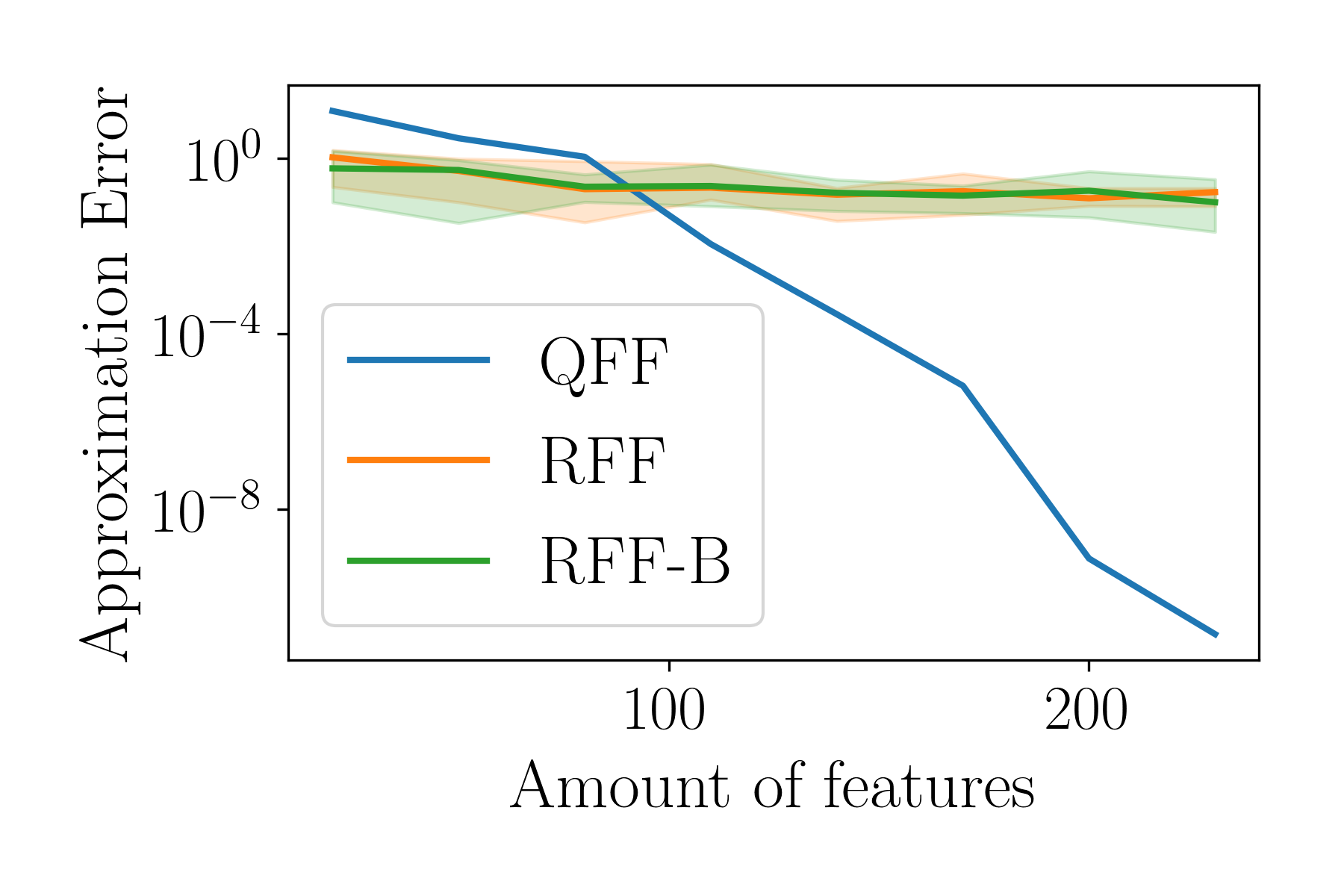}
		\caption{$\mu_0$}
	\end{subfigure}
	\hfill
	\begin{subfigure}[t]{0.24\textwidth}
		\centering
		\includegraphics[width=\textwidth]{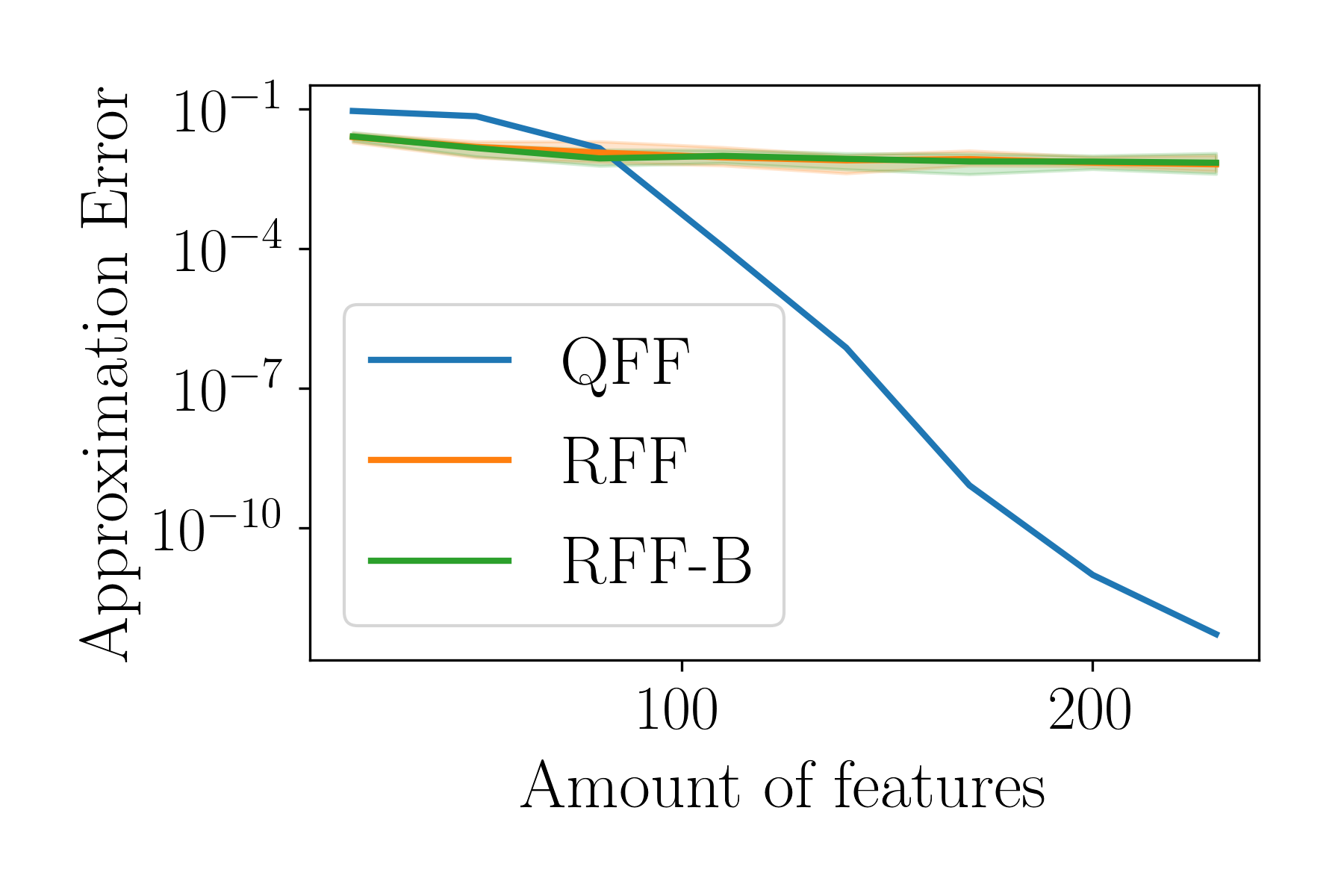}
		\caption{$\Sigma_0$}
	\end{subfigure}
	\hfill
	\begin{subfigure}[t]{0.24\textwidth}
		\centering
		\includegraphics[width=\textwidth]{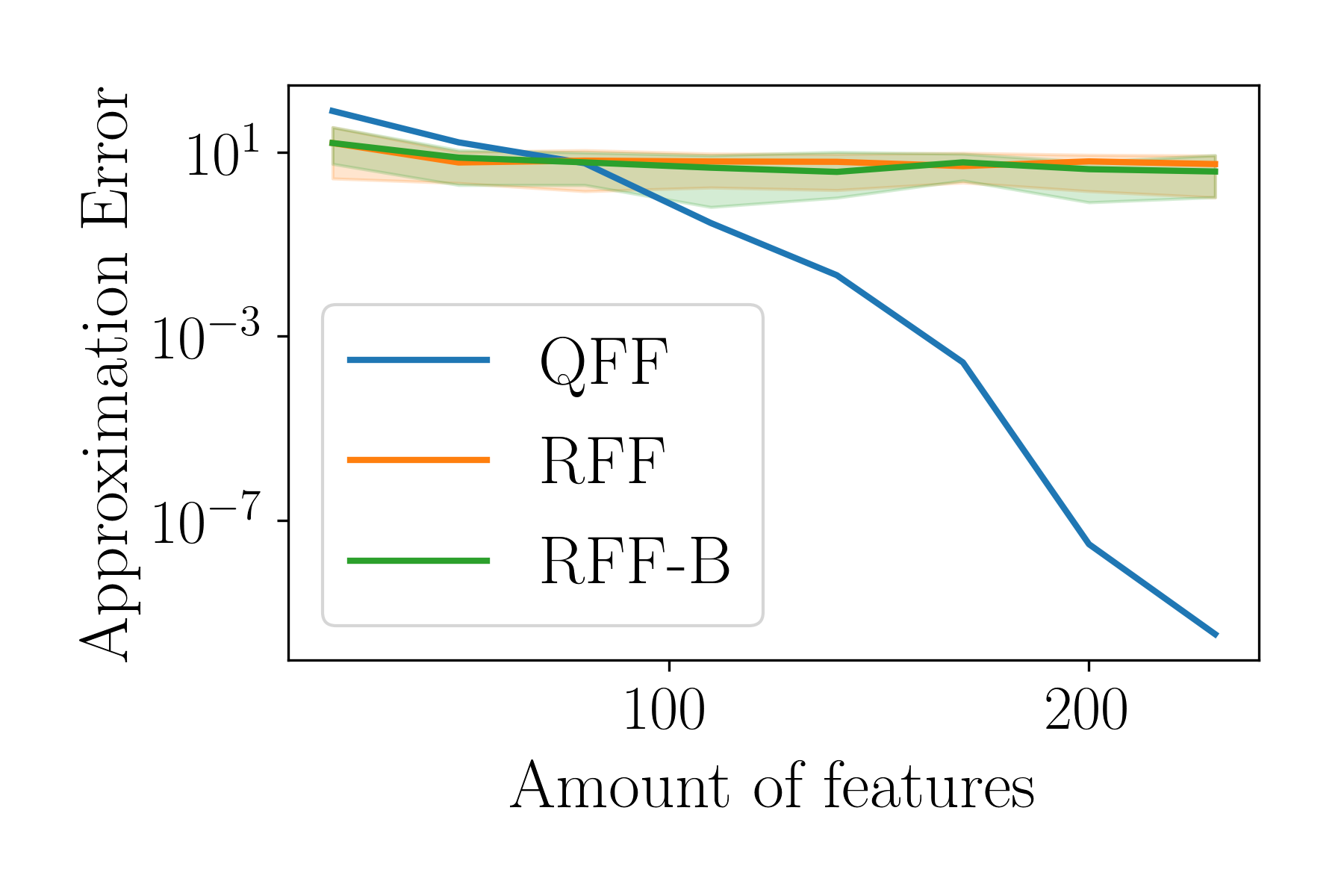}
		\caption{$\mu'_0$}
	\end{subfigure}
	\hfill
	\begin{subfigure}[t]{0.24\textwidth}
		\centering
		\includegraphics[width=\textwidth]{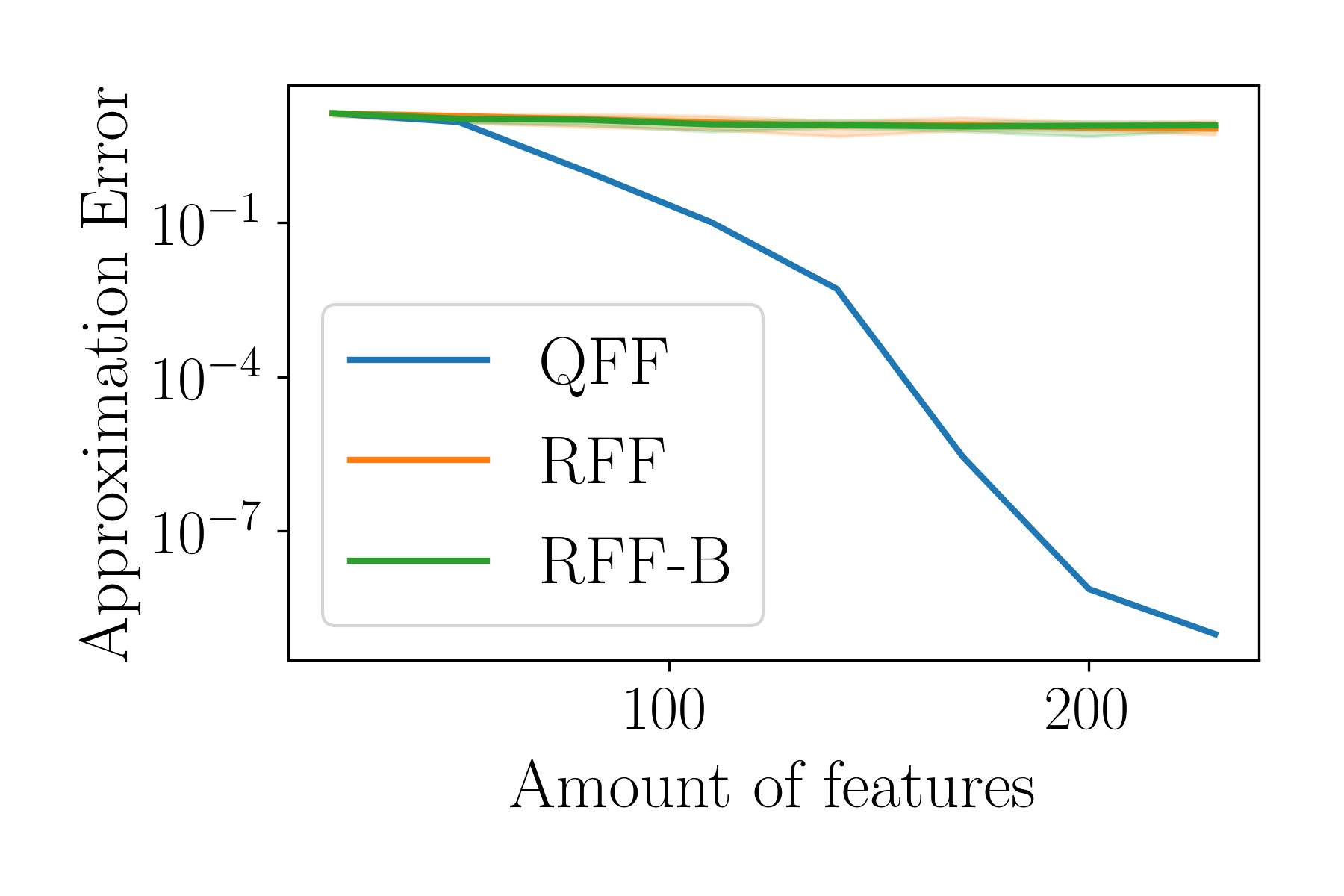}
		\caption{$\Sigma'_0$}
	\end{subfigure}\\
	\begin{subfigure}[t]{0.24\textwidth}
		\centering
		\includegraphics[width=\textwidth]{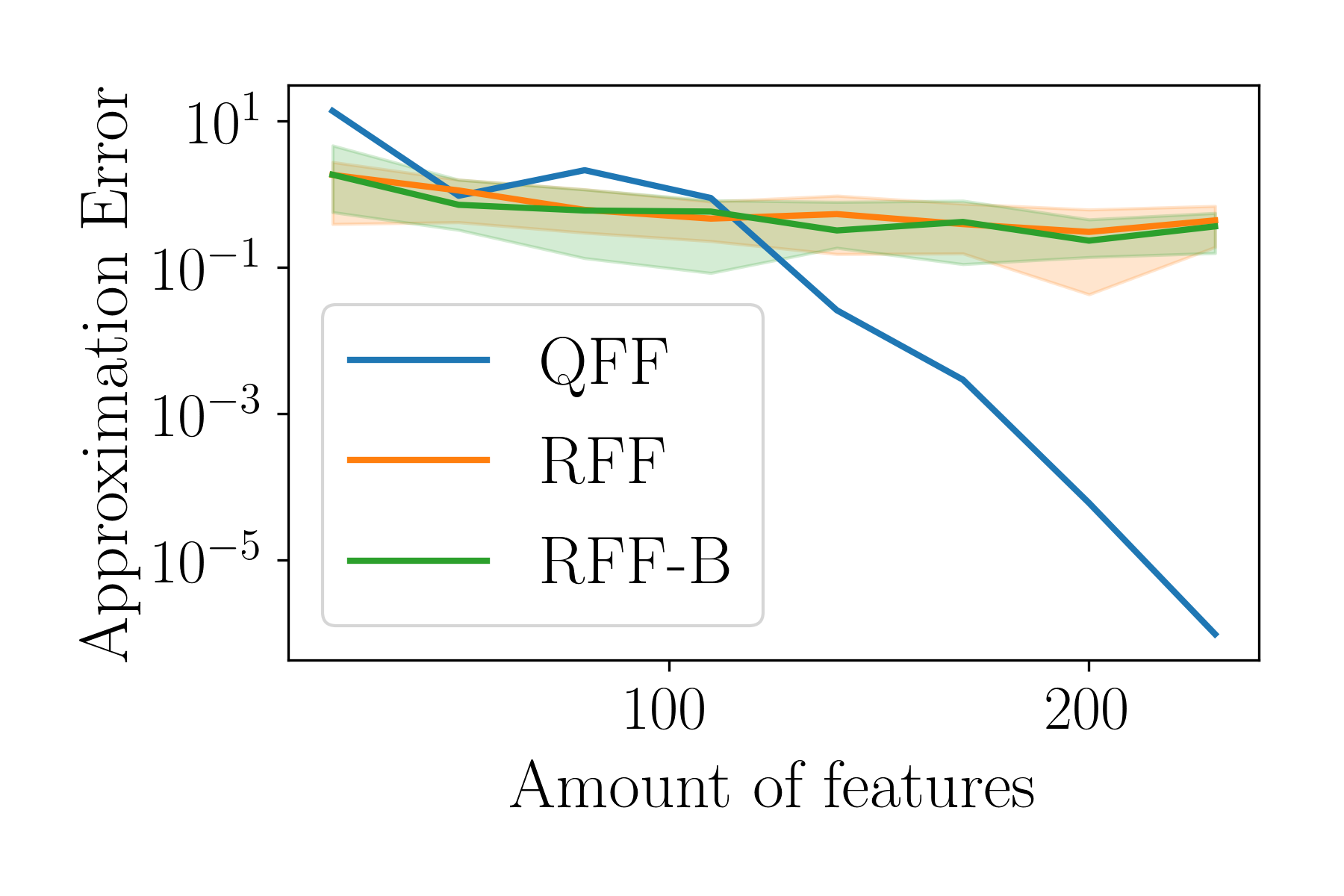}
		\caption{$\mu_1$}
	\end{subfigure}
	\hfill
	\begin{subfigure}[t]{0.24\textwidth}
		\centering
		\includegraphics[width=\textwidth]{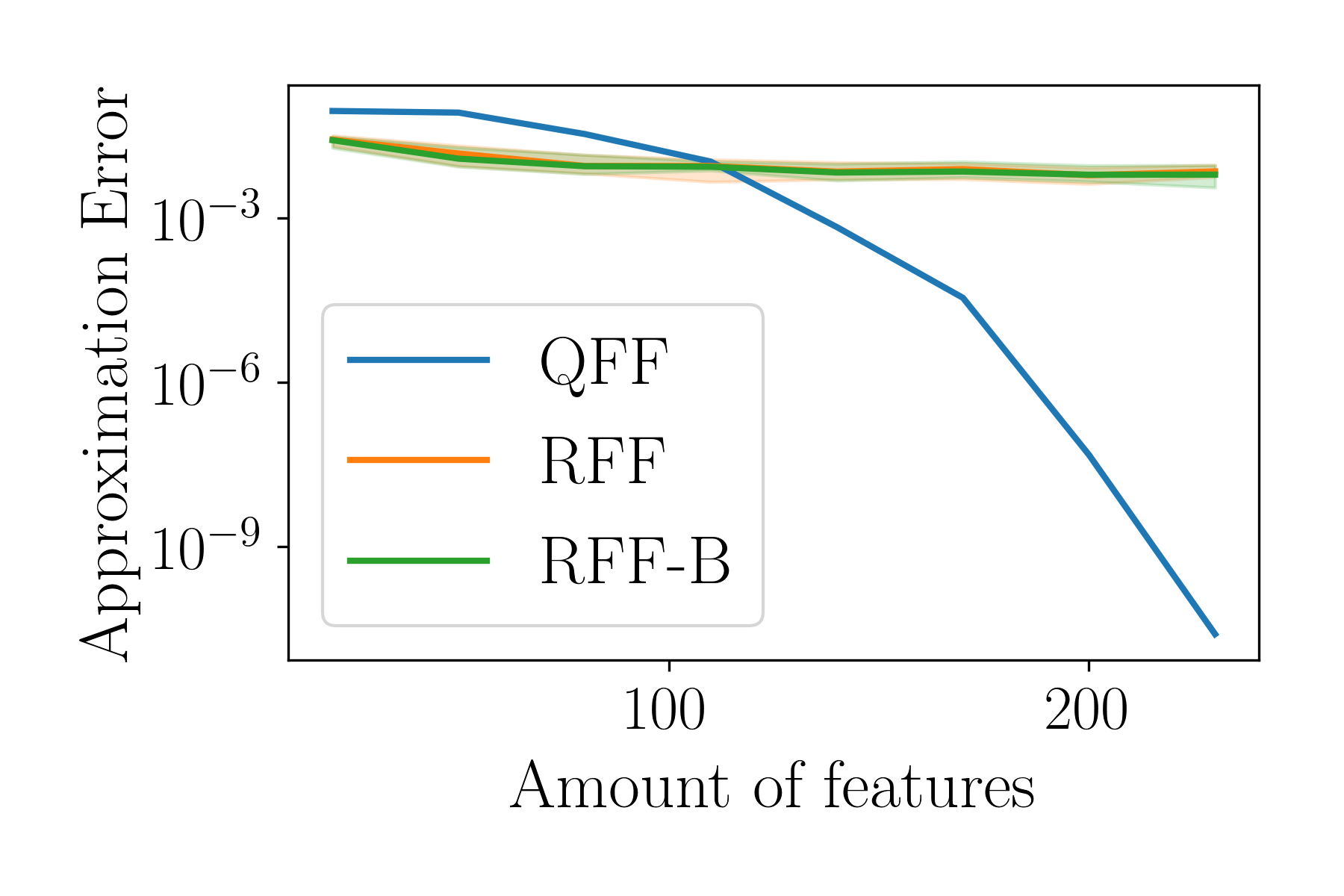}
		\caption{$\Sigma_1$}
	\end{subfigure}
	\hfill
	\begin{subfigure}[t]{0.24\textwidth}
		\centering
		\includegraphics[width=\textwidth]{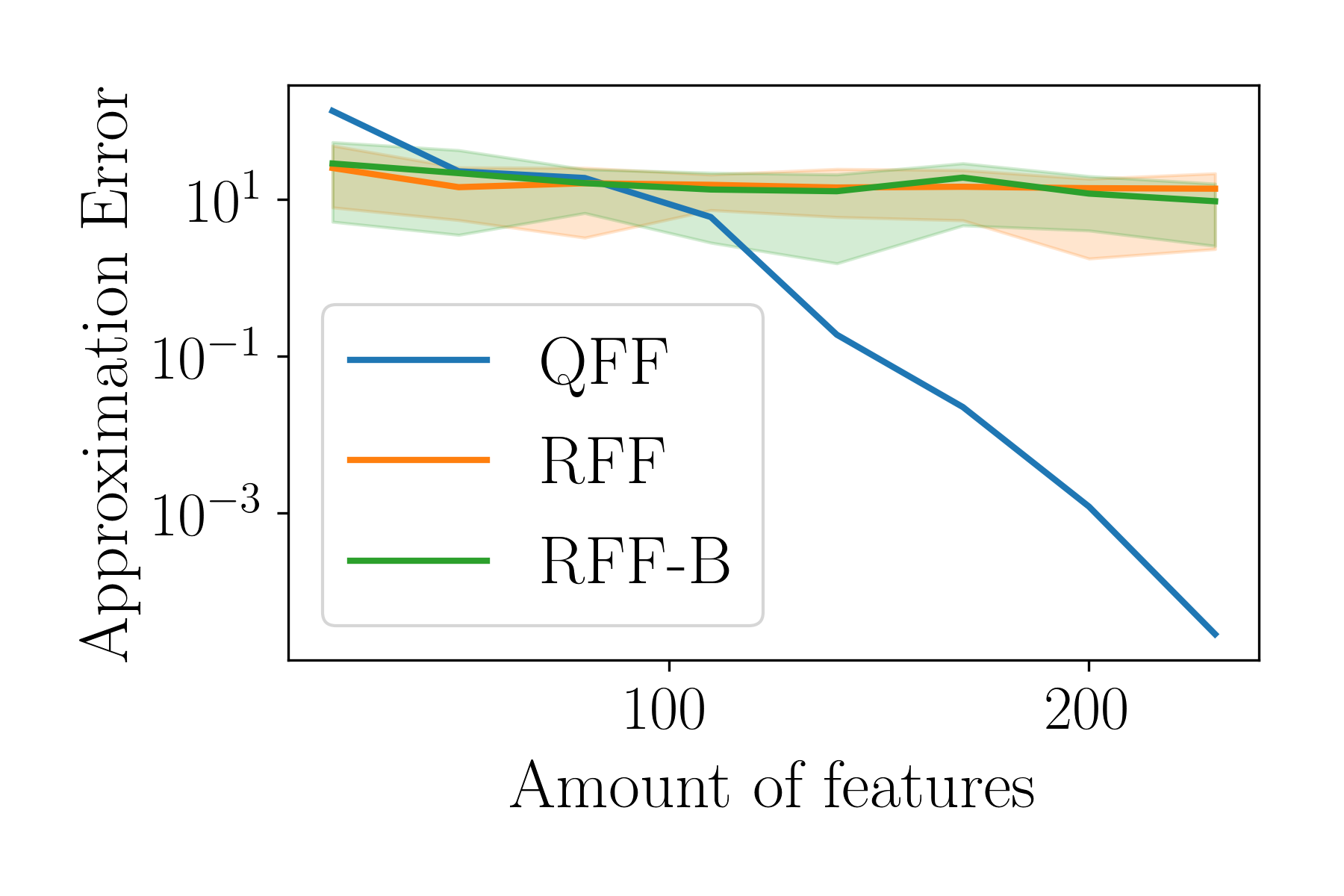}
		\caption{$\mu'_1$}
	\end{subfigure}
	\hfill
	\begin{subfigure}[t]{0.24\textwidth}
		\centering
		\includegraphics[width=\textwidth]{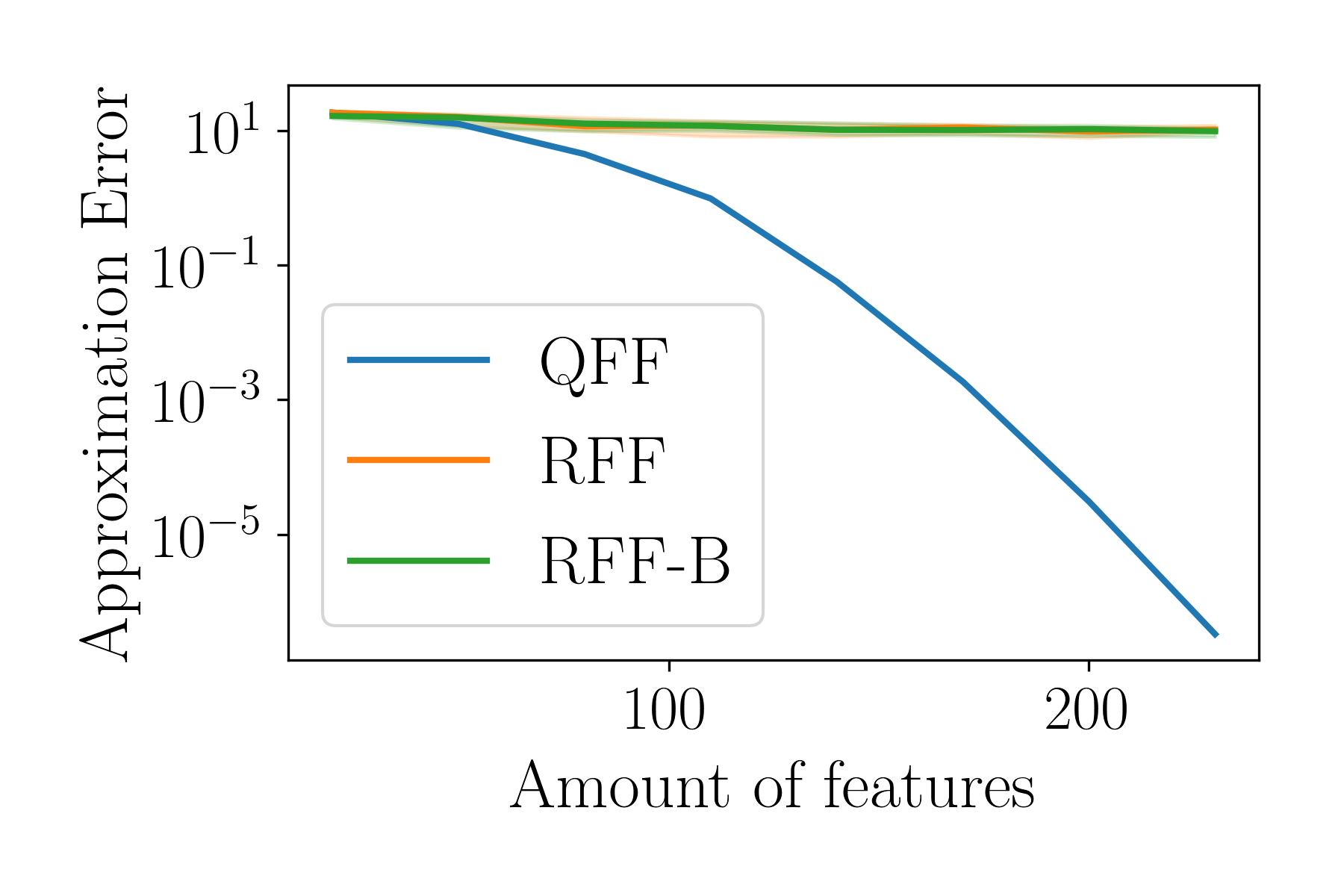}
		\caption{$\Sigma'_1$}
	\end{subfigure}
	\caption{Approximation error of the different feature approximations compared to the accurate GP, evaluated at $t=0.8$ for the Lorenz system with 1000 observations and an SNR of 5. For each feature, we show the median as well as the 12.5\% and 87.5\% quantiles over 10 independent noise realizations, separately for each state dimension.}
\end{figure}

\newpage

\section{Risk Approximation Error Bounds}
\label{sec:AppendixRiskApproxProof}
Let 
\begin{align}
\mathcal{R}_{\lambda \gamma \bm{\phi}}(\bm{x}, \bm{\theta}, \bm{y}) &= \bm{x}^{T}(\bm{C}_{\bm{\phi}} + \lambda\bm{\bm{I}})^{-1}\bm{x}\\
&+ (\bm{x} - \bm{y})^T \sigma^{-2}(\bm{x} - \bm{y}) \\
&+ (\bm{f}(\bm{x}, \bm{\theta}) - \bm{\bm{D}}\bm{x})^{T} (\bm{A} + \gamma \bm{\bm{I}})^{-1}(\bm{f}(\bm{x}, \bm{\theta}) - \bm{\bm{D}}\bm{x})
\end{align}

where $\bm{C}_{\bm{\phi}}[i,j]= \rho e^{-\frac{r_{ij}^2}{2l^2}}$ for some fixed hyperparameters $\bm{\phi}=(\rho,l)$, which denote the variance and the lengthscale. Let $n$ be the number of data points and $m$ be the order of the Quadrature scheme used to approximate the kernel. By writing $\bm{\tilde{C}}_{\bm{\phi}}$, $\tilde{\bm{A}}$ and $\tilde{\bm{D}}$ for the approximated quantities as described in Section \ref{sec:Application}, we get get the approximate risk function
\begin{align}
\tilde{\mathcal{R}}_{\lambda \gamma \bm{\phi}}(\bm{x},\bm{\theta}) :=& \bm{x}^{T}(\tilde{\bm{C}}_{\bm{\phi}} + \lambda \bm{I})^{-1}\bm{x} \\
&+ (\bm{x} - \bm{y})^{T}\sigma^{-2}(\bm{x} - \bm{y})\\
&+ (f(\bm{x},\bm{\theta}) -\tilde{\bm{D}}\bm{x})^{T}(\tilde{\bm{A}}+\gamma \bm{I})^{-1}(\bm{f}(\bm{x},\bm{\theta}) -\tilde{\bm{D}}\bm{x}).
\end{align}

To prove Theorem \ref{theo:Consistency}, we show that $\tilde{\mathcal{R}}$ converges to $\mathcal{R}$ in the relative error sense: As $m$ increases, the relative error $\frac{|\mathcal{R}-\tilde{\mathcal{R}}|}{\mathcal{R}}$ will become arbritrarily small. Theorem \ref{theo:Consistency} is restated here as Theorem \ref{thm:Proof bound}.

\begin{theorem}\label{thm:Proof bound}
	Let $\mathcal{R}$ and $\tilde{\mathcal{R}}$ be defined as above. The parameters $\lambda$ and $\gamma$, the kernel hyperparametes $\bm{\phi}=(\rho,l)$ and the number of data points $n$ are considered fixed. We assume $n\geq 60$. Consider $1 >\epsilon > 0$. If $m$, the order of the quadrature scheme, is at least
	\begin{equation}
	m\geq 10 + \max \lbrace \frac{e}{2l^2},\log_2(\frac{\rho^2 n^3}{\lambda ^ 2 \gamma l^4 \epsilon}) \rbrace
	\end{equation}
	then we have
	\begin{equation}
	\frac{|\mathcal{R}_{\lambda \gamma \bm{\phi}}(\bm{x},\bm{\theta})-\tilde{\mathcal{R}}_{\lambda \gamma \bm{\phi}}(\bm{x},\bm{\theta})|}{\mathcal{R}_{\lambda \gamma \bm{\phi}}(\bm{x},\bm{\theta})} \leq \epsilon
	\end{equation}
	for any configuration of the variables $\bm{x}$ and $\bm{\theta}$.
\end{theorem}
The lengthscale corresponds to time observations normalized in the $[0,1]$ interval. 
In order to make things a bit less complicated in the calculations, we shall assume that $\gamma \leq 1$,$\lambda \leq 1$,$\rho \geq 1$, $l\leq 1$ (and specifically $l\leq \frac{e}{4}$). If any of these assumptions is violated, we can just substitute the corresponding parameter with $1$ in the previous bound, and the resulting bound will still be valid. Moreover, the logarithm is the binary one and we will simply use $\log$ from now on.
Before we prove the theorem, we will introduce some notations and some preliminary results.

Let $|\bm{K}|_F = \sqrt{\text{tr}(\bm{K}\bm{K}^T)}$ be the Frobenius norm of a matrix $\bm{K}$, $|\bm{K}|_{\max} = \max_{i,j}|\bm{K}_{ij}|$ the max norm of $\bm{K}$ and $\sigma_1(\bm{K}):=|\bm{K}|_2$ the spectral norm of $\bm{K}$ (which is by definition the largest singular value). It holds that $|\bm{K}\bm{x}| \leq \sigma_1(\bm{K})|\bm{x}|$. Moreover, for a $n \times n$ matrix  $\bm{K}$ we have 

\begin{equation}
\sigma_1(\bm{K}) \leq \sqrt{ \sum \sigma_i^2(\bm{K}) } = \sqrt{|\bm{K}|^2_F} \leq \sqrt{n^2|\bm{K}|_{max}^2} = n|\bm{K}|_{max} 
\end{equation}

Specifically, if $\bm{C}_{\bm{\phi}}[i,j]= \rho e^{-\frac{r_{ij}^2}{2l^2}}$, then
\begin{equation}
\sigma_1(\bm{C}_{\bm{\phi}}) \leq \rho n
\end{equation}
while
\begin{equation}
\sigma_1(\bm{C}'_{\bm{\phi}})  \leq \frac{\rho}{l  }n
\end{equation}  
since $\bm{C}'_{\bm{\phi}}[i,j] = -\frac{\rho}{l} \frac{r_{ij}}{l} e^{-\frac{r_{ij}^2}{2l^2}}$ as $xe^{-\frac{x^2}{2}} \leq \frac{1}{\sqrt{e}} \leq 1$.\\
From Woodbury's identity for matrix inversion, for invertible $\bm{K}$ and $\bm{K}+\bm{E}$ we have:
\begin{equation}
\label{eq:Woodbury}
(\bm{K}+\bm{E})^{-1} = \bm{K}^{-1} - \bm{K}^{-1}(\bm{I}+\bm{E}\bm{K}^{-1})^{-1}\bm{E}\bm{K}^{-1}
\end{equation}

Let $\bm{E}_1 :=\bm{E}_1(m)= \tilde{\bm{C}}_{\bm{\phi}} - \bm{C}_{\bm{\phi}}$. We know that $|\bm{E}_1|_{max} \leq \sqrt{\frac{\pi}{2}}\frac{1}{m^m}(\frac{e}{4l^2})^m $  so for $a:=\frac{e}{4l^2} $ we have 
\begin{equation}
\sigma_1(\bm{E}_1) \leq 2(\frac{a}{m})^mn
\end{equation}
Now let us bound the first term of the risk
\begin{lemma}\label{lem:First term}
	Consider the term $\bm{x}^{T}(\bm{C}_{\bm{\phi}} + \lambda \bm{I})^{-1}\bm{x}$ approximated by $x^{T}(\tilde{\bm{C}}_{\bm{\phi}} + \lambda \bm{I})^{-1}x$. Then for $m\geq M := 7 + \max \lbrace \frac{e}{2l^2},\log(\frac{\rho^2 n^3}{\lambda ^ 2 \gamma l^4 }) \rbrace$ we get
	\begin{equation}
	|\bm{x}^{T}(\bm{C}_{\bm{\phi}} + \lambda \bm{I})^{-1}\bm{x} - \bm{x}^{T}(\tilde{\bm{C}}_{\bm{\phi}} + \lambda \bm{I})^{-1}\bm{x}| \leq \epsilon_1 \bm{x}^{T}(\bm{C}_{\bm{\phi}} + \lambda \bm{I})^{-1}\bm{x}
	\end{equation}
	where $\epsilon_1 = \frac{4n}{\lambda} (\frac{a}{m})^m$.
	\begin{proof}
		From Woodbury's inversion furmula (\ref{eq:Woodbury}), if we set $\bm{K}= \bm{C}_{\bm{\phi}} + \lambda \bm{I}$ and $\bm{E}=\bm{E}_1$ we have $\sigma_1(\bm{K}^{-1}) < \frac{1}{\lambda}$. Regarding the matrix $\bm{I}+\bm{E}\bm{K}^{-1}$, we would like to find a lower bound on the smallest singular value $\sigma_n(\bm{I}+\bm{E}\bm{K}^{-1})$. For any vector $u$ with $|u|=1$ we have
		\begin{equation}
		|(\bm{I}+\bm{E}\bm{K}^{-1})u| \geq |u| - | \bm{E}\bm{K}^{-1}u| \geq 1 - \sigma_1(\bm{E})\sigma_1(\bm{K}^{-1}) \geq 1- \frac{2n}{\lambda}(\frac{a}{m})^m \geq \frac{1}{2}
		\end{equation}
		
		for $m \geq \max \lbrace \frac{e}{2l^2},\log(\frac{4n}{\lambda}) \rbrace$, which is fullfiled if $m\geq M$. So $\sigma_n(\bm{I}+\bm{E}\bm{K}^{-1}) \geq \frac{1}{2}$ and consequently $\sigma_1((\bm{I}+\bm{E}\bm{K}^{-1})^{-1}) \leq 2$. Moreover, $\sigma_1(\bm{K}^{-1}(\bm{I}+\bm{E}\bm{K}^{-1})^{-1}\bm{E}\bm{K}^{-1}) \leq \sigma_1( \bm{K}^{-1})^2\sigma_1 ((\bm{I}+\bm{E}\bm{K}^{-1})^{-1})\sigma_1(\bm{E}) \leq \frac{4n}{\lambda ^ 2} (\frac{a}{m})^m $. So
		\begin{equation}\label{eq:E22 term}
		\sigma_1((\bm{C}_{\bm{\phi}} + \lambda \bm{I})^{-1} - (\tilde{\bm{C}}_{\bm{\phi}} + \lambda \bm{I})^{-1}) \leq \frac{4n}{\lambda ^ 2} (\frac{a}{m})^m
		\end{equation}
		
		In total we have 
		\begin{align}
		&|\bm{x}^{T}(\bm{C}_{\bm{\phi}} + \lambda \bm{I})^{-1}\bm{x} - \bm{x}^{T}(\tilde{\bm{C}}_{\bm{\phi}} + \lambda \bm{I})^{-1}\bm{x}| = |\bm{x}^{T}\bm{K}^{-1}(\bm{I}+\bm{E}\bm{K}^{-1})^{-1}\bm{E}\bm{K}^{-1}  \bm{x}| \leq \\ &|\bm{x}^{T}\bm{K}^{-\frac{1}{2}}| \sigma_1( \bm{K}^{-\frac{1}{2}}(\bm{I}+\bm{E}\bm{K}^{-1})^{-1}\bm{E}\bm{K}^{-\frac{1}{2}} ) |\bm{x}^{T}\bm{K}^{-\frac{1}{2}}| \leq  \\& |\bm{x}^{T}\bm{K}^{-\frac{1}{2}}| |\bm{x}^{T}\bm{K}^{-\frac{1}{2}}| \sigma_1( \bm{K}^{-\frac{1}{2}})^2\sigma_1 ((\bm{I}+\bm{E}\bm{K}^{-1})^{-1})\sigma_1(\bm{E}) \leq \\ & |\bm{x}^{T}(\bm{C}_{\bm{\phi}} + \lambda \bm{I})^{-1}\bm{x}| \frac{4n}{\lambda} (\frac{a}{m})^m
		\end{align}
		
	\end{proof}
\end{lemma} 

Now we define $\bm{E}_2 =\bm{E}_2(m) := \tilde{\bm{A}} - \bm{A}$ and $\bm{E}_3 = \bm{E}_3(m) := \tilde{\bm{D}} - \bm{D}$. We need to find an upper bound for the spectral norm of these two error terms.

\begin{lemma}
	Consider the error terms $\bm{E}_2$ and $\bm{E}_3$ as defined above. Then for $m\geq M := 7 + \max \lbrace \frac{e}{2l^2},\log(\frac{\rho^2 n^3}{\lambda ^ 2 \gamma l^4 }) \rbrace$ we get
	\begin{equation}
	\sigma_1(\bm{E}_2(m+3)) \leq 20\frac{\rho^2 n^3 a^2}{\lambda ^ 2}(\frac{a}{m})^m \quad \sigma_1(\bm{E}_3(m+3)) \leq 10\frac{n^2 \rho a}{\lambda ^ 2} (\frac{a}{m})^m
	\end{equation}
	
	\begin{proof}
		We have 
		\begin{equation}
		\bm{E}_2 =  \tilde{\bm{C}}^{''}_{\bm{\phi}} - \bm{C}^{''}_{\bm{\phi}} + {}^{\prime}\tilde{\bm{C}}_{\bm{\phi}}(\tilde{\bm{C}}_{\bm{\phi}} + \lambda \bm{I})^{-1}\tilde{\bm{C}}'_{\bm{\phi}} - {}^{\prime}\bm{C}_{\bm{\phi}}(\bm{C}_{\bm{\phi}} + \lambda \bm{I})^{-1}\bm{C}'_{\bm{\phi}}
		\end{equation}
		and we define $\bm{E}_{21} := \tilde{\bm{C}}'_{\bm{\phi}} - \bm{C}'_{\bm{\phi}} $, $\bm{E}_{22} := (\tilde{\bm{C}}_{\bm{\phi}} + \lambda \bm{I})^{-1} -(\bm{C}_{\bm{\phi}} + \lambda \bm{I})^{-1} $ and $ \bm{E}_{23} :=  \tilde{\bm{C}}^{''}_{\bm{\phi}} - \bm{C}^{''}_{\bm{\phi}}$. We know that $ |\bm{E}_{21}(m+2)|_{max} \leq 16 a(\frac{a}{m})^m $  so  $\sigma_1(\bm{E}_{21}(m+3)) \leq 16 a(\frac{a}{m})^m n $. Moreover, we know $ |\bm{E}_{23}(m+3)|_{max} \leq 32a^2 (\frac{a}{m})^m$ so  $\sigma_1(\bm{E}_{23}(m+3)) = 32a^2 (\frac{a}{m})^m n $. Finally, $\sigma_1(E_{22}(m)) \leq \frac{4n}{\lambda ^ 2} (\frac{a}{m})^m $ (by (\ref{eq:E22 term})). So
		
		\begin{align}
		& \sigma_1(\bm{E}_2(m+3)) = \\ & \sigma_1(\bm{E}_{23} + ({}^{\prime}\bm{C}_{\bm{\phi}} + \bm{E}^{T}_{21})((\bm{C}_{\bm{\phi}} + \lambda \bm{I})^{-1} + \bm{E}_{22})(\bm{C}'_{\bm{\phi}} + \bm{E}_{21}) - {}^{\prime}\bm{C}_{\bm{\phi}}(\bm{C}_{\bm{\phi}} + \lambda \bm{I})^{-1}\bm{C}'_{\bm{\phi}} ) \leq \\ &
		\sigma_1(\bm{E}_{23}) + 2\sigma_1({}^{\prime}\bm{C}_{\bm{\phi}}) \sigma_1((\bm{C}_{\bm{\phi}} + \lambda \bm{I})^{-1})\sigma_1(\bm{E}_{21}) +
		\sigma_1(\bm{E}_{21})^2\sigma_1((\bm{C}_{\bm{\phi}} + \lambda \bm{I})^{-1}) + \\ & \sigma_1({}^{\prime}\bm{C}_{\bm{\phi}})^2 \sigma_1(\bm{E}_{22}) 
		+  2\sigma_1(\bm{E}_{21})\sigma_1(\bm{E}_{22})\sigma_1({}^{\prime}\bm{C}_{\bm{\phi}}) + \sigma_1(\bm{E}_{21})^2\sigma_1(\bm{E}_{22}) \leq\\
		& 20\frac{\rho^2 n^3 a^2}{\lambda ^ 2}(\frac{a}{m})^m
		\end{align}
		
		Now we will prove for each summand in the last sum that it is at most $\frac{\rho^2 n^3 a^2}{\lambda ^ 2}(\frac{a}{m})^m$ times a constant. We will use the fact that $m \geq M$. Indeed:
		\begin{align}
			 \sigma_1(\bm{E}_{23}) &\leq  32a^2 (\frac{a}{m})^mn \leq \frac{1}{2} \frac{\rho^2 n^3 a^2}{\lambda ^ 2}(\frac{a}{m})^m\\
			 2\sigma_1({}^{\prime}\bm{C}_{\bm{\phi}}) \sigma_1((\bm{C}_{\bm{\phi}} + \lambda \bm{I})^{-1})\sigma_1(\bm{E}_{21}) &\leq 32\frac{\rho a}{l \lambda } (\frac{a}{m})^m n^2 \leq \frac{\rho^2 n^3 a^2}{\lambda ^ 2}(\frac{a}{m})^m\\
			  \sigma_1(\bm{E}_{21})^2\sigma_1((\bm{C}_{\bm{\phi}} + \lambda \bm{I})^{-1}) &\leq 16^2 \frac{a^2 n^2}{\lambda} (\frac{a}{m})^{2m} \leq \frac{1}{2} \frac{\rho^2 n^3 a^2}{\lambda ^ 2}(\frac{a}{m})^m\\
			 \sigma_1({}^{\prime}\bm{C}_{\bm{\phi}})^2 \sigma_1(\bm{E}_{22}) &\leq 16 \frac{\rho^2 n^3 a}{\lambda^2} (\frac{a}{m})^m \leq 16 \frac{\rho^2 n^3 a^2}{\lambda ^ 2}(\frac{a}{m})^m\\
			 2\sigma_1(\bm{E}_{21})\sigma_1(\bm{E}_{22})\sigma_1({}^{\prime}\bm{C}_{\bm{\phi}}) &\leq 16 a(\frac{a}{m})^m n \frac{4n}{\lambda ^ 2} (\frac{a}{m})^m \frac{n\rho}{l} \leq 2^9 \frac{a^2 n^2 \rho^2}{\lambda^2}(\frac{a}{m})^{2m} \leq  \frac{\rho^2 n^3 a^2}{\lambda ^ 2}(\frac{a}{m})^m\\
			 \sigma_1(\bm{E}_{21})^2\sigma_1(\bm{E}_{22}) &\leq  2^{10} \frac{a^2n^3}{\lambda^2} (\frac{a}{m})^{3m} \leq  \frac{\rho^2 n^3 a^2}{\lambda ^ 2}(\frac{a}{m})^m
		\end{align}
		Similarly,
		\begin{equation}
		\bm{E}_3 = {}^{\prime}\tilde{\bm{C}_{\bm{\phi}}} (\tilde{\bm{C}}_{\bm{\phi}} + \lambda \bm{I})^{-1} - {}^{\prime}\bm{C}_{\bm{\phi}}(\bm{C}_{\bm{\phi}} + \lambda \bm{I})^{-1} = {}^{\prime}\bm{C}_{\bm{\phi}}\bm{E}_{22} + \bm{E}_{21}^T(\bm{C}_{\bm{\phi}} + \lambda \bm{I})^{-1} + \bm{E}_{21}^T \bm{E}_{22}
		\end{equation}
		Using again that $m \geq M$ we have:
		\begin{align}
			\sigma_1({}^{\prime}\bm{C}_{\bm{\phi}}\bm{E}_{22}) &\leq \frac{4n^2 \rho}{\lambda ^ 2l} (\frac{a}{m})^m \leq 8 \frac{n^2 \rho a}{\lambda ^ 2} (\frac{a}{m})^m\\
			\sigma_1(\bm{E}_{21}^T(\bm{C}_{\bm{\phi}} + \lambda \bm{I})^{-1}) &\leq 16 \frac{n}{\lambda} a(\frac{a}{m})^m \leq \frac{n^2 \rho a}{\lambda ^ 2} (\frac{a}{m})^m\\
			\sigma(\bm{E}_{21}^T \bm{E}_{22}) &\leq 16 n a(\frac{a}{m})^m \frac{4n}{\lambda ^ 2} (\frac{a}{m})^m = 64 \frac{n^2a}{\lambda^2} (\frac{a}{m})^{2m} \leq \frac{n^2 \rho a}{\lambda ^ 2} (\frac{a}{m})^m\\
		\end{align} 
		Thus, we have shown that $\sigma_1(\bm{E}_3)\leq  10\frac{n^2 \rho a}{\lambda ^ 2} (\frac{a}{m})^m$
	\end{proof}
\end{lemma}
For the third term, $(\bm{f}(\bm{x},\bm{\theta}) -\bm{D}\bm{x})^{T}(\bm{A}+\gamma \bm{I})^{-1}(\bm{f}(\bm{x},\bm{\theta}) -\bm{D}\bm{x})$ we have

\begin{align}
& | (\bm{f}(\bm{x},\bm{\theta}) -\tilde{\bm{D}}\bm{x})^{T}(\tilde{\bm{A}}+\gamma \bm{I})^{-1}(\bm{f}(\bm{x},\bm{\theta}) -\tilde{\bm{D}}\bm{x})  - (\bm{f}(\bm{x},\bm{\theta}) -\bm{D}\bm{x})^{T}(\bm{A}+\gamma \bm{I})^{-1}(\bm{f}(\bm{x},\bm{\theta}) -\bm{D}\bm{x}) | = \\ & 
| (\bm{f}(\bm{x},\bm{\theta}) -\tilde{\bm{D}}\bm{x})^{T}(\tilde{\bm{A}}+\gamma \bm{I})^{-1}(\bm{f}(\bm{x},\bm{\theta}) -\tilde{\bm{D}}\bm{x}) - (\bm{f}(\bm{x},\bm{\theta}) -\bm{D}\bm{x})^{T}(\tilde{\bm{A}}+\gamma \bm{I})^{-1}(\bm{f}(\bm{x},\bm{\theta}) -\bm{D}\bm{x}) + \\
& (\bm{f}(\bm{x},\bm{\theta}) -\bm{D}\bm{x})^{T}(\tilde{\bm{A}}+\gamma \bm{I})^{-1}(\bm{f}(\bm{x},\bm{\theta}) -\bm{D}\bm{x}) - (\bm{f}(\bm{x},\bm{\theta}) -\bm{D}\bm{x})^{T}(\bm{A}+\gamma \bm{I})^{-1}(\bm{f}(\bm{x},\bm{\theta}) -\bm{D}\bm{x})| \leq \\
&
| (\bm{f}(\bm{x},\bm{\theta}) -\tilde{\bm{D}}\bm{x})^{T}(\tilde{\bm{A}}+\gamma \bm{I})^{-1}(\bm{f}(\bm{x},\bm{\theta}) -\tilde{\bm{D}}\bm{x}) - (\bm{f}(\bm{x},\bm{\theta}) -\bm{D}\bm{x})^{T}(\tilde{\bm{A}}+\gamma \bm{I})^{-1}(\bm{f}(\bm{x},\bm{\theta}) -\bm{D}\bm{x})| + \\
& |(\bm{f}(\bm{x},\bm{\theta}) -\bm{D}\bm{x})^{T}(\tilde{\bm{A}}+\gamma \bm{I})^{-1}(\bm{f}(\bm{x},\bm{\theta}) -\bm{D}\bm{x}) - (\bm{f}(\bm{x},\bm{\theta}) -\bm{D}\bm{x})^{T}(\bm{A}+\gamma \bm{I})^{-1}(\bm{f}(\bm{x},\bm{\theta}) -\bm{D}\bm{x})|\\
&:= T_1 + T_2
\end{align}
This sum can be bounded by bounding each summand separately.

We start with $T_2$:

\begin{lemma}\label{lem:Third term second}
	Consider the term $T_2$ as defined above. Then for $m \geq M := 7 + \max \lbrace \frac{e}{2l^2},\log(\frac{\rho^2 n^3}{\lambda ^ 2 \gamma l^4 }) \rbrace$, if we use for the approximations quadrature schemes of order $m+3$ we have:
	\begin{equation}\label{eq:T2 bound}
	T_2 \leq \epsilon_{32}(\bm{f}(\bm{x},\bm{\theta}) -\bm{D}\bm{x})^{T}(\bm{A}+\gamma \bm{I})^{-1}(\bm{f}(\bm{x},\bm{\theta}) -\bm{D}\bm{x})
	\end{equation}
	where
	\begin{equation}
	\epsilon_{32} = 40\frac{\rho^2 n^3 a^2}{\lambda ^ 2 \gamma}(\frac{a}{m})^m
	\end{equation}
	
\end{lemma}
\begin{proof}
	As in the proof of Lemma \ref{lem:First term}, we can use Woodbury's identity. Now, we use $\bm{A} + \gamma \bm{I}$ instead of $\bm{C}_{\bm{\phi}} + \lambda \bm{I}$, $\bm{E}_2$ instead of $\bm{E}_1$ and $(\bm{f}(x,\bm{\theta}) -\bm{D}\bm{x})$ instead of $\bm{x}$. This directly yields
	\begin{equation}
	\sigma_1( \bm{K}^{-\frac{1}{2}})^2\sigma_1 ((\bm{I}+\bm{E}\bm{K}^{-1})^{-1})\sigma_1(\bm{E}_2) \leq 40\frac{\rho^2 n^3 a^2}{\lambda ^ 2 \gamma}(\frac{a}{m})^m.
	\end{equation} 
\end{proof}

\begin{lemma}\label{lem:Third term first}
	Consider the term $T_1$ as defined above. Then for $m \geq M := 7 + \max \lbrace \frac{e}{2l^2},\log(\frac{\rho^2 n^3}{\lambda ^ 2 \gamma l^4 }) \rbrace$, if we use for the approximations quadrature schemes of order $m+3$ we have:
	\begin{equation}
	T_1 \leq \epsilon_{31}(\bm{f}(\bm{x},\bm{\theta}) -\bm{D}\bm{x})^{T}(\bm{A}+\gamma \bm{I})^{-1}(\bm{f}(\bm{x},\bm{\theta}) -\bm{D}\bm{x}) +\epsilon_{31}\bm{x}^{T}(\bm{C}_{\bm{\phi}} + \lambda \bm{I})^{-1}\bm{x}
	\end{equation}
	where $\epsilon_{31} = 30 \frac{n^{\frac{5}{2}} \rho^{\frac{3}{2}} a}{\lambda ^ 2 \gamma^{\frac{1}{2}}} (\frac{a}{m})^m $.
	\begin{proof}
		It holds that
		\begin{align}
		&\sigma_1((\tilde{\bm{A}}+\gamma \bm{I})^{-\frac{1}{2}})  \sigma_1( \bm{E}_3) \sigma_1((\bm{C}_{\bm{\phi}} + \lambda \bm{I})^{\frac{1}{2}})
		\leq 10\frac{n^2 \rho a}{\lambda ^ 2} (\frac{a}{m})^m (\frac{\rho n +\lambda}{\gamma})^{\frac{1}{2}} \leq \\ & 15 \frac{n^{\frac{5}{2}} \rho^{\frac{3}{2}} a}{\lambda ^ 2 \gamma^{\frac{1}{2}}} (\frac{a}{m})^m =  \frac{\epsilon_{31}}{2}
		\end{align}
		
		Simply using that $\bm{E}_3 = \tilde{\bm{D}} - \bm{D}$ we get:
		\begin{flalign}
		&T_1=\\
		&| (\bm{f}(\bm{x},\bm{\theta}) -\tilde{\bm{D}}\bm{x})^{T}(\tilde{\bm{A}}+\gamma \bm{I})^{-1}(\bm{f}(\bm{x},\bm{\theta}) -\tilde{\bm{D}}\bm{x}) - (\bm{f}(\bm{x},\bm{\theta}) -\bm{D}\bm{x})^{T}(\tilde{\bm{A}}+\gamma \bm{I})^{-1}(\bm{f}(\bm{x},\bm{\theta}) -\bm{D}\bm{x})| = \\
		& | 2(\bm{E}_3\bm{x})^T(\tilde{\bm{A}}+\gamma \bm{I})^{-1}(\bm{f}(\bm{x},\bm{\theta}) -\bm{D}\bm{x}) + (\bm{E}_3\bm{x})^T(\tilde{\bm{A}}+\gamma \bm{I})^{-1}(\bm{E}_3\bm{x}) | \leq \\
		& 2|\bm{E}_3\bm{x}|\sigma_1((\tilde{\bm{A}}+\gamma \bm{I})^{-\frac{1}{2}})|(\tilde{\bm{A}}+\gamma \bm{I})^{-\frac{1}{2}}(\bm{f}(\bm{x},\bm{\theta}) -\bm{D}\bm{x})| + \sigma_1((\tilde{\bm{A}}+\gamma \bm{I})^{-1})|\bm{E}_3\bm{x}|^2
		\end{flalign}
		
		Since $|\bm{E}_3\bm{x}|=|\bm{E}_3(\bm{C}_{\bm{\phi}} + \lambda \bm{I})^{\frac{1}{2}} (\bm{C}_{\bm{\phi}} + \lambda \bm{I})^{-\frac{1}{2}} \bm{x}| \leq \sigma_1(\bm{E}_3) \sigma_1((\bm{C}_{\bm{\phi}} + \lambda \bm{I})^{\frac{1}{2}}) |(\bm{C}_{\bm{\phi}} + \lambda \bm{I})^{-\frac{1}{2}} \bm{x}| $ we have:
		\begin{align}
		&\sigma_1((\tilde{\bm{A}}+\gamma \bm{I})^{-1})|\bm{E}_3\bm{x}|^2 \leq \sigma_1((\tilde{\bm{A}}+\gamma \bm{I})^{-1}) \sigma_1(\bm{E}_3)^2 \sigma_1((\bm{C}_{\bm{\phi}} + \lambda \bm{I})^{\frac{1}{2}})^2 |(\bm{C}_{\bm{\phi}} + \lambda \bm{I})^{-\frac{1}{2}} \bm{x}|^2 \leq \\ & \frac{\epsilon_{31}^2}{4} |(\bm{C}_{\bm{\phi}} + \lambda \bm{I})^{-\frac{1}{2}} \bm{x}|^2 \leq \frac{\epsilon_{31}}{2} |(\bm{C}_{\bm{\phi}} + \lambda \bm{I})^{-\frac{1}{2}} \bm{x}|^2
		\end{align}
		
		Note that $|(\bm{C}_{\bm{\phi}} + \lambda \bm{I})^{-\frac{1}{2}} \bm{x}|^2$ is the first term of $\mathcal{R}_{\lambda \gamma \bm{\phi}}(\bm{x},\bm{\theta})$
		
		Moreover 
		\begin{align}
		&2|\bm{E}_3\bm{x}|\sigma_1((\tilde{\bm{A}}+\gamma \bm{I})^{-\frac{1}{2}})|(\tilde{\bm{A}}+\gamma \bm{I})^{-\frac{1}{2}}(f(\bm{x},\bm{\theta}) -\bm{D}\bm{x})| = \\ &
		\sigma_1((\tilde{\bm{A}}+\gamma \bm{I})^{-\frac{1}{2}}) 2 |\bm{E}_3(\bm{C}_{\bm{\phi}} + \lambda \bm{I})^{\frac{1}{2}} (\bm{C}_{\bm{\phi}} + \lambda \bm{I})^{-\frac{1}{2}} \bm{x}|| (\tilde{\bm{A}}+\gamma \bm{I})^{-\frac{1}{2}}(\bm{f}(\bm{x},\bm{\theta}) -\bm{D}\bm{x})|\leq \\
		&\sigma_1((\tilde{\bm{A}}+\gamma \bm{I})^{-\frac{1}{2}})  \sigma_1( \bm{E}_3) \sigma_1((\bm{C}_{\bm{\phi}} + \lambda \bm{I})^{\frac{1}{2}}) 2 | (\bm{C}_{\bm{\phi}} + \lambda \bm{I})^{-\frac{1}{2}} \bm{x}| | (\tilde{\bm{A}}+\gamma \bm{I})^{-\frac{1}{2}}(\bm{f}(\bm{x},\bm{\theta}) -\bm{D}\bm{x})|
		\leq \\ 
		& \frac{\epsilon_{31}}{2} (| (\bm{C}_{\bm{\phi}} + \lambda \bm{I})^{-\frac{1}{2}} \bm{x}| ^2 + | (\tilde{\bm{A}}+\gamma \bm{I})^{-\frac{1}{2}}(\bm{f}(\bm{x},\bm{\theta}) -\bm{D}\bm{x})|^2) \leq \\
		&  \frac{\epsilon_{31}}{2} (| (\bm{C}_{\bm{\phi}} + \lambda \bm{I})^{-\frac{1}{2}} \bm{x}| ^2 + (1+\epsilon_{32}) | (\bm{A}+\gamma \bm{I})^{-\frac{1}{2}}(\bm{f}(\bm{x},\bm{\theta}) -\bm{D}\bm{x})|^2) \quad \text{from  (\ref{eq:T2 bound}) }  \leq \\
		&  \frac{\epsilon_{31}}{2}   | (\bm{C}_{\bm{\phi}} + \lambda \bm{I})^{-\frac{1}{2}} \bm{x}| ^2 +  \epsilon_{31} | (\bm{A}+\gamma \bm{I})^{-\frac{1}{2}}(\bm{f}(\bm{x},\bm{\theta}) -\bm{D}\bm{x})|^2 \quad \text{($\epsilon_{32} \leq 1$ for $m\geq M$ )}
		\end{align}
	\end{proof}
\end{lemma}

Combining all of these pieces, we can now proof the original Theorem.
\begin{proof}[Proof of Theorem \ref{thm:Proof bound}]
	
	Consider $m \geq M := 7 + \max \lbrace \frac{e}{2l^2},\log(\frac{\rho^2 n^3}{\lambda ^ 2 \gamma l^4 }) \rbrace $ and let $m^{\prime} = m+3$. Then, if we apply quadrature schemes of order $m^\prime$ for $\tilde{\mathcal{R}}$ ,using the results from Lemmata ((\ref{lem:First term}), (\ref{lem:Third term first}) and (\ref{lem:Third term second})) and that $\epsilon_{1} \leq \epsilon_{32} $ and $\epsilon_{31} \leq \epsilon_{32} $ we get
	\begin{equation}
	\frac{|\mathcal{R}_{\lambda \gamma \bm{\phi}}(x,\bm{\theta})-\tilde{\mathcal{R}}_{\lambda \gamma \bm{\phi}}(\bm{x},\bm{\theta})|}{\mathcal{R}_{\lambda \gamma \bm{\phi}}(\bm{x},\bm{\theta})} \leq 2\epsilon_{32} = 80\frac{\rho^2 n^3 a^2}{\lambda ^ 2 \gamma}(\frac{a}{m})^m \leq 
	50\frac{\rho^2 n^3}{\lambda ^ 2 \gamma l^4}(\frac{a}{m})^m
	\end{equation}
	In order to make that smaller than $\epsilon$ it suffices $m\geq  \max \lbrace \frac{e}{2l^2},\log(50\frac{\rho^2 n^3}{\lambda ^ 2 \gamma l^4 \epsilon}) \rbrace$. So we can choose for the order of the quadrature scheme $m^{\prime}$ to be
	\begin{equation}
	m^{\prime} = 10 + \max \lbrace \frac{e}{2l^2},\log(\frac{\rho^2 n^3}{\lambda ^ 2 \gamma l^4 \epsilon}) \rbrace
	\end{equation}
	
\end{proof}

\newpage

\section{Experimental Setups}
\label{sec:AppendixExperiments}
Here, we will give a brief overview of the experimental setups we used to evaluate $\operatorname{ODIN-S}$.
\subsection{Basic Definitions}
The trajectory RMSE has proven to be an efficient metric to evaluate the quality of a parameter inference scheme, especially in the context of non-identifiable systems. Here, we restate its definition exactly as provided by \citet{wenk2019odin}.
\begin{definition}[Trajectory RMSE]
	\label{def:trajRMSE}
	Let $\hat{\bm{\theta}}$ be the parameters estimated by an algorithm. Let $\mathbf{t}$ be the vector collecting the observation times. Define $\tilde{\mathbf{x}}(t)$ as the trajectory one obtains by integrating the ODEs using the estimated parameters, but the true initial value, i.e.
	\begin{align}
	\tilde{x}(0) &= \mathbf{x}^*(0)\\
	\tilde{x}(t) &= \int_{0}^{t}f(\tilde{x}(s), \hat{\bm{\theta}}) ds
	\end{align}
	and define $\tilde{\mathbf{x}}$ element-wise as its evaluation at observation times $\mathbf{t}$, i.e. $\tilde{\mathbf{x}}_i = \tilde{x}(t_i)$. The trajectory RMSE is then defined as 
	\begin{equation}
	\textrm{tRMSE} \coloneqq \frac{1}{N}||\tilde{\mathbf{x}} - \mathbf{x}||_2
	\end{equation}
	where $||.||_2$ denotes the standard Euclidean 2-norm.
\end{definition}
Additionally, we would like to restate the definition of the signal-to-noise ratio, as it was used in our work to create the observation noise for the Quadrocopter system.
\begin{definition}[Signal to Noise Ratio]
	Let $x(t)$ be a time-continuous signal for a closed time interval. Let $\sigma_x^2$ denote its variance across time. Furthermore, let $\sigma^2$ be the variance of an additive, Gaussian noise signal. Then we define the SNR as the ration of these two variances, i.e.
	\begin{equation}
	\text{SNR} = \frac{\sigma_x^2}{\sigma^2}.
	\end{equation}
\end{definition}
\subsection{Lotka Volterra}
\begin{alignat}{2}
\dot{x}_1(t) &= \enskip && \theta_1 x_1(t) - \theta_2 x_1(t) x_2(t) \nonumber \\
\dot{x}_2(t) &= \enskip-&& \theta_3 x_2(t) + \theta_4 x_1(t) x_2(t) \label{eq:LV_dynamics}.
\end{alignat}
The Lotka Volterra system \citep{lotka1932growth} has become a widely used benchmarking system. Due to its locally linear dynamics \citep{gorbach2017scalable} and relatively tame trajectories, it is a system many algorithms can solve. We follow the standard setting in the literature and use $\bm{\theta}=[2, 1, 4, 1]$ and $\mathbf{x}(0) = [5, 3]$ to generate trajectories over the time interval $[0, 2]$. The dynamics are shown in Equation \eqref{eq:LV_dynamics}.

\subsection{Protein Transduction}

\begin{alignat}{2}
&\dot{S} &&= -\theta_1 S - \theta_2 S R + \theta_3 R_S \nonumber \\
&\dot{dS} &&= \theta_1 S \nonumber \\
&\dot{R} &&= -\theta_2 S R + \theta_3 R_S + \theta_5 \frac{R_{pp}}{\theta_6 + R_{pp}} \nonumber \\
&\dot{R}_S &&= \theta_2 S R - \theta_3 R_S - \theta_4 R_S \nonumber \\
&\dot{R}_{pp} &&= \theta_4 R_S - \theta_5 \frac{R_{pp}}{\theta_6 + R_{pp}} \label{eq:PT_dynamics}
\end{alignat}
A more challenging system was introduced by \citet{vyshemirsky2007bayesian}. Its nonlinear terms and non-stationarity introduce interesting challenges for many collocation methods. We follow the standard setting in the literature and use $\bm{\theta} = [0.07,0.6,0.05,0.3,0.017,0.3]$ and $\mathbf{x}(0) = [1,0,1,0,0]$, but change the time interval to generate trajectories over the time interval $[0, 50]$, since they stay pretty much constant for $t>50$. The dynamics are shown in Equation \eqref{eq:PT_dynamics}.

\subsection{Lorenz 63}
\begin{alignat}{2}
&\dot{x} &&= \theta_0 (y - x)\\
&\dot{y} &&= x (\theta_1 - z) - y\\
&\dot{z} &&= xy - \theta_2 z 
\label{eq:Lorenz_dynamics}
\end{alignat}
The Lorenz 63 system was introduced by \citet{lorenz1963deterministic} to model atmospheric flows. It is an optimal test bed for parameter inference algorithms, as it exhibits chaotic behavior for the parameter settings we chose. Working with chaotic dynamics is notoriously challenging due to high sensitivity to parameter changes and the presence of many local optima. We follow standard literature and use $\bm{\theta} = [10, 28, 8/3]$ and $\mathbf{x}(0)=[1, 1, 1]$ to generate trajectories over the time interval $[0, 1]$. The dynamics are shown in Equation \eqref{eq:Lorenz_dynamics}.

\subsection{Quadrocopter}
\begin{alignat}{2}
&\dot{x}_0 &&= -g \sin(x_7) + x_5 x_1 - x_4 x_2 \nonumber \\
&\dot{x}_1 &&= g \sin(x_6)\cos(x_7) - x_0 x_5 + x_2 x_3 \nonumber \\
&\dot{x}_2 &&= - \frac{u_0 + u_1 + u_2 + u_3}{\theta_0} + g\cos(x_6)\cos(x_7) + x_0 x_4 - \theta_4 x_1 \nonumber \\
&\dot{x}_3 &&= \frac{1}{\theta_1} (\theta_5(-u_0 + u_1 + u_2 - u_3)) + (\theta_2 - \theta_3(\theta_2 + \theta_1))) x_4 x_5)  \nonumber \\
&\dot{x}_4 &&= \frac{1}{\theta_2} (\theta_4(u_0 - u_1 + u_3 - u_4) + (\theta_3(\theta_2 + \theta_1) - \theta_1) x_3 x_5)  \nonumber \\
&\dot{x}_5 &&= \frac{(\theta_1 - \theta_2) x_3 x_4}{\theta_3(\theta_2 + \theta_1)}   \nonumber \\
&\dot{x}_6 &&= x_3 + (x_4\sin(x_6) + \frac{x_5\cos(x_6) \sin(x_7)} {\cos(x_7)}  \nonumber \\
&\dot{x}_7 &&= x_4 \cos(x_6) - x_5 \sin(x_6) \nonumber \\
&\dot{x}_8 &&= \frac{x_4 \sin(x_6) + x_5 \cos(x_6)}{\cos(x_7)} \nonumber \\
&\dot{x}_9 &&= \cos(x_7)\cos(x_8)x_0 + (-\cos(x_6)\sin(x_8) + \sin(x_6)\sin(x_7)\cos(x_8)) x_1 \nonumber \\
& &&+ (\sin(x_6)\sin(x_8)+\cos(x_6)\sin(x_7)\cos(x_8)) x_2 \nonumber \\
&\dot{x}_{10} &&= \cos(x_7)\sin(x_8) x_0 + (\cos(x_6)\cos(x_8)+\sin(x_6)\sin(x_7)\sin(x_8)) x_1 \nonumber \\
& &&+ (\cos(x_6)\sin(x_7)\sin(x_8)-\sin(x_6)\cos(x_8)) x_2 \nonumber \\
&\dot{x}_{11} &&= \sin(x_7) x_0 - \sin(x_6)\cos(x_7) x_1 - \cos(x_6)\cos(x_7) x_2 \label{eq:Quadro_dynamics}
\end{alignat}
As an ultimate benchmark, we introduce a parametric model describing the dynamics of a 6DOF quadrocopter, shown in Equation \eqref{eq:Quadro_dynamics}. Its strongly nonlinear dynamics and the presence of inputs make it a formidable challenge. The states of this system are representing the linear velocities ($x_0, x_1, x_2$), the angular velocities ($x_3, x_4, x_5$), the angles ($x_6, x_7, x_8$) and the position ($x_9, x_{10}, x_{11}$) of the quadrocopter. The four inputs represent the forces applied at the four different propellers. While in principle any input commands could be incorporated, we keep the inputs constant at $u=[0.248, 0.2475, 0.24775, 0.24775]$. This input leads to interesting nonstationary climbing, pitching and rolling behavior. We use $\bm{\theta}=[0.1, 0.00062, 0.00113, 0.9, 0.114, 0.0825, 9.85]$ and $x_i(0)=0$ for $i=0\dots11$ to generate trajectories over the time interval $[0, 15]$.

\newpage

\section{Additional Empirical Evaluation SLEIPNIR}
\label{sec:AppendixExpPlots}
\subsection{tRMSE vs Features}

\subsubsection{Lotka Volterra}

\begin{figure}[!h]
	\centering
	\begin{subfigure}[t]{0.325\textwidth}
		\centering
		\includegraphics[width=\textwidth]{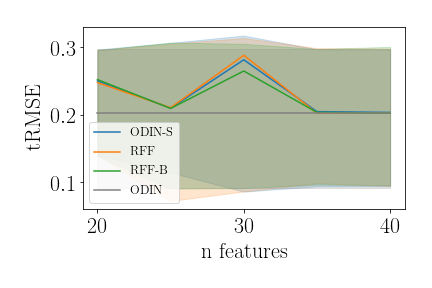}
		\caption{100 obs}
	\end{subfigure}
	\hfill
	\begin{subfigure}[t]{0.325\textwidth}
		\centering
		\includegraphics[width=\textwidth]{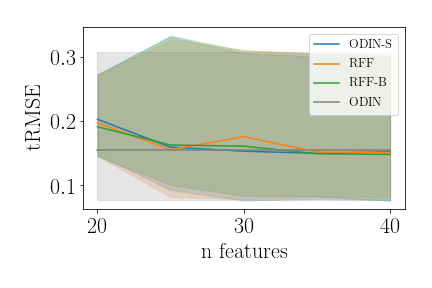}
		\caption{200 obs}
	\end{subfigure}
	\hfill
	\begin{subfigure}[t]{0.325\textwidth}
		\centering
		\includegraphics[width=\textwidth]{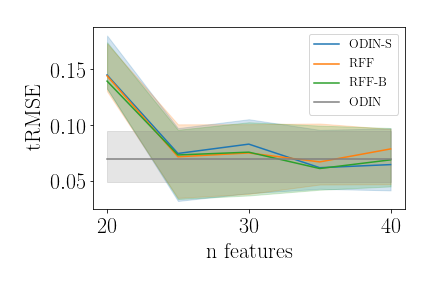}
		\caption{500 obs}
	\end{subfigure}
	\\
	\begin{subfigure}[t]{0.325\textwidth}
		\centering
		\includegraphics[width=\textwidth]{graphs/LV/RMSEVsFeatures/0.1/OneRep_1000_ALL.png}
		\caption{1000 obs}
	\end{subfigure}
	\hfill
	\begin{subfigure}[t]{0.325\textwidth}
		\centering
		\includegraphics[width=\textwidth]{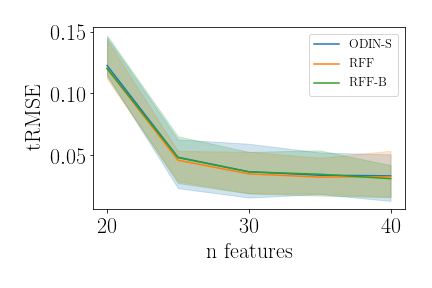}
		\caption{2000 obs}
	\end{subfigure}
	\hfill
	\begin{subfigure}[t]{0.325\textwidth}
		\centering
		\includegraphics[width=\textwidth]{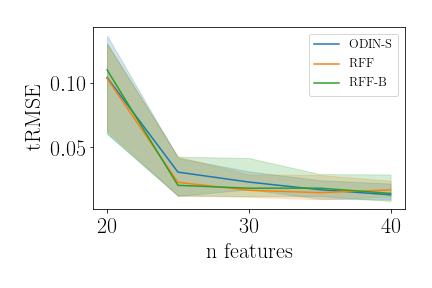}
		\caption{5000 obs}
	\end{subfigure}
	\caption{tRMSE vs features for the Lotka Volterra system using additive observation noise with $\sigma^2=0.1$.}
\end{figure}

\begin{figure}[!h]
	\centering
	\begin{subfigure}[t]{0.325\textwidth}
		\centering
		\includegraphics[width=\textwidth]{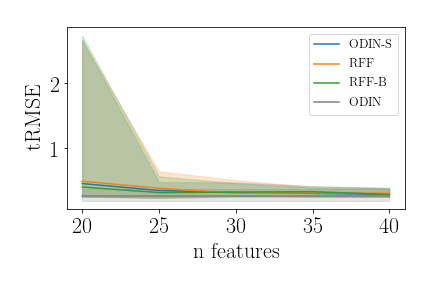}
		\caption{100 obs}
	\end{subfigure}
	\hfill
	\begin{subfigure}[t]{0.325\textwidth}
		\centering
		\includegraphics[width=\textwidth]{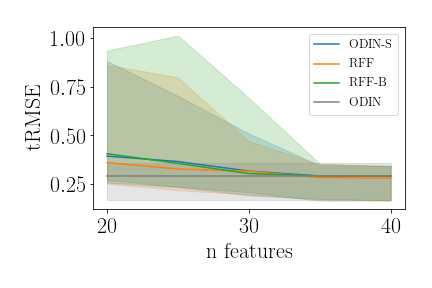}
		\caption{200 obs}
	\end{subfigure}
	\hfill
	\begin{subfigure}[t]{0.325\textwidth}
		\centering
		\includegraphics[width=\textwidth]{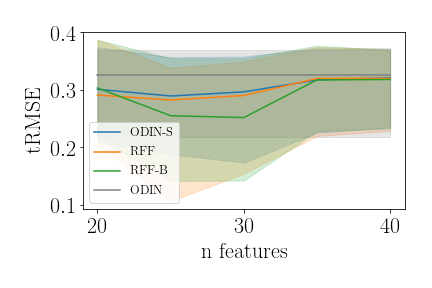}
		\caption{500 obs}
	\end{subfigure}
	\\
	\begin{subfigure}[t]{0.325\textwidth}
		\centering
		\includegraphics[width=\textwidth]{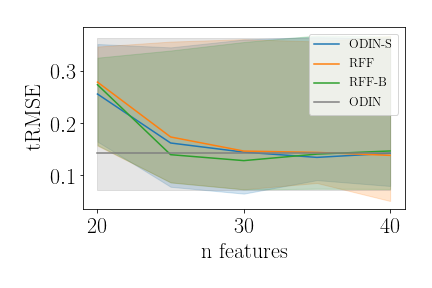}
		\caption{1000 obs}
	\end{subfigure}
	\hfill
	\begin{subfigure}[t]{0.325\textwidth}
		\centering
		\includegraphics[width=\textwidth]{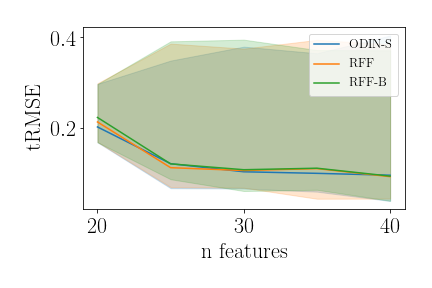}
		\caption{2000 obs}
	\end{subfigure}
	\hfill
	\begin{subfigure}[t]{0.325\textwidth}
		\centering
		\includegraphics[width=\textwidth]{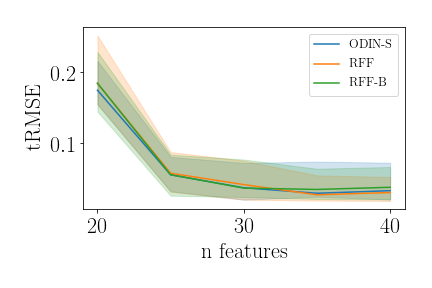}
		\caption{5000 obs}
	\end{subfigure}
	\caption{tRMSE vs features for the Lotka Volterra system using additive observation noise with $\sigma^2=0.5$.}
\end{figure}

\newpage

\subsubsection{Protein Transduction}

\begin{figure}[!h]
	\centering
	\begin{subfigure}[t]{0.24\textwidth}
		\centering
		\includegraphics[width=\textwidth]{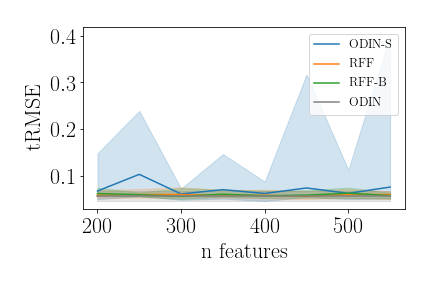}
		\caption{500 obs}
	\end{subfigure}
	\hfill
	\begin{subfigure}[t]{0.24\textwidth}
		\centering
		\includegraphics[width=\textwidth]{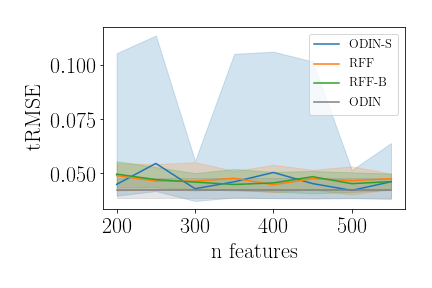}
		\caption{1000 obs}
	\end{subfigure}
	\hfill
	\begin{subfigure}[t]{0.24\textwidth}
		\centering
		\includegraphics[width=\textwidth]{graphs/PT/RMSEVsFeatures/0.01/OneRep_2000_ALL.png}
		\caption{2000 obs}
	\end{subfigure}
	\hfill
	\begin{subfigure}[t]{0.24\textwidth}
		\centering
		\includegraphics[width=\textwidth]{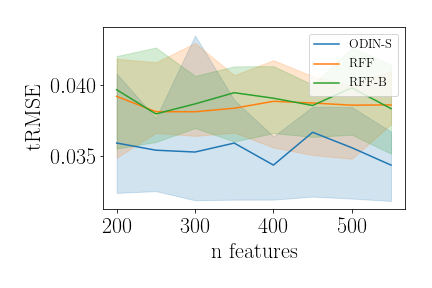}
		\caption{3000 obs}
	\end{subfigure}
	\caption{tRMSE vs features for the Protein Transduction system using additive observation noise with $\sigma^2=0.01$.}
\end{figure}

\begin{figure}[!h]
	\centering
	\begin{subfigure}[t]{0.24\textwidth}
		\centering
		\includegraphics[width=\textwidth]{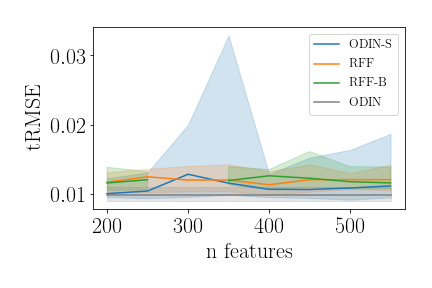}
		\caption{500 obs}
	\end{subfigure}
	\hfill
	\begin{subfigure}[t]{0.24\textwidth}
		\centering
		\includegraphics[width=\textwidth]{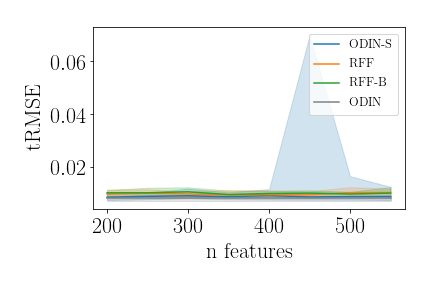}
		\caption{1000 obs}
	\end{subfigure}
	\hfill
	\begin{subfigure}[t]{0.24\textwidth}
		\centering
		\includegraphics[width=\textwidth]{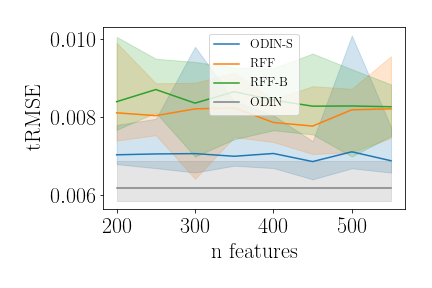}
		\caption{2000 obs}
	\end{subfigure}
	\hfill
	\begin{subfigure}[t]{0.24\textwidth}
		\centering
		\includegraphics[width=\textwidth]{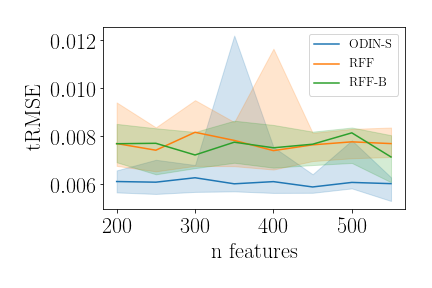}
		\caption{3000 obs}
	\end{subfigure}
	\caption{tRMSE vs features for the Protein Transduction system using additive observation noise with $\sigma^2=0.0001$.}
\end{figure}

\newpage

\subsubsection{Lorenz}

\begin{figure}[!h]
	\centering
	\begin{subfigure}[t]{0.325\textwidth}
		\centering
		\includegraphics[width=\textwidth]{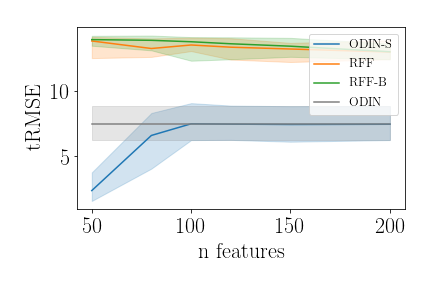}
		\caption{100 obs}
	\end{subfigure}
	\hfill
	\begin{subfigure}[t]{0.325\textwidth}
		\centering
		\includegraphics[width=\textwidth]{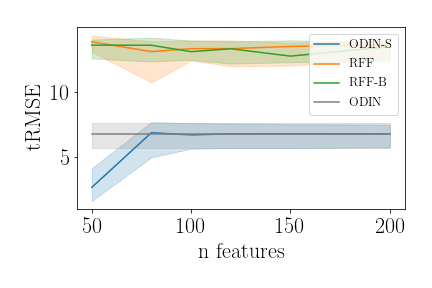}
		\caption{200 obs}
	\end{subfigure}
	\hfill
	\begin{subfigure}[t]{0.325\textwidth}
		\centering
		\includegraphics[width=\textwidth]{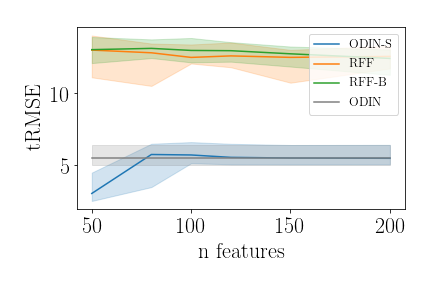}
		\caption{500 obs}
	\end{subfigure}
	\\
	\begin{subfigure}[t]{0.325\textwidth}
		\centering
		\includegraphics[width=\textwidth]{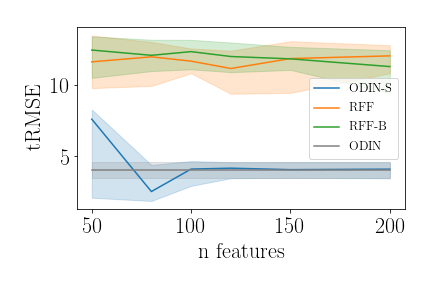}
		\caption{1000 obs}
	\end{subfigure}
	\hfill
	\begin{subfigure}[t]{0.325\textwidth}
		\centering
		\includegraphics[width=\textwidth]{graphs/Lorenz/RMSEVsFeatures/5.0/OneRep_2000_ALL.png}
		\caption{2000 obs}
	\end{subfigure}
	\hfill
	\begin{subfigure}[t]{0.325\textwidth}
		\centering
		\includegraphics[width=\textwidth]{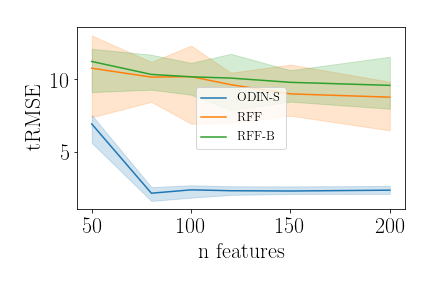}
		\caption{5000 obs}
	\end{subfigure}
	\caption{tRMSE vs features for the Lorenz system with noise created using a signal-to-noise ratio of 5.}
\end{figure}

\begin{figure}[!h]
	\centering
	\begin{subfigure}[t]{0.325\textwidth}
		\centering
		\includegraphics[width=\textwidth]{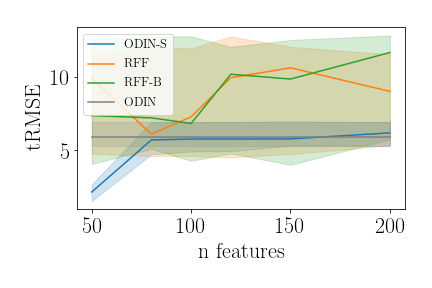}
		\caption{100 obs}
	\end{subfigure}
	\hfill
	\begin{subfigure}[t]{0.325\textwidth}
		\centering
		\includegraphics[width=\textwidth]{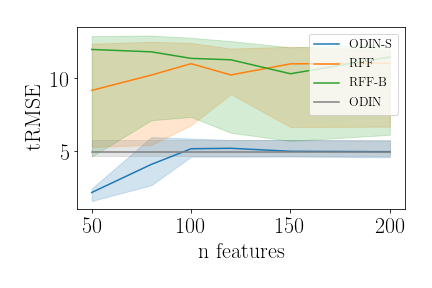}
		\caption{200 obs}
	\end{subfigure}
	\hfill
	\begin{subfigure}[t]{0.325\textwidth}
		\centering
		\includegraphics[width=\textwidth]{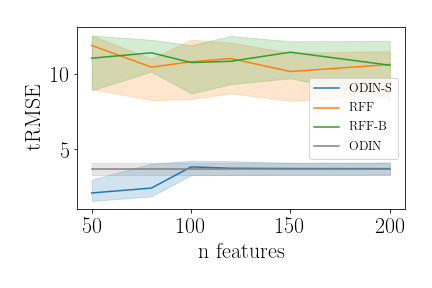}
		\caption{500 obs}
	\end{subfigure}
	\\
	\begin{subfigure}[t]{0.325\textwidth}
		\centering
		\includegraphics[width=\textwidth]{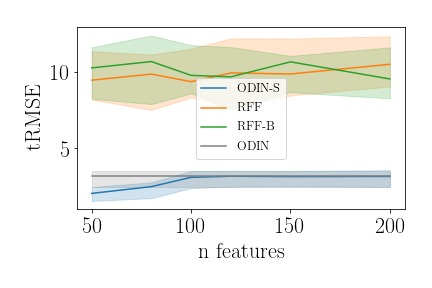}
		\caption{1000 obs}
	\end{subfigure}
	\hfill
	\begin{subfigure}[t]{0.325\textwidth}
		\centering
		\includegraphics[width=\textwidth]{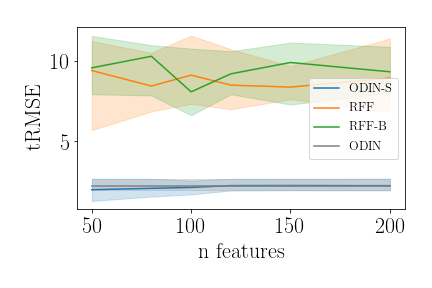}
		\caption{2000 obs}
	\end{subfigure}
	\hfill
	\begin{subfigure}[t]{0.325\textwidth}
		\centering
		\includegraphics[width=\textwidth]{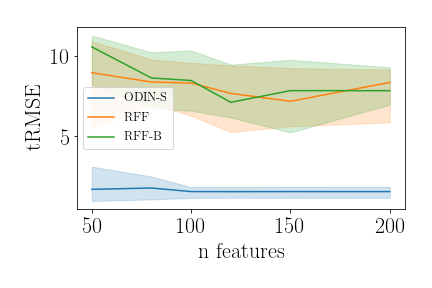}
		\caption{5000 obs}
	\end{subfigure}
	\caption{tRMSE vs features for the Lorenz system with noise created using a signal-to-noise ratio of 10.}
\end{figure}

\newpage

\begin{figure}[!h]
	\centering
	\begin{subfigure}[t]{0.325\textwidth}
		\centering
		\includegraphics[width=\textwidth]{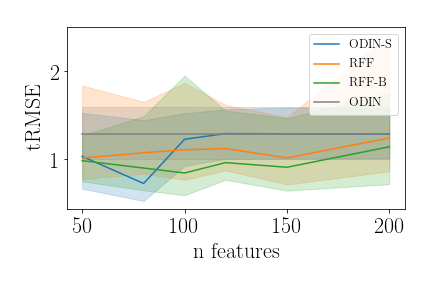}
		\caption{100 obs}
	\end{subfigure}
	\hfill
	\begin{subfigure}[t]{0.325\textwidth}
		\centering
		\includegraphics[width=\textwidth]{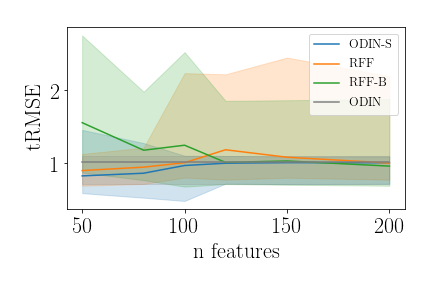}
		\caption{200 obs}
	\end{subfigure}
	\hfill
	\begin{subfigure}[t]{0.325\textwidth}
		\centering
		\includegraphics[width=\textwidth]{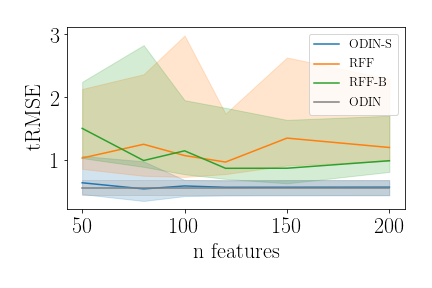}
		\caption{500 obs}
	\end{subfigure}
	\\
	\begin{subfigure}[t]{0.325\textwidth}
		\centering
		\includegraphics[width=\textwidth]{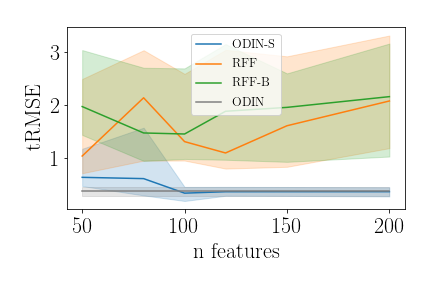}
		\caption{1000 obs}
	\end{subfigure}
	\hfill
	\begin{subfigure}[t]{0.325\textwidth}
		\centering
		\includegraphics[width=\textwidth]{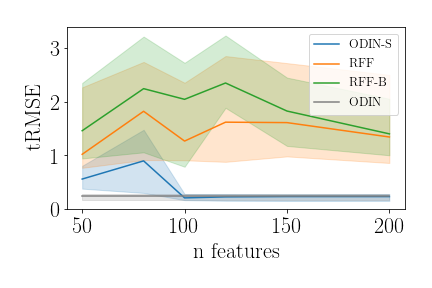}
		\caption{2000 obs}
	\end{subfigure}
	\hfill
	\begin{subfigure}[t]{0.325\textwidth}
		\centering
		\includegraphics[width=\textwidth]{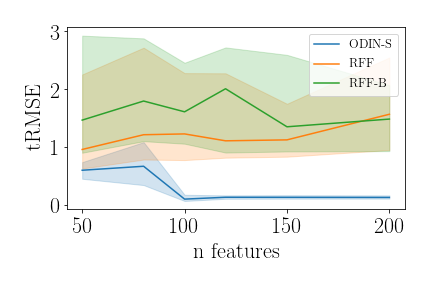}
		\caption{5000 obs}
	\end{subfigure}
	\caption{tRMSE vs features for the Lorenz system with noise created using a signal-to-noise ratio of 100.}
\end{figure}

\newpage

\subsection{Learning Curves}
\subsubsection{Lotka Volterra}

\begin{figure}[!h]
	\centering
	\begin{subfigure}[t]{0.325\textwidth}
		\centering
		\includegraphics[width=\textwidth]{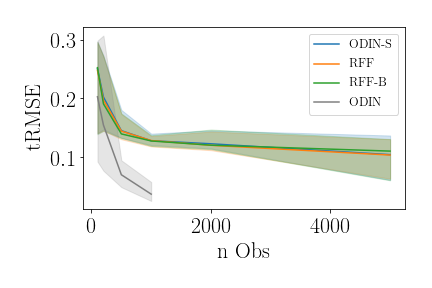}
		\caption{20 features}
	\end{subfigure}
	\hfill
	\begin{subfigure}[t]{0.325\textwidth}
		\centering
		\includegraphics[width=\textwidth]{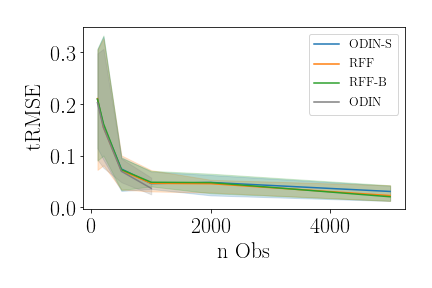}
		\caption{25 features}
	\end{subfigure}
	\hfill
	\begin{subfigure}[t]{0.325\textwidth}
		\centering
		\includegraphics[width=\textwidth]{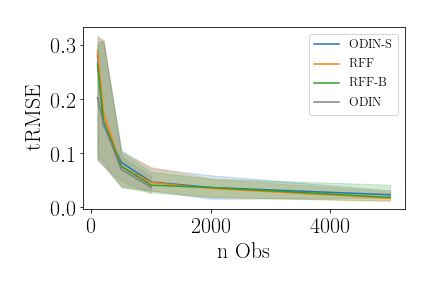}
		\caption{30 features}
	\end{subfigure}
	\\
	\hspace*{\fill}
	\hfill
	\begin{subfigure}[t]{0.325\textwidth}
		\centering
		\includegraphics[width=\textwidth]{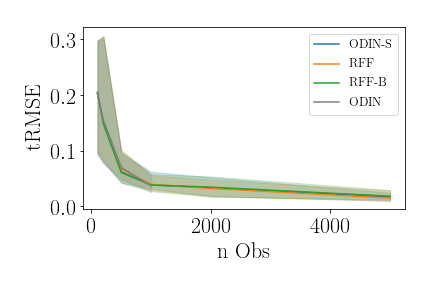}
		\caption{35 features}
	\end{subfigure}
	\hfill
	\begin{subfigure}[t]{0.325\textwidth}
		\centering
		\includegraphics[width=\textwidth]{graphs/LV/RMSEVsObs/0.1/OneRep_40_ALL.png}
		\caption{40 features}
	\end{subfigure}	\hfill \hspace*{\fill}
	\caption{tRMSE vs amount of observations for the Lotka Volterra system with additive noise with $\sigma^2=0.1$.}
\end{figure}

\begin{figure}[!h]
	\centering
	\begin{subfigure}[t]{0.325\textwidth}
		\centering
		\includegraphics[width=\textwidth]{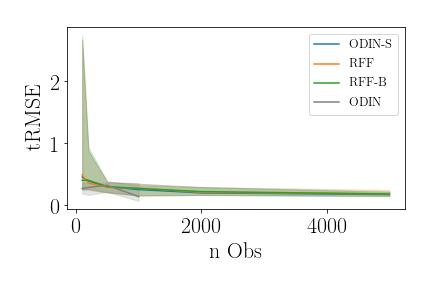}
		\caption{20 features}
	\end{subfigure}
	\hfill
	\begin{subfigure}[t]{0.325\textwidth}
		\centering
		\includegraphics[width=\textwidth]{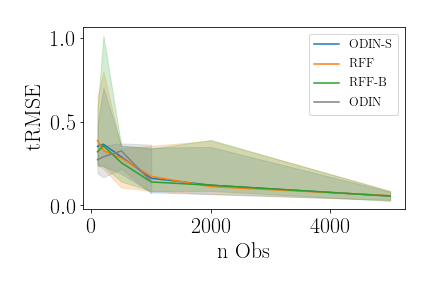}
		\caption{25 features}
	\end{subfigure}
	\hfill
	\begin{subfigure}[t]{0.325\textwidth}
		\centering
		\includegraphics[width=\textwidth]{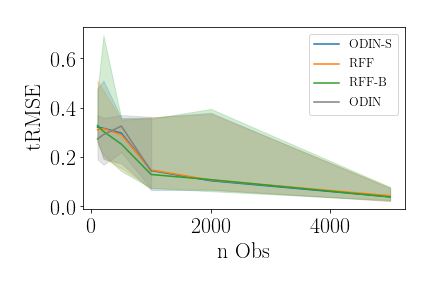}
		\caption{30 features}
	\end{subfigure}
	\\
	\hspace*{\fill}
	\hfill
	\begin{subfigure}[t]{0.325\textwidth}
		\centering
		\includegraphics[width=\textwidth]{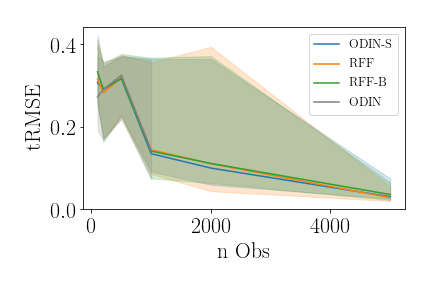}
		\caption{30 features}
	\end{subfigure}
	\hfill
	\begin{subfigure}[t]{0.325\textwidth}
		\centering
		\includegraphics[width=\textwidth]{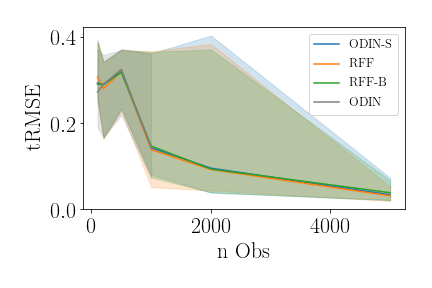}
		\caption{35 features}
	\end{subfigure}	\hfill \hspace*{\fill}
	\caption{tRMSE vs amount of observations for the Lotka Volterra system with additive noise with $\sigma^2=0.5$.}
\end{figure}

\newpage

\subsubsection{Protein Transduction}

\begin{figure}[!h]
	\centering
	\begin{subfigure}[t]{0.325\textwidth}
		\centering
		\includegraphics[width=\textwidth]{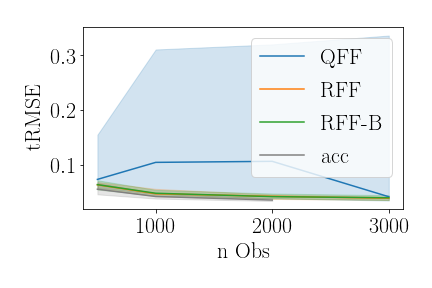}
		\caption{100 features}
	\end{subfigure}
	\hfill
	\begin{subfigure}[t]{0.325\textwidth}
		\centering
		\includegraphics[width=\textwidth]{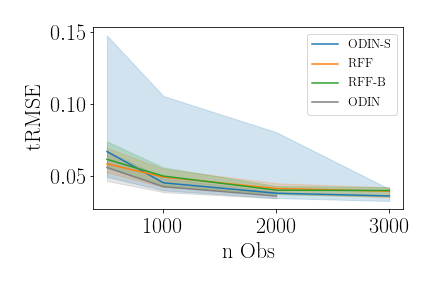}
		\caption{200 features}
	\end{subfigure}
	\hfill
	\begin{subfigure}[t]{0.325\textwidth}
		\centering
		\includegraphics[width=\textwidth]{graphs/PT/RMSEVsObs/0.01/OneRep_300_ALL.png}
		\caption{300 features}
	\end{subfigure}
	\hspace*{\fill}
	\hfill
	\begin{subfigure}[t]{0.325\textwidth}
		\centering
		\includegraphics[width=\textwidth]{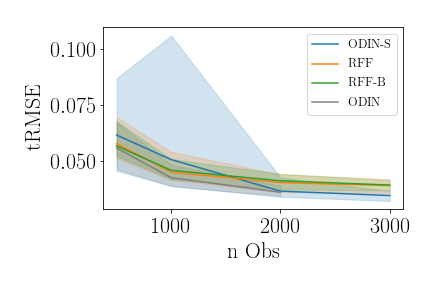}
		\caption{400 features}
	\end{subfigure}
	\hfill
	\begin{subfigure}[t]{0.325\textwidth}
		\centering
		\includegraphics[width=\textwidth]{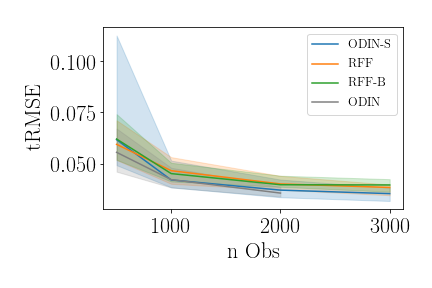}
		\caption{500 features}
	\end{subfigure}
	\hfill
	\hspace*{\fill}
	\caption{tRMSE vs amount of observations for the Protein Transduction system using additive Gaussian noise with $\sigma^2=0.01$.}
\end{figure}

\begin{figure}[!h]
	\centering
	\begin{subfigure}[t]{0.325\textwidth}
		\centering
		\includegraphics[width=\textwidth]{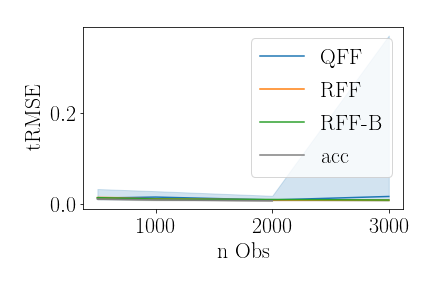}
		\caption{100 features}
	\end{subfigure}
	\hfill
	\begin{subfigure}[t]{0.325\textwidth}
		\centering
		\includegraphics[width=\textwidth]{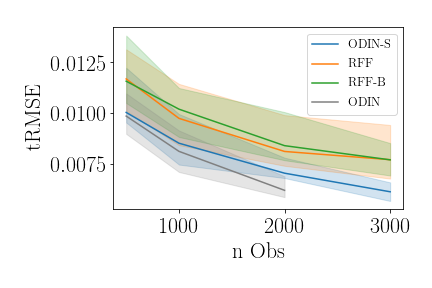}
		\caption{200 features}
	\end{subfigure}
	\hfill
	\begin{subfigure}[t]{0.325\textwidth}
		\centering
		\includegraphics[width=\textwidth]{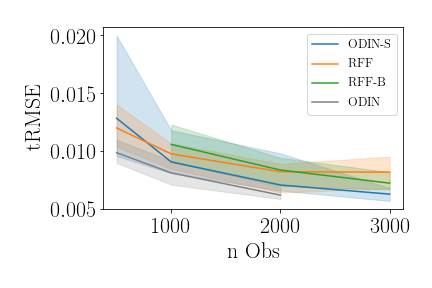}
		\caption{300 features}
	\end{subfigure}
	\hspace*{\fill}
	\hfill
	\begin{subfigure}[t]{0.325\textwidth}
		\centering
		\includegraphics[width=\textwidth]{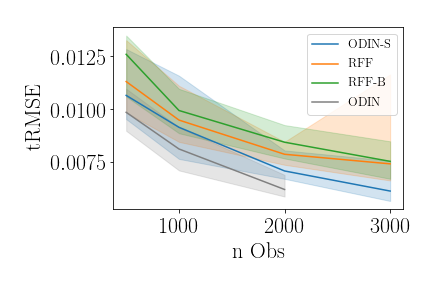}
		\caption{400 features}
	\end{subfigure}
	\hfill
	\begin{subfigure}[t]{0.325\textwidth}
		\centering
		\includegraphics[width=\textwidth]{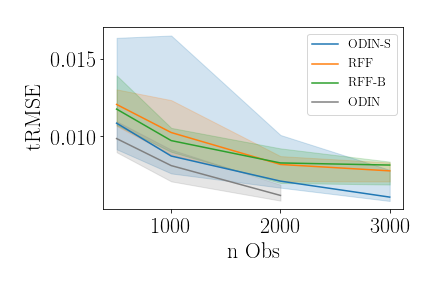}
		\caption{500 features}
	\end{subfigure}
	\hfill
	\hspace*{\fill}
	\caption{tRMSE vs amount of observations for the Protein Transduction system using additive Gaussian noise with $\sigma^2=0.0001$.}
\end{figure}

\newpage

\subsubsection{Lorenz}

\begin{figure}[!h]
	\centering
	\begin{subfigure}[t]{0.325\textwidth}
		\centering
		\includegraphics[width=\textwidth]{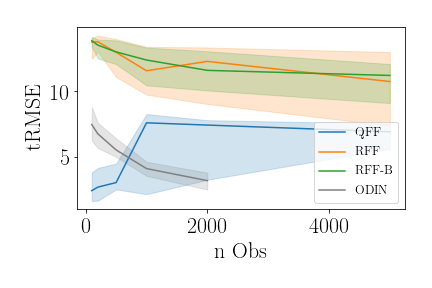}
		\caption{50 features}
	\end{subfigure}
	\hfill
	\begin{subfigure}[t]{0.325\textwidth}
		\centering
		\includegraphics[width=\textwidth]{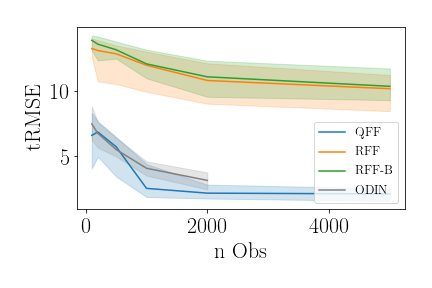}
		\caption{80 features}
	\end{subfigure}
	\hfill
	\begin{subfigure}[t]{0.325\textwidth}
		\centering
		\includegraphics[width=\textwidth]{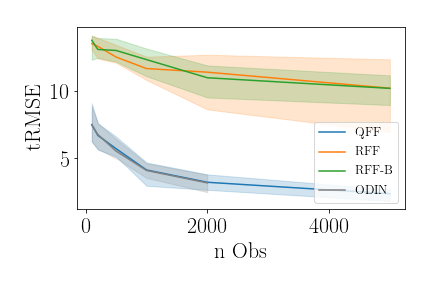}
		\caption{100 features}
	\end{subfigure}
	\\
	\begin{subfigure}[t]{0.325\textwidth}
		\centering
		\includegraphics[width=\textwidth]{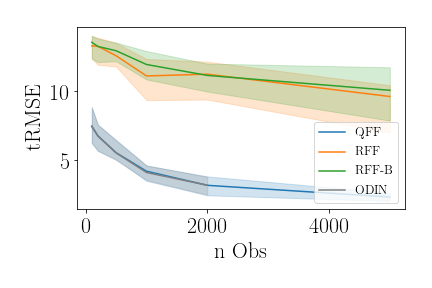}
		\caption{120 features}
	\end{subfigure}
	\hfill
	\begin{subfigure}[t]{0.325\textwidth}
		\centering
		\includegraphics[width=\textwidth]{graphs/Lorenz/RMSEVsObs/5.0/OneRep_150_ALL.png}
		\caption{150 features}
	\end{subfigure}
	\hfill
	\begin{subfigure}[t]{0.325\textwidth}
		\centering
		\includegraphics[width=\textwidth]{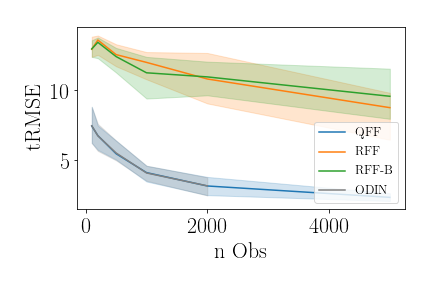}
		\caption{200 features}
	\end{subfigure}
	\caption{tRMSE vs amount of observations for the Lorenz system with noise created using a signal-to-noise ratio of 5.}
\end{figure}

\begin{figure}[!h]
	\centering
	\begin{subfigure}[t]{0.325\textwidth}
		\centering
		\includegraphics[width=\textwidth]{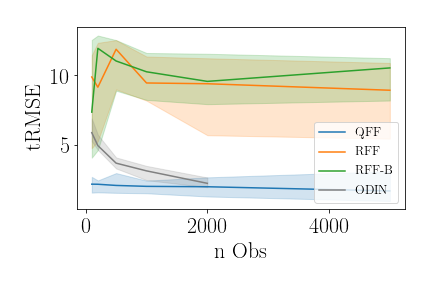}
		\caption{50 features}
	\end{subfigure}
	\hfill
	\begin{subfigure}[t]{0.325\textwidth}
		\centering
		\includegraphics[width=\textwidth]{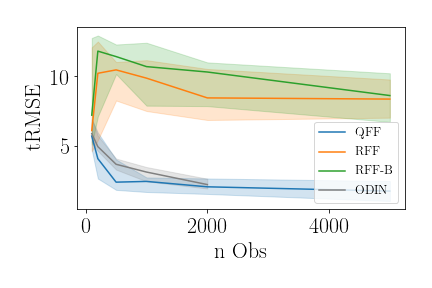}
		\caption{80 features}
	\end{subfigure}
	\hfill
	\begin{subfigure}[t]{0.325\textwidth}
		\centering
		\includegraphics[width=\textwidth]{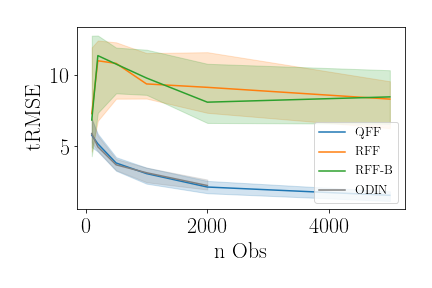}
		\caption{100 features}
	\end{subfigure}
	\\
	\begin{subfigure}[t]{0.325\textwidth}
		\centering
		\includegraphics[width=\textwidth]{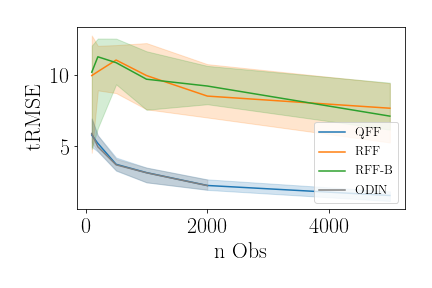}
		\caption{120 features}
	\end{subfigure}
	\hfill
	\begin{subfigure}[t]{0.325\textwidth}
		\centering
		\includegraphics[width=\textwidth]{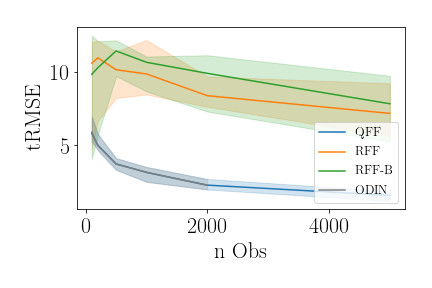}
		\caption{150 features}
	\end{subfigure}
	\hfill
	\begin{subfigure}[t]{0.325\textwidth}
		\centering
		\includegraphics[width=\textwidth]{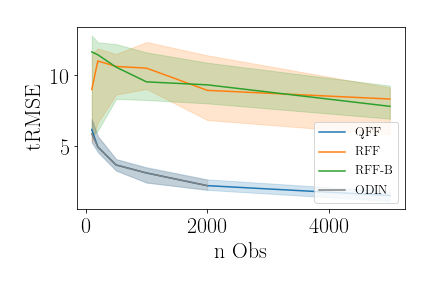}
		\caption{200 features}
	\end{subfigure}
	\caption{tRMSE vs amount of observations for the Lorenz system with noise created using a signal-to-noise ratio of 10.}
\end{figure}

\newpage

\begin{figure}[!h]
	\centering
	\begin{subfigure}[t]{0.325\textwidth}
		\centering
		\includegraphics[width=\textwidth]{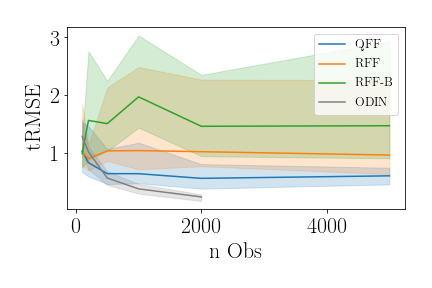}
		\caption{50 features}
	\end{subfigure}
	\hfill
	\begin{subfigure}[t]{0.325\textwidth}
		\centering
		\includegraphics[width=\textwidth]{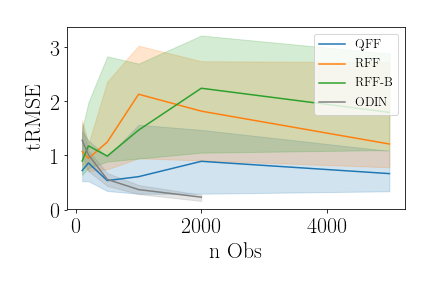}
		\caption{80 features}
	\end{subfigure}
	\hfill
	\begin{subfigure}[t]{0.325\textwidth}
		\centering
		\includegraphics[width=\textwidth]{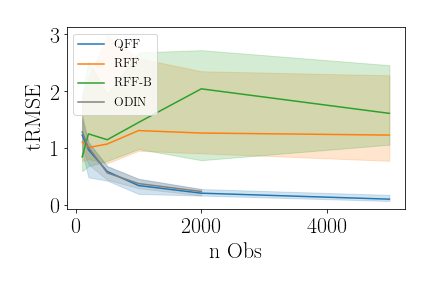}
		\caption{100 features}
	\end{subfigure}
	\\
	\begin{subfigure}[t]{0.325\textwidth}
		\centering
		\includegraphics[width=\textwidth]{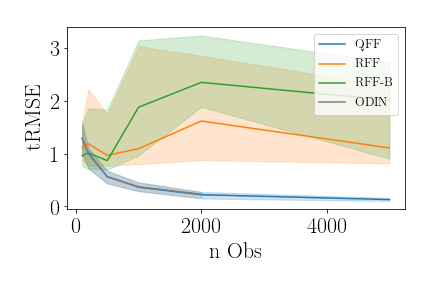}
		\caption{120 features}
	\end{subfigure}
	\hfill
	\begin{subfigure}[t]{0.325\textwidth}
		\centering
		\includegraphics[width=\textwidth]{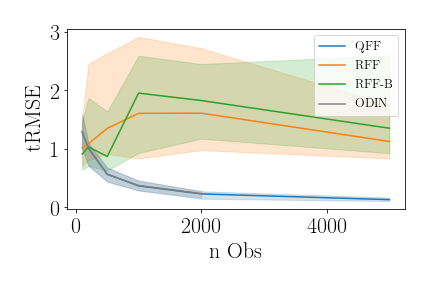}
		\caption{150 features}
	\end{subfigure}
	\hfill
	\begin{subfigure}[t]{0.325\textwidth}
		\centering
		\includegraphics[width=\textwidth]{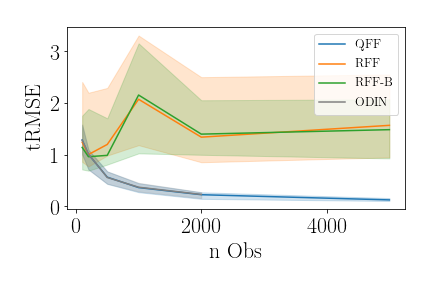}
		\caption{200 features}
	\end{subfigure}
	\caption{tRMSE vs amount of observations for the Lorenz system with noise created using a signal-to-noise ratio of 100.}
\end{figure}

\newpage

\subsubsection{Run Time}

\begin{figure}[!h]
	\centering
	\begin{subfigure}[t]{0.325\textwidth}
		\centering
		\includegraphics[width=\textwidth]{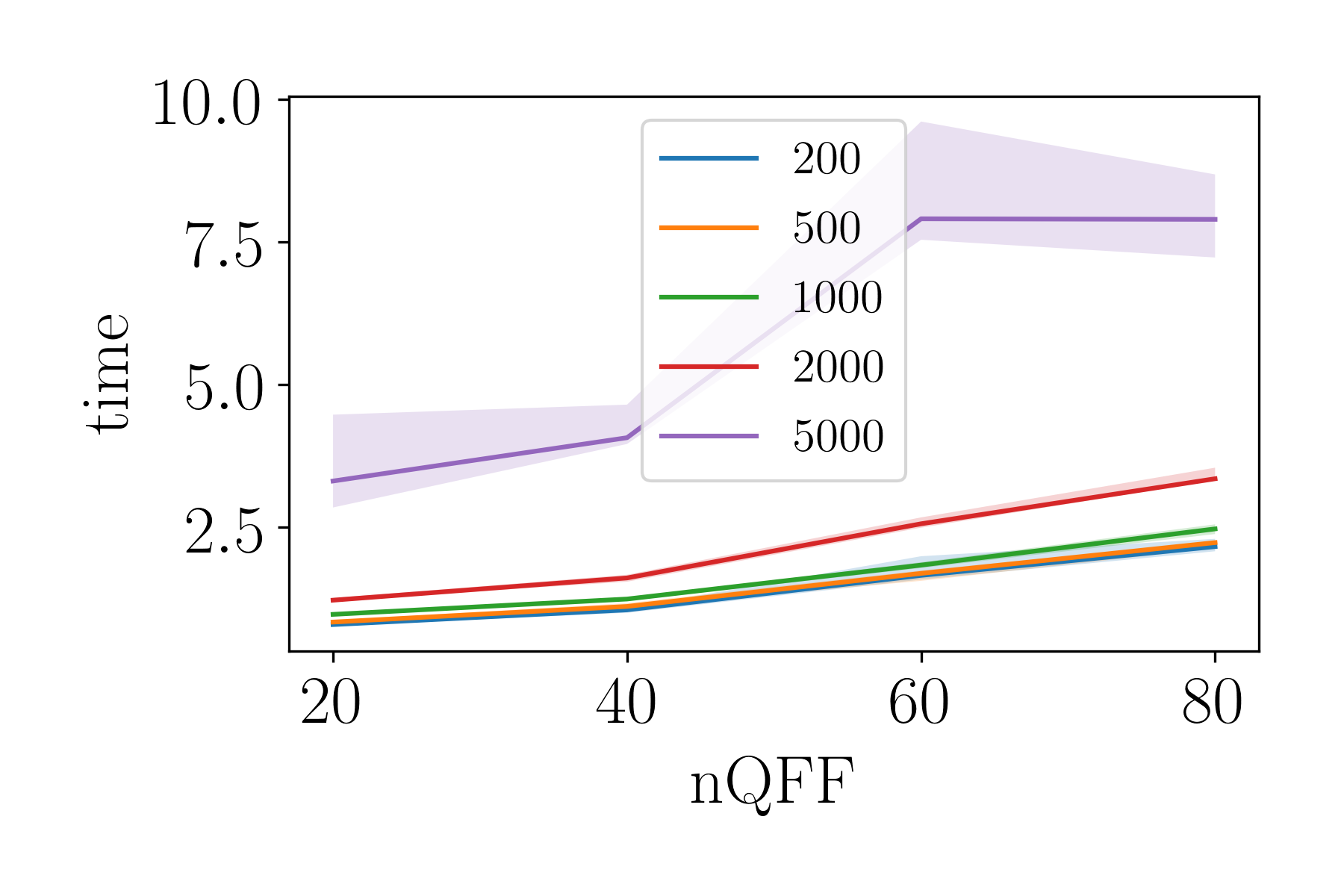}
		\caption{LV}
	\end{subfigure}
	\hfill
	\begin{subfigure}[t]{0.325\textwidth}
		\centering
		\includegraphics[width=\textwidth]{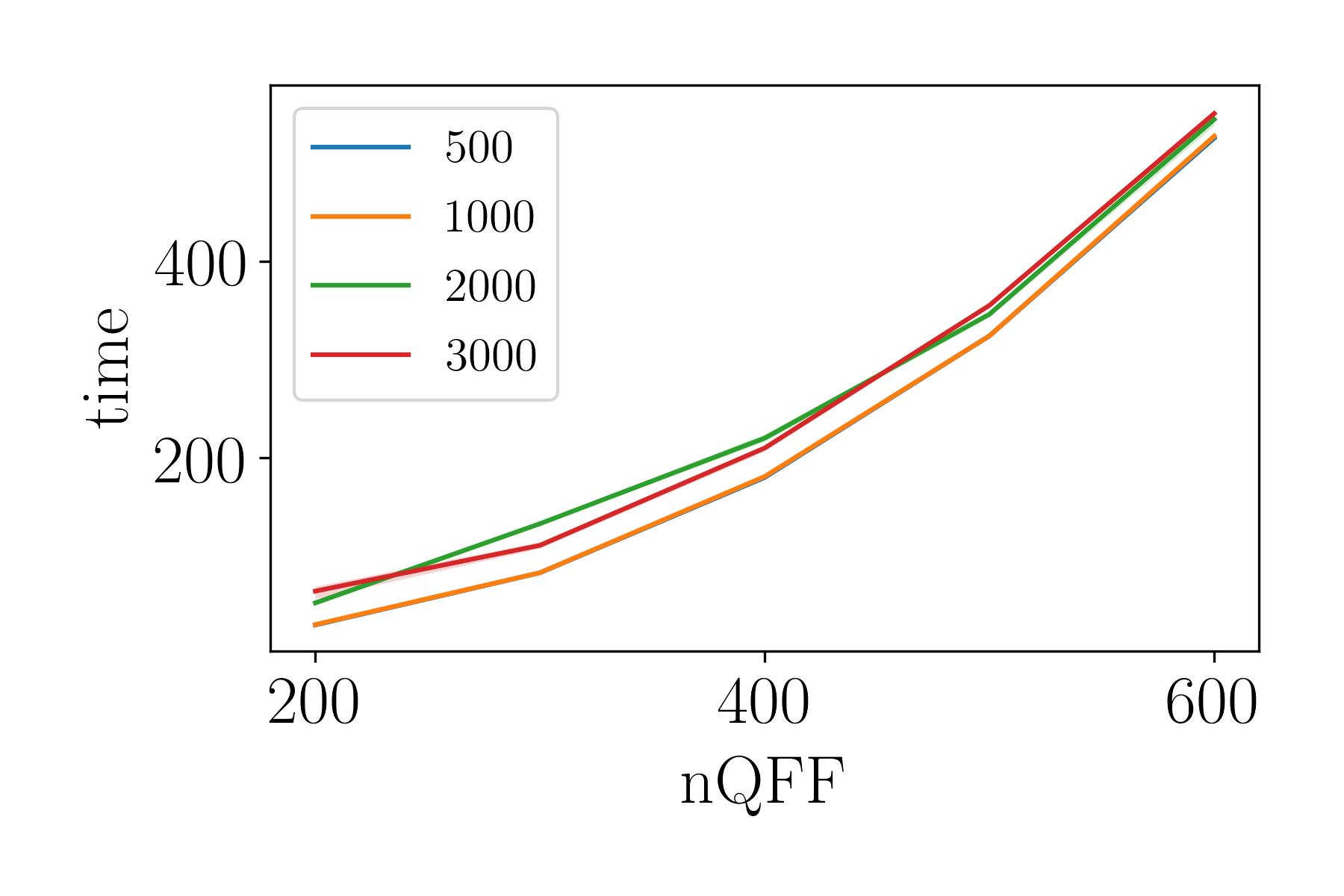}
		\caption{PT}
	\end{subfigure}
	\hfill
	\begin{subfigure}[t]{0.325\textwidth}
		\centering
		\includegraphics[width=\textwidth]{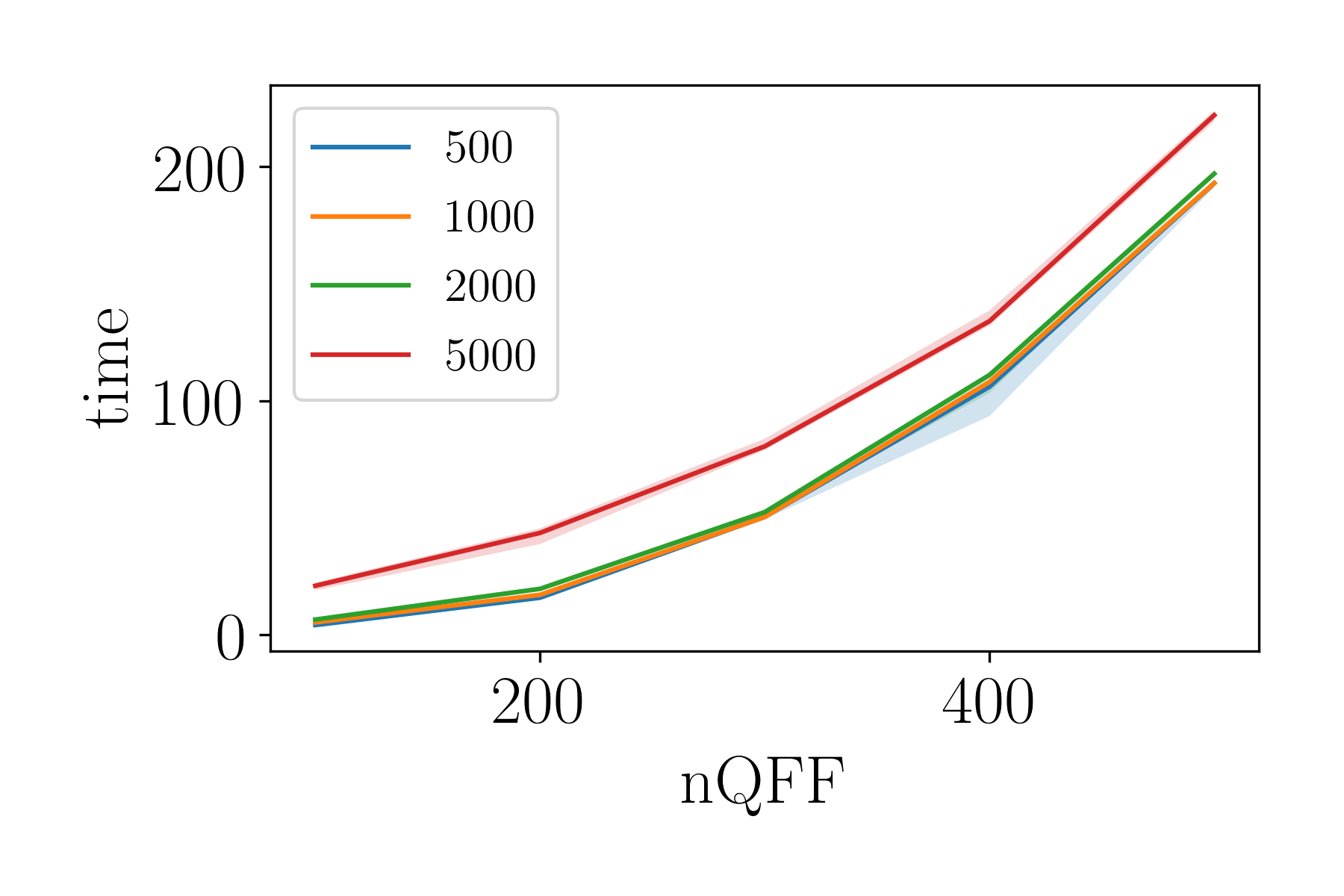}
		\caption{Lorenz}
	\end{subfigure}
	\caption{Run time per iteration in ms vs amount of features for different amounts of observations. As expected from theoretical analysis, the run time per iteration scales approximately cubic.}
	\label{fig:AppendixTimeVsFeatures}
\end{figure}

\begin{figure}[!h]
	\centering
	\begin{subfigure}[t]{0.325\textwidth}
		\centering
		\includegraphics[width=\textwidth]{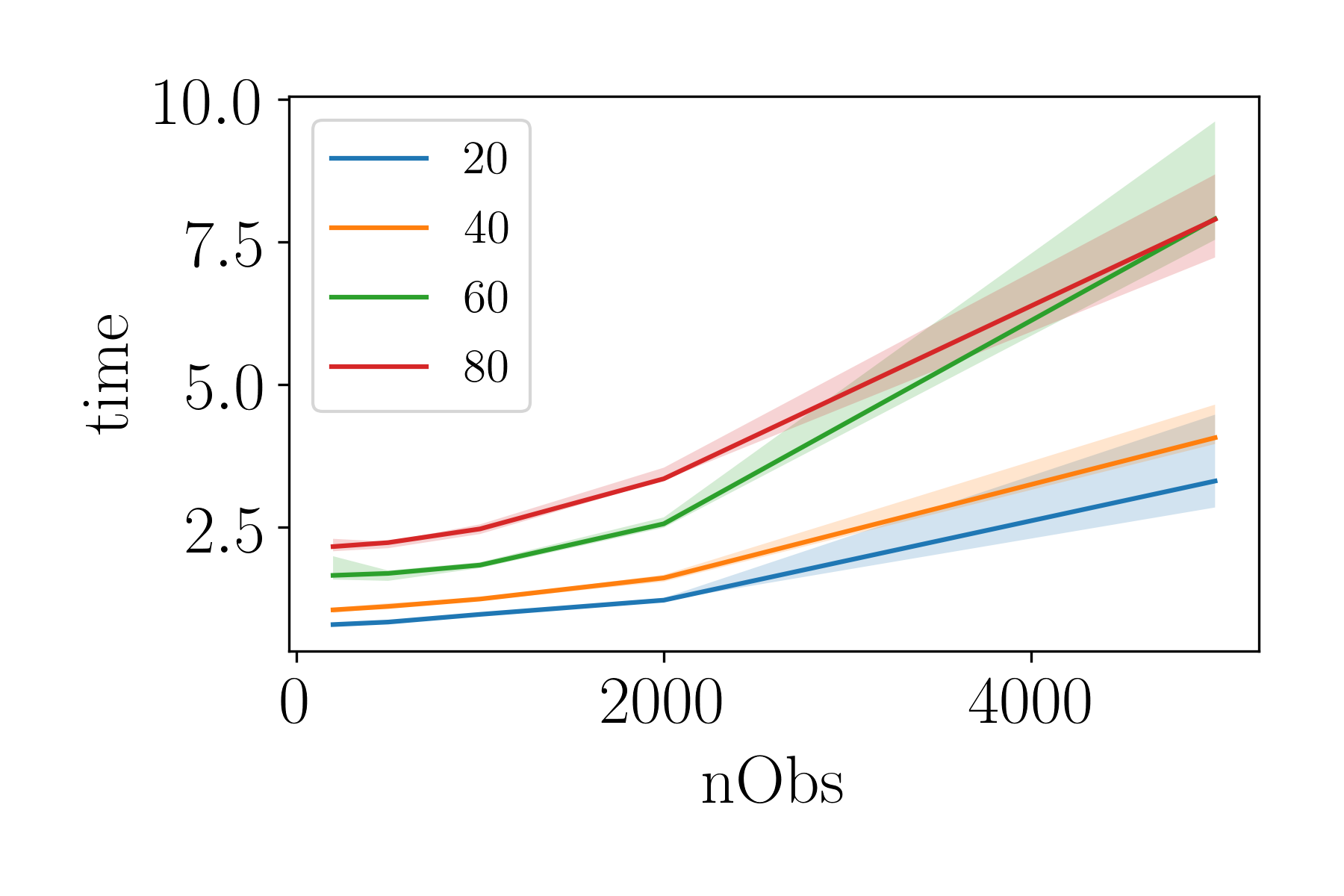}
		\caption{LV}
	\end{subfigure}
	\hfill
	\begin{subfigure}[t]{0.325\textwidth}
		\centering
		\includegraphics[width=\textwidth]{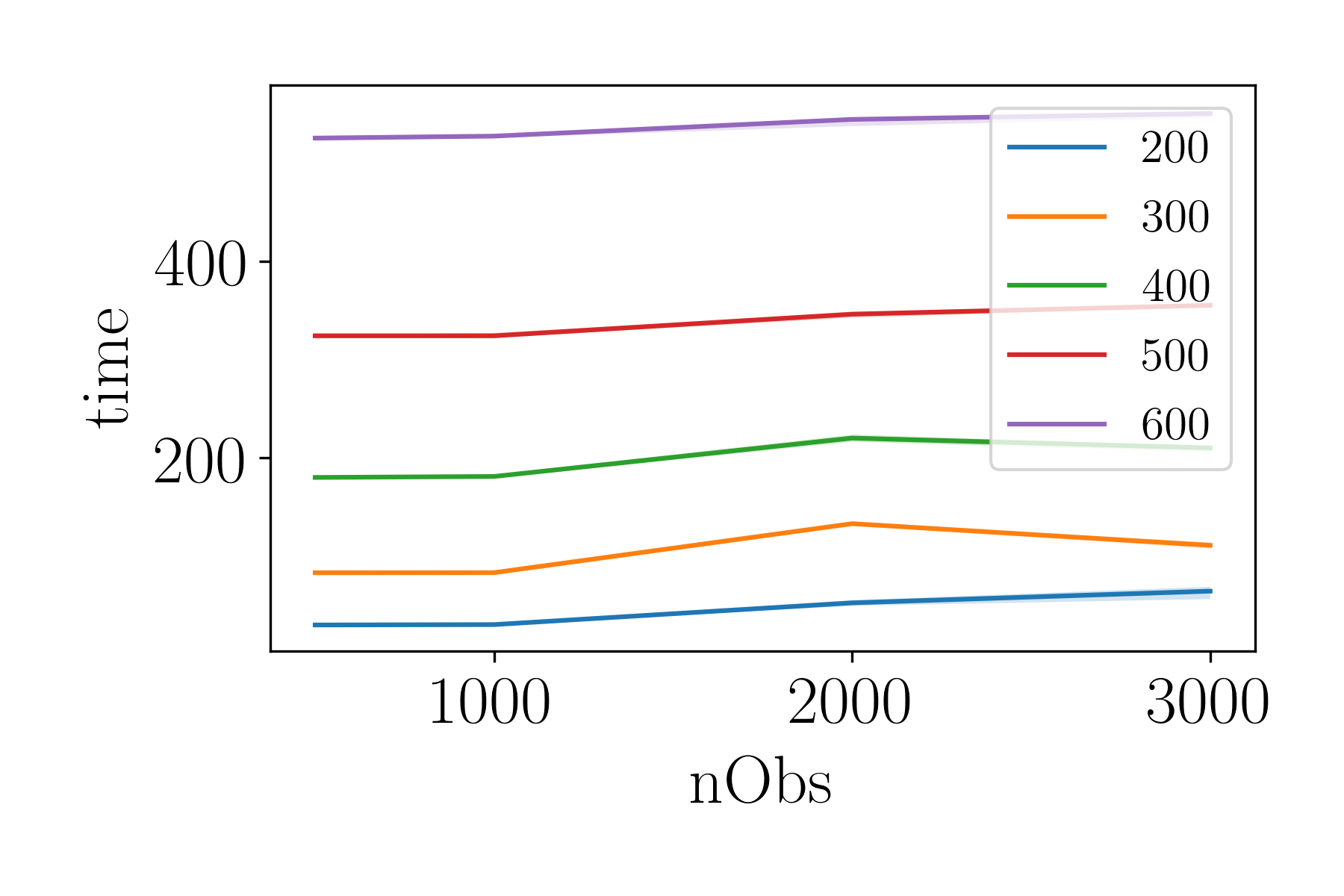}
		\caption{PT}
	\end{subfigure}
	\hfill
	\begin{subfigure}[t]{0.325\textwidth}
		\centering
		\includegraphics[width=\textwidth]{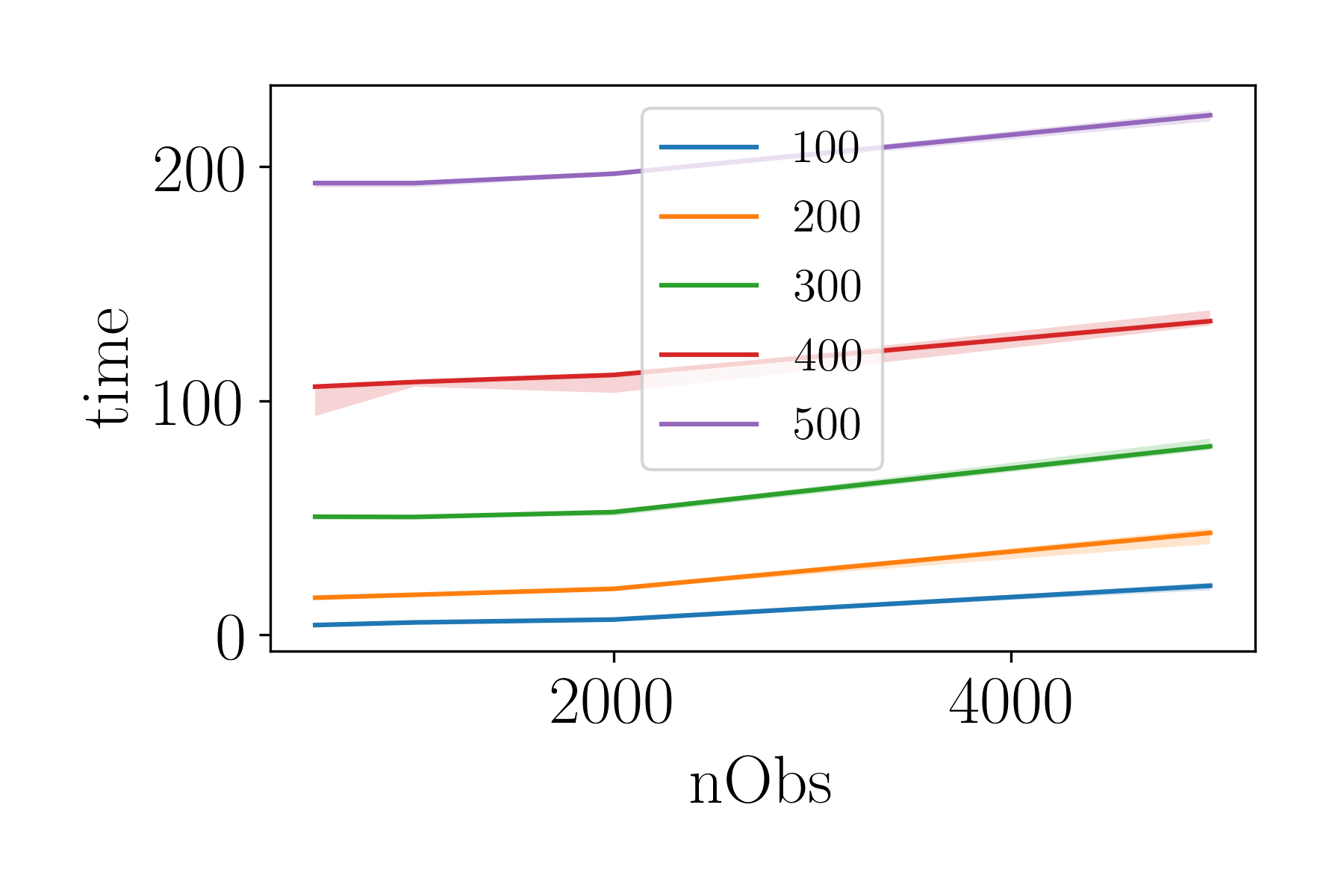}
		\caption{Lorenz}
	\end{subfigure}
	\caption{Run time per iteration in ms vs amount of observations for different amounts of Fourier features. As expected from theoretical analysis, the run time per iteration scales approximately linear, even though there is a strong bias term.}
	\label{fig:AppendixTimeVsObs}
\end{figure}

\newpage

\subsection{Quadrocopter State Inference}
\label{subsec:QuadroStates}
\begin{figure*}[!h]
	\centering
	\begin{subfigure}[t]{0.325\textwidth}
		\centering
		\includegraphics[width=\textwidth]{graphs/Quadro/state0.png}
	\end{subfigure}
	\hfill
	\begin{subfigure}[t]{0.325\textwidth}
		\centering
		\includegraphics[width=\textwidth]{graphs/Quadro/state1.png}
	\end{subfigure}
	\hfill
	\begin{subfigure}[t]{0.325\textwidth}
		\centering
		\includegraphics[width=\textwidth]{graphs/Quadro/state2.png}
	\end{subfigure}
	\\
	\begin{subfigure}[t]{0.325\textwidth}
		\centering
		\includegraphics[width=\textwidth]{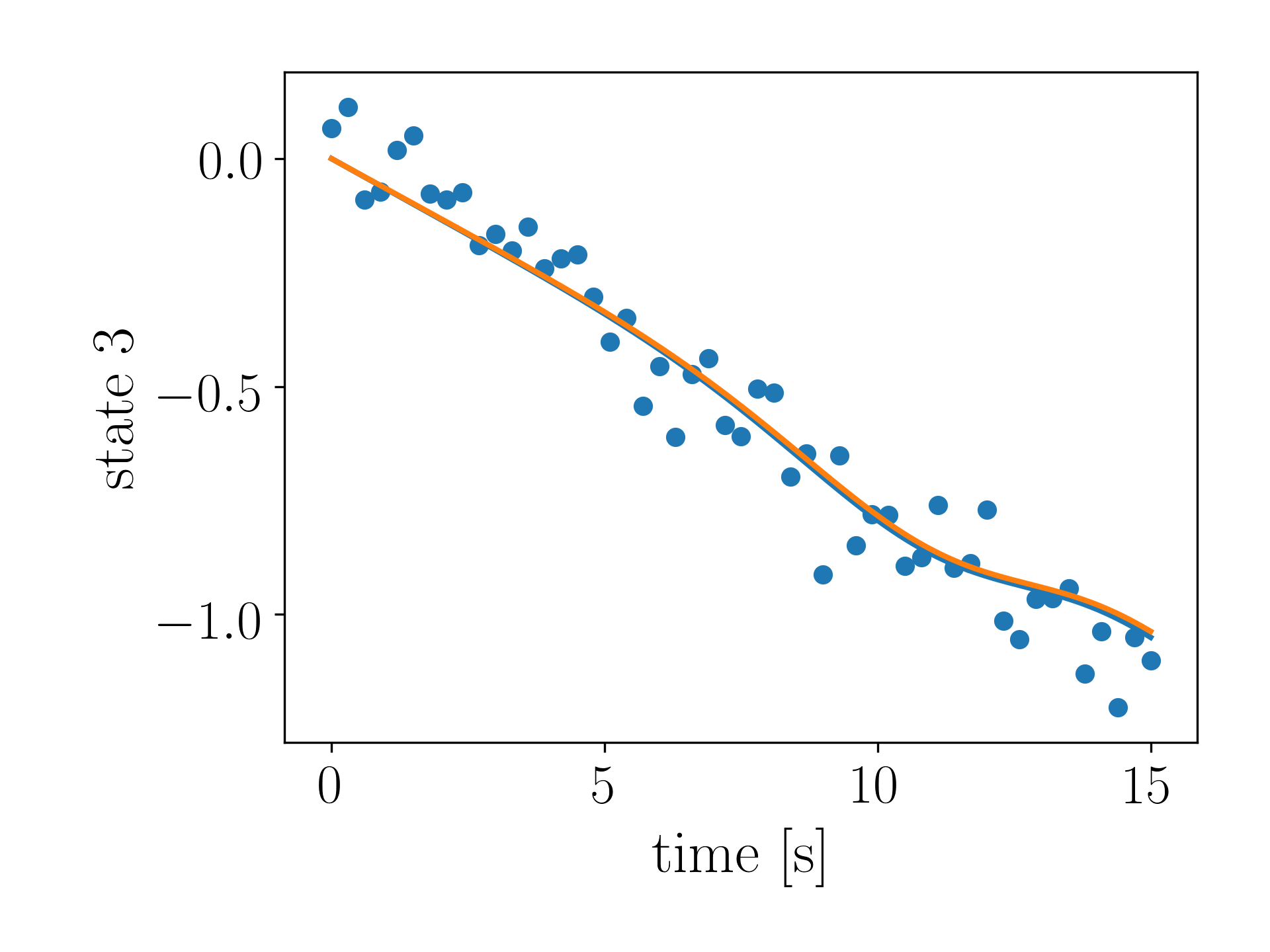}
	\end{subfigure}
	\hfill
	\begin{subfigure}[t]{0.325\textwidth}
		\centering
		\includegraphics[width=\textwidth]{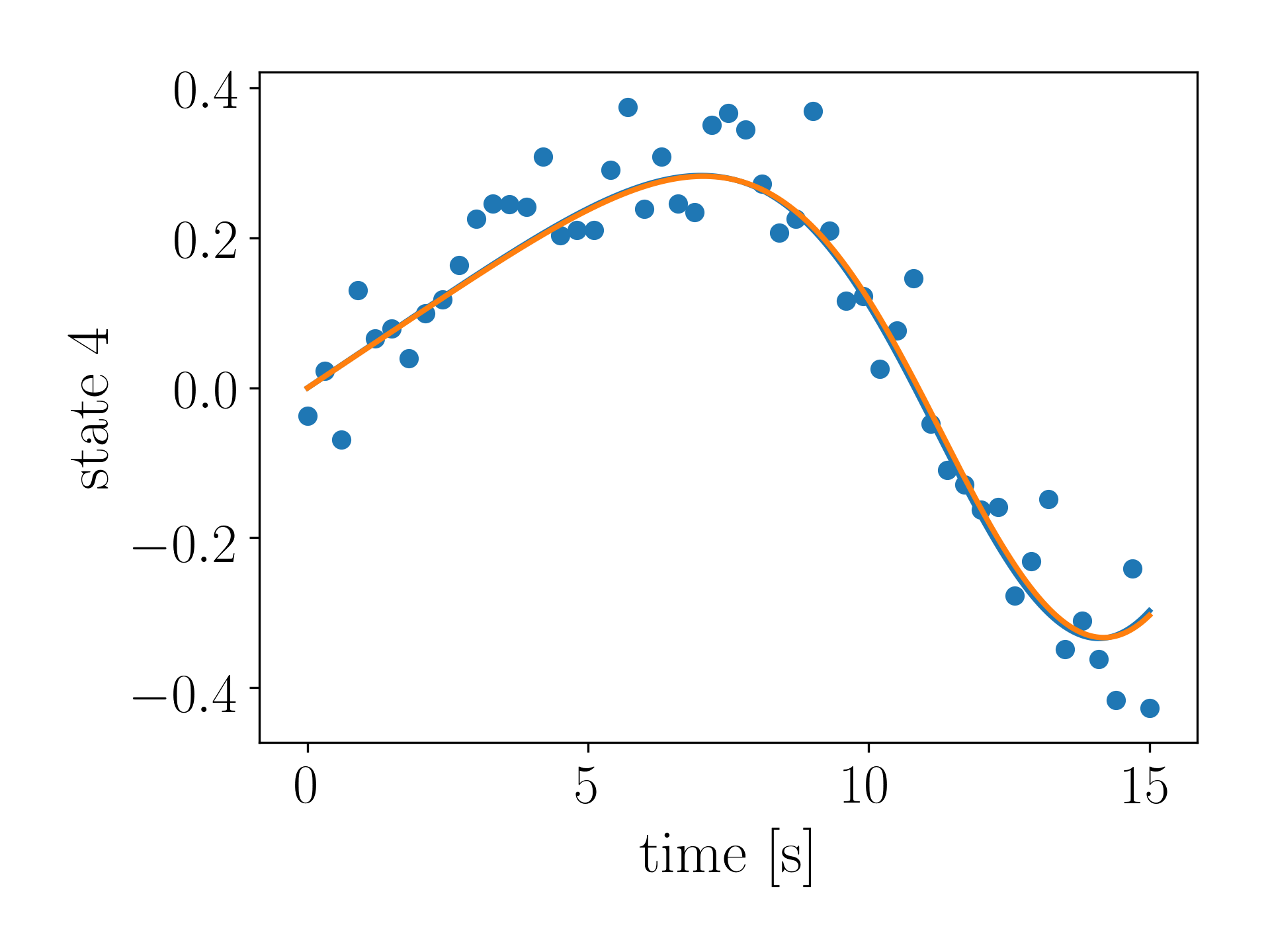}
	\end{subfigure}
	\hfill
	\begin{subfigure}[t]{0.325\textwidth}
		\centering
		\includegraphics[width=\textwidth]{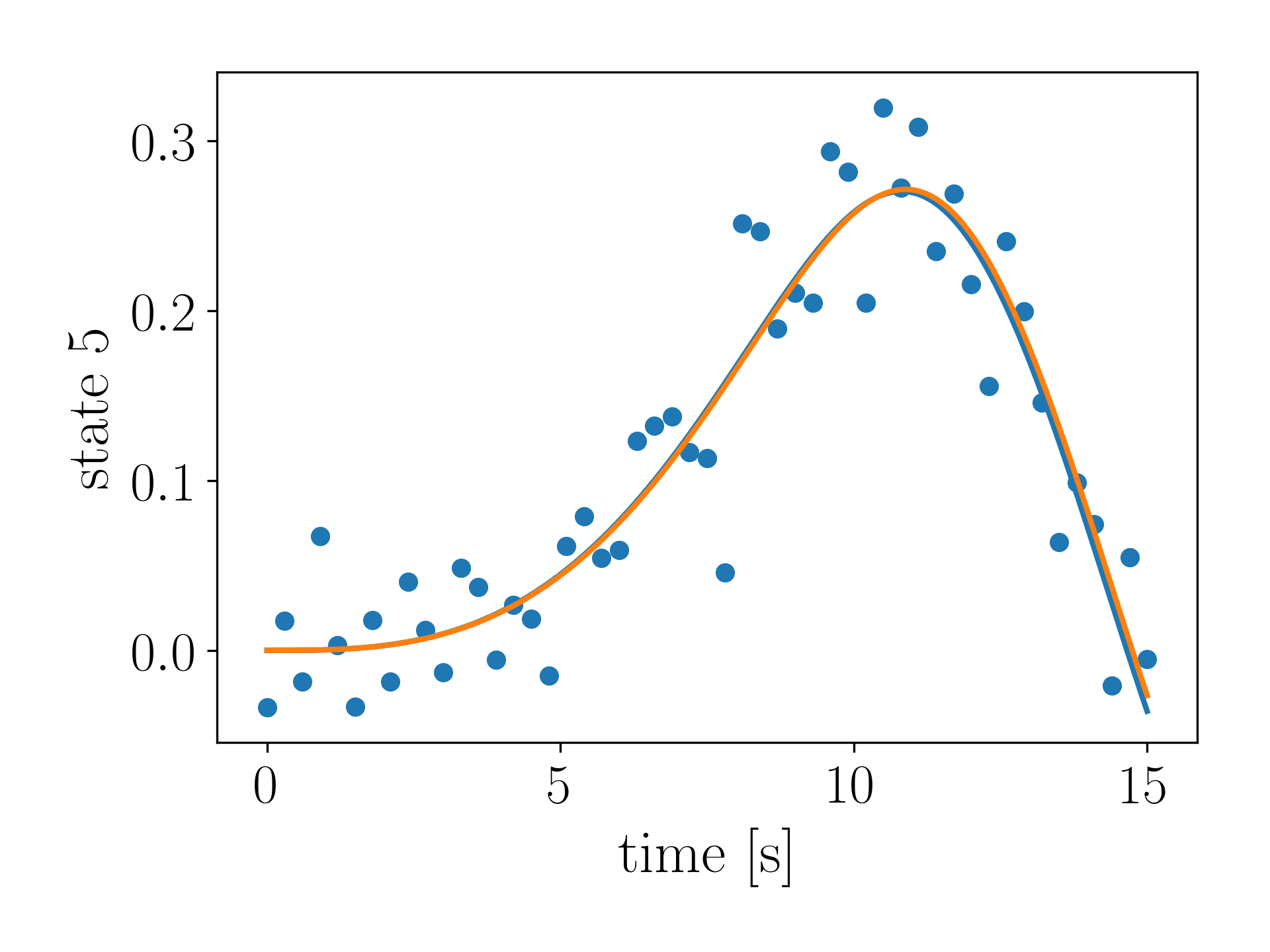}
	\end{subfigure}
	\\
	\begin{subfigure}[t]{0.325\textwidth}
		\centering
		\includegraphics[width=\textwidth]{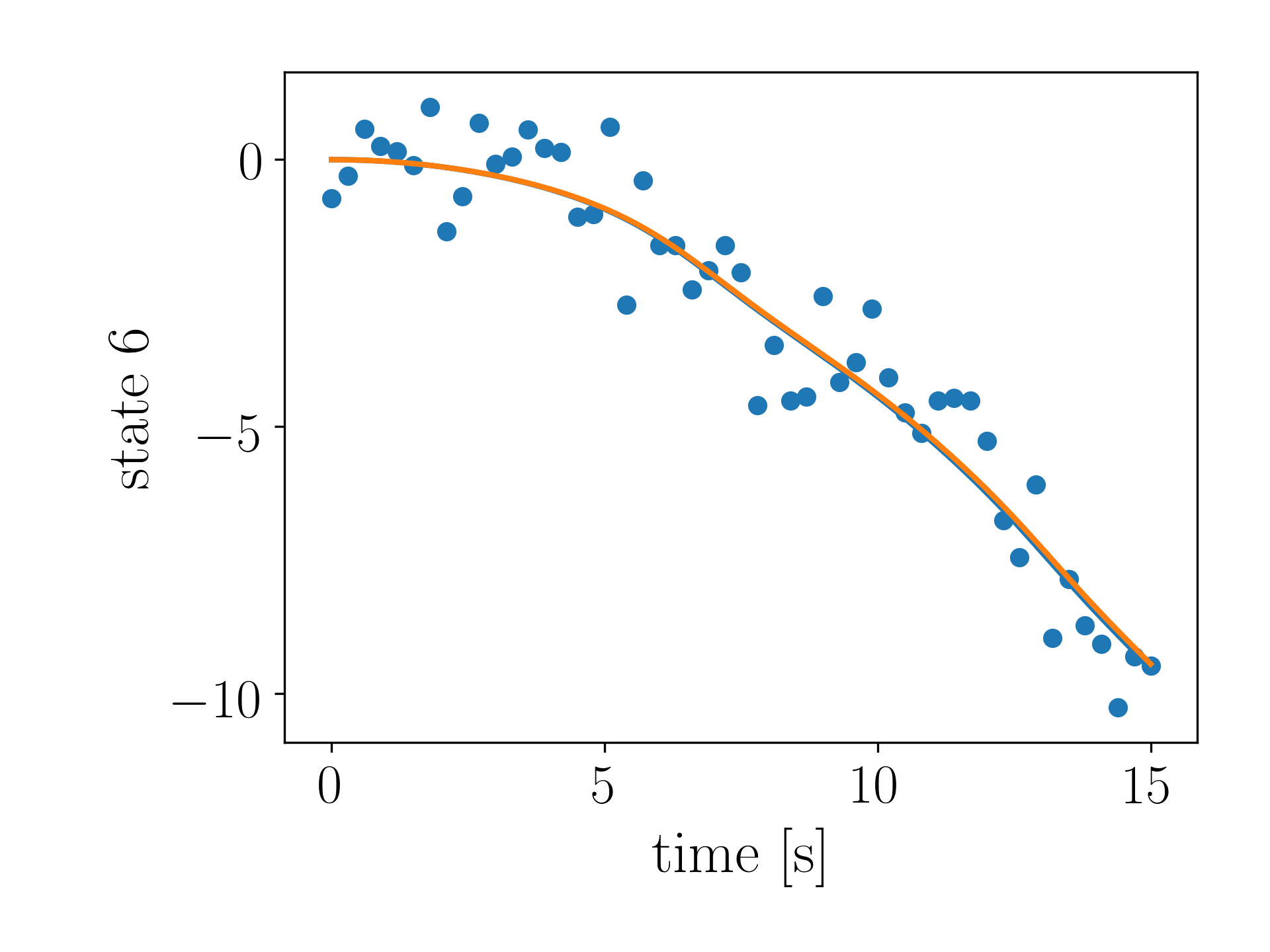}
	\end{subfigure}
	\hfill
	\begin{subfigure}[t]{0.325\textwidth}
		\centering
		\includegraphics[width=\textwidth]{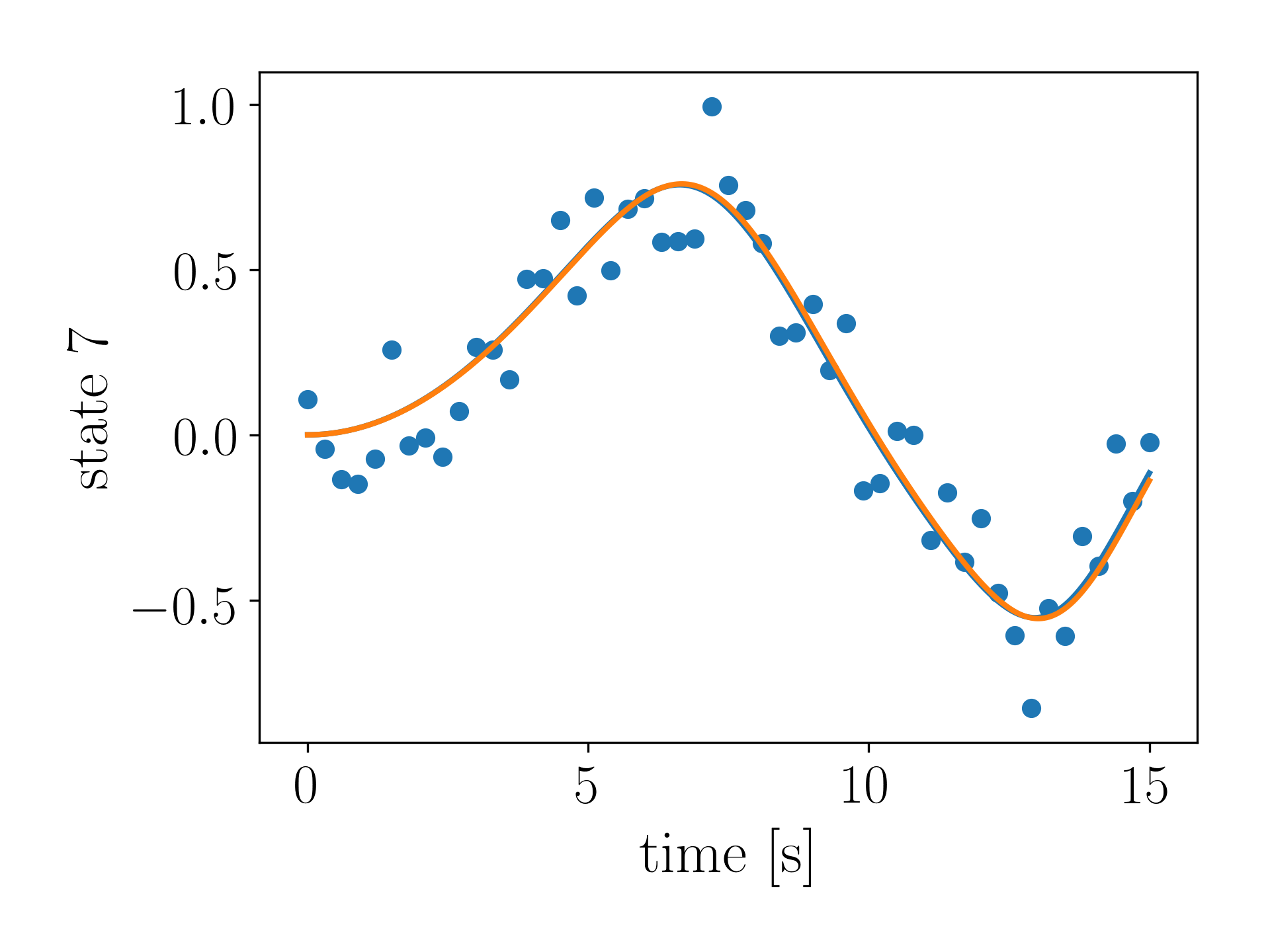}
	\end{subfigure}
	\hfill
	\begin{subfigure}[t]{0.325\textwidth}
		\centering
		\includegraphics[width=\textwidth]{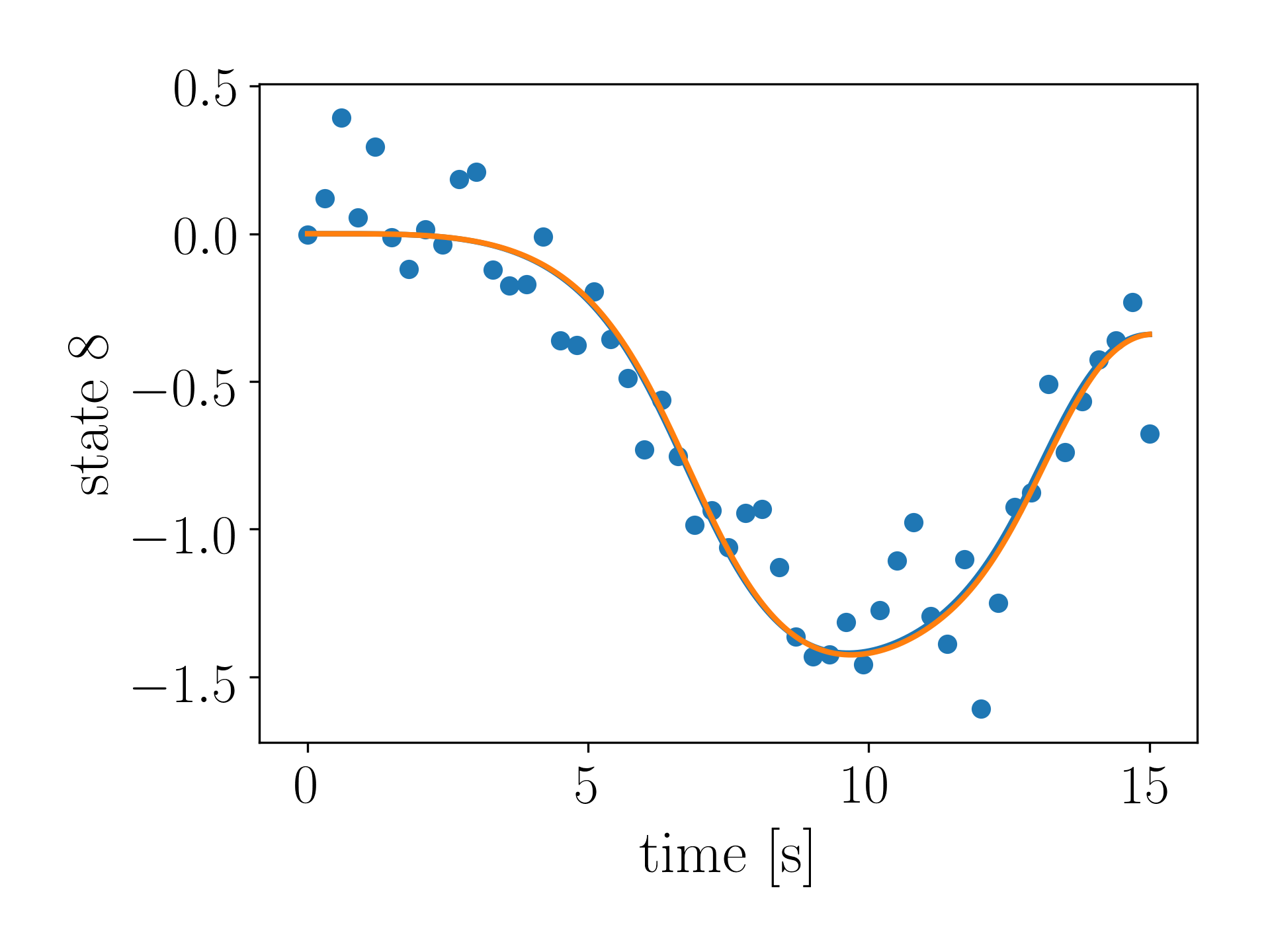}
	\end{subfigure}
	\\
	\begin{subfigure}[t]{0.325\textwidth}
		\centering
		\includegraphics[width=\textwidth]{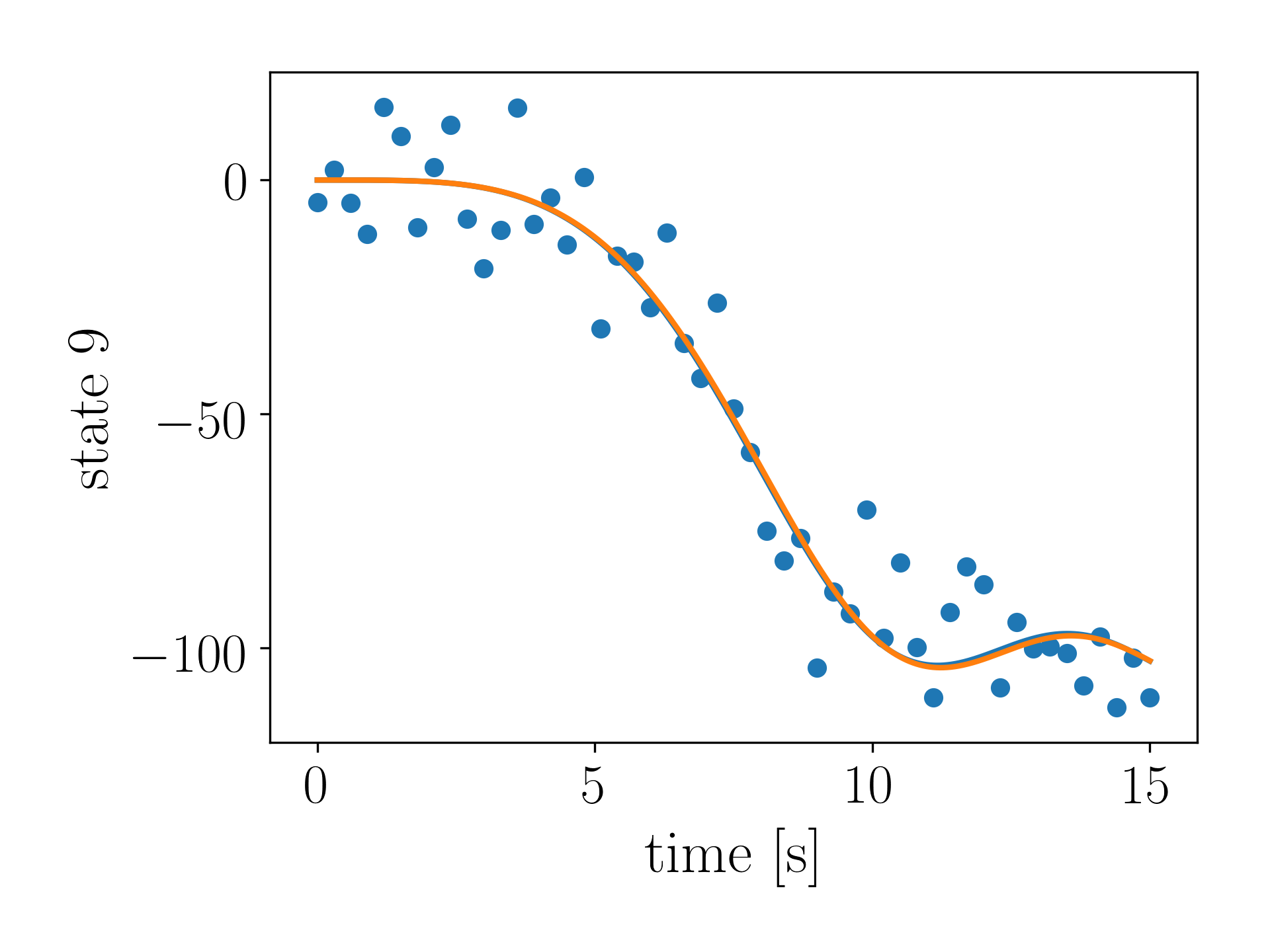}
	\end{subfigure}
	\hfill
	\begin{subfigure}[t]{0.325\textwidth}
		\centering
		\includegraphics[width=\textwidth]{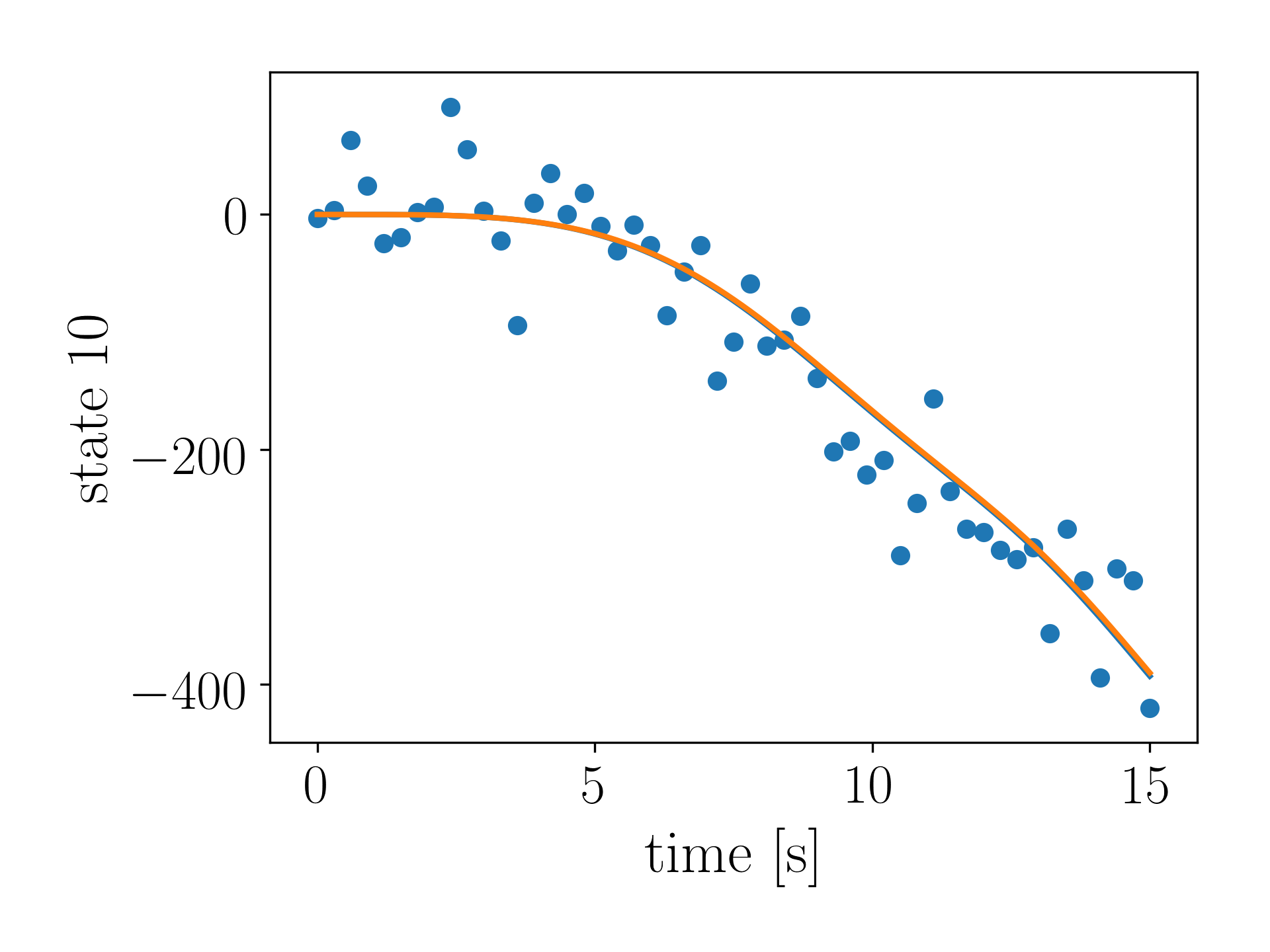}
	\end{subfigure}
	\hfill
	\begin{subfigure}[t]{0.325\textwidth}
		\centering
		\includegraphics[width=\textwidth]{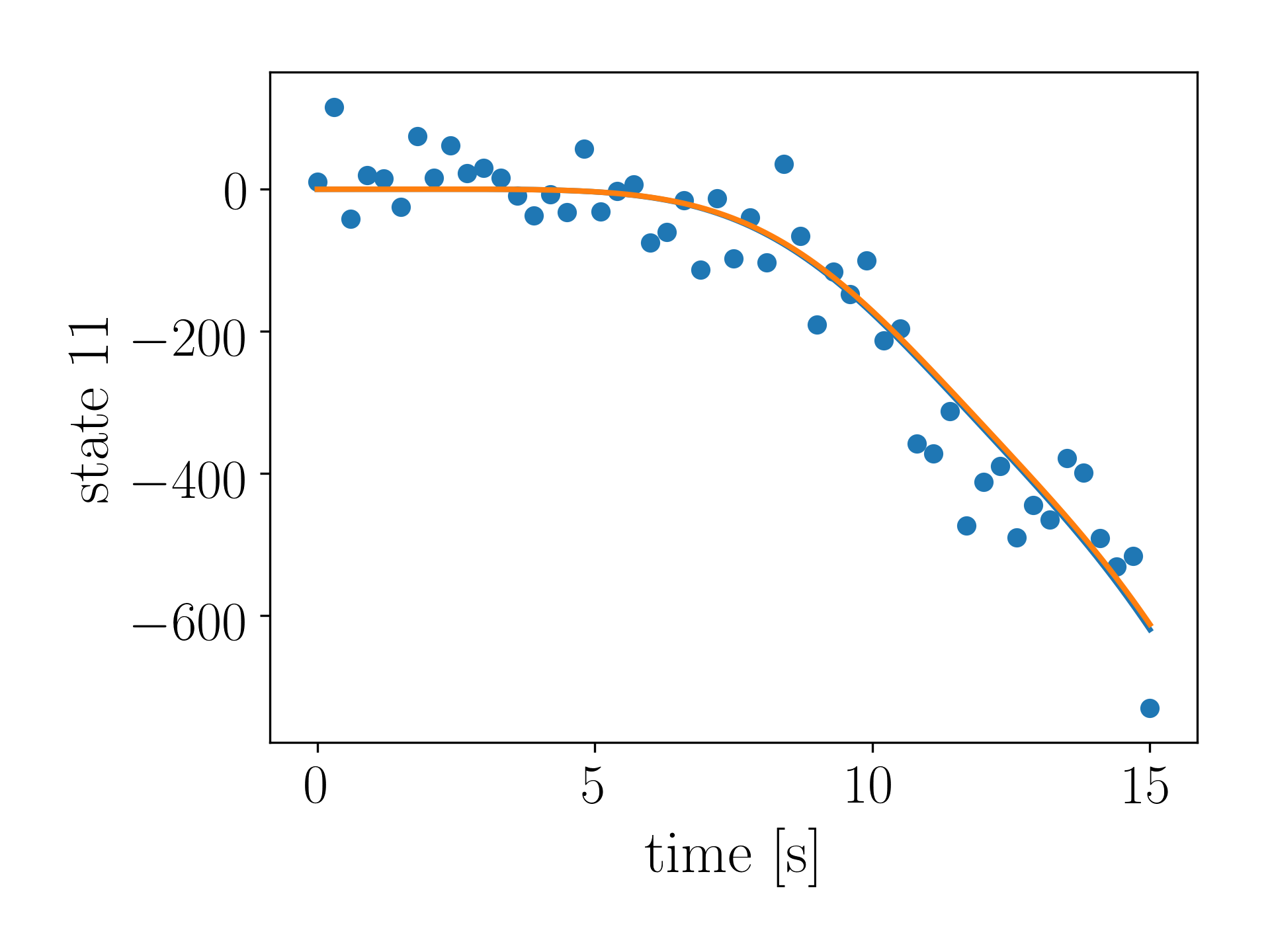}
	\end{subfigure}	
	%	\vspace{-0.3cm}
	\caption{State trajectories obtained by integrating the parameters inferred by $\operatorname{ODIN-S}$ (orange). The blue line represents the ground truth, while the blue dots show every 300-th observation for a signal-to-noise ratio of 10.}
	\label{fig:QuadroRMSEFull}
\end{figure*}

\vskip 0.2in

\end{document}